\pdfoutput=1 % Use PDFLaTeX

\documentclass{phdthesis}

\usepackage[english]{babel}
\usepackage[utf8]{inputenc}
\usepackage{lipsum}
\usepackage{bbding}
\usepackage{amsmath}
\usepackage{amssymb}
\usepackage{mathtools}
\usepackage{tabularx}
\usepackage{pbox}
\usepackage{verbatim}
\usepackage{subcaption}
\usepackage{colortbl}
\usepackage{etoolbox}
\usepackage{booktabs}
\usepackage{arydshln}
\usepackage{makecell}
\usepackage{tabu}
\usepackage{multirow}
\usepackage{enumitem}
\usepackage{setspace}
\usepackage{textcomp}
\usepackage{csquotes}
\usepackage[percent]{overpic}

\usepackage[font=small,labelsep=period]{caption}
\usepackage{makeidx}
\usepackage[toc]{glossaries}
\makeglossaries

\newcolumntype{Y}{>{\centering\arraybackslash}X}
\newcolumntype{R}{>{\raggedleft\arraybackslash}X}
\newcolumntype{L}{>{\raggedright\arraybackslash}X}

\newcommand\customparagraph[1]{\vspace{0.7em}\noindent\textbf{#1}}

\makeatletter
\newcommand\footnoteref[1]{\protected@xdef\@thefnmark{\ref{#1}}\@footnotemark}
\makeatother

\newcommand{\win}[1]{\textcolor{NavyBlue}{\textbf{#1}}}
\newcommand{\wincat}[1]{\textcolor{black}{\textbf{#1}}}
\newcommand\myeq{\mkern1.5mu{=}\mkern1.5mu}

\definecolor{bblue}{rgb}{0.0,0.25,0.65}
\definecolor{ccol}{rgb}{0.2,0.2,0.2}
\definecolor{ccol2}{rgb}{0.9,0.9,0.9}
\definecolor{avgcol}{rgb}{1.0,0.839,0.4}
\definecolor{arcol}{rgb}{0.341,0.733,0.541}
\definecolor{timecol}{rgb}{0.902,0.486,0.451}
\definecolor{lightgray}{rgb}{0.935, 0.935, 0.935}

\newcommand{\light}{\textcolor[rgb]{0.85,0.85,0.85}}

\newcommand{\mytilde}{\raise.17ex\hbox{$\scriptstyle\sim$}}

\usepackage{xspace}
\makeatletter
\DeclareRobustCommand\onedot{\futurelet\@let@token\@onedot}
\def\@onedot{\ifx\@let@token.\else.\null\fi\xspace}

\newcommand{\mlcellc}[2][c]{%
	\begin{tabular}[#1]{@{}c@{}}#2\end{tabular}}

\def\eg{e.g\onedot} 
\def\ie{i.e\onedot} 
 
\def\etc{etc\onedot} \def\vs{vs\onedot}
\def\wrt{w.r.t\onedot}

\def\etal{et al\onedot}
\makeatother

\makeatletter
\def\adl@drawiv#1#2#3{%
	\hskip.5\tabcolsep
	\xleaders#3{#2.5\@tempdimb #1{1}#2.5\@tempdimb}%
	#2\z@ plus1fil minus1fil\relax
	\hskip.5\tabcolsep}
\newcommand{\cdashlinelr}[1]{%
	\noalign{\vskip\aboverulesep
		\global\let\@dashdrawstore\adl@draw
		\global\let\adl@draw\adl@drawiv}
	\cdashline{#1}
	\noalign{\global\let\adl@draw\@dashdrawstore
		\vskip\belowrulesep}}
\makeatother

\makeatletter
\newlength{\qrr@dimen@}
\expandafter\pretocmd\csname tabular*\endcsname{\setlength{\qrr@dimen@}{#1}}{}{}
\newcommand*{\Rowcolor}[2][\tabcolsep]{%
	\ifx\relax#1\relax\else
	\kern-\the\dimexpr#1\relax
	\fi
	\makebox[0pt][l]{%
		\fboxsep=0pt
		\colorbox{#2}{%
			\strut\kern\qrr@dimen@
		}%
	}%
	\ifx\relax#1\relax\else
	\kern\the\dimexpr#1\relax
	\fi
	\ignorespaces
}
\makeatother

\usepackage{fancyhdr}
\usepackage{titlesec}
\titleformat{\chapter}[display]
{\LARGE\bfseries}
{\filright\Large{\chaptertitlename} \Large\thechapter}
{1.0ex}
{\titlerule\vspace{1.2ex}}
[\vspace{1ex}]

\titlespacing*{\chapter}{0pt}{0pt}{40pt}

\titleformat*{\section}{\Large\bfseries}
\titleformat*{\subsection}{\large\bfseries}
\titleformat*{\subsubsection}{\large\bfseries}

\bibliography{references}

\title{Pose Estimation of Specific Rigid Objects}
\author{Ing.\ Tom{\'a}{\v{s}} Hoda{\v{n}}}

\supervisor{Prof.\ Ing.\ Ji{\v{r}}{\'i} Matas, Ph.D.}
\supervisorAffiliation{
	Czech Technical University in Prague\\
	Faculty of Electrical Engineering\\
	Department of Cybernetics\\
	Karlovo n{\'a}m{\v{e}}st{\'i} 13\\
	121 35 Prague 2\\
	Czech Republic}

\placeyear{Prague, February 2021}
\date{February 2021}

\begin{document}
	
	\spacing{1.15}
	
	\maketitle

	\newpage
\chapter*{Abstract}
\addcontentsline{toc}{chapter}{Abstract}

In this thesis, we address the problem of estimating the 6D pose of %
rigid objects from a single RGB or RGB-D input image, assuming that 3D models of the objects are available. This problem is of great importance to many application fields such as robotic manipulation, augmented reality, and autonomous driving.

First, we propose EPOS, a method for 6D object pose estimation from an RGB \mbox{image}. The key idea is to represent an object by compact surface fragments and predict the pro\-ba\-bi\-li\-ty distribution of corresponding fragments at each pixel of the input image by a neural network. Each pixel is linked with a data-dependent number of fragments,
which allows systematic handling of symmetries, and the 6D poses are estimated from the %
links by a RANSAC-based fitting method. EPOS is applicable to a broad range of objects, including challenging ones with global or partial symmetries and without any texture, and outperformed all RGB and most RGB-D and D methods on several standard datasets.

Second, we present HashMatch,
an RGB-D method that slides a window over the input image and searches for a match against
templates, which are pre-generated by rendering 3D object models in different orientations.
The method applies a cascade of evaluation stages to each window location, which avoids exhaustive matching against all templates.
The key is a voting stage based on hashing that generates a small set of
candidate templates, which are then processed by more expensive verification and refinement stages.

Third, we propose ObjectSynth, an approach to synthesize
photorealistic images of 3D object models for training methods based on neural networks.
The 3D object models are arranged in 3D models of indoor scenes and the images are synthesized by physically-based rendering.
The resulting images yield substantial improvements compared to commonly used images of objects rendered on top of random photographs.

Fourth, we introduce T-LESS, a dataset for 6D object pose estimation that includes 3D models and training and test RGB-D images of thirty
electrical parts.
These objects have no significant texture or discriminative color, exhibit symmetries and similarities in shape and size, and some objects are a composition of others.
T-LESS is the first dataset to include objects of such properties which are common in industrial environments.

Fifth, we define BOP, a benchmark that captures the status quo in the field.
The benchmark currently comprises eleven datasets in a unified format,
an evaluation methodology, an online evaluation system, and public challenges held at
international workshops organized annually at the ICCV and ECCV conferences.

\newpage
\chapter*{Abstrakt}

Tématem této disertační práce je odhad 3D pozice a 3D orientace rigidních objektů z jediného RGB nebo RGB-D snímku, kdy 3D modely objektů jsou předem známé. Řešení této úlohy počítačového vidění má široké uplatnění v mnoha aplikacích, jako je například robotická manipulace, rozšířená realita nebo autonomní řízení vozidel.

Prvním přínosem práce je metoda EPOS pro odhad pozice a orientace objektů z RGB snímku. Hlavní myšlenkou je reprezentovat povrch objektů množinou kompaktních fragmentů a pro každý fragment a každý bod na snímku odhadnout pomocí neuronové~sítě prav\-dě\-po\-do\-bnost, s jakou daný bod leží na projekci daného fragmentu.
Každý bod~na snímku je na základě těchto odhadů provázán s potenciálně mnoha fragmenty, což umož\-ňu\-je zachytit případné symetrie objektu. Pozice a orientace objektů jsou odhadnuty z na\-vá\-za\-ných korespondencí robustní metodou založenou na algoritmu RANSAC. Metoda EPOS je použitelná pro celou řadu objektů, včetně symetrických objektů a objektů bez textury, a překonala všechny metody pro odhad z RGB snímku a většinu metod pro odhad z RGB-D a D snímku na několika standardních datasetech.

Druhým přínosem je metoda HashMatch, která prochází vstupní RGB-D snímek posuvným oknem a pro každou pozici okna hledá odpovídající šablonu z množiny získané syntézou různých pohledů na 3D modely objektů. Každá pozice okna je vyhodnocena kaskádou kroků,
díky které se metoda vyhýbá porovnávání se všemi šablonami.
Klíčovým krokem je rychlá identifikace malého počtu potenciálně odpovídajících šablon pomocí hlasování, kde každý hlas je vypočítán z hloubkové informace z několika bodů v okně.

Třetím přínosem je přístup ObjectSynth pro syntézu fotorealistických snímků
pro trénování metod využívajících neuronové sítě. 3D modely objektů jsou rozmístěny v 3D modelech místností a snímky jsou získány fyzikálně založeným renderováním.
Metody trénované na těchto snímcích dosahují výrazného zlepšení v porovnání s běžně používanými snímky zobrazujícími 3D modely objektů na náhodných fotografiích.

Čtvrtým přínosem je dataset T-LESS, který obsahuje 3D modely a trénovací a testovací snímky třiceti elektronických součástek. Tyto součástky nemají výraznou texturu nebo barvu, v mnoha případech jsou symetrické, vzájemně si podobné tvarem či velikostí, a některé součástky jsou složeninami z ostatních. T-LESS je prvním datasetem obsahujícím objekty těchto vlastností, které jsou časté v průmyslovém prostředí.

Pátým přínosem je srovnávací systém BOP, který zachycuje status quo v odhadu pozice a orientace objektů. BOP aktuálně obsahuje jedenáct datasetů v jednotném formátu, vyhodnocovací metodologii, webový vyhodnocovací portál, a veřejné soutěže pořádané na mezinárodních seminářích organizovaných 
každoročně
na konferencích ICCV a ECCV.

	\newpage
\chapter*{Acknowledgements}
\addcontentsline{toc}{chapter}{Acknowledgements}

\begin{figure}[h!]
	\centering
	\includegraphics[width=1.0\linewidth]{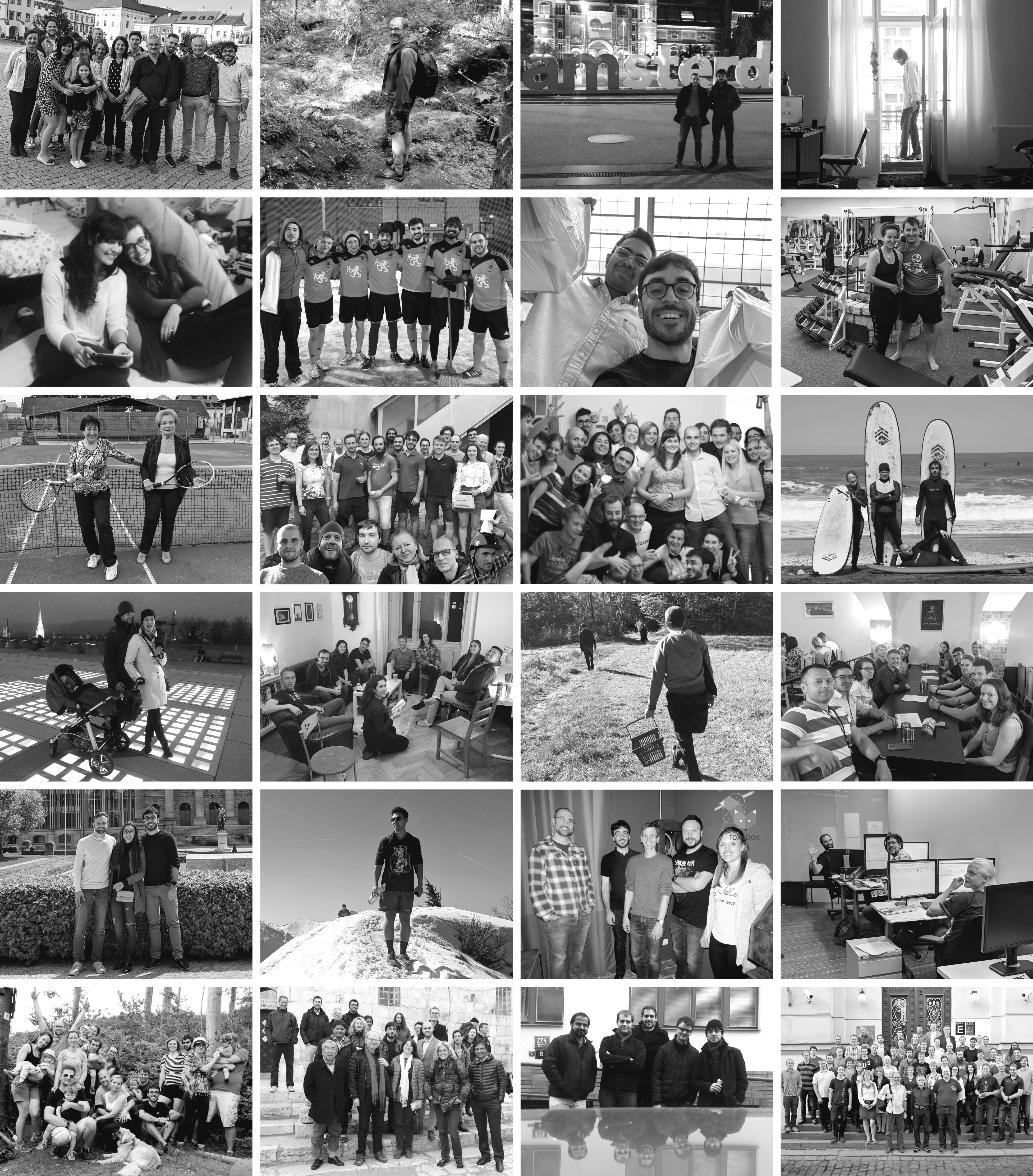} \\
\end{figure}

	\tableofcontents
	
	\mainmatter
	
	\chapter{Introduction} \label{sec:introdution}

As humans, we heavily rely on visual perception when exploring and navigating the world around us.
With a flash glance, we can estimate the three-dimensional structure of the surrounding scene, recognize the scene elements, and understand the semantic context.
Researchers in computer vision aim to bring such perception ability to computers, \ie, to create a ``seeing machine''~\cite{prince2012computer}, by applying mathematical techniques to
digital~images.

One classical area of computer vision is object recognition, which focuses on extracting information about objects from images.
An object is commonly defined as a
thing with a well-defined boundary and center~\cite{alexe2010object}, and may be of various materials and flexibility properties.
Multiple object instances may be visible in an image, with each instance belonging to a possibly different object class, which may represent an object category (dogs,~cars, drinks, \etc) or a specific object (a~black Ford~T from 1908, a~0.33l Coca-Cola can, \etc).

Depending on the required information about object classes or instances visible in the input image, we can distinguish several object recognition problems. The basic problem is image classification~\cite{russakovsky2015imagenet}, where the goal is to predict only labels of visible object classes without requiring any information about individual object instances.
The number of visible object instances is additionally required in object counting~\cite{arteta2016counting}, their bounding boxes in object detection~\cite{huang2017speed}, and pixel-accurate delineations in object segmentation~\cite{he2017mask}.

\begin{figure}[t!]
	\begin{center}
		\includegraphics[width=0.63\linewidth]{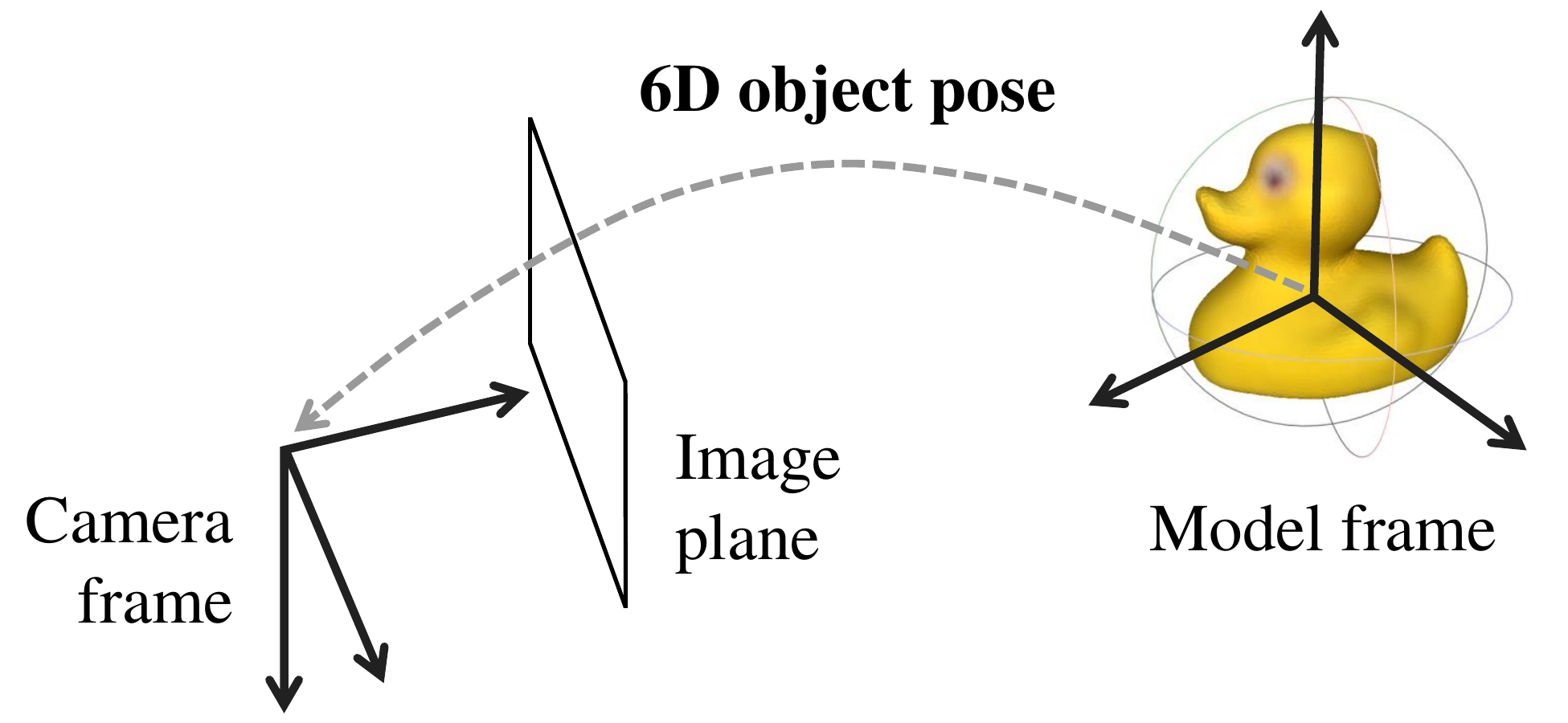} \\
		\vspace{1.0ex}
		\caption{\label{fig:object_pose} \textbf{6D pose of a rigid object.}
			The size and shape of a rigid object are fixed and remain unaltered when forces are applied. %
			The pose of a rigid object is therefore fully defined by a rigid transformation with six degrees of freedom (three in translation and three in rotation) from the 3D coordinate frame of the object model to the 3D coordinate frame of the camera.}
	\end{center}
\end{figure}

The problem of object pose estimation goes
further and requires estimating poses of visible object instances in the 3D space. In the case of rigid objects, the pose has six~degrees of freedom, \ie, three in translation and three in rotation, and is referred to as the 6D object pose (Figure~\ref{fig:object_pose}).
This problem is of great importance to numerous application fields.
For example, in robotic manipulation, the information about object poses allows an end effector to act upon the objects.
In augmented reality, this information enables enhancing one's perception of reality by visually augmenting the objects with extra information such as hints for assembly.
In autonomous driving, the poses of surrounding vehicles, pedestrians, and obstacles facilitate the planning of the next actions.

\begin{figure}[t!]
	\begin{center}
		\includegraphics[width=0.9\linewidth]{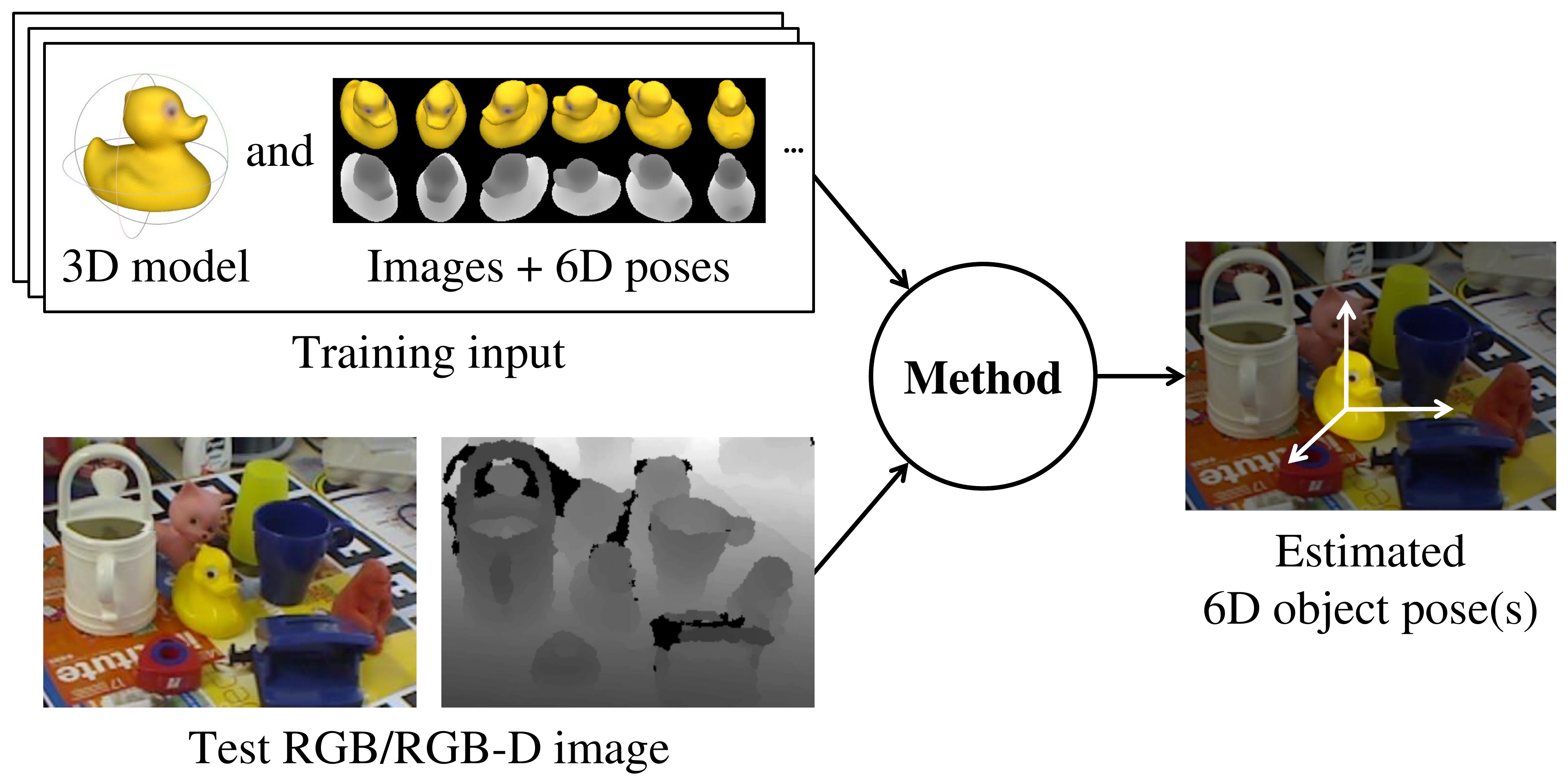} \\
		\vspace{1.0ex}
		\caption{\label{fig:object_pose_estimation}
			\textbf{Estimating 6D pose of specific rigid objects.}\
			At training time, a method is given 3D object models and images annotated with 6D object poses. At test time, the goal of the method is to estimate the 6D poses of visible object instances from a single input image. The images may have RGB or RGB-D~channels and the intrinsic camera parameters are known.}
	\end{center}
\end{figure}

\section{Problem Formulation} \label{sec:problem_formulation}

In this thesis, we consider the problem of estimating the 6D pose
of specific rigid objects with available
3D mesh models.
In particular, given the 3D models and a set of training images showing object instances in known 6D poses, the considered problem is to estimate the 6D poses of possibly multiple instances of possibly multiple object classes from a single test image.
The training and test images may have RGB or RGB-D (aligned color and depth) channels and the intrinsic camera parameters are assumed known (Figure~\ref{fig:object_pose_estimation}).

Depending on whether prior information about the object instances visible in the test image is provided,
we distinguish two variants of the 6D object pose estimation problem: (1)~6D object localization, where the identifiers of visible object instances are provided, and (2)~6D object detection, where no prior information is provided. The first problem is relevant for robotic assembly, where the robot needs to find a specific set of objects which are guaranteed to be present in the workspace,
while the second is relevant for augmented reality and autonomous driving, where the surrounding scene is typically unknown.

We primarily focus on the 6D object localization problem under a practical scenario where no real images of the objects are available at training time, only the 3D object models and images synthesized using the models. While capturing real images of objects under various conditions and annotating the images with 6D object poses requires a~sig\-ni\-ficant human effort~\cite{hodan2017tless}, the 3D models are either available before the physical objects, which is often the case for manufactured objects, or can be reconstructed at an admissible cost.\footnote{3D reconstruction approaches for opaque, matte, and mildly specular objects are established~\cite{newcombe2011kinectfusion} and promising approaches for transparent and highly specular ones are emerging~\cite{qian20163d,wu2018full,godard2015multi}.}
Moreover, besides being useful for
estimating object poses, the 3D models enable further reasoning about the objects. For example, a robot can use the estimated poses together with the
model surface to compute grasps of the objects.

\section{Limitations of Existing Methods} \label{sec:limitations_of_existing_methods}

A popular approach to 6D object pose estimation is to extract regions of the input image that possibly contain an object, and for each region holistically predict the object class, the object pose, and a confidence score.
The image regions can be extracted by instance segmentation~\cite{he2017mask}, object proposal generation~\cite{hosang2015makes}, or by exhaustive enumeration of regions over the whole image.
The per-region prediction is traditionally made by matching the region against prototype image representations, called templates, which exhaustively capture different object appearances~\cite{brunelli2009template,hinterstoisser2012accv}.
The time complexity of such an approach is linear in the number of templates and the processing time quickly becomes prohibitive with an increasing number of objects and with a growing variety of object appearances.
In this thesis, we propose an extension of the template matching approach which makes the time complexity largely unaffected by the number of templates.

Recently, efficient per-region prediction has been achieved with neural networks that learn object appearances implicitly through the network parameters~\cite{kehl2017ssd,li2018unified,sundermeyer2019augmented}.
The time complexity of these methods is constant in the number of learned objects, although learning new objects requires re-training the network and the accuracy tends to degrade with the number of learned objects~\cite{kaskman2019homebreweddb}.
As shown by the failure cases presented in~\cite{kehl2017ssd,sundermeyer2019augmented}, estimating pose of partially occluded objects remains challenging for the recent holistic methods, even when occlusions are simulated at training time~\cite{sundermeyer2019augmented}.

Another common approach to 6D object pose estimation, which tends to be more robust to occlusions, is to establish 2D-3D correspondences between pixels of the input image and locations on the 3D object model, and robustly estimate the pose by the P\emph{n}P-RANSAC algorithm~\cite{fischler1981random,lepetit2009epnp}.
Early methods establish the correspondences by matching local image features, such as SIFT~\cite{lowe1999object}, while recent methods mostly rely on neural networks and predict either the corresponding 3D location at each pixel~\cite{brachmann2014learning,park2019pix2pose,zakharov2019dpod} or the 2D projections of pre-selected 3D keypoints~\cite{rad2017bb8,oberweger2018making,peng2019pvnet}.

Establishing 2D-3D correspondences is challenging for objects with global or partial symmetries~\cite{mitra2006partial} in shape and texture.
The visible part of such objects, which is determined by self-occlusions and occlusions by other objects, may have multiple fits to the object model. 
Consequently, the potentially corresponding 2D and 3D locations
form a many-to-many relationship, \ie, a 2D image location may correspond to multiple 3D locations on the object model surface and vice versa.
The existing correspondence-based methods assume a one-to-one relationship and therefore cannot reliably handle objects with symmetries.
Additionally,
methods relying on local image features have a poor performance on texture-less objects as the feature detectors often fail to provide a sufficient number of reliable locations and the descriptors are no longer discriminative enough~\cite{tombari2013bold,hodan2015detection}.
In this thesis, we propose a method that can handle the many-to-many relationship of potentially corresponding locations and is applicable to a broad range of objects, including objects with symmetries and objects without any texture.

One of the critical factors to the success of neural networks is the availability of a large number of training images~\cite{goodfellow2016deep}. Researchers often resort to training on synthetic images as obtaining real training images annotated with 6D object poses is expensive.
The images are typically synthesized by rendering the 3D object models and pasting the renderings on top of random photographs~\cite{su2015render,rad2017bb8,kehl2017ssd,hinterstoisser2017pre}.
However, the evident domain gap between these ``render\;\&\;paste'' training images and real test images limits the potential of neural networks. In this thesis, we propose an approach to generate highly photorealistic training images and show significant improvements over the ``render\;\&\;paste'' images.

Although many methods for 6D object pose estimation have been published recently, it has been largely unclear which methods perform well and in which scenarios.
This has been the case because new methods have usually been compared with only a few~competitors on a small subset of publicly available datasets. Moreover, evaluating the 6D object pose estimates is not straightforward. Due to object symmetries and occlusions, there may be multiple 6D poses consistent with the image and no standard evaluation methodology that can reliably handle such ambiguities has emerged. In this thesis, we define a benchmark that includes a new evaluation methodology and multiple datasets covering different practical scenarios. We also present a new industry-relevant dataset. %

\section{Contributions} \label{sec:contributions}

In this thesis, we make the following contributions:

\customparagraph{EPOS.} We present a novel method for 6D object pose estimation from an RGB image
applicable to a broad range of objects, including objects with global or partial symmetries and objects without any texture.
The key idea is to represent an object by a set of compact surface fragments, predict the probability distribution of corresponding fragments at each pixel of the input image by a neural network, and link each pixel with possibly multiple surface fragments.
The 6D poses are estimated from the
predicted many-to-many 2D-3D correspondences
by a RANSAC-based robust fitting procedure.
In the BOP Challenge 2019~\cite{bop19challenge,hodan2018bop}, the method outperformed all RGB and most RGB-D and D methods on the T-LESS~\cite{hodan2017tless}, YCB-V~\cite{xiang2017posecnn}, and LM-O~\cite{brachmann2014learning} datasets.

\customparagraph{HashMatch.}
Another proposed method extends the traditional template matching pa\-ra\-digm with
a cascade of evaluation stages that is applied to each location on an RGB-D input image.
The cascade avoids exhaustive matching of every location against all templates and makes the time complexity of the method
sub-linear in the number of templates.
The key stage of the cascade is a voting procedure based on hashing that generates a small set of candidate templates, which are then processed by more expensive verification and refinement stages.
The method placed fourth out of the 15 participants in the BOP Challenge 2017~\cite{hodan2018bop} and was successfully deployed in a robotic assembly project.

\customparagraph{ObjectSynth.}
We propose a procedural approach to synthesize
photorealistic images of 3D object models, which can be used to effectively train neural networks for object pose estimation from real images.
The key ingredients are:
(1)~3D models of objects are arranged in 3D models of indoor scenes with realistic materials and lighting, (2)~plausible geometric configurations of objects and cameras are generated by physics simulation, and (3)~a high level of photorealism is achieved by physically-based rendering.
The effectiveness of the synthesized training images is demonstrated on object detection experiments and confirmed in the BOP Challenge 2020~\cite{hodan2020bop}, where strong object pose estimation results are achieved with images synthesized by a refined version of the proposed approach.

\customparagraph{T-LESS.} By creating the T-LESS dataset, we introduce industry-relevant objects to the academic community.
The included objects have no significant texture or discriminative color, exhibit symmetries and similarities in shape and size, and some objects are a composition of others.
T-LESS provides two types of 3D models for each object (a manually created CAD model and a semi-automatically reconstructed one), and training and test images captured with three synchronized sensors (a~structured-light and a time-of-flight RGB-D sensor and a high-resolution RGB camera) and annotated with 6D object poses.

\customparagraph{BOP.} To capture the status quo in 6D object pose estimation, we created the BOP benchmark that currently comprises of (1)~eleven datasets in a unified format covering different practical scenarios, (2)~an evaluation methodology with three new pose-error functions, which address limitations of the previously used functions, (3)~an online evaluation system at \texttt{\href{http://bop.felk.cvut.cz}{bop.felk.cvut.cz}}, which is open for continuous submission of new results and reports the up-to-date state of the art, and (4)~public challenges held at the
Workshops on Recovering 6D Object Pose~\cite{hodan2020r6d}, which we organize at the ICCV and ECCV conferences.

\section{Structure of the Thesis}

The rest of the thesis is organized as follows.
Chapter~\ref{ch:related_work} reviews existing methods,~da\-ta\-sets, and evaluation methodologies for 6D object pose estimation.\
Chapter~\ref{ch:method_template} proposes Hash\-Match, a method for efficient template matching.\
Chapter~\ref{ch:method_epos} presents EPOS, a~method that can systematically handle objects with symmetries.\
Chapter~\ref{ch:synthesis} presents Object\-Synth, an approach for the synthesis of photo\-realistic training images.\
Chapter~\ref{ch:tless} describes T-LESS, the first dataset with industry-relevant objects.\
Chapter~\ref{ch:bop} introduces BOP, a widely-accepted benchmark for 6D object pose estimation.\
Finally, Chapter~\ref{ch:conclusion} concludes the thesis and suggests topics for future work.

\section{Publications} \label{sec:research_papers}

The thesis builds on the following publications (chronologically ordered):

\begin{itemize}
\item[\cite{hodan2015detection}] \textbf{Tom{\'a}{\v{s}} Hoda{\v{n}}}, Xenophon Zabulis, Manolis Lourakis, {\v{S}}t{\v{e}}p{\'a}n Obdr{\v{z}}{\'a}lek, Ji{\v{r}}{\'\i} Matas.
\emph{Detection and Fine 3D Pose Estimation of Texture-less Objects in RGB-D Images}.
International Conference on Intelligent Robots and Systems (IROS), 2015.

\item[\cite{hodan2016evaluation}] \textbf{Tom{\'a}{\v{s}} Hoda{\v{n}}}, Ji{\v{r}}{\'\i} Matas, {\v{S}}t{\v{e}}p{\'a}n Obdr{\v{z}}{\'a}lek.
\emph{On Evaluation of {6D} Object Pose Estimation}.
European Conference on Computer Vision Workshops (ECCVW), 2016.

\item[\cite{hodan2017tless}] \textbf{Tom{\'a}{\v{s}} Hoda{\v{n}}}, Pavel Haluza, {\v{S}}t{\v{e}}p{\'a}n Obdr{\v{z}}{\'a}lek, Ji{\v{r}}{\'\i} Matas, Manolis Lourakis, Xenophon Zabulis.
\emph{{T-LESS}: An {RGB-D} Dataset for {6D} Pose Estimation of Texture-less Objects}.
Winter Conference on Applications of Computer Vision (WACV), 2017.

\item[\cite{hodan2018bop}] \textbf{Tom{\'a}{\v{s}} Hoda{\v{n}}}, Frank Michel, Eric Brachmann, Wadim Kehl, Anders Glent Buch, Dirk Kraft, Bertram Drost, Joel Vidal, Stephan Ihrke, Xenophon Zabulis, Caner Sahin, Fabian Manhardt, Federico Tombari, Tae-Kyun Kim, Ji{\v{r}}{\'i} Matas, Carsten Rother.
\emph{{BOP}: Benchmark for {6D} Object Pose Estimation}.
European Conference on Computer Vision (ECCV), 2018.

\item[\cite{hodan2018workshop}] \textbf{Tom{\'a}{\v{s}} Hoda{\v{n}}}, Rigas Kouskouridas, Tae-Kyun Kim, Federico Tombari, Kostas Bekris, Bertram Drost, Thibault Groueix, Krzysztof Walas, Vincent Lepetit, Ales Leonardis, Carsten Steger, Frank Michel, Caner Sahin, Carsten Rother, Ji{\v{r}}{\'i} Matas. \emph{A Summary of the 4th International Workshop on Recovering {6D} Object Pose}. European Conference on Computer Vision Workshops (ECCVW), 2018.

\item[\cite{hodan2019photorealistic}] \textbf{Tom{\'a}{\v{s}} Hoda{\v{n}}}, Vibhav Vineet, Ran Gal, Emanuel Shalev, Jon Hanzelka, Treb Connell, Pedro Urbina, Sudipta N. Sinha, Brian Guenter.
\emph{Photorealistic Image Synthesis for Object Instance Detection}.
International Conference on Image Processing (ICIP), 2019.

\item[\cite{hodan2020epos}] \textbf{Tom{\'a}{\v{s}} Hoda{\v{n}}}, D{\'a}niel Bar{\'a}th, Ji{\v{r}}{\'i} Matas.
\emph{{EPOS}: Estimating {6D} Pose of Objects with Symmetries}.
Conference on Computer Vision and Pattern Recognition (CVPR), 2020.

\item[\cite{denninger2020blenderproc}] Maximilian Denninger, Martin Sundermeyer, Dominik Winkelbauer, Dmitry Olefir, \textbf{Tom{\'a}{\v{s}} Hoda{\v{n}}}, Youssef Zidan, Mohamad Elbadrawy, Markus Knauer, Harinandan Katam, Ahsan Lodhi. \emph{BlenderProc: Reducing the Reality Gap with Photorealistic Rendering}.	Robotics: Science and Systems (RSS) Workshops, 2020.

\item[\cite{hodan2020bop}] \textbf{Tom{\'a}{\v{s}} Hoda{\v{n}}}, Martin Sundermeyer, Bertram Drost, Yann Labb{\'e}, Eric Brachmann, Frank Michel, Carsten Rother, Ji{\v{r}}{\'i} Matas.
\emph{BOP Challenge 2020 on 6D Object Localization}.
European Conference on Computer Vision Workshops (ECCVW), 2020.

\end{itemize}

\noindent Other publications not explicitly discussed in the thesis:

\begin{itemize}
	
\item[\cite{hodan2015efficient}] \textbf{Tom{\'a}{\v{s}} Hoda{\v{n}}}, Dima Damen, Walterio Mayol-Cuevas, Ji{\v{r}}{\'\i} Matas.
\emph{Efficient Texture-less Object Detection for Augmented Reality Guidance}.
International Symposium on Mixed and Augmented Reality Workshops (ISMARW), 2015.
	
\item[\cite{patel2020learning}] Yash Patel, \textbf{Tom{\'a}{\v{s}} Hoda{\v{n}}}, Ji{\v{r}}{\'\i} Matas. \emph{Learning Surrogates via Deep Embedding}. European Conference on Computer Vision (ECCV), 2020.
\end{itemize}

	\chapter{Related Work} \label{ch:related_work}

Research in object pose estimation has closely followed trends common in the whole field of computer vision.
Early methods, dating back to the work of Roberts from 1963~\cite{roberts1963machine}, had to deal with a limited computational power and typically aim to fit 3D object models to the intensity edges extracted from the input image, which provide an efficient and relatively stable representation (Section~\ref{sec:rel_methods_edge}).
Following the success of SIFT-like local features from the turn of the century~\cite{lowe1999object}, later methods estimate the object pose from correspondences that are established between the input image and the object model by matching the local features (Section~\ref{sec:rel_methods_2d_features}).
With the introduction of Microsoft Kinect in 2010, the attention of the research field was steered towards estimating the object pose from RGB-D or D-only images, yielding methods based on 3D local features (Section~\ref{sec:rel_methods_3d_features}), particularly successful point pair features (Section~\ref{sec:rel_methods_ppf}), and methods based on RGB-D template matching (Section~\ref{sec:rel_methods_template}).

Significant improvements in object pose estimation have been recently achieved by machine learning techniques.
The contributions of most of the ``classical'' methods mentioned above are related to extracting and matching representations suitable for the problem at hand.
In contrast, recent methods typically count on representations learned by general function approximators, such as random forests or deep neural networks, and focus on more abstract, conceptual aspects of the problem such as formulating the training loss function, handling object symmetries, %
or bridging the domain gap between synthetic training and real test images (Section~\ref{sec:rel_methods_learning}).

\section{Classical Methods} \label{sec:rel_methods_classical}

The classical, non-learning-based methods for object pose estimation are split in this~sec\-tion according to the image representations they rely on, as the chosen image representation often determines the type of objects and scenes for which the method is suitable. Sections~\ref{sec:rel_methods_edge}--\ref{sec:rel_methods_ppf} describe methods based on matching local or semi-local features extracted from an image or a point cloud,
and Section~\ref{sec:rel_methods_template} describes methods based on matching holistic templates of the objects against regions of the input image.

\subsection{Edge-Based Features} \label{sec:rel_methods_edge}

\begin{figure}[b!]
	\begin{center}
		\begingroup
		\small
		\begin{tabular}{ @{}c@{ } @{}c@{ } @{}c@{ } @{}c@{ } }
			Input image & Detected edges & \multicolumn{2}{c}{Two views at the fitted 3D models} \vspace{0.5ex} \\
			\includegraphics[width=0.243\linewidth]{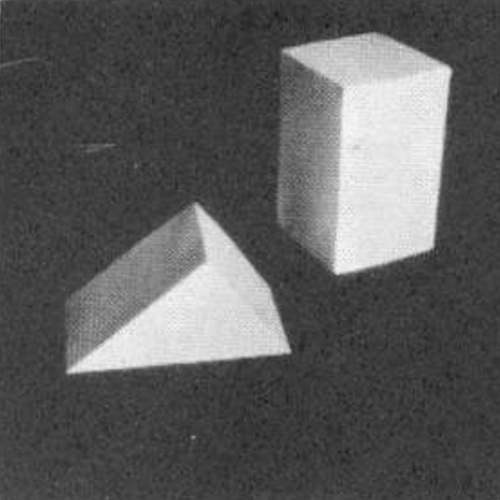} &
			\includegraphics[width=0.243\linewidth]{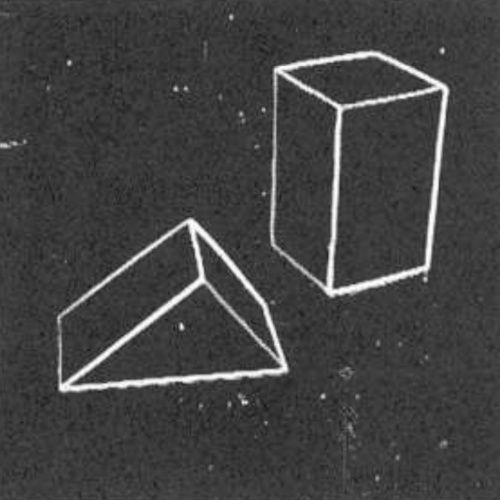} &
			\includegraphics[width=0.243\linewidth]{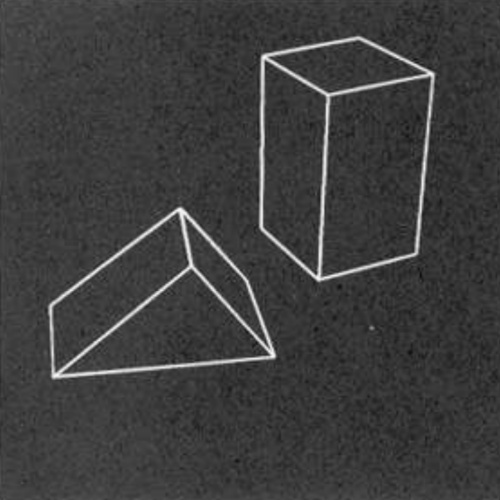} &
			\includegraphics[width=0.243\linewidth]{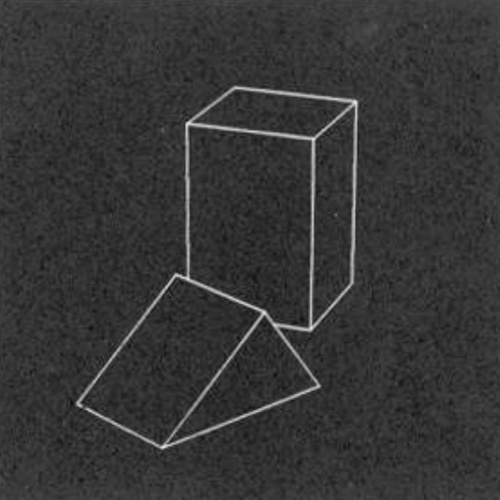} \\
		\end{tabular}
		\endgroup
		\caption{\label{fig:rel_roberts} \textbf{Roberts~\cite{roberts1963machine} fits simple 3D models to intensity edges.} The models are represented by a set of line segments in 3D, junctions of the segments are matched with junctions of intensity edges detected in the input image, and the object pose is estimated from the established correspondences.
		}
	\end{center}
\end{figure}

In his Ph.D. thesis, Roberts~\cite{roberts1963machine} assumes that objects can be constructed from transformations of known simple 3D wire-frame models. His approach is to detect intensity edges in the input image, match junctions of the intensity edges with junctions of the model edges, and solve for the pose
by minimizing the re-projection error (Figure~\ref{fig:rel_roberts}).
A~similar approach was proposed by Lowe~\cite{lowe1987three}, who assumes the 3D object model is known and establishes correspondences between groups of edges that are likely to be invariant over a wide range of viewpoints (examples include instances of collinearity, proximity, or parallelism).
The scalability of the matching stage to multiple objects is addressed by Lamdan and Wolfson~\cite{lamdan1988geometric} with geometric hashing, and by Beis and Lowe~\cite{beis1999indexing} with an approximate nearest-neighbour search.
Damen \etal~\cite{damen2012real} propose a scalable approach which traces constellations of short edge segments using pre-defined paths, matches the constellations against training images of the objects using a pre-calculated hash table,
and verifies the hypotheses by the distance transform.
A similar approach based on grouping of neighboring edge segments is proposed by Tombari \etal~\cite{tombari2013bold}, who also demonstrates the superiority of edge-based features for detection of texture-less objects over the
texture-based local features described in Section~\ref{sec:rel_methods_2d_features}.

Edge-based methods have also been proposed for the related task of object category recognition. For example, Carmichael and Hebert~\cite{carmichael2002object} apply weak classifiers to assign individual edge pixels to the target object or to clutter, based on the configuration of edges in the vicinity.
Chia \etal~\cite{chia2010object} extract shape primitives composed of edge segments and ellipses from the input image, match the primitives against a pre-calculated codebook, and cast votes for the object center.
Similarly, Opelt \etal~\cite{opelt2006boundary} apply boosting to classify contour fragments which then vote for the object center.
Danielsson \etal~\cite{danielsson2009automatic} learn consistent constellations of edge segments over object categories from training images. The most consistent pair of edge segments
is selected as the aligning pair and exhaustively matched against all pairs in the input image. An extension to multiple edge segments forming a fully connected clique was proposed by Leordeanu \etal~\cite{leordeanu2007beyond}.

Many of the edge-based methods above rely on edge maps computed with traditional edge detectors such as Canny~\cite{canny1986computational}. In~\cite{hodan2015efficient}, we show that the performance of the method by Damen \etal~\cite{damen2012real} can be improved if the the Canny detector is replaced with the detector by Doll{\'a}r and Zitnick~\cite{dollar2013structured}, which produces cleaner edge maps and favor the object outline while eschewing noise and higher-frequency patterns.
The improved method works well on texture-less objects in clean scenes but struggles from spurious edges of shadows and background clutter -- this limitation is typical for edge-based methods.

\subsection{2D Local Features} \label{sec:rel_methods_2d_features}

Methods from this category represent an object by a set of discriminative 2D local features, which are extracted from training (color or grayscale) images of the object and need to be registered in a common 3D coordinate system to enable estimating the 6D object pose.
The local features are typically invariant to changes in illumination and to similarity or affine image transformations. Examples of such features include SIFT~\cite{lowe1999object}, MSER~\cite{matas2004robust}, and SURF~\cite{bay2006surf}. At test time, the training features are matched against features extracted from the test image, and the 6D object poses are estimated from the~established 2D-3D correspondences typically by the P\emph{n}P-RANSAC algorithm~\cite{fischler1981random,lepetit2009epnp}.

\begin{figure}[t!]
	\begin{center}
		\begingroup
		\setlength{\tabcolsep}{2pt} %
		\renewcommand{\arraystretch}{1} %
		\small
		\begin{tabular}{c c c c c c}
			Object 1 & Model 1 & Object 2 & Model 2 & Input image & Estimated poses \vspace{0.5ex} \\
			\includegraphics[height=0.143\linewidth]{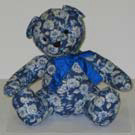} &
			\includegraphics[height=0.143\linewidth]{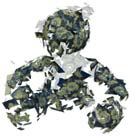} &
			\includegraphics[height=0.143\linewidth]{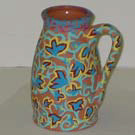} &
			\includegraphics[height=0.143\linewidth]{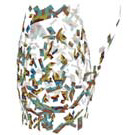} &
			\includegraphics[height=0.143\linewidth]{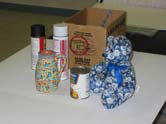} &
			\includegraphics[height=0.143\linewidth]{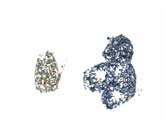} \\
		\end{tabular}
		\endgroup
		\caption{\label{fig:rel_ponce} \textbf{Ponce \etal~\cite{ponce2004toward} represent an object by small patches registered in~3D.}
		This representation is created automatically from a small set of unordered training images. Registration in 3D enables estimating the 6D object pose from correspondences between the model patches and patches detected in the input image.
		}
	\end{center}
\end{figure}

Lowe~\cite{lowe1999object}
matches the SIFT features between a set of training images and the test image without recovering any 3D information but demonstrates
the robustness of the approach against occlusion and clutter in the case of objects with a distinct and non-repeatable shape or texture.
A survey of subsequent approaches that aim to recognize the objects but not to estimate their poses in the 3D space is provided by Matas and Obdr{\v{z}}{\'a}lek~\cite{matas2004object}.
Ponce \etal~\cite{ponce2004toward} represent an object by a set of small affine-covariant image patches and a description of their 3D spatial relationship. This representation is created automatically from a small set of unordered training images and enables estimating the 6D object pose (Figure~\ref{fig:rel_ponce}).
Muja and Lowe~\cite{muja2009fast} created the widely used FLANN library for an approximate nearest neighbor search with automatic algorithm configuration, which provides a significant improvement in speed of the feature matching stage.
Collet \etal~\cite{collet2011moped} achieve robust performance with a novel algorithm that iteratively combines feature clustering with robust pose estimation -- the feature clustering quickly partitions the scene and produces object hypotheses that are used to refine the feature clusters, and the two steps iterate until convergence.

\begin{figure}[t!]
	\begin{center}
		
		\begingroup
		\small
		\renewcommand{\arraystretch}{0.9} %
		
		\begin{tabular}{ @{}c@{ } @{}c@{ } @{}c@{ } @{}c@{ } @{}c@{ } }
			\multicolumn{2}{c}{Textured objects} & & \multicolumn{2}{c}{Texture-less objects} \vspace{0.5ex} \\
			
			\includegraphics[width=0.239\columnwidth]{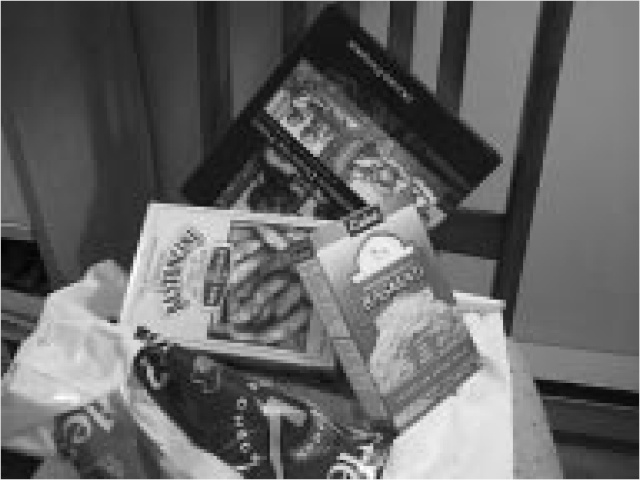} &
			\includegraphics[width=0.239\columnwidth]{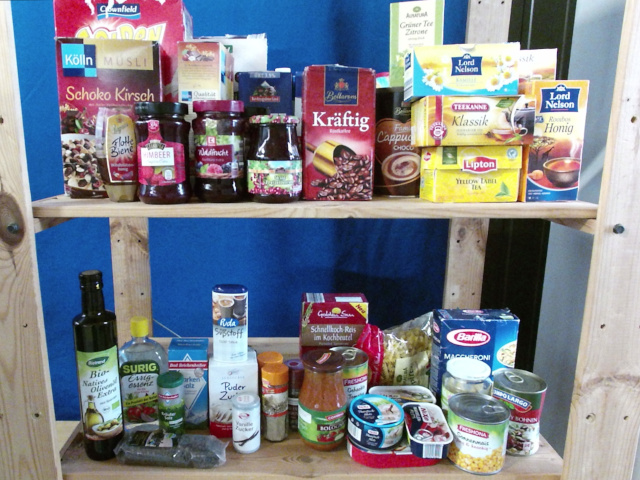} &
			\hspace{1.0ex} &
			\includegraphics[width=0.239\columnwidth]{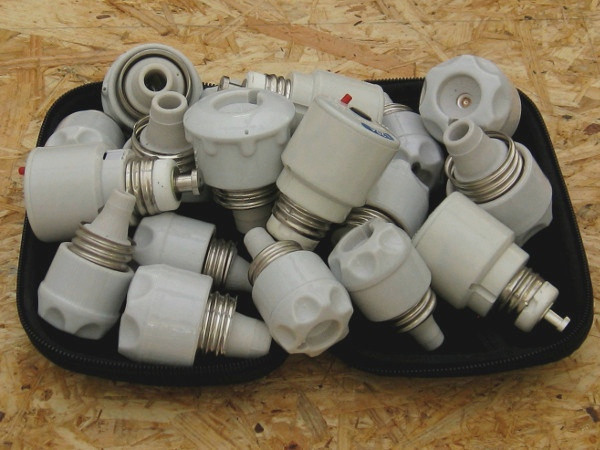} &
			\includegraphics[width=0.239\columnwidth]{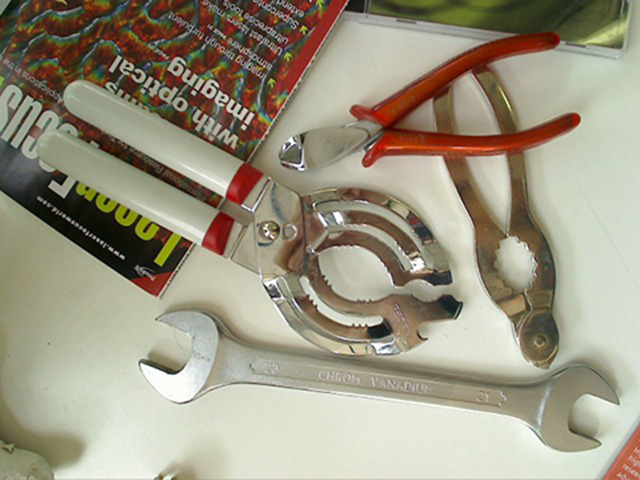} \\
			
			\includegraphics[width=0.239\columnwidth]{images/thesis_proposal/textured/ycbv_59_001093.jpg} &
			\includegraphics[width=0.239\columnwidth]{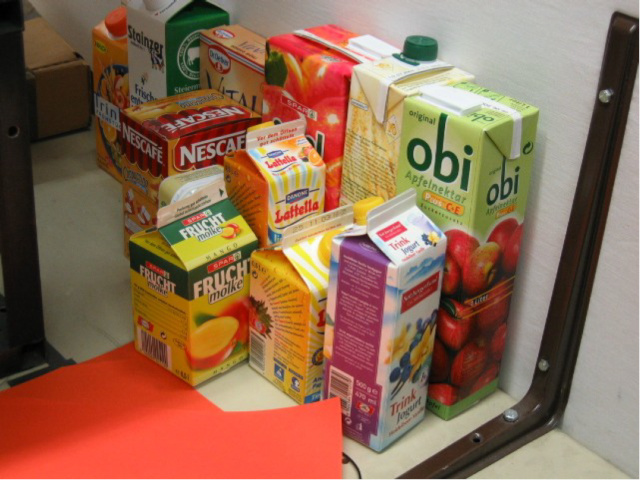} &
			&
			\includegraphics[width=0.239\columnwidth]{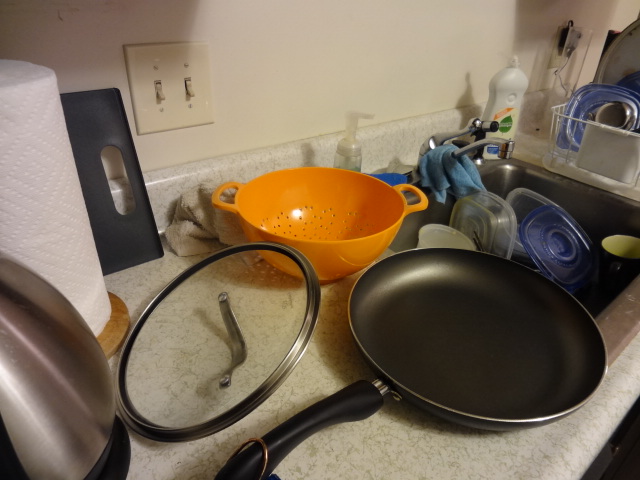} &
			\includegraphics[width=0.239\columnwidth]{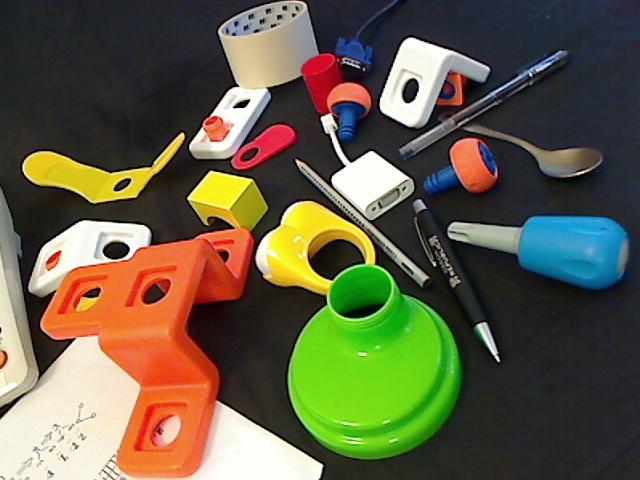} \\
		\end{tabular}
		\endgroup
		
		\caption{\label{fig:intro_textureless_objects} \textbf{Textured \vs texture-less objects.} Objects without a significant texture (right) are common in household, office, and industrial environments and are challenging for methods relying on 2D local features. This is because the feature detectors often fail to provide a sufficient number of reliable locations and the descriptors are not discriminative enough~\cite{tombari2013bold}.
			The shown images originate from~\cite{lowe1999object,jund2016freiburg,xiang2017posecnn,obdrzalek2006object,hodan2017tless,tombari2013bold,hsiao2014occlusion,cai2013fast}.
		}
	\end{center}
\end{figure}

Methods based on 2D local features excel on objects with a distinct and non-repeatable
texture
but have difficulties with texture-less objects (Figure~\ref{fig:intro_textureless_objects}),
where the feature detectors often fail to provide a sufficient number of reliable locations and the descriptors are not discriminative enough to provide reliable correspondences~\cite{tombari2013bold}.
Moreover, methods based on 2D local features tend to have a poor performance on \mbox{objects} with symmetries,
where the features are in a many-to-many relationship, \ie, a feature in the input~\mbox{image} potentially corresponds to multiple features on the object model, and vice versa (Section \ref{sec:rel_handling_ambiguity}). This degrades the performance of
the methods which usually assume a one-to-one relationship.
In Chapter~\ref{ch:method_epos}, we present a correspondence-based method that learns the many-to-many relationship and can handle both symmetric and texture-less objects.

\subsection{3D Local Features} \label{sec:rel_methods_3d_features}

A number of techniques to detect and describe repeatable and distinctive local features on 3D surfaces have been proposed after the introduction of consumer-grade depth sensors such as Microsoft Kinect, Primesense Carmine, or Intel RealSense. Examples of the 3D local features include SHOT~\cite{salti2014shot}, KPQ~\cite{mian2010repeatability}, ISS~\cite{zhong2009intrinsic}, and MeshDoG~\cite{unnikrishnan2008multi}, and their
comprehensive evaluation is provided in~\cite{tombari2013performance,guo2016comprehensive}.
Methods based on these features often assume that the 3D object models are available, typically in the form of mesh models. Features extracted from the 3D models are organized into a database, matched against features extracted from the input point cloud (calculated from the depth image channel and known camera parameters), and the poses are estimated from the established correspondences, \eg, by a RANSAC-based estimation technique~\cite{papazov2010efficient}. The pose hypotheses can be further refined individually by a surface registration method, such as Iterative Closest Point (ICP)~\cite{besl1992method,rusinkiewicz2001efficient}, or globally by optimizing all hypotheses together~\cite{papazov2010efficient,aldoma2012global}.
Guo \etal~\cite{guo20143d} provide a detailed survey of methods based on 3D local features.

Similarly to methods based on 2D local features (Section~\ref{sec:rel_methods_2d_features}), methods based on 3D local features tend to have a poor performance on objects with symmetries. This was also shown by the relatively low accuracy scores of the methods by Buch \etal~\cite{buch2016local,buch2017rotational} on the symmetric objects from the T-LESS dataset (Section~\ref{sec:bop_challenge_2017}).

\subsection{Point Pair Features} \label{sec:rel_methods_ppf}

\begin{figure}[b!]
	\begin{center}
		\begingroup
		\setlength{\tabcolsep}{2pt} %
		\renewcommand{\arraystretch}{0.9} %
		\small
		\begin{tabular}{ c c c }
			\includegraphics[height=0.2285\linewidth]{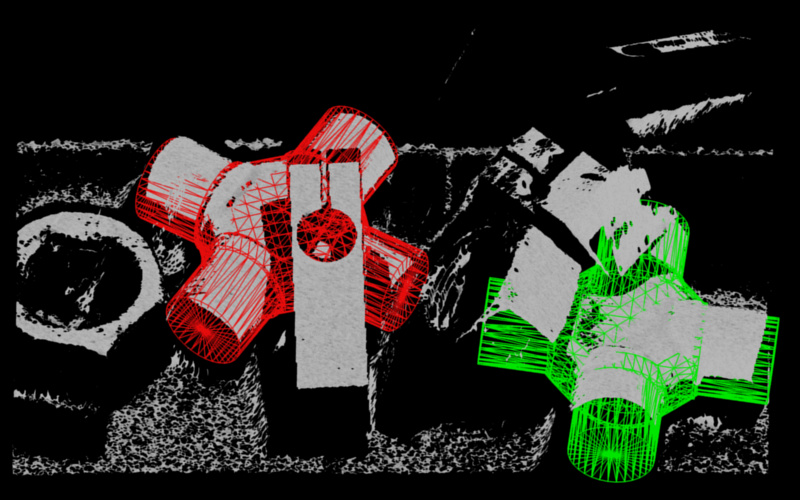} &
			\includegraphics[height=0.2285\linewidth]{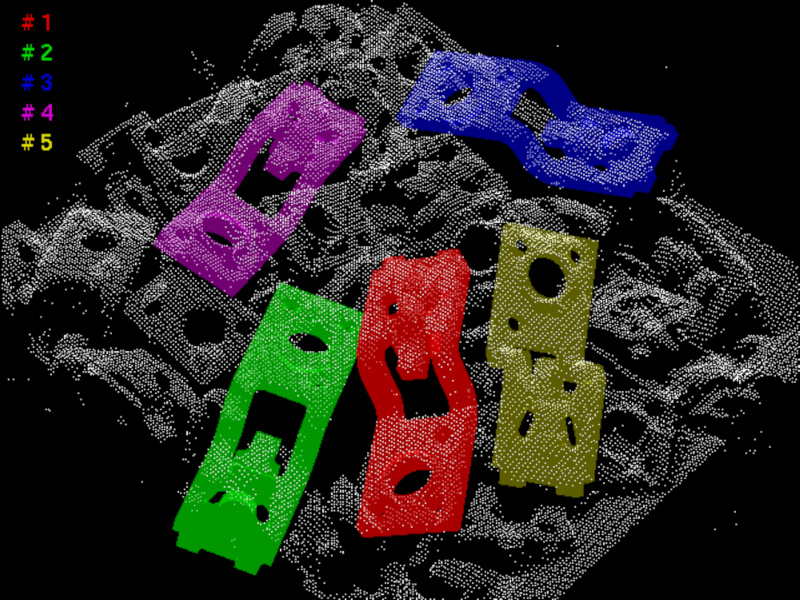} &
			\includegraphics[height=0.2285\linewidth]{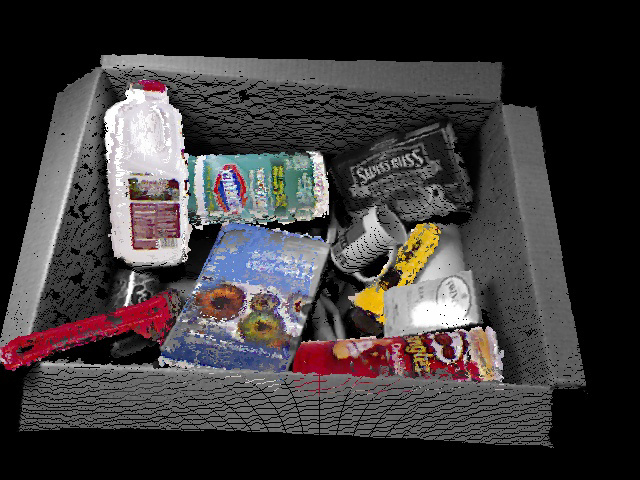} \\
		\end{tabular}
		\endgroup
		\caption{\label{fig:rel_ppf} \textbf{3D models aligned to point clouds using point pair features~\cite{drost2010model,choi2012voting,choi2012pose}.}
			The point clouds are calculated from the depth image channel and known camera parameters.
		}
	\end{center}
\end{figure}

In 2010, Drost \etal~\cite{drost2010model} introduced a depth-based method that even in 2020 still belongs to the top-performing methods. Based on the idea of surflet pairs~\cite{wahl2003surflet}, the method matches pairs of oriented points (given by the 3D coordinates and the surface normals) between the point cloud of the test scene and the object model, and groups the matches via a voting scheme.
At training time, point pairs from the model are exhaustively sampled and stored in a hash table.
At test time, reference points are sampled in the scene and a low-dimensional voting space is defined for each reference point by restricting to those poses that align the reference point with a model point (in both the location and the orientation).
Point pairs are then created from the reference point and other points sampled in the scene, similar point pairs are retrieved from the hash table, and each retrieved point pair casts a vote for an object pose.
The votes are accumulated to produce a set of pose hypotheses, which are finally refined by a coarse-to-fine ICP algorithm and re-scored by the relative amount of the visible model surface (Figure~\ref{fig:rel_ppf}).

Multiple improvements of the method by Drost \etal~\cite{drost2010model} have been proposed. Kim and Medioni~\cite{kim20113d} include visible context information, differentiating visible points, points on the surface, and invisible points. Choi and Christensen~\cite{choi2012pose} augment the point pair feature with color information. Drost and Ilic~\cite{drost20123d} propose a multimodal extension including edge information extracted from RGB image channels. Birdal and Ilic~\cite{birdal2015point} reduce the search space by a coarse-to-fine segmentation, favor points with higher expected visibility, and filter pose hypotheses by an occlusion-aware ranking. Hinterstoisser \etal~\cite{hinterstoisser2016going} introduce novel sampling and voting schemes that significantly reduce the influence of clutter and sensor noise.
Inspired by~\cite{hinterstoisser2016going}, Vidal \etal~\cite{vidal2018method} propose further improvements, including a novel view-dependent verification process, and achieved the top scores in the first two editions of the BOP Challenge in 2017 and 2019 (Sections~\ref{sec:bop_challenge_2017} and \ref{sec:bop_challenge_2020}). In the 2020 edition of the challenge, the point pair features were represented by a hybrid method by K\"onig and Drost~\cite{koenig2020hybrid}, which placed second overall and first among fast methods with the average processing time below $1\,$s per image. This method starts by segmenting object instances by a deep neural network and then uses the point pair features to estimate the object poses from subsets of the point cloud defined by the instance masks.

Kiforenko \etal~\cite{kiforenko2018performance} present a comprehensive evaluation of the point pair features, including a comparison with 3D local features such as SHOT~\cite{salti2014shot}. The comparison shows that the point pair features perform better in general but are inferior on point clouds with a higher degree of noise and with a lower resolution.

\subsection{Template Matching} \label{sec:rel_methods_template}

\begin{figure}[t!]
	\begin{center}
		\begingroup
		\setlength{\tabcolsep}{2pt} %
		\renewcommand{\arraystretch}{0.9} %
		\small
		\begin{tabular}{c c}
			\includegraphics[height=0.295\linewidth]{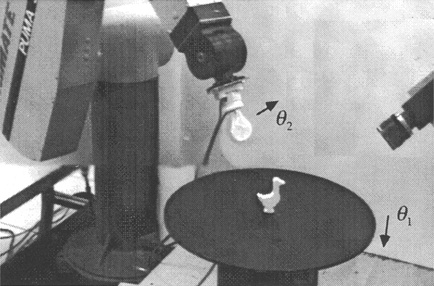} &
			\includegraphics[height=0.295\linewidth]{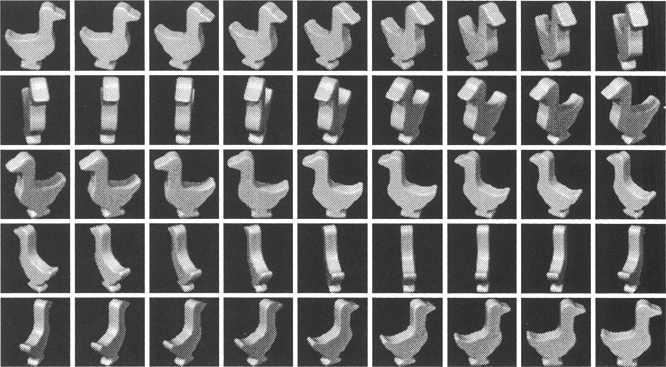} \\
		\end{tabular}
		\endgroup
		\caption{\label{fig:rel_templates} \textbf{An acquisition setup (left) for capturing template images (right)~\cite{murase1995visual}.} Capturing the templates in such a controlled environment allows to automatically annotate the templates with 6D object poses. Alternatively, if 3D models of the objects are available, the templates can be obtained by rendering the models, as in~\cite{hinterstoisser2012accv}.
		}
	\end{center}
\end{figure}

These methods compare prototype image representations, called templates, against regions of the input image that are of the same size as the templates and are usually extracted by an exhaustive sliding window technique.
The templates show an object under different conditions, such as different viewpoints and illumination, and are obtained by capturing the object in a controlled environment (Figures~\ref{fig:rel_templates} and~\ref{fig:tless_acquisition_setup})
or by rendering the 3D model of the object~\cite{hinterstoisser2012accv}.
In both cases, the templates are automatically annotated with 6D object poses. When a template is detected, the associated 6D pose can be used as a starting point for an edge-based or a depth-based refinement procedure~\cite{cai2013fast,hinterstoisser2012accv,hodan2015detection}.

Murase and Nayar~\cite{murase1995visual} avoid exhaustive matching of an image region against all templates by compressing the templates into a low-dimensional eigenspace
where each object is represented by a manifold.
The identity of the object in an image region is determined by projecting the region to the eigenspace and finding the corresponding manifold,
and the object pose is determined by the position on the manifold.
Cai \etal~\cite{cai2013fast} avoid exhaustive template matching by an efficient edge-based voting procedure that for each image region rapidly retrieves a small set of candidate templates, which are then verified by the oriented chamfer matching~\cite{shotton2005contour}. The voting procedure is based on hashing of distances and orientations of intensity edges that are the closest to fixed reference points, which are located on a regular grid attached to the sliding window. Multiple hash tables are used to increase the robustness against occlusion, with the hash key for each table calculated by discretizing measurements from a different subset of reference points. %
In Chapter~\ref{ch:method_template}, we present an RGB-D method that avoids exhaustive template matching by applying a cascade of evaluation stages to each image region. The key stage is a voting procedure inspired by~\cite{cai2013fast} and based on hashing of depth measurements and orientations of surface normals. %
A similar RGB-D method was concurrently proposed by Kehl \etal~\cite{kehl2015hashmod} who hash orientations of image gradients and surface normals. %
Hinterstoisser \etal~\cite{hinterstoisser2011linemod,hinterstoisser2012pami} represent each template by a map of color gradient orientations and, if the depth image channel is available, by a map of surface normal orientations. The orientation angles are quantized and represented as bit vectors, allowing for fast matching via binary operations. Robustness to misalignments is achieved by comparing the binarized representation with pixels in a local neighborhood.
Hinterstoisser \etal~\cite{hinterstoisser2012accv} later made the method more efficient by matching the orientation of color gradients at only a subset of locations with large gradient magnitude, and the orientation of surface normals at only a subset of locations with locally stable normals.
Rios-Cabrera \etal~\cite{rios2013discriminatively} extend the template representation of~\cite{hinterstoisser2012pami} by learning weights of template parts based on their discriminative power.

In general, the template matching methods tend to be sensitive to a cluttered background and partial object occlusions~\cite{murase1995visual,leonardis1996dealing}.
Moreover, since a large number of templates is required to exhaustively capture possible object appearances, the application of these methods is typically limited to industrial setups with a controlled variation of the image formation conditions~\cite{hodan2015detection}. %

\section{Learning-Based Methods} \label{sec:rel_methods_learning}

With the rise of machine learning techniques in computer vision, most 6D object pose estimation methods started to rely on image features learned by deep neural networks. Such features have been shown versatile and suitable for any type of objects, including textured and texture-less objects~\cite{hodan2020epos}.
The methods reviewed in this section are therefore split according to the conceptual approach rather than the used features.

\subsection{Learning to Vote for 6D Object Pose}

\begin{figure}[h!]
	\vspace{1.5ex}
	\begin{center}
		\includegraphics[width=1.0\linewidth]{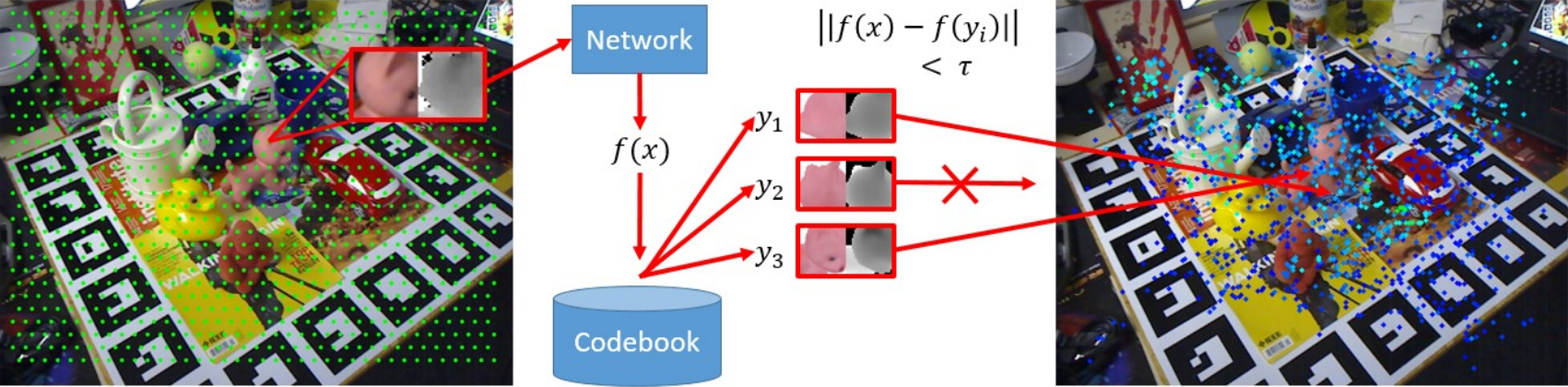} \\
		\vspace{1.5ex}
		\caption{\label{fig:rel_kehl_voting} \textbf{Patch-wise 6D voting by Kehl \etal~\cite{kehl2016deep}.}
		Auto-encoder features are calculated for scale-invariant patches extracted from the input RGB-D image, and the nearest features retrieved from a codebook cast votes for the 6D object pose.
		}
		\vspace{-2.0ex}
	\end{center}
\end{figure}

Tejani \etal~\cite{tejani2014latent} adapt the template representation of~\cite{hinterstoisser2012accv} into a scale-invariant patch representation and train a random forest with the split function at each node defined by a random patch and a random threshold on the similarity with that patch. During the inference phase, each patch of the input image is processed by the forest, going to the left subtree of a node if the similarity with the split-function patch is below the threshold, and to the right subtree otherwise. Each tree in the forest maps a patch to a leaf that stores a set of 6D pose votes, and the pose estimates are obtained by aggregating the votes from all patches.
Doumanoglou \etal~\cite{doumanoglou2016recovering} improve the method by replacing the representation of~\cite{hinterstoisser2012accv}
with features calculated with an auto-encoder neural network. Kehl \etal~\cite{kehl2016deep} follow a similar strategy but instead of using a random forest they calculate auto-encoder features for the image patches and find the nearest neighbors in a codebook, where each entry is associated with 6D pose votes (Figure~\ref{fig:rel_kehl_voting}).

\subsection{Predicting 3D Object Coordinates} \label{sec:related_object_coordinates}

\begin{figure}[t!]
	\begin{center}
		\includegraphics[width=0.8\linewidth]{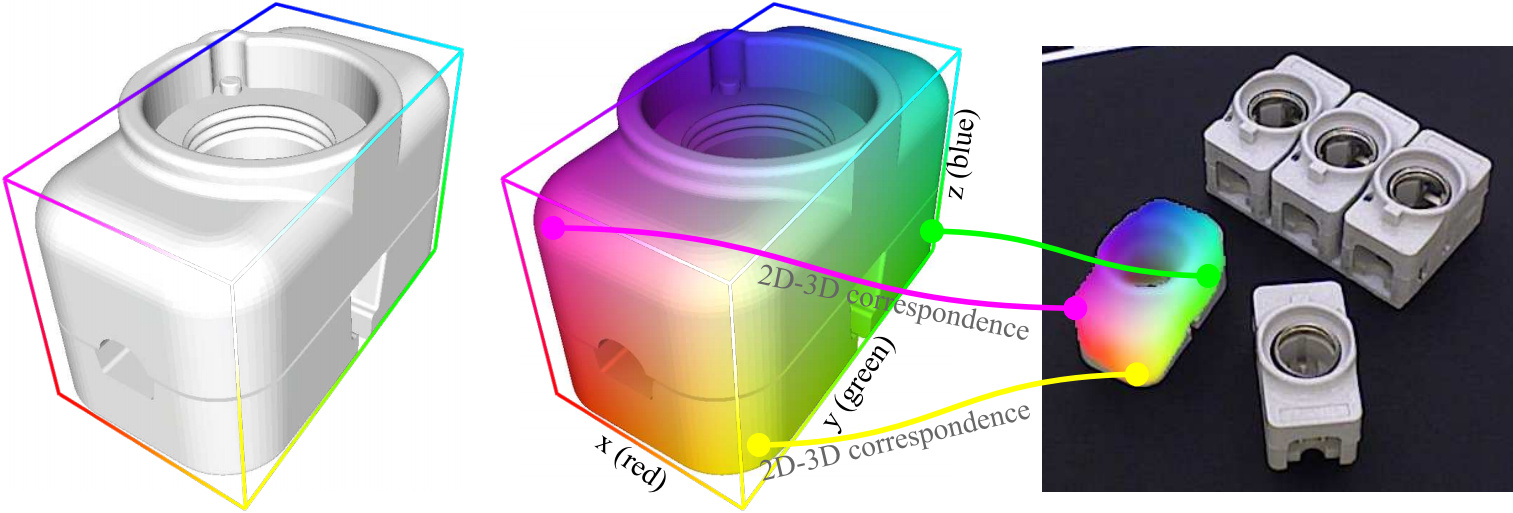} \\
		\vspace{1.0ex}
		\caption{\label{fig:rel_3d_obj_coords} \textbf{Densely predicting 3D object coordinates~\cite{brachmann2014learning,park2019pix2pose}.} The 3D points on the model surface (left) are mapped to RGB (middle) for visualization. The 3D object coordinates are predicted for densely sampled pixels of the input image (right) to establish 2D-3D correspondences (or 3D-3D if the depth image channel is available). The 6D object poses are then estimated from the correspondences by a RANSAC-based procedure.
		}
	\end{center}
\end{figure}

Inspired by the work of Taylor \etal~\cite{taylor2012vitruvian} on
human pose estimation, Brachmann \etal~\cite{brachmann2014learning} apply a random forest to the RGB-D input image to densely predict
the object identity and the 3D object coordinates, \ie the 3D location in the coordinate frame of the object model (Figure~\ref{fig:rel_3d_obj_coords}).
Split functions at the tree nodes are defined by differences in RGB and depth. A pool of pose hypotheses is generated by a RANSAC-based procedure, with each hypothesis calculated from a triplet of predicted 3D-3D correspondences by the Kabsch algorithm.
The top hypotheses
are iteratively refined to maximize the alignment of the predicted correspondences as well as the alignment of the observed depth with the object model, and the best hypothesis
is chosen as the final pose estimate.

Brachmann \etal~\cite{brachmann2016uncertainty} later extended the method in three ways.
Firstly, instead of a single random forest, a stack of random forests is trained using the auto-context framework~\cite{tu2009auto}.
Secondly, when the identities of objects visible in the input image are unknown, the RANSAC-based procedure generates not only the object pose but also the object identity. Thirdly, to represent uncertainty, the random forest predicts at each pixel a distribution over 3D object coordinates.
The distribution is defined by a Gaussian mixture model, which is suitable for objects with no or a few discrete symmetries, but not for objects with continuous symmetries. A pixel from the silhouette of objects with continuous symmetries corresponds to a circle on the 3D object model, which is problematic to represent with several Gaussian distributions.
The method is further extended by Michel \etal~\cite{michel2017global} who substitute the RANSAC-based procedure with a conditional random field to identify geometrically consistent clusters of predicted object coordinates.

Multiple subsequent methods focus primarily on RGB images, predict the 3D object coordinates with a neural network instead of the random forest, and estimate the object poses from 2D-3D correspondences by the P\emph{n}P-RANSAC algorithm~\cite{fischler1981random,lepetit2009epnp}.
Jafari \etal~\cite{jafari2018ipose} and Nigam \etal~\cite{nigam2018detect} segment object instances
to eliminate surrounding clutter and occluders, and predict the 3D object coordinates only for pixels in the instance masks.
Zakharov \etal~\cite{zakharov2019dpod} predict the UV texture coordinates instead of the 3D object coordinates.
Park \etal~\cite{park2019pix2pose} detect the object instances with 2D bounding boxes and for each detection predict the object silhouette, the 3D object coordinates, and the expected per-pixel error of the coordinates. Additionally, they apply adversarial training to improve prediction in the occluded parts and propose a novel loss function that can handle objects with global symmetries by guiding predictions to the closest symmetric pose.
Li \etal~\cite{li2019cdpn} estimate the 3D rotation from predicted 2D-3D correspondences and the 3D translation by directly regressing a scale-invariant 3D translation vector.
Instead of the 3D object coordinates, Pitteri \etal~\cite{pitteri20203d} predict
an embedding of local 3D geometry and match it to 3D locations on the object model that have a similar embedding vector.

\begin{figure}[b!]
	\vspace{1.5ex}
	\begin{center}
		\begingroup
		\small
		\begin{tabular}{ @{}c@{ } @{}c@{ } @{}c@{ } @{}c@{ } }
			Input image & Predicted heatmaps & Extracted maxima & Estimated pose \vspace{0.5ex} \\
			\includegraphics[width=0.243\linewidth]{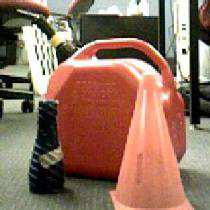} &
			\includegraphics[width=0.243\linewidth]{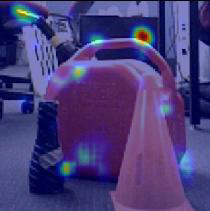} &
			\includegraphics[width=0.243\linewidth]{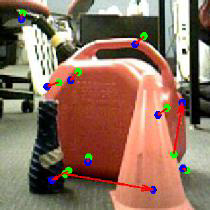} &
			\includegraphics[width=0.243\linewidth]{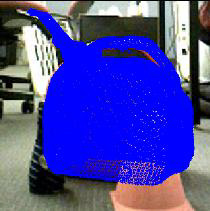} \\
		\end{tabular}
		\endgroup
		\caption{\label{fig:rel_2d_proj_pred} \textbf{Predicting 2D projections of 3D keypoints.} The 2D projections are usually obtained either by extracting maxima of predicted keypoint-specific heatmaps, as done in~\cite{rad2017bb8,pavlakos20176,oberweger2018making,tremblay2018deep} and shown in this figure, or by directly regressing the 2D coordinates, as done in~\cite{tekin2018real,hu2019segmentation,peng2019pvnet}. In the third image, the green dots show the extracted maxima and the blue dots show the ground-truth projections of the 3D keypoints associated with the 3D object model.
		}
	\end{center}
\end{figure}

\subsection{Predicting 2D Projections of 3D Keypoints} \label{sec:related_keypoint_projection}

Another approach to establish the 2D-3D correspondences is to predict the 2D projections of a fixed set of 3D keypoints, which are pre-selected for each object model (Figure~\ref{fig:rel_2d_proj_pred}).
Rad and Lepetit~\cite{rad2017bb8} define the 3D keypoints by the corners of the 3D bounding box of the object model, for each keypoint predict a probability distribution over all pixels of a 2D object detection, and link each 3D keypoint with the maximum of the predicted distribution. Pavlakos \etal~\cite{pavlakos20176} define the 3D keypoints manually on the model surface.
Oberweger~\cite{oberweger2018making} increase the robustness of the approach to partial occlusions by predicting the probability distribution from multiple small patches independently and accumulating the predictions before establishing the correspondences.
Fu and Zhou~\cite{fu2019deephmap++} present a similar idea.
Tremblay \etal~\cite{tremblay2018deep} predict over the whole image nine probability distributions, one for each corner of the 3D bounding box and one for its centroid, and eight vector fields pointing from the projections of the eight corners to the projection of the corresponding centroid. The vector fields enable recovering poses of multiple instances of the same object without the 2D object detection stage.
Tekin \etal~\cite{tekin2018real} achieve a speed of $50$ frames per second with a method inspired by the YOLO object detector~\cite{redmon2016you}.
They split the input image into regular cells, and for each cell predict the probability of each object's presence and regress the 2D projection coordinates of the 3D keypoints.
Hu \etal~\cite{hu2019segmentation} follow a similar strategy but use smaller cells and aggregate the predictions over the cells.
Peng \etal~\cite{peng2019pvnet} densely regress 2D unit vectors pointing
to the 2D projections of the 3D keypoints and find the 2D projections via a RANSAC-based procedure.
Pitteri \etal~\cite{pitteri2019cornet} focus on objects with prominent corners, detect generic 3D corners by predicting 2D projections of virtual keypoints around the corners, and estimate object poses by fitting 3D object models to the detected 3D corners. The advantage of this approach is that no re-training is necessary when adding new objects.

\subsection{Classifying Into Discrete 3D Viewpoints} \label{sec:rel_classifying_discrete}

This group includes methods that apply a 2D object detector such as SSD~\cite{liu2016ssd} or Faster R-CNN~\cite{ren2017faster} to localize object instances with amodal 2D bounding boxes (covering the whole instances including the occluded parts), and classify each box into a set of discrete 3D viewpoints. The 6D object poses are calculated from the predicted 3D viewpoints and from the scale and location of the 2D bounding boxes.
Because the predicted 3D viewpoints are discrete and estimating accurate amodal 2D bounding boxes is challenging~\cite{kehl2017ssd}, a post-refinement step is usually necessary to achieve a competitive performance.

\begin{figure}[h!]
	\vspace{1.5ex}
	\begin{center}
		\includegraphics[width=0.9\linewidth]{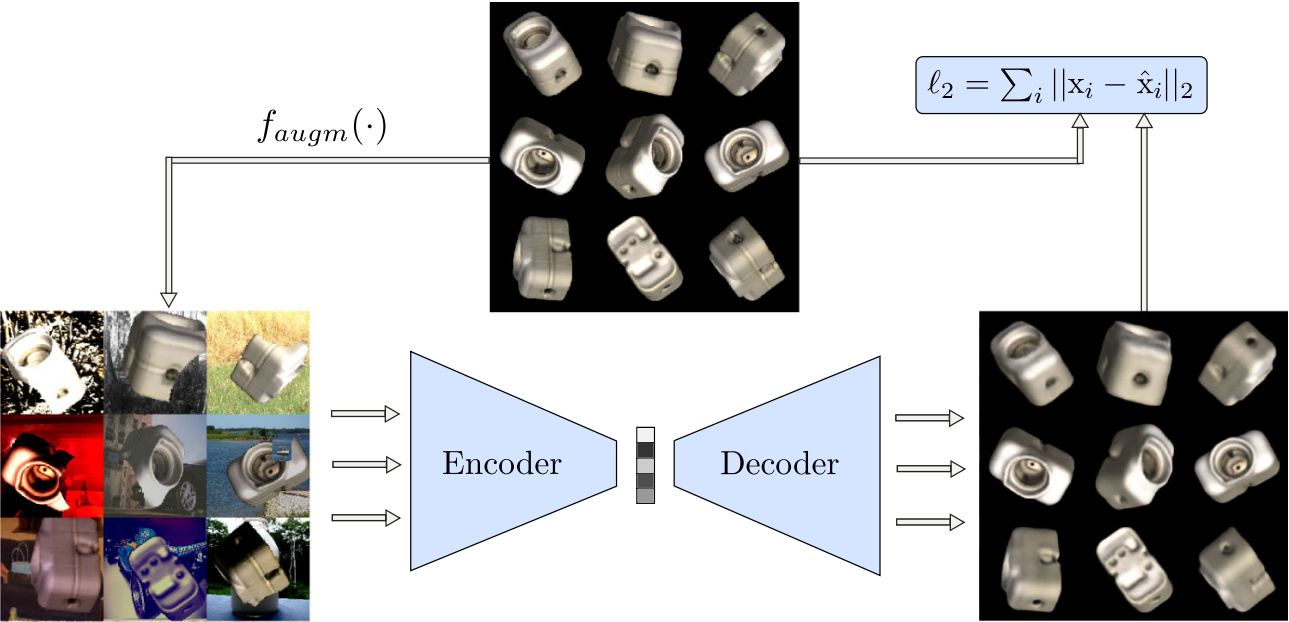} \\
		\vspace{1.0ex}
		\caption{\label{fig:rel_aae} \textbf{Augmented Autoencoder by Sundermeyer \etal~\cite{sundermeyer2019augmented}.}
		The network is trained to reconstruct clean object views from their augmented versions. The encoder is used to calculate descriptors of 2D object detections that are matched against a codebook.}
	\end{center}
\end{figure}

Kehl, Manhardt \etal~\cite{kehl2017ssd} extend the SSD object detector~\cite{liu2016ssd} to predict for each anchor box also a probability distribution over a set of discrete 3D viewpoints.
Sundermeyer \etal~\cite{sundermeyer2019augmented}, who received the best paper award at ECCV 2018, propose to calculate a global descriptor of a localized object instance by the so-called Augmented Autoencoder network. The network consists of an encoder, which maps the image region to a latent descriptor space, and a decoder, which maps the descriptor to a denoised reconstruction of the image region. %
The network is trained on renderings of the 3D object models that are heavily augmented by randomizing the light positions and reflectance properties, inserting random background images, varying image contrast and brightness, adding Gaussian blur and color distortions, and by creating random occlusions. %
At inference time, the nearest neighbor of the descriptor is retrieved from a pre-calculated codebook, where each entry is associated with a 3D orientation (Figure~\ref{fig:rel_aae}).
Instead of training a specific encoder and decoder per object, Sundermeyer \etal~\cite{sundermeyer2020multi} propose to share a single encoder among multiple objects, which dramatically improves the scalability of the method.

\begin{figure}[b!]
	\begin{center}
		\begingroup
		\setlength{\tabcolsep}{2pt} %
		\renewcommand{\arraystretch}{0.9} %
		\small
		\begin{tabular}{ c c c c c c }
			Input image & \hspace{3.5ex} & Object labels & Distances & \hspace{3.5ex} & ROI \vspace{0.5ex} \\
			\includegraphics[width=0.219\linewidth]{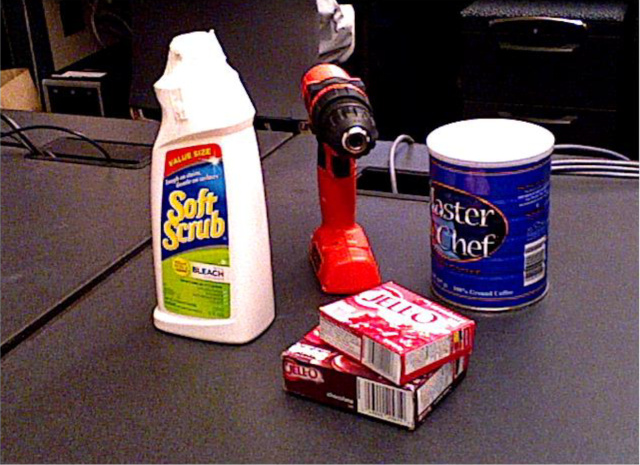} & &
			\includegraphics[width=0.219\linewidth]{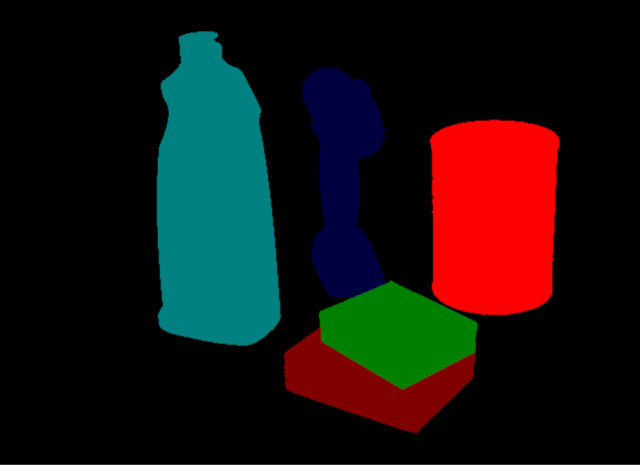} &
			\includegraphics[width=0.219\linewidth]{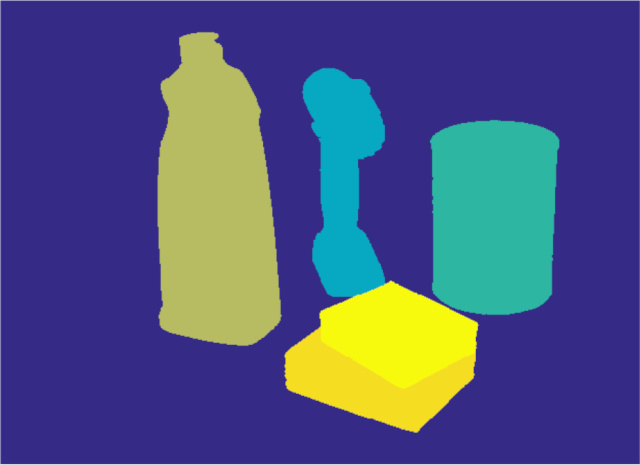} & &
			\includegraphics[width=0.219\linewidth]{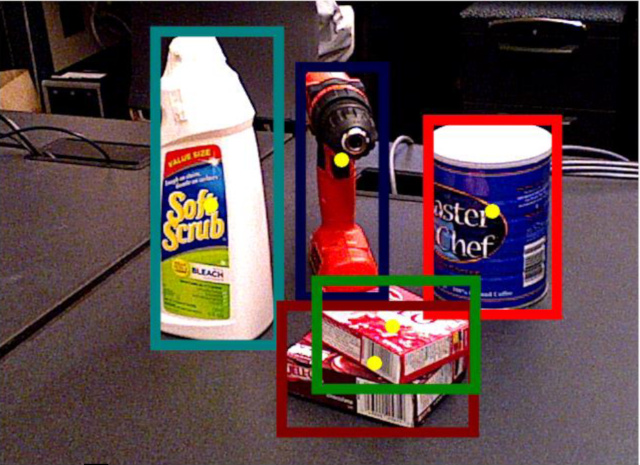} \\
			\vspace{-2ex} & & & & & \\
			& & Instance center X & Instance center Y & & Estimated poses \vspace{0.5ex} \\
			& &
			\includegraphics[width=0.219\linewidth]{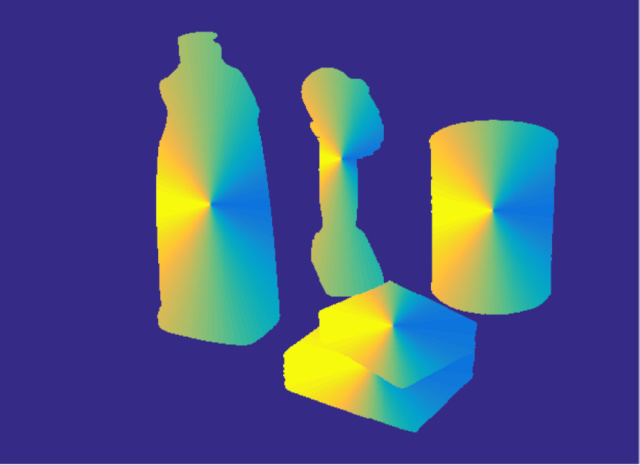} &
			\includegraphics[width=0.219\linewidth]{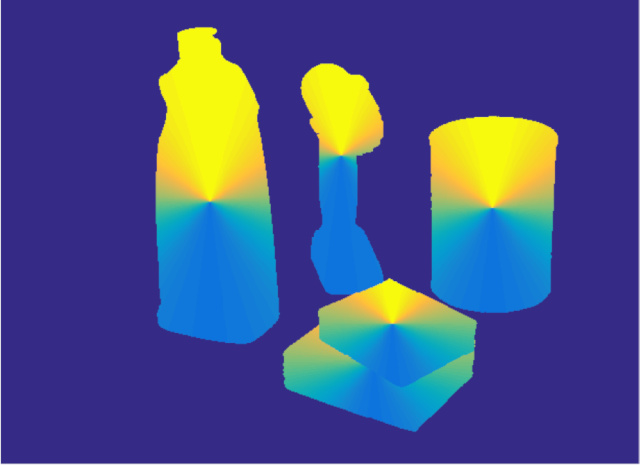} & &
			\includegraphics[width=0.219\linewidth]{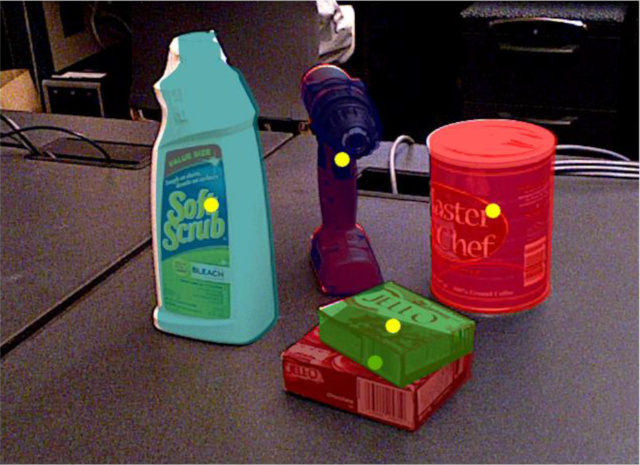} \\
		\end{tabular}
		\endgroup
		\caption{\label{fig:rel_posecnn} \textbf{Regressing 3D orientation and 3D location by Xiang \etal~\cite{xiang2017posecnn}.}
		Object instances are first found by densely predicting object labels, unit vectors to the 2D projection of the object center, and the distance from the camera.
		The 3D orientation of an instance is then predicted by cropping the corresponding region and regressing to a quaternion representation.
		}
	\end{center}
\end{figure}

\subsection{Regressing 3D Orientation and 3D Location} \label{sec:rel_regressing_pose}

Methods that aim to directly regress the 3D orientation and/or 3D location of the objects have also emerged.
Xiang \etal~\cite{xiang2017posecnn} first densely predict object labels, unit vectors pointing to the 2D projection of the corresponding object center, and the distance of the object center from the camera. The 2D projections of object centers are found by Hough voting, and masks of object instances are estimated by clustering pixels that vote for the same object center.
The 3D location of the object center is calculated from the predicted 2D projection of the center and the predicted distances from the camera, which are averaged over the instance mask. The 3D orientation is then predicted by cropping a region determined by the instance mask and regressing to a quaternion representation.
Manhardt \etal~\cite{manhardt2019explaining} regress multiple quaternions for each 2D object detection to estimate the distribution of possible poses induced by object symmetries.
Do \etal~\cite{do2019real} predict the 3D orientation by regressing a Lie algebra representation.
Li \etal~\cite{li2018unified} and Mahendran \etal~\cite{mahendran2018mixed} predict the 3D orientation via a mixed classification-regression scheme -- a 2D object detection is classified into 3D orientation bins and refined by regressing a continuous delta \wrt the center of a bin. This scheme allows capturing distributions that are non-unimodal due to object symmetries and does not suffer from quantization errors. Li \etal~\cite{li2018unified} use this scheme to predict also the 3D object location.
Bui \etal~\cite{bui2017x} directly regress the 3D orientation and 3D location of the object from X-ray images and show that the accuracy can be improved if the network is trained by jointly minimizing a pose error, as in~\cite{kendall2015posenet}, and a re-projection error.
Wang \etal~\cite{wang2019densefusion} aggregate deep per-pixel features over an object instance mask to obtain a global instance-level feature. The 6D object pose and its confidence are then predicted at each pixel of the instance mask from a concatenation of the per-pixel features and the global feature, and the pose with the highest confidence is selected as the final estimate.
Labb{\'e} \etal~\cite{labbe2020cosypose} propose a method that detects the objects by Mask R-CNN~\cite{he2017mask} and to each detection applies a neural network for coarse pose estimation followed by a neural network for iterative refinement. Both networks have the same structure inspired by DeepIM~\cite{li2018deepim}, but the first network is trained to predict the relative transformation \wrt the canonical pose and the second network is trained to predict small deltas \wrt the current estimate. Instead of regressing the discontinuous 4D quaternion representation, as done in~\cite{li2018deepim}, Labb{\'e} \etal regress a continuous 6D representation from~\cite{zhou2019continuity} which is more suitable for training.

\subsection{Handling the Pose Ambiguity} \label{sec:rel_handling_ambiguity}

\begin{figure}[t!]
	\begin{center}
		\begingroup
		\renewcommand{\arraystretch}{0.9} %
		\small
		\begin{tabular}{ @{}c@{ } @{}c@{ } @{}c@{ } @{}c@{ } @{}c@{ } }
			(a) & (b) & (c) & (d) & (e)  \vspace{0.5ex} \\
			\includegraphics[width=0.193\columnwidth]{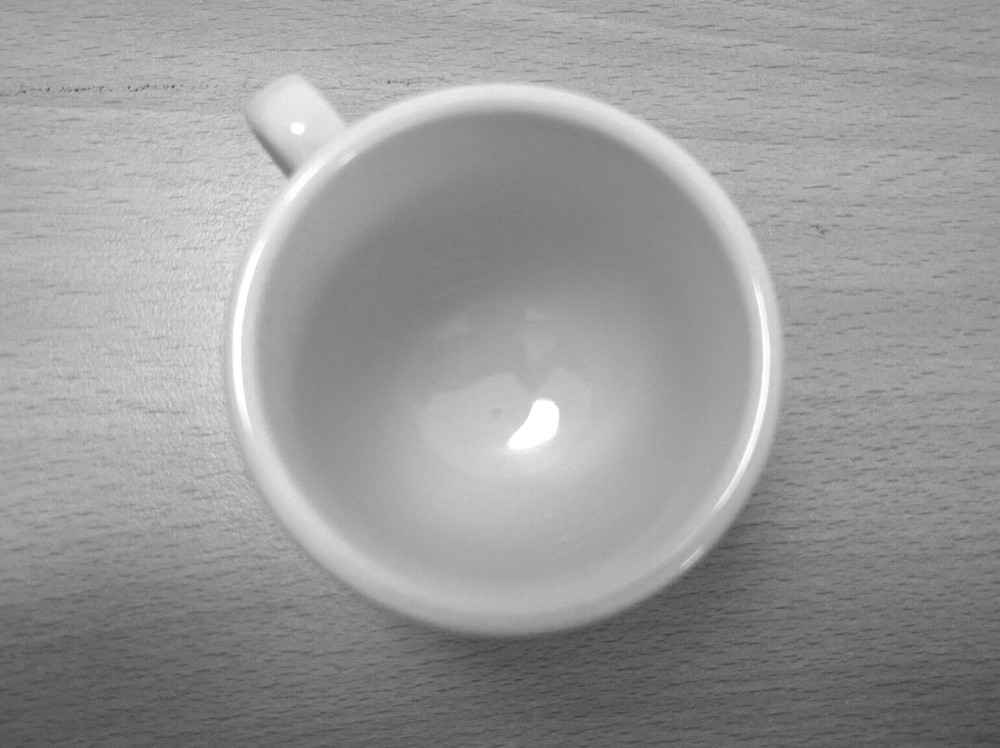} &
			\includegraphics[width=0.193\columnwidth]{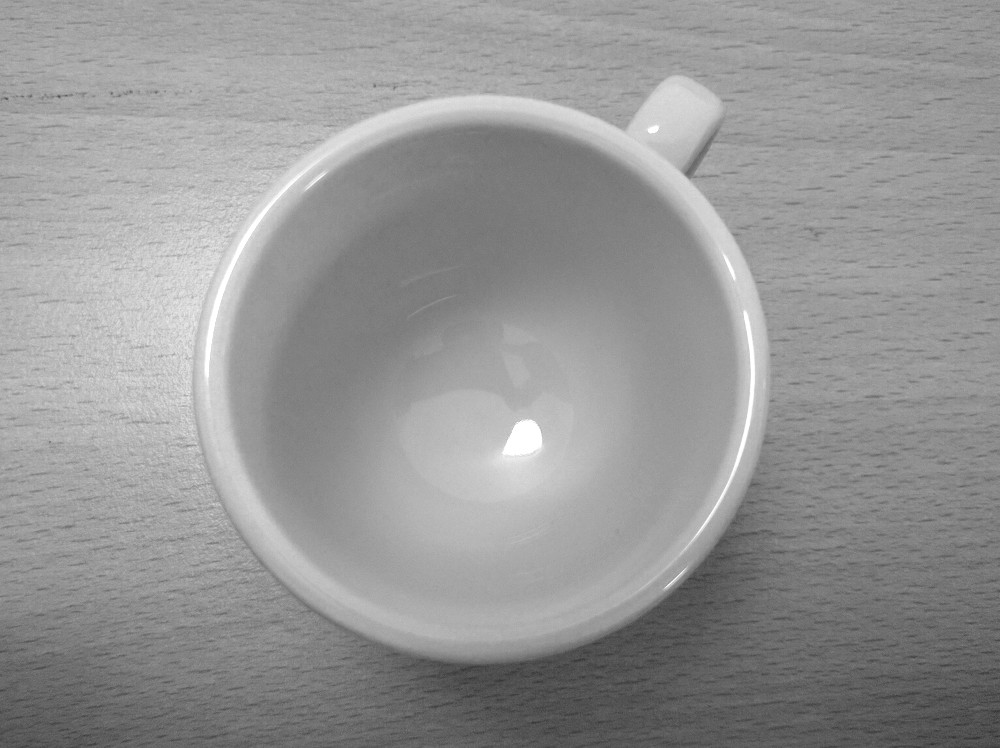} &
			\includegraphics[width=0.193\columnwidth]{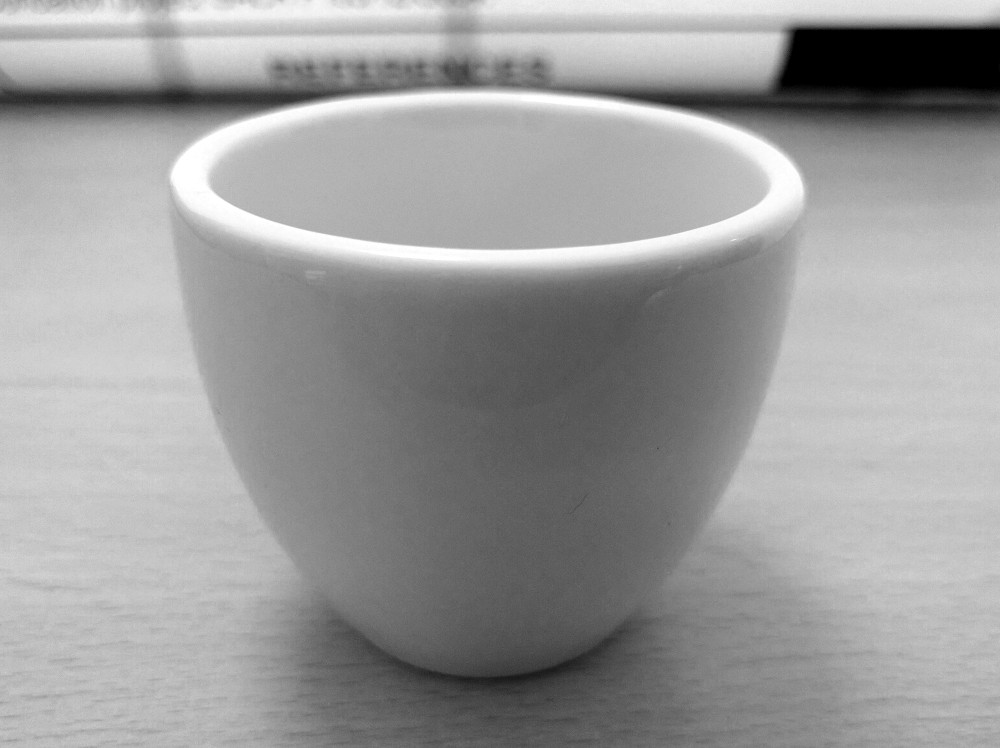} &
			\includegraphics[width=0.193\columnwidth]{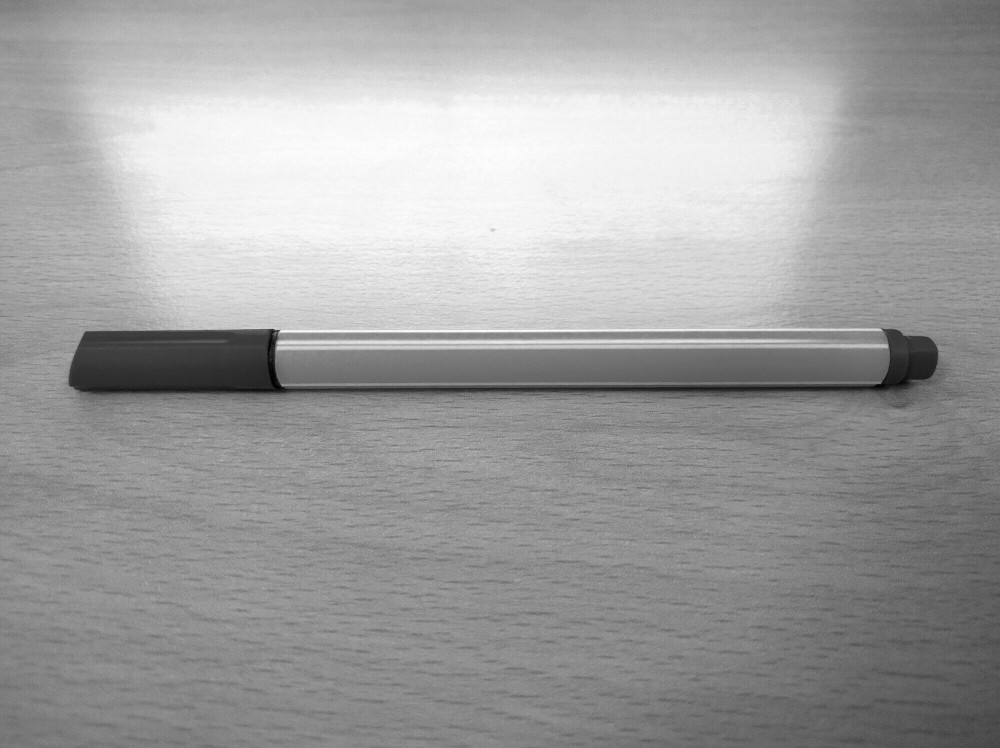} &
			\includegraphics[width=0.193\columnwidth]{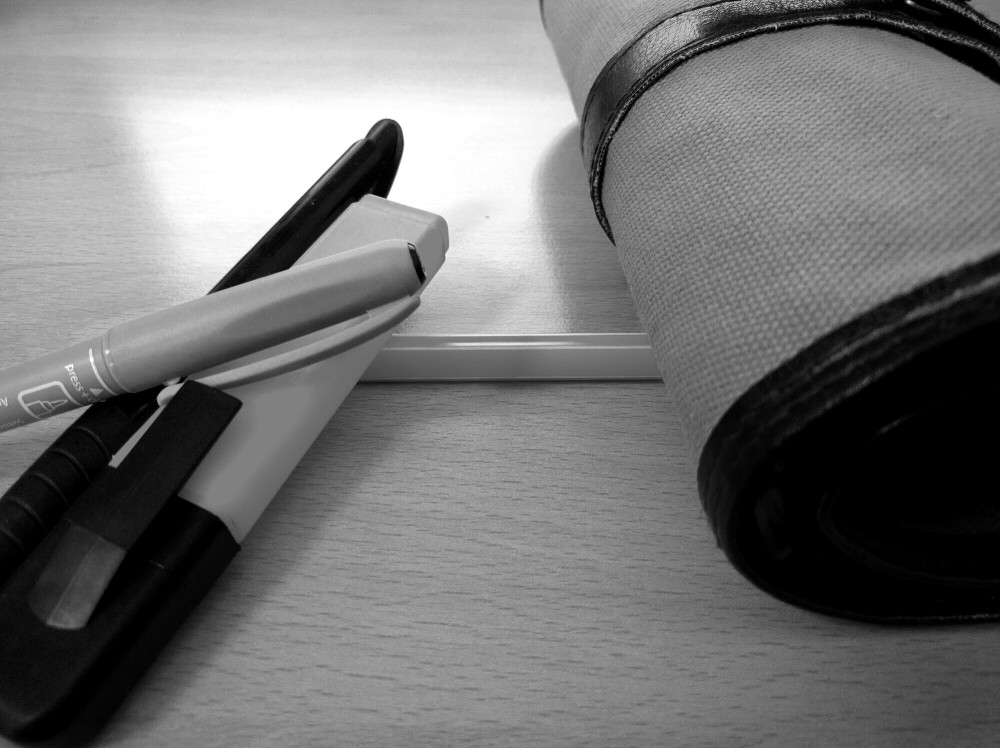} \\
			\vspace{-2.5ex} & & & & \\
		\end{tabular}
		\endgroup
		\caption{\label{fig:pose_ambiguity} \textbf{Object pose ambiguity.} Different poses of a cup (a, b) cannot be distinguished if the handle is not visible due to self-occlusion (c). The pose of a pen (d) is ambiguous if its discriminative ends are occluded by other objects (e).}
	\end{center}
\end{figure}

The object pose may be ambiguous, \ie, there may be (infinitely) many poses that are consistent with the image. The ambiguity is caused by the existence of multiple fits of the visible part of the object surface to the object model. The visible part is determined by self-occlusion and occlusion by other objects and the multiple fits are induced by global or partial object symmetries~\cite{hodan2016evaluation,mitra2006partial}. See Figure~\ref{fig:pose_ambiguity} for examples of ambiguous poses.

\begin{figure}[t!]
	\begin{center}
		
		\begingroup
		\small
		\renewcommand{\arraystretch}{0.9} %
		
		\begin{tabular}{ @{}c@{ } @{}c@{ } @{}c@{ } @{}c@{ } @{}c@{ } }
			\multicolumn{2}{c}{Predicting 3D object coordinates} & & \multicolumn{2}{c}{Predicting 2D projections of 3D keypoints} \vspace{0.5ex} \\
			
			\includegraphics[width=0.239\columnwidth]{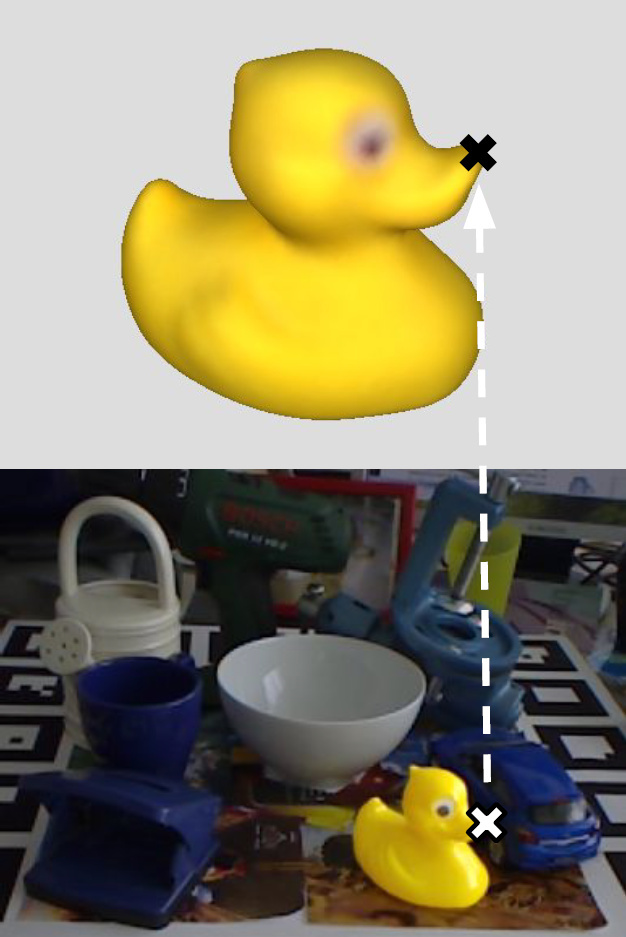} &
			\includegraphics[width=0.239\columnwidth]{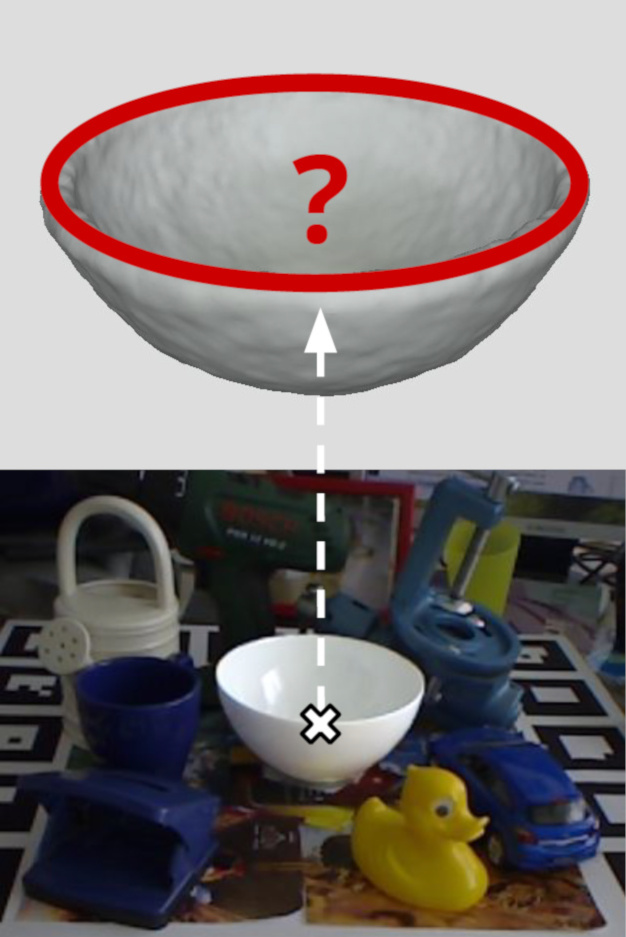} &
			\hspace{1.0ex} &
			\includegraphics[width=0.239\columnwidth]{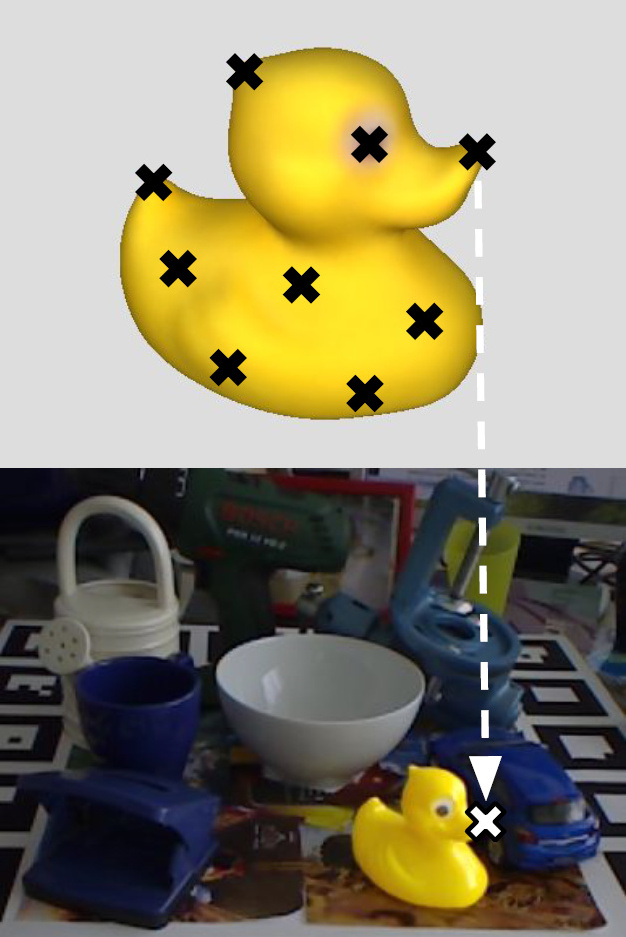} &
			\includegraphics[width=0.239\columnwidth]{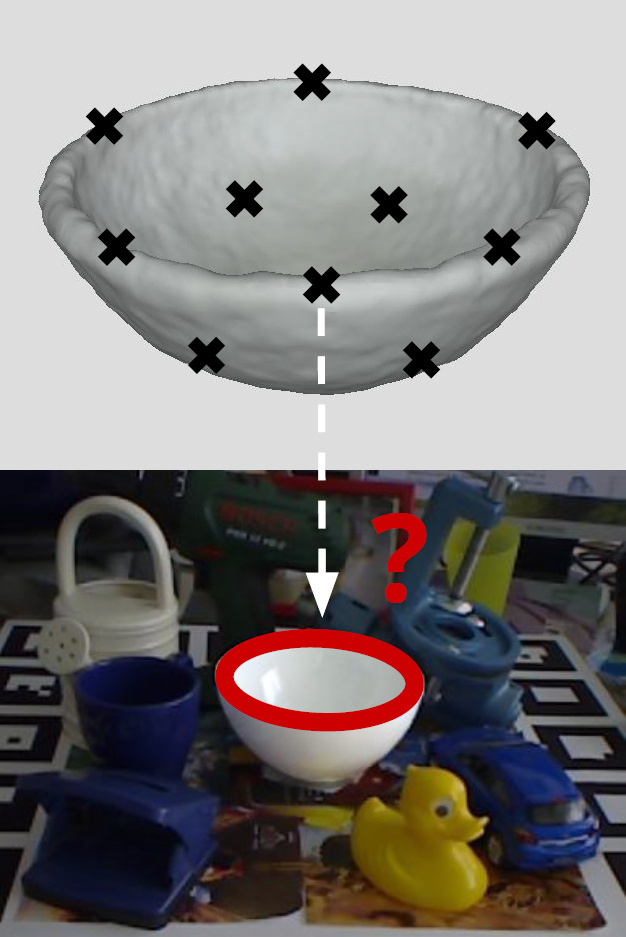} \\
		\end{tabular}
		\endgroup
		
		\caption{\label{fig:rel_corr_ambiguity} \textbf{Ambiguity of 2D-3D correspondences.}
			A 2D image location corresponds to a~single 3D location on the object model in the case of distinct object parts, but to one of multiple indistinguishable 3D locations in the case of global or partial symmetries (left two columns).
			Existing correspondence-based methods
			(i) compromise among the possible 3D locations and tend to return the average, which is often not a valid solution, or (ii) consider only the most confident 3D location, which is also problematic
			as this yields a limited support for each of the possible object poses. A similar problem arises when predicting 2D projections of 3D keypoints.
		}
	\end{center}
\end{figure}

When the object pose is ambiguous, a 2D image location
potentially corresponds to multiple 3D locations on the
object model, and vice versa.
Such a many-to-many relationship degrades the performance of the correspondence-based methods (Sections \ref{sec:related_object_coordinates} and \ref{sec:related_keypoint_projection}) which assume a one-to-one relationship.
The correspondence-based methods can be split into a classification-based and a regression-based group.
The classification-based methods predict up to one corresponding 3D location for each 2D location by classifying
into 3D bins~\cite{brachmann2014learning,nigam2018detect}, or predict up to one 2D location for each 3D keypoint, where the 2D location is typically given by the maximum response in a predicted heatmap~\cite{pavlakos20176,oberweger2018making,fu2019deephmap++} (Figure~\ref{fig:rel_2d_proj_pred}).
Both approaches yield a set of correspondences which carries only a limited support for each of the possible object poses.
On the other hand, the regression-based methods~\cite{tekin2018real,zakharov2019dpod,peng2019pvnet} need to compromise among the potentially corresponding locations and tend to return the average, which is often not a valid solution.
For example, the average of all points on a sphere is the center of the sphere, which is not a valid surface location. The problem is illustrated in Figure \ref{fig:rel_corr_ambiguity}.
Besides, the pose ambiguity also causes problems to methods aiming to regress the pose directly (Section \ref{sec:rel_regressing_pose}),~as~these methods usually assume a unimodal pose distribution.

The problem of pose ambiguity
Rad and Lepetit~\cite{rad2017bb8} assume that the global object symmetries are known and propose a pose normalization applicable to the case when the projection of the axis of symmetry is close to vertical.
Pitteri \etal~\cite{pitteri2019object} introduce a pose normalization that is not limited to this special case.
Kehl \etal~\cite{kehl2017ssd} train a classifier for only a subset of viewpoints defined by global object symmetries.
Corona \etal~\cite{corona2018pose} show that predicting the order of rotational symmetry can improve the accuracy of pose estimation.
Xiang \etal~\cite{xiang2017posecnn} optimize a loss function that is invariant to global object symmetries.
Park \etal~\cite{park2019pix2pose} and Labb{\'e} \etal~\cite{labbe2020cosypose} guide pose regression by calculating the loss \wrt to the closest symmetric pose.
However, all of these approaches cover only pose ambiguities due to global object symmetries, not
due to partial object symmetries.

As the EPOS method, which we present in Chapter~\ref{ch:method_epos}, the methods by Li \etal~\cite{li2018unified}, Manhardt \etal~\cite{manhardt2019explaining}, Sundermeyer \etal~\cite{sundermeyer2019augmented}, and Ammirato \etal~\cite{ammirato2020symgan} can handle pose ambiguities due to both global and partial object symmetries without requiring any prior information about the symmetries.
Li \etal~\cite{li2018unified} deals with the possibly non-unimodal pose distribution by a classification and regression scheme applied to the rotation and translation space.
Manhardt \etal~\cite{manhardt2019explaining} predicts multiple poses for each object instance to estimate the distribution of possible poses induced by symmetries.
Sundermeyer \etal~\cite{sundermeyer2019augmented} detects object instances in the input image, for each detected region calculates a descriptor, and find its nearest neighbor among pre-calculated descriptors obtained by rendering the object model from different viewpoints and under different occlusions. This approach handles object symmetries implicitly and allows to enumerate all poses that are consistent with the image by considering all nearby descriptors~\cite{deng2019poserbpf}.
Ammirato \etal~\cite{ammirato2020symgan} learn to predict object orientation via an adversarial training framework that consists of a generator and a discriminator.
The generator predicts the object orientation from the input image and the discriminator evaluates the visual similarity of the input image and the object model rendered in the predicted orientation. Since the visual similarity is used as a training signal for the generator, the generator is encouraged to predict any orientation that is consistent with the input image.
Nevertheless, all of these methods rely on localizing the object instances with amodal 2D bounding boxes, which is challenging when the instances are partially occluded~\cite{kehl2017ssd}. EPOS does not rely on such localization.

\subsection{Bridging the Domain Gap}

Deep neural networks have become the standard tool for tackling many computer vision problems including object detection and object pose estimation.
Besides the progress in optimization and architecture design of neural networks and the development of graphics cards that accelerate the training process, another critical factor to the success of neural networks is the availability of a large number of training images~\cite{goodfellow2016deep}. However, capturing and annotating real training images for problems such as object pose estimation requires a significant human effort or a specialized acquisition setup~\cite{hodan2017tless}.

\begin{figure}[b!]
	\begin{center}
		\includegraphics[width=1.0\linewidth]{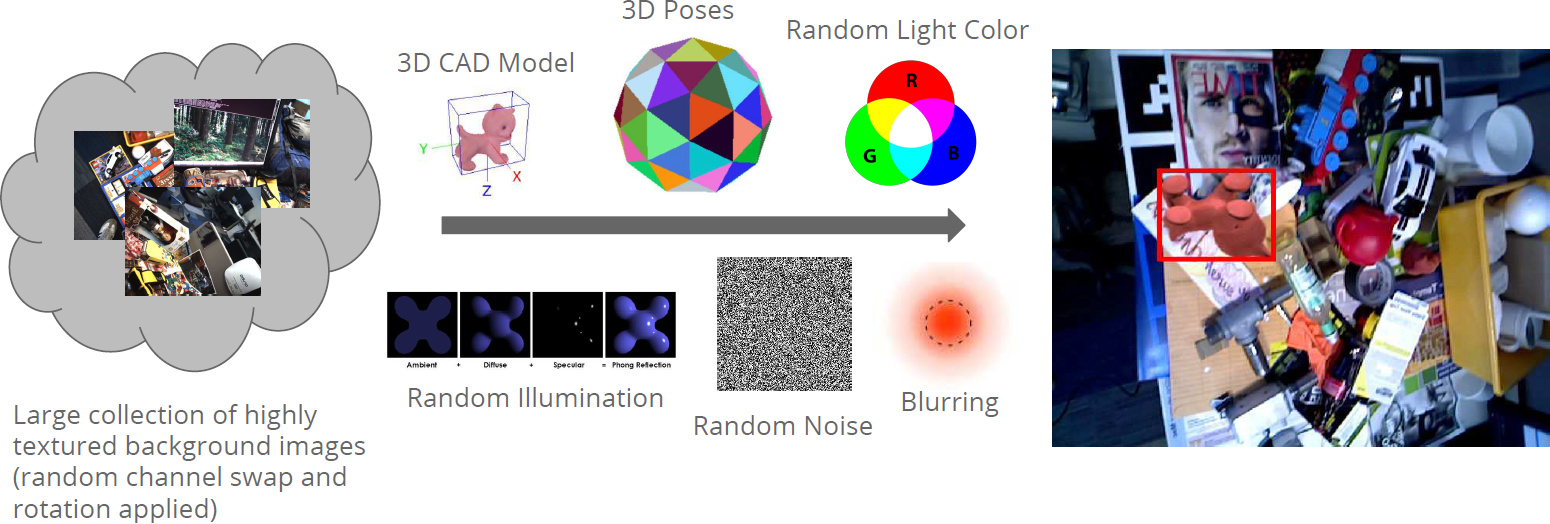} \\
		\caption{\label{fig:rel_hinterstoisser_synthesis} \textbf{Image synthesis pipeline by Hinterstoisser~\cite{hinterstoisser2017pre}.}
			The 3D object models are rendered by OpenGL on top of randomly selected photographs, with randomized parameters of the Phong illumination model, random light color, and random image noise and blur.
		}
	\end{center}
\end{figure}

An alternative to acquiring real training images is synthesizing images by computer graphics.
This approach scales well as it requires only a minimal human effort, which may include 3D modeling.
Su \etal~\cite{su2015render} synthesize images of 3D object models
for viewpoint estimation, Hinterstoisser \etal~\cite{hinterstoisser2017pre} for object instance detection
(Figure~\ref{fig:rel_hinterstoisser_synthesis}), and Dosovitskiy \etal~\cite{dosovitskiy2015flownet} for optical flow estimation. They all use a fixed OpenGL pipeline and paste the rendered pixels over randomly selected real photographs. 
Rad \etal~\cite{rad2017bb8}, Tekin \etal~\cite{tekin2018real} and Dwibedi \etal~\cite{dwibedi2017cut}
paste segments of objects from real images on other real images for object detection and pose estimation.
Dvornik \etal~\cite{dvornik2018modeling} demonstrate the importance of selecting suitable background images, and Hinterstoisser \etal~\cite{hinterstoisser2019annotation} achieve improvements when both the objects and the background are synthetic.
While these approaches are easy to implement, the resulting images are not realistic. Objects often have inconsistent shading with respect to the background scene, interreflections and shadows are missing, and the object pose and context are not natural.

Another line of work explore rendering of complete scenes and generating corresponding ground-truth maps.
Richter \etal~\cite{richter2016playing,richter2017playing} leverage existing commercial game engines to acquire training data for several tasks.
Synthia~\cite{ros2016synthia} and Virtual KITTI~\cite{gaidon2016virtual} were generated using virtual cities modeled from scratch. Handa \etal~\cite{hand2016understanding} and Zhang \etal~\cite{zhang2017physically} model 3D scenes for semantic scene understanding and McCormac \etal~\cite{mccormac2017scenenet} synthesize
RGB-D video frames from simulated cameras moving within static scenes.

\begin{figure}[b!]
	\begin{center}
		\begingroup
		\small
		\setlength{\tabcolsep}{4pt} %
		\renewcommand{\arraystretch}{0.9} %
		\begin{tabular}{ @{}c@{ } @{}c@{ } @{}c@{ } }
			Examples of training images & \hspace{1.0ex} & Test image \vspace{0.5ex} \\
			\includegraphics[height=0.242\columnwidth]{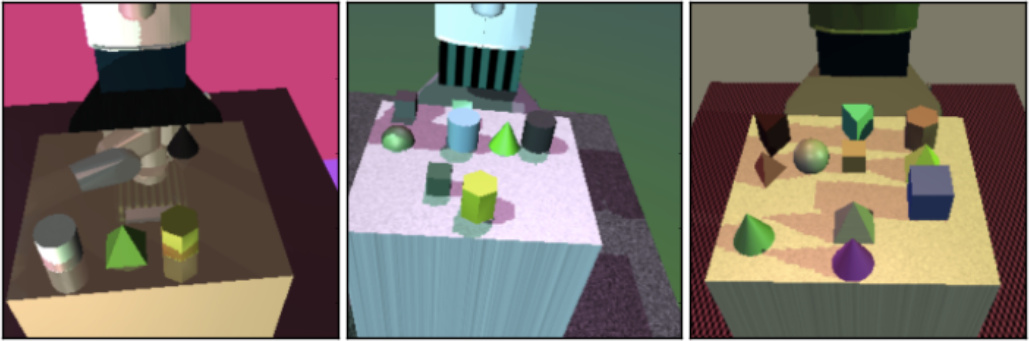} & &
			\includegraphics[height=0.242\columnwidth]{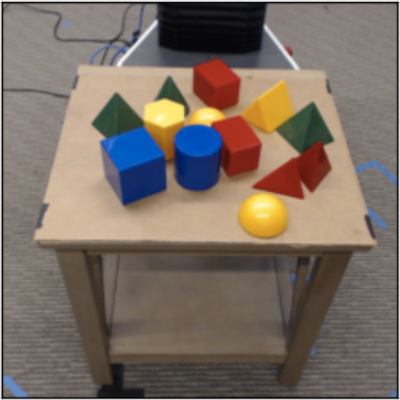} \\
		\end{tabular}
		\endgroup
		
		\caption{\label{fig:rel_domain_randomization} \textbf{Domain randomization by Tobin \etal~\cite{tobin2017domain}.}
			An object detector is trained on hundreds of
			thousands of low-fidelity rendered images with randomized camera/object positions, textures, and lighting conditions.
			At test time, the detector is applied to real images.
		}
	\end{center}
\end{figure}

Despite training neural networks on massive datasets of diverse synthetic images, a~large drop in performance was observed when models trained only on synthetic images were tested on real images~\cite{richter2016playing,su2015render,rozantsev2018sharing}.
The domain gap between the synthetic and real images can be reduced by domain adaptation techniques, which aim to learn domain invariant representations or to transfer trained models from one domain to another~\cite{csurka2017comprehensive}. Besides, the issue can be mitigated by domain randomization techniques, which randomize rendering parameters and were shown beneficial for training object detection and pose estimation models~\cite{tobin2017domain,tremblay2018training,sundermeyer2019augmented,zakharov2019deceptionnet} (Figure~\ref{fig:rel_domain_randomization}).

A different line of work, presumably complementary to the domain adaptation and randomization techniques, has recently tried to reduce the domain gap by synthesizing training images with a higher degree of visual realism. The use of physically-based rendering has been considered with this motivation and shown promising results~\cite{li2018cgintrinsics,zhang2017physically}.

Physically-based rendering (PBR) techniques, \eg, Arnold~\cite{georgiev2018arnold}, accurately simulate the flow of light energy in the scene by ray tracing. This approach naturally accounts for complex illumination effects such as scattering, refraction and reflection, including diffuse and specular interreflection between the objects and the scene and between the objects themselves. The rendered images look realistic and are often difficult to differentiate from real photographs~\cite{pharr2016physically}. Rendering techniques based on rasterization, \eg, OpenGL~\cite{shreiner2009opengl}, can approximate the complex effects in an ad hoc way through custom shaders, but the approximations cause physically incorrect artifacts that are difficult to eliminate~\cite{marschner2015fundamentals}. Physically-based rendering has been
noticeably slower than rasterization, however, the recently introduced Nvidia RTX ray tracing GPU promises a substantial reduction of the rendering time. Moreover, the rendering time can be reduced by tracing fewer rays per pixel and applying a denoiser to effectively eliminate noise in the rendered image~\cite{inteldenoise}.

Li and Snavely \cite{li2018cgintrinsics} use PBR images to train models for intrinsic image decomposition and Zhang \etal~\cite{zhang2017physically} for semantic segmentation, normal estimation and boundary detection.
Wood \etal~\cite{wood2015rendering} use PBR images of eyes for training gaze estimation models.
Attias \etal~\cite{movshovitz2016useful} render photorealistic images of 3D car models placed within 3D scene models and show the benefit over naive rendering methods, whereas Tremblay \etal~\cite{tremblay2018deep} render objects in physically plausible poses in diverse scenes, but do not use physically-based rendering.
Other ways to generate photorealistic images~\cite{wood2016learning,shrivastava2017learning} and methods to synthesize realistic depth images~\cite{planche2017depthsynth} have also been proposed.

In Chapter~\ref{ch:synthesis}, we present an approach to synthesize PBR images of 3D object models arranged in physically plausible poses inside 3D scene models, and show that the images can be used to effectively train networks for object detection or object pose estimation.

\subsection{Using the Depth Image Channel}

While most existing methods for 6D object pose estimation that are based on neural networks apply the networks only to the RGB channels, promising methods using the depth channel as an additional input start to emerge.
Wang \etal~\cite{wang2019densefusion} segment objects
in RGB 
channels, in each mask calculate per-pixel color features by an auto-encoder network and per-pixel depth features by PointNet~\cite{qi2017pointnet}, fuse the two types of features, and regress object poses as described in Section~\ref{sec:rel_regressing_pose}.
He \etal~\cite{he2020pvn3d} fuse color and depth features (the latter are calculated by PointNet++~\cite{qi2017pointnet++}) and regress 3D coordinates of model keypoints. Similarly, Qi and Chen~\cite{qi2020imvotenet} use fused features to vote for 3D object centers and predict 3D bounding boxes.
Hagelskj{\ae}r and Buch~\cite{hagelskjaer2020pointvotenet} process a color point cloud with PointNet to find candidate object locations and to establish 3D-3D correspondences at each such location. Chen \etal~\cite{chen2020g2l} train PointNet to predict the mask and translation of an object instance from the depth channel.
Li \etal~\cite{li2018unified} process the RGB and depth channels with two convolutional branches and fuse them before predicting the object pose via a mixed classification-regression scheme (Section~\ref{sec:rel_regressing_pose}). Sock \etal~\cite{sock2018multi} process the depth channel with a convolutional neural network that is trained jointly for 2D detection, 6D pose estimation, and hypotheses verification.
F{\'e}lix and Neves~\cite{rodrigues2019deep,raposo2017using} and K\"onig and Drost~\cite{koenig2020hybrid} propose hybrid methods which segment object instances using a neural network
and estimate the object pose from each mask using the point pair features.

\section{Refinement Methods} %

The accuracy of pose estimates produced by any of the methods reviewed above can be often improved by methods specialized on estimating small refinement transformations. %
The refinement is crucial especially for template-matching methods (Sections~\ref{sec:rel_methods_template}) and methods predicting the pose by classification into discrete 3D viewpoints (Sections~\ref{sec:rel_classifying_discrete}).

\begin{figure}[b!]
	\begin{center}
		\includegraphics[width=0.9\linewidth]{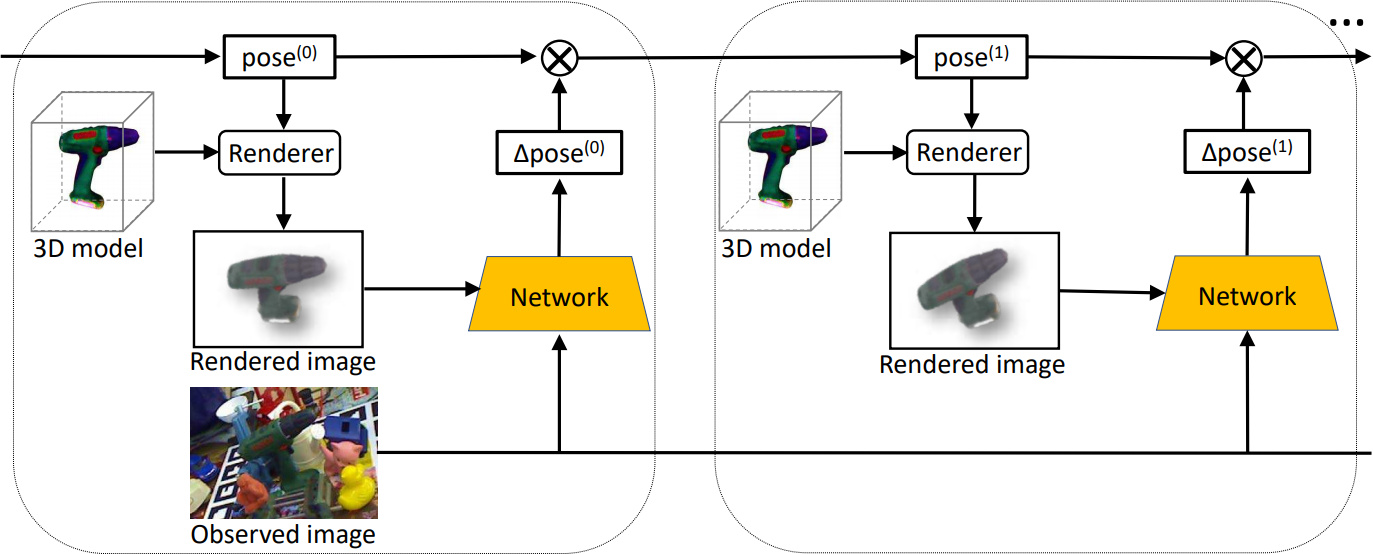} \\
		\vspace{1.5ex}
		\caption{\label{fig:rel_deepim} \textbf{Deep iterative refinement by Li \etal~\cite{li2018deepim}.}
			A neural network is trained to predict a relative rigid transformation from a cropped region of the input RGB image and a rendering of the 3D object model in the current pose estimate. %
		}
	\end{center}
\end{figure}

The Iterative Closest Point (ICP) algorithm~\cite{besl1992method} is the golden standard for pose refinement from point clouds. ICP aims to find an aligning transformation between two sets of points by alternating between establishing putative correspondences between the closest points and estimating a transformation that aligns them. When refining object poses, the source point cloud is defined by vertices of the 3D object model in the estimated pose or calculated from rendered depth image of the model. The target point cloud is calculated from the depth channel of the input image. ICP converges to the correct transformation when the source point cloud is initially sufficiently close to the target point cloud.
However, when the two point sets are relatively far apart or have a small overlap, the strategy of matching closest points generates large numbers of incorrect correspondences. The situation is aggravated by the inevitable presence of noise and outliers. Consequently, ICP can easily get stuck in local minima and its performance largely depends on the quality of initialization. To counter this, numerous enhancements to the basic ICP have been proposed that aim to improve the speed of convergence or increase the robustness to local minima, outlying points, and noise~\cite{rusinkiewicz2001efficient}. %
These enhancements often require considerable trial and error for tuning their parameters.
ICP can be applied also to the intensity edges to align the projection of the object model to the image content~\cite{zhang1994iterative,damen2012real}.

An alternative to ICP is Particle Swarm Optimization (PSO)~\cite{poli2007swarm,oikonomidis2011efficient,ivekovic2008human}, which stochastically evolves a population of candidate pose estimates over multiple iterations and was demonstrated to be less prone to local minima than ICP~\cite{zabulis2015object}.
PSO is applied for post-refinement in the template-matching method proposed in Chapter~\ref{ch:method_template}.

Learning-based refinement methods have emerged as well. Oberweger \etal~\cite{oberweger2015training} use a neural nework to iteratively refine a hand pose estimate. The network predicts a pose update from the input image and a rendered image of the hand in the current pose. The update is applied to the current pose and the process is repeated. A conceptually similar procedure was used to refine object pose estimates by Rad and Lepetit~\cite{rad2017bb8}, Li \etal~\cite{li2018deepim} (Figure~\ref{fig:rel_deepim}), Manhardt and Kehl \etal~\cite{manhardt2018deep}, and Wang \etal~\cite{wang2019densefusion}. Bauer \etal~\cite{bauer2020verefine} additionally integrate physics simulation
to achieve physically plausible pose estimates.

\section{Datasets} \label{sec:rel_datasets}

This section reviews datasets for estimating the 6D pose of specific rigid objects, grouped by the type of included images, and also mentions datasets for similar problems. If not stated otherwise, the datasets include ground truth
in the form of 6D object poses.

\subsection{RGB-D Datasets}

Most datasets used for research in object pose estimation are captured with consumer-grade RGB-D sensors which became widely available with the launch of Microsoft Kinect in 2010. The first generation of Kinect is based on the structured-light principle -- a light speckle pattern is projected onto the scene using a near-infrared laser emitter and the light reflected back to a standard off-the-shelf infrared camera is analyzed to estimate the depth of the scene surfaces~\cite{khoshelham2012accuracy,kramer2012hacking}. Although the exact details are not publicly available as the technology is patented, it is known that the sensor relies upon established computer vision techniques such as depth from focus and depth from stereo. In 2014, the second generation of Kinect was released. This completely new sensor is based on the time-of-flight principle in which the depth of a scene is measured by the absolute time needed by a light wave to travel into the scene and, after reflection, back to the sensor.

For the evaluation of methods based on the 3D local features, Aldoma \etal~\cite{aldoma2012global} created a dataset with 3D mesh models without color of 35 household objects. The objects are
often symmetric and mutually similar in shape and size.
The dataset includes 50 test RGB-D images of table-top scenes with multiple objects in single instances, with no clutter and various levels of occlusion.
Several small datasets that were used for evaluation of the SHOT descriptor are provided by Salti \etal~\cite{salti2014shot}. These datasets include synthetic data as well as data acquired with a spacetime-stereo method and an RGB-D sensor.

The Challenge and Willow datasets~\cite{xie2013multimodal}, which were collected for the 2011 ICRA Solutions in Perception Challenge, share a set of 35 textured household objects. Training data for each object is given in the form of 37 RGB-D training images that show the object from different viewpoints, plus a color point cloud obtained by merging the training images. The Challenge and Willow datasets respectively contain 176 and 353 test RGB-D images of several objects in single instances placed on top of a turntable. The Willow dataset also features distractor objects and object occlusion.
Similar is the TUW dataset~\cite{aldoma2014automation} that presents 17 textured and texture-less objects shown in 224 test RGB-D images. Instead of a turntable setup, images were obtained by moving a robot around a static cluttered environment with some objects appearing in multiple instances.
The BigBIRD dataset~\cite{singh2014bigbird} includes images of 125 mostly textured objects that were captured in isolation on a turntable with multiple calibrated RGB-D and DSLR sensors, with fixed lighting and object-sensor distance, and with no occlusions or clutter. For each object, the dataset provides 600 RGB-D point clouds, 600 high-resolution RGB images, and a color 3D mesh model reconstructed from the point clouds.
Georgakis \etal~\cite{georgakis2016multiview} provide 6735 test RGB-D images from kitchen scenes including a subset of the BigBIRD objects, with the ground truth in the form of 2D bounding boxes and 3D point labeling.
Lai \etal~\cite{lai2011large} created an extensive dataset with 250K segmented RGB-D images of 300 household objects captured on a turntable from three elevations. The dataset also contains 22 test video sequences with a few hundred RGB-D frames in each showing the objects in household scenes. The ground truth is provided only in the form of approximate rotation angles for training images and in the form of 3D point labeling for test images.
Schlette \etal~\cite{schlette2014new} synthesized RGB-D images from simulated object manipulation scenarios involving 4 texture-less objects from the Cranfield assembly benchmark~\cite{collins1985development}.

The RGB-D datasets by Hinterstoisser \etal~\cite{hinterstoisser2012accv}, Brachmann \etal~\cite{brachmann2014learning}, Tejani \etal~\cite{tejani2014latent}, Doumanoglou \etal~\cite{doumanoglou2016recovering}, Rennie \etal~\cite{rennie2016dataset}, Drost \etal~\cite{drost2017introducing}, Xiang \etal \cite{xiang2017posecnn}, and Kaskman \etal~\cite{kaskman2019homebreweddb} are included in the BOP benchmark and reviewed in Section~\ref{sec:bop_datasets}. The T-LESS dataset, which features industry-relevant texture-less objects with symmetries and mutual similarities and is also included in BOP, is presented in Chapter~\ref{ch:tless}.

\subsection{Depth-Only and RGB-Only Datasets}

The depth-only dataset of Mian \etal~\cite{mian2006three} includes 3D mesh models of 5 objects and 50 test depth images from an industrial range scanner which show the modeled objects occluding each other.
A similar dataset is provided by Taati \etal~\cite{taati2007variable}.
The Desk3D dataset by Bonde \etal~\cite{bonde2014robust} includes 3D mesh models of 6 objects
captured in over 850 test depth images with occlusion, clutter and similarly looking distractor objects.
The images were taken with an RGB-D sensor but only the depth images are publicly available.

The IKEA dataset by Lim \etal~\cite{lim2013parsing} includes RGB images of objects that are aligned with their 3D models. Crivellaro \etal~\cite{crivellaro2015novel}
supply 3D CAD models and annotated RGB sequences with 3 highly occluded and texture-less objects. Mu\~{n}oz \etal~\cite{munoz2016fast} provide RGB sequences of 6 texture-less objects captured in isolation against a clean background and without occlusion.
Further to the above, there exist RGB datasets such as \cite{damen2012real,tombari2013bold,rios2013discriminatively,hsiao2014occlusion} for which the ground truth is provided only in the form of 2D bounding boxes.

\subsection{Datasets for Similar Problems}

The RGB-D dataset of Michel \etal~\cite{michel2015pose} is focused on articulated objects with the goal to estimate the 6D pose of each object part, subject to the constraints introduced by their joints.
There are also datasets for categorical pose estimation.
For example, the 3DNet~\cite{wohlkinger20123dnet} and the UoB-HOOC~\cite{walas2016uob} contain generic 3D models and RGB-D images annotated with 6D object poses. The UBC VRS~\cite{meger2011mobile}, the RMRC (a subset of NYU Depth v2~\cite{silberman2012indoor} with annotations derived from~\cite{guo2013support}), the B3DO~\cite{janoch2013category}, and the SUN RGB-D~\cite{song2015sun} provide no 3D models and ground truth only in the form of bounding boxes.
The PASCAL3D+~\cite{xiang2014beyond} and the ObjectNet3D~\cite{xiang2016objectnet3d} provide generic 3D models and RGB images annotated with ground-truth 6D poses.
A comprehensive overview of RGB-D datasets for various computer vision tasks is provided by Firman~\cite{firman2016rgbd}.

\section{Evaluation Methodologies} \label{sec:rel_eval}

This section reviews common approaches to evaluate 6D object pose estimation, including the problem formulation and definition of functions measuring error of the pose estimates. %

\subsection{6D Object Pose Estimation Problems}

Methods for 6D object pose estimation usually report their predictions on the basis of two sources of information. Firstly, at training time, a method is given 3D object models and/or training images showing the objects in known poses. Secondly, at test time, the method is provided with an input image and possibly a list of object instances visible in the image. If the list is provided, the problem is referred to as 6D object localization and the evaluated method reports pose estimates of the listed instances.
Otherwise, the problem is referred to as 6D object detection and the method reports pose estimates of instances which the method detects. In both cases, the method also reports confidences of the estimates. The intrinsic camera parameters are assumed known.

In the 6D object localization problem, the performance of a method is measured by the recall rate, \ie, the fraction of annotated object instances for which a correct pose was estimated (commonly used criteria of pose correctness are described in Section~\ref{sec:rel_mesuring_error}). In the 6D object detection problem, the performance is typically measured by scores based on the precision and recall, such as the F1-score~\cite{tejani2014latent}.

The aspect which is evaluated on the 6D object detection but not on the 6D object localization problem is the capability of the method to calibrate the predicted confidence scores across all object models, \ie, whether the same score represents the same level of confidence no matter what object model the pose is estimated for.
This calibration is important for achieving a good performance \wrt to both the precision and recall.
The 6D object localization problem still requires the method to sort the pose estimates, although only within the set of pose estimates associated with the same object model
--~the method needs to output the top $n$ pose estimates for a given object model which are then evaluated against $n$ ground-truth poses associated with that model.

In the literature on 6D object pose estimation, methods are mostly evaluated on the 6D object localization problem, mainly because the accuracy scores on this simpler problem are still far from being saturated (the current state of the art is reported in Section~\ref{sec:bop_challenge_2020}).

\subsection{Measuring Pose Error} \label{sec:rel_mesuring_error}

The object pose is defined by a rigid transformation from the 3D coordinate system of the
model to the 3D coordinate system of the camera and represented by a $3\times4$ matrix $\textbf{P} = [\mathbf{R}\,|\, \mathbf{t}]$, where $\mathbf{R}$ is a $3\times3$ rotation matrix and $\mathbf{t}$ is a $3\times1$ translation vector.

Evaluating object pose estimates is not straightforward. This is because each object instance visible in an image is annotated with a single ground-truth pose in all commonly used datasets, but the object pose may be ambiguous and there may be (infinitely) many poses consistent with the image (Section~\ref{sec:rel_handling_ambiguity}).
The indistinguishable poses should be treated as equivalent, but explicitly enumerating all of them is difficult as one would need to identify the visible object parts in each image and find their fits to the object model.

In several papers, \eg, \cite{drost2010model,choi2012pose,brachmann2016uncertainty}, the error of an estimate $\hat{\mathbf{P}} = [\hat{\mathbf{R}}\,|\,\hat{\mathbf{t}}]$ \wrt the ground-truth $\bar{\mathbf{P}} = [\bar{\mathbf{R}}\,|\,\bar{\mathbf{t}}]$ is measured by the translational and rotational errors (the latter is defined by the angle from the axis-angle representation of the relative rotation~\cite{morawiec2004orientations}):
\begin{equation} \label{eq:trarot_dist}
	e_\mathrm{TE}\big(\hat{\mathbf{t}}, \bar{\mathbf{t}}\big) = \big\Vert\bar{\mathbf{t}} - \hat{\mathbf{t}}\big\Vert_2, \; \;
	e_\mathrm{RE}\big(\hat{\mathbf{R}}, \bar{\mathbf{R}}\big) = \arccos\big(\big(\text{Tr}\big(\hat{\mathbf{R}}\bar{\mathbf{R}}^{-1}\big) - 1\big) \, / \, 2\big)
\end{equation}

\noindent The estimated pose $\hat{\mathbf{P}}$ is considered correct if both $e_\mathrm{TE}$ and $e_\mathrm{RE}$ are below a respective threshold.
Choi and Christensen~\cite{choi2012pose} set the thresholds to $1\,$cm and $10^{\circ}$, Brachmann \etal~\cite{brachmann2016uncertainty} to $5\,$cm and $5^{\circ}$, and Drost \etal~\cite{drost2010model} to $0.1 d\,$cm and $12^{\circ}$, where $d$ is the object diameter, \ie, the largest distance between any pair of points on the model surface.

The rotational error could be extended to take into account the axes of
symmetry, which can be identified as
in Section~\ref{sec:symmetries}.
The downside of measuring the error directly in rotation is that even a small deviation in rotation can yield a large misalignment of the object surface in the case of elongated objects or objects with a complex surface. %
The rotational error is therefore less relevant for applications such as robotic grasping or augmented reality, where the surface alignment is the main indicator of the pose quality.

The most widely used pose-error function has been the
Average Distance (AD) by Hinterstoisser \etal~\cite{hinterstoisser2012accv}, defined as the average distance from vertices $V_M$ of the object model $M$ in the ground-truth pose $\bar{\mathbf{P}}$ to the vertices
in the estimated pose $\hat{\mathbf{P}}$. If all views at the object are distinguishable, \ie, the object looks different from every viewpoint, the distance is measured between the corresponding vertices (in homogeneous coordinates):
\begin{equation} \label{eq:ad_sym}
	e_\mathrm{ADD}\big(\hat{\mathbf{P}}, \bar{\mathbf{P}}, V_M\big) = \frac{1}{|V_M|}\sum_{\mathbf{x} \in V_M} \big\Vert\bar{\mathbf{P}} \mathbf{x} - \hat{\mathbf{P}} \mathbf{x}\big\Vert_2
\end{equation}

\noindent Otherwise, for an object with indistinguishable views (\ie, the object looks the same from two or more viewpoints), the distance is measured to the closest vertex, which may not necessarily be the corresponding vertex:
\begin{equation} \label{eq:ad_nonsym}
	e_\mathrm{ADI}\big(\hat{\mathbf{P}}, \bar{\mathbf{P}}, V_M\big) = \frac{1}{|V_M|}\sum_{\mathbf{x_1} \in V_M} \min\limits_{\mathbf{x}_2 \in V_M} \big\Vert\bar{\mathbf{P}} \mathbf{x}_1 - \hat{\mathbf{P}} \mathbf{x}_2\big\Vert_2
\end{equation}

\noindent The estimated pose~$\hat{\mathbf{P}}$ is usually considered correct if $e \leq 0.1 d$, where $e$ is $e_\mathrm{ADD}$ or $e_\mathrm{ADI}$, and $d$ is the object diameter defined as above.

The downside of ADD and ADI
is their high dependence on the geometry of the object model and on the sampling density of its surface -- the average distance tends to be~dominated by higher-frequency surface parts.
Besides, the existence of indistinguishable object views does not imply that the whole object 
is symmetric, as parts that break the symmetry may be occluded. ADI penalizes misalignment of such invisible parts, which may not be desirable for applications such as robotic manipulation with suction cups where only alignment of the visible part is relevant.
For example, if (c) in Figure~\ref{fig:pose_ambiguity} was the input image showing a cup, and (a) and (b) were the top views at the cup in the ground-truth and the estimated pose, ADI would yield a non-zero error because the invisible handle has a different position in the two poses.
Another example of invisible parts whose misalignment would be penalized are inner parts of the object models, which are often included in CAD models used in industry.
Moreover, the ADI variant may yield unintuitively low errors for poses that are clearly distinguishable due to a many-to-one vertex matching that may be established by the search for the closest vertex, and due to the possibility of matching the outer and the inner model parts -- see Section~\ref{sec:bop_vsd} for examples.

In Section~\ref{sec:bop_methodology}, we propose a new methodology which is used in BOP and includes three pose-error functions that address the downsides of the functions reviewed above.

\chapter[HashMatch:\ Hashing for Efficient Template Matching]{HashMatch\\ {\Large Hashing for Efficient Template Matching}}{} \label{ch:method_template}

The method presented in this chapter, referred to as HashMatch, aims at accurate 6D pose estimation of texture-less objects with available 3D models from a single RGB-D image.
The method slides a window of a fixed size over the input image in multiple scales and searches for a match against a set of object templates.
The templates are pre-generated by rendering isolated 3D object models in different orientations and annotated with 6D object poses.
Instead of matching each window location to all templates, HashMatch applies an efficient cascade of evaluation stages to each window location, which makes the computational complexity of the method sub-linear in the number of templates.

The evaluation cascade starts with fast filtering that rapidly rejects most window locations by a simple objectness check based on the number of discontinuities in the depth image channel. %
For each remaining window location, a set of candidate templates is retrieved by an efficient voting procedure based on hashing of depth measurements and surface normals, which are sampled on a regular grid.
The candidate templates are then verified by a sequence of tests evaluating the consistency of color, depth, image gradients, and surface normals. Finally, the approximate 6D object pose associated with each detected template is used as a starting point for a stochastic, population-based optimization procedure that refines the pose by fitting the 3D object model to the depth image channel. %

\begin{figure}[t!]
	\begingroup
	\renewcommand{\arraystretch}{0.9}
	
	\begin{center}
	\includegraphics[width=\textwidth]{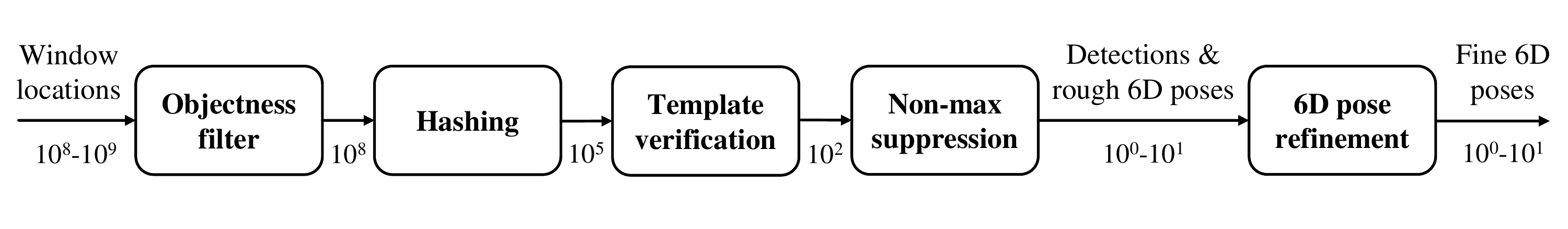}
	\caption{\label{fig:iros15_pipeline} \textbf{HashMatch pipeline.} The method applies an %
	cascade-style evaluation to each sliding window location. Note the typical numbers of detection candidates, each defined by a template identifier and a window location, advancing through individual stages (for a $5\,$px sliding step, $108$$\times$$108\,$px templates, a VGA input image, and an image pyramid with four larger and four smaller scales with a scaling factor of $1.2$). %
	The hashing stage efficiently reduces the number of candidates usually by three orders of magnitude before the more expensive	verification stage.
	}
	\end{center}
	
	\begin{center}
	\footnotesize
	\begin{tabular}{ @{}c@{ } @{}c@{ } @{}c@{ } @{}c@{ } }
		Input RGB-D image & All windows ($4 \cdot 10^8$) & Objectness ($1.7 \cdot 10^8$) & Hashing ($5.2 \cdot 10^5$) \vspace{0.5ex} \\
		\includegraphics[width=0.243\linewidth]{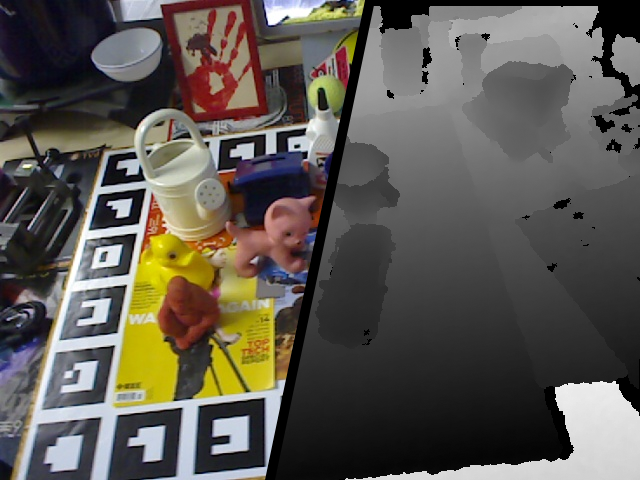} &
		\includegraphics[width=0.243\linewidth]{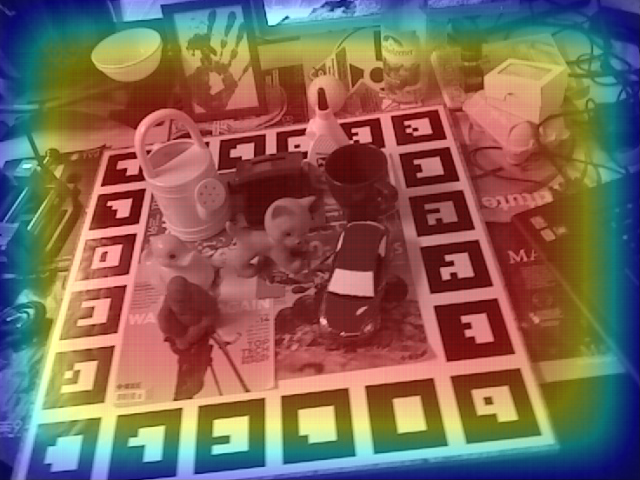} &
		\includegraphics[width=0.243\linewidth]{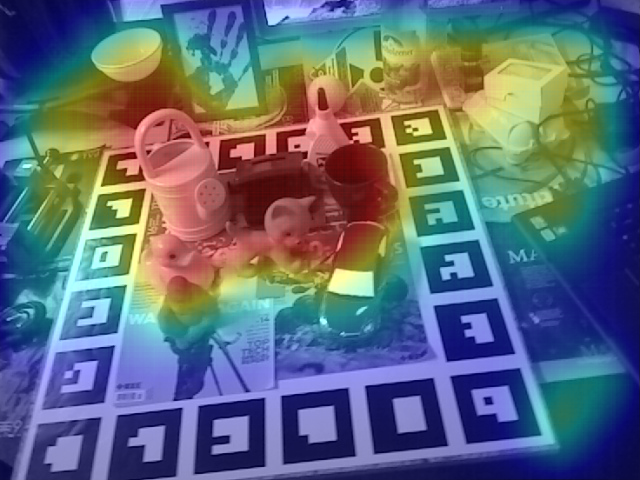} &
		\includegraphics[width=0.243\linewidth]{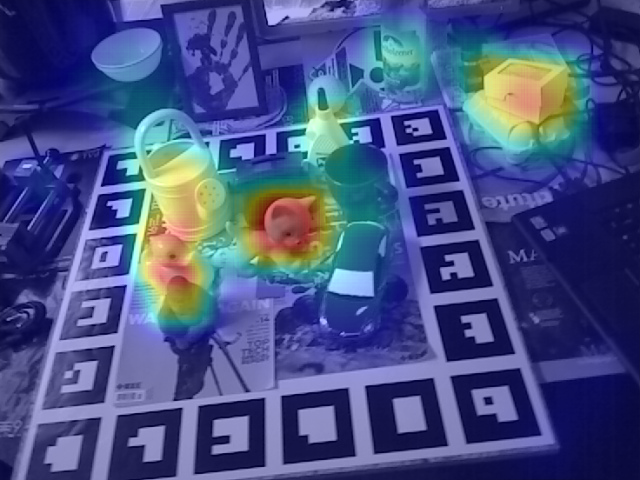} \\
		
		\multicolumn{4}{c}{}\vspace{-2.0ex} \\
		Verification ($44$) & Non-max supp. ($1$) & Rough 6D pose & Refined 6D pose \vspace{0.5ex} \\
		\includegraphics[width=0.243\linewidth]{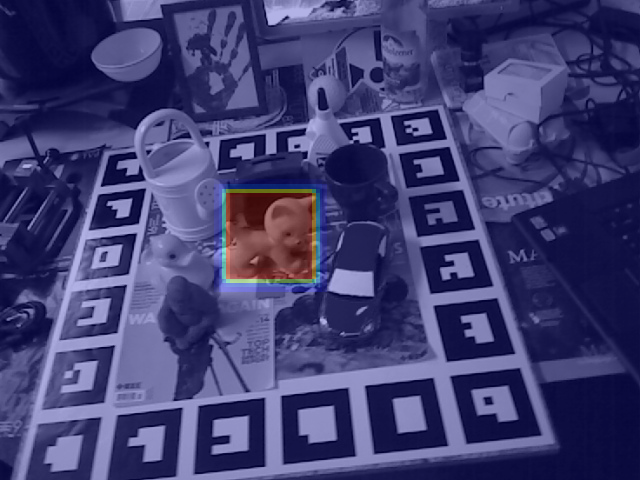} &
		\includegraphics[width=0.243\linewidth]{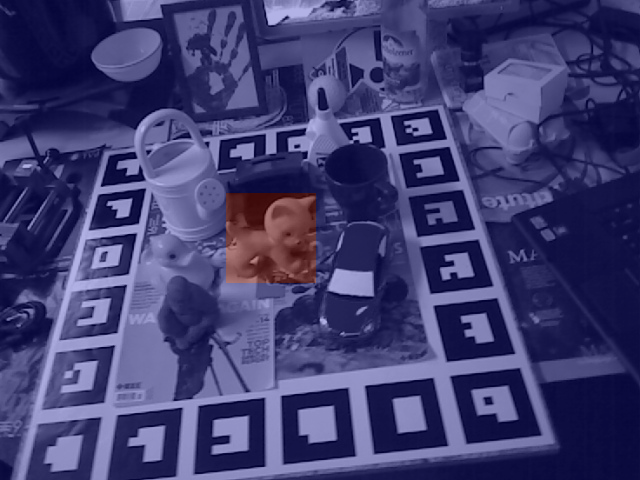} &
		\includegraphics[width=0.243\linewidth]{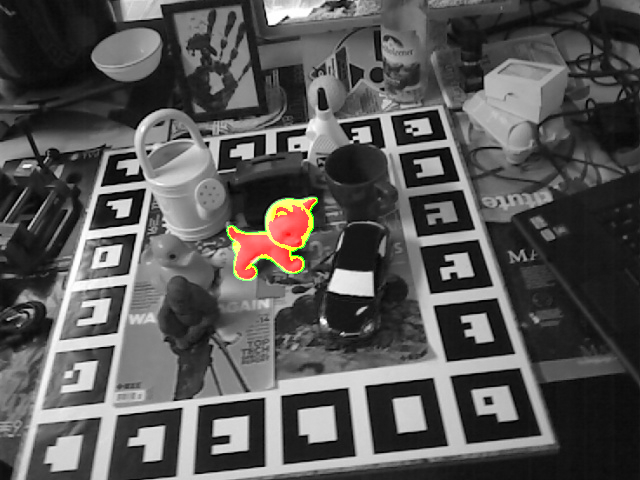} &
		\includegraphics[width=0.243\linewidth]{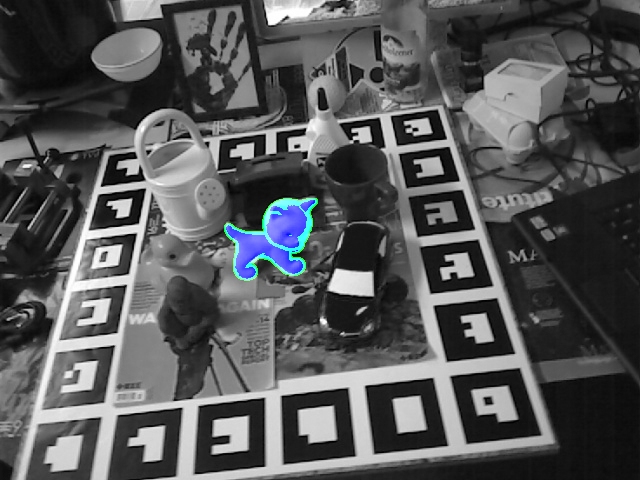} \\
	\end{tabular}
	\caption{\label{fig:iros15_pipeline_example} \textbf{6D localization of a cat in a sample RGB-D image from~\cite{hinterstoisser2012accv}.} The numbers of detection candidates advancing through the pipeline are in brackets. The heatmaps show the density -- red indicates a high and blue indicates a low number of candidates overlapping a pixel.}
	\end{center}

	\endgroup
\end{figure}

The pipeline of HashMatch is illustrated in Figures~\ref{fig:iros15_pipeline} and \ref{fig:iros15_pipeline_example} together with the typical numbers of detection candidates advancing through individual stages. The key is the voting stage based on hashing, which substantially reduces the number of candidates, usually by three orders of magnitude, before the more expensive template verification stage.
The object detection part of the method is detailed in Section~\ref{sec:iros15_detection} and the pose refinement part in Section~\ref{sec:iros15_refinement}. An experimental evaluation is provided in Section~\ref{sec:iros15_experiments}, demonstrating the proposed method to achieve accuracy comparable to the state of the art, while improving the computational complexity \wrt the number of templates.

The HashMatch method was published in \cite{hodan2015detection}, used to report baseline results on T-LESS (Chapter~\ref{ch:tless}), placed fourth out of the 15 participants in the BOP Challenge 2017 (Section~\ref{sec:bop_challenge_2017}), and was successfully deployed
in a robotic assembly project (Section~\ref{sec:iros15_app}).

\section{Object Detection by Template Matching} \label{sec:iros15_detection}

\begin{figure}[t!]
	\begingroup
	\renewcommand{\arraystretch}{0.9}

	\begin{center}
		\footnotesize
		\begin{tabular}{ @{}c@{ } @{}c@{ } @{}c@{ } @{}c@{ } @{}c@{ } @{}c@{ } }
			\includegraphics[width=0.16\linewidth]{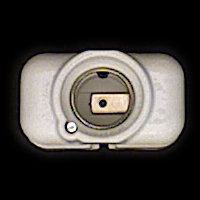} &
			\includegraphics[width=0.16\linewidth]{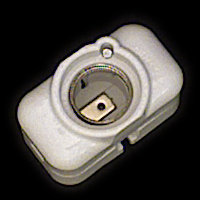} &
			\includegraphics[width=0.16\linewidth]{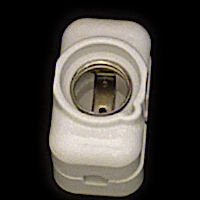} &
			\includegraphics[width=0.16\linewidth]{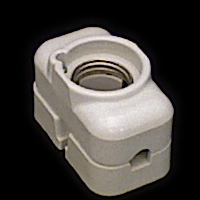} &
			\includegraphics[width=0.16\linewidth]{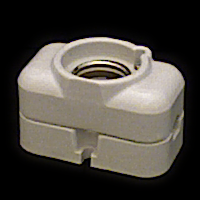} &
			\includegraphics[width=0.16\linewidth]{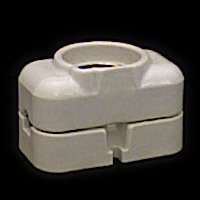} \\
			
			\includegraphics[width=0.16\linewidth]{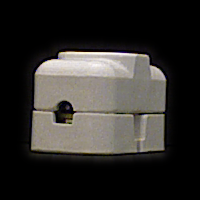} &
			\includegraphics[width=0.16\linewidth]{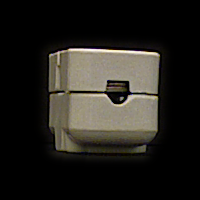} &
			\includegraphics[width=0.16\linewidth]{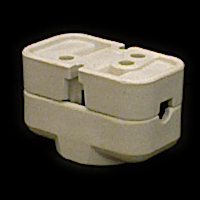} &
			\includegraphics[width=0.16\linewidth]{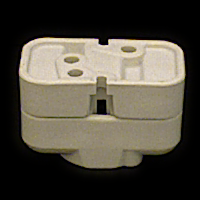} &
			\includegraphics[width=0.16\linewidth]{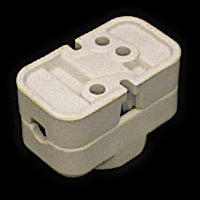} &
			\includegraphics[width=0.16\linewidth]{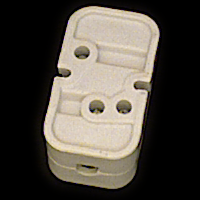} \\
		\end{tabular}
		\caption{\label{fig:iros15_templates} \textbf{RGB channels of sample templates from~\cite{hodan2017tless}.} Each object is represented by multiple (usually several thousands) templates showing the object under different 3D orientations (\ie, under different azimuth, elevation and in-plane rotation angles).
		}
	\end{center}
	
	\endgroup
\end{figure}

Detection of objects in the input RGB-D image is based on a sliding window approach operating on an image pyramid constructed from the input image. Let $W$ denote a set of all tested window locations. Each window location is defined by a triplet $w=(u_w,v_w,s_w)$, where $(u_w,v_w)$ are the 2D image coordinates of the center of the window and $s_w$ is the image scale. The window is a square with its size (in pixels) fixed across all image scales. The number of tested window locations $|W|$ is a function of the image resolution, sliding step,
size of the sliding window, scale range (\eg, two or four octaves),
and scale-space discretization. The objects are represented with a set $T$ of RGB-D templates which are of the same fixed size as the sliding window.
For each object, the set $T$ typically contains several thousands of templates showing the object at the same distance but in different 3D orientations (Figure~\ref{fig:iros15_templates}).
The distance is defined between the camera center and the object centroid\footnote{The object centroid is defined by the center of the 3D bounding box of the object model and is aligned with the origin of the 3D coordinate system of the object model.}, and chosen so that the object silhouette from all viewing angles fits the template.
A template $t \in T$ is associated with the object identifier $o_t$ and the object pose given by a
rotation matrix ${\bf R}_t$ and a
translation vector $\mathbf{t}_t$. Templates can be obtained
by rendering the 3D
models or by capturing the objects using a specialized
setup~\cite{hodan2017tless}.

In general, each window location $w \in W$ needs to be compared against all templates, with the asymptotic computational complexity being $O(|W||T|)$. This is
expensive even for a moderate number of objects. We therefore propose an evaluation cascade where the
locations $W$ are quickly reduced (Section~\ref{sec:iros15_pre-filtering}) and the candidate templates~$T$ are substantially pruned (Section~\ref{sec:iros15_hypothesis_generation}) before the template matching itself (Section~\ref{sec:iros15_hypothesis_verification}).

\subsection{Objectness Filter}
\label{sec:iros15_pre-filtering}

Each window location $w$ is first assessed with a simple \emph{objectness} measure, \ie, the likelihood that the location contains an object~\cite{alexe2012measuring,cheng2014bing}.
This is a binary classification problem of labeling $w$ as a background or a potential foreground, with the foreground defined by~$T$.

The proposed objectness measure is based on the number of depth-discontinuity edgels inside the window and is computed via an integral image for efficiency. Depth-discontinuity edgels arise at pixels where the response of the Sobel operator computed over the depth channel is larger than a threshold $\tau$, which is in the same units as the depth channel. A window location is classified as a potential foreground if the number of depth edgels is larger than a threshold $\theta$.
In the experiments presented in Section~\ref{sec:iros15_experiments}, the threshold $\tau$ is set to $30\%$ of the diameter\footnote{The object diameter is defined as the largest distance between any pair of points on the model surface.} of the smallest object in the database,
and
$\theta$ is set to $30\%$ of the minimum number of depth edgels
present in any template.
This setting is tolerant to partial occlusions but still strong enough to prune most of the window locations -- depending on the scene clutter, roughly $60\%$ to $90\%$ window locations are pruned in images of the considered robot workspace depicted in Figure~\ref{fig:iros15_app}. Only window locations classified as a potential foreground are processed further.

\subsection{Hypothesis Generation by Hashing-Based Voting} \label{sec:iros15_hypothesis_generation}

In this stage, a small subset of potentially matching templates is quickly identified for each window location $w$ that passed the objectness filter.
The set of candidate templates is found using multiple hash tables $H$, which is an operation with a constant computational complexity in the number of stored templates $|T|$. Each hash table $h \in H$ is indexed by a hash key obtained by discretizing a feature vector $F_h$, which is extracted from a window location~$w$ or a template~$t$.
Multiple hash tables are constructed to increase the robustness, each table with a different definition of the associated feature vector $F_h$. %
At training time, the hash tables are filled with identifiers of templates, with possibly multiple but typically a low number of identifiers stored at each hash key.
At test time, the hash tables are used to vote for the potentially matching templates.
A template $t$ can receive up to $|H|$ votes, in which case all of the feature vectors from $t$ are discretized into the same hash keys as the feature vectors from $w$. Up to $n$ templates with the highest number of votes, which received at least $m$ votes, are passed onward to the next stage of the cascade ($n=100$ and $m=3$ in the experiments presented in Section~\ref{sec:iros15_experiments}).

\begin{figure}[t!]
	\begingroup
	\renewcommand{\arraystretch}{0.9}
	
	\begin{center}
		\begin{tabular}{ @{}c@{ } @{}c@{ } }
			\includegraphics[width=0.243\linewidth]{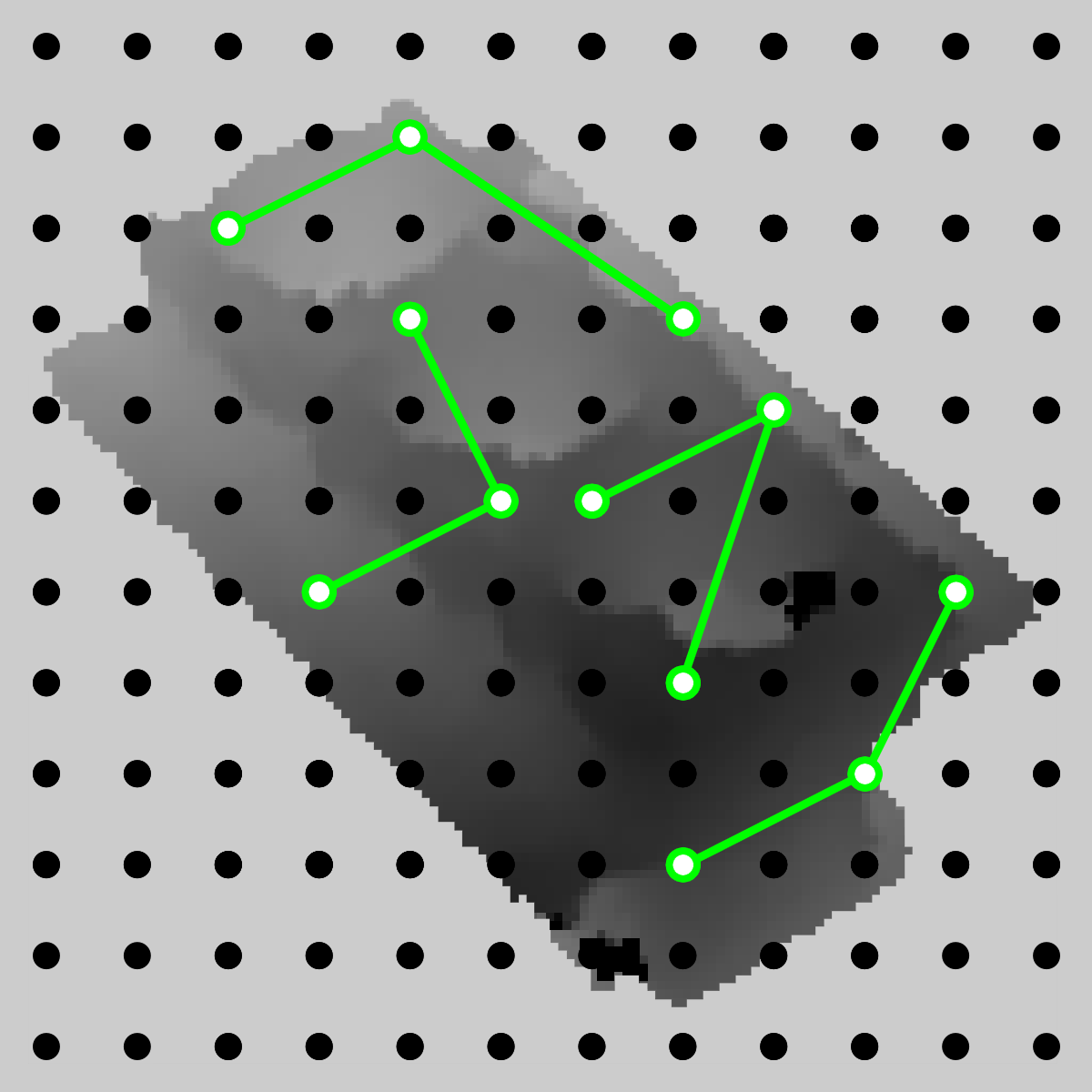} &
			\includegraphics[width=0.243\linewidth]{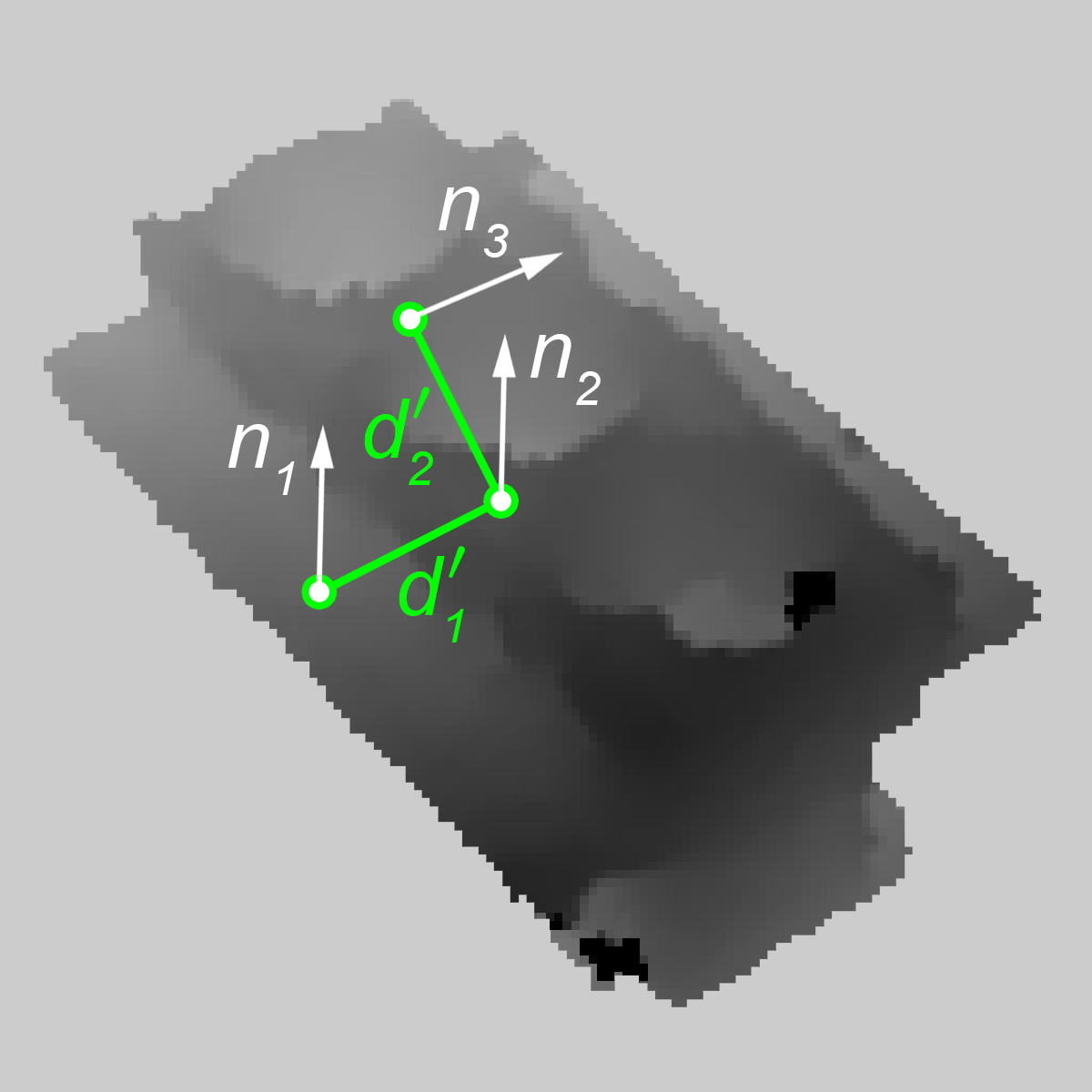}
		\end{tabular}
		\caption{\label{fig:iros15_hashing}
		\textbf{Feature extraction.}
		A grid of reference points is placed over the templates at training time and the sliding window at test time.
		A feature vector $F_h$ is defined by depth differences $\{d'_1, d'_2\}$ and normal vectors $\{{\bf n}_1, {\bf n}_2, {\bf n}_3\}$ measured at three reference points associated with the hash table~$h$.
		The feature vector is discretized and used to index the hash table $h$ -- to insert a template at training time and retrieve potentially matching templates at test time. The feature vector is valid only when the depth value is available at all three reference points.
		}
	\end{center}

	\begin{center}
		\begin{tabular}{ @{}c@{ } @{}c@{ } @{}c@{ } @{}c@{ } }
			\includegraphics[width=0.243\linewidth]{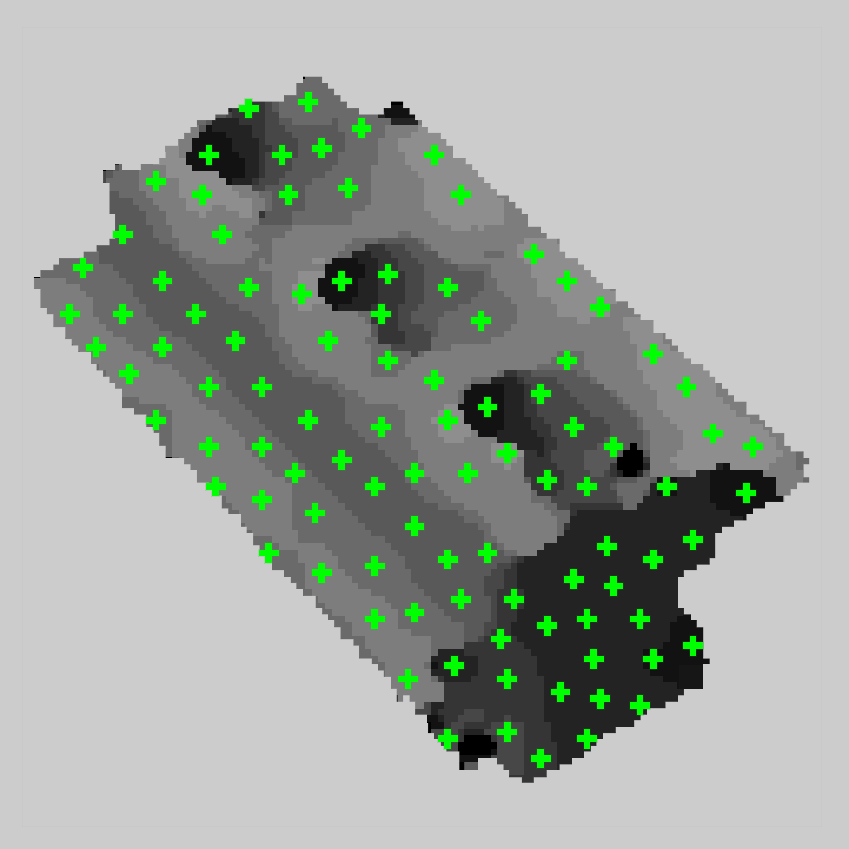} &
			\includegraphics[width=0.243\linewidth]{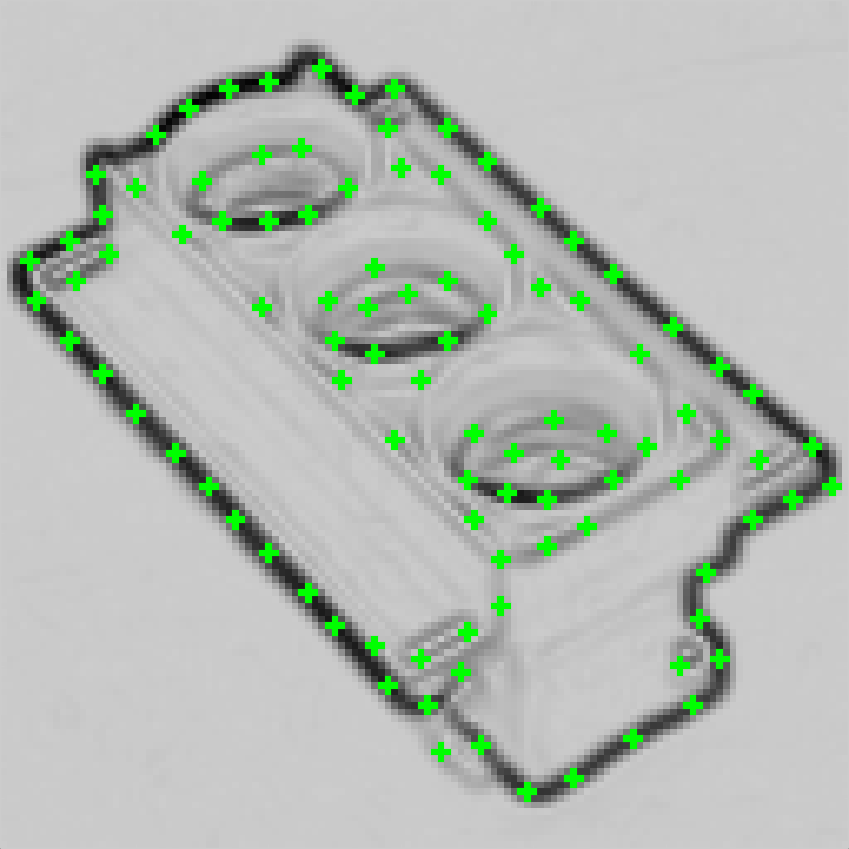} &
			\includegraphics[width=0.243\linewidth]{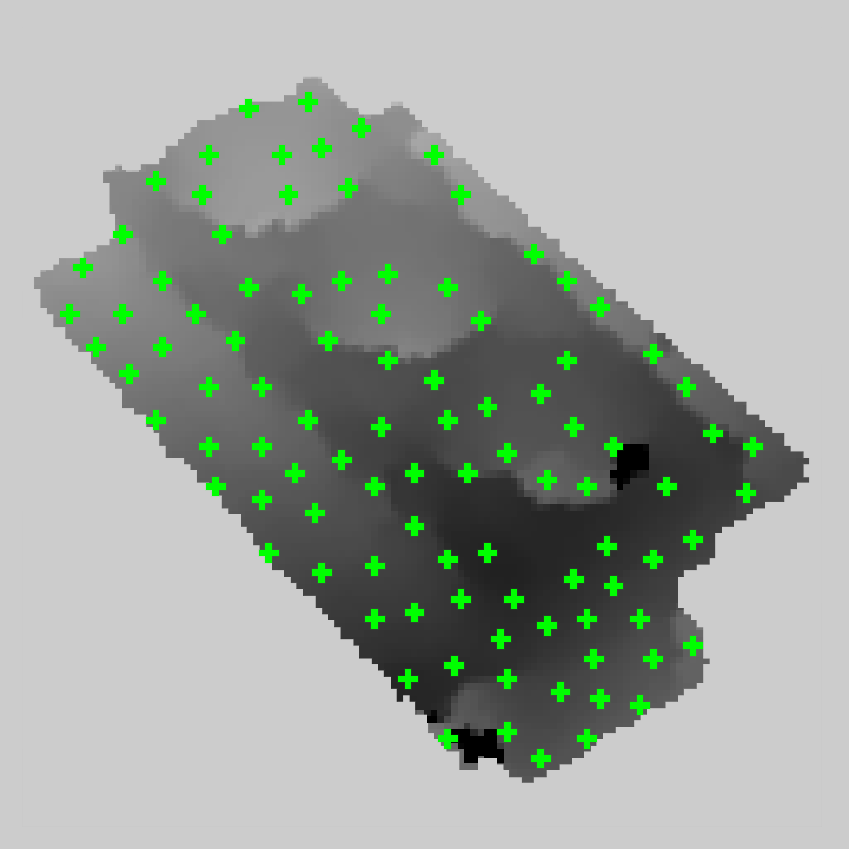} &
			\includegraphics[width=0.243\linewidth]{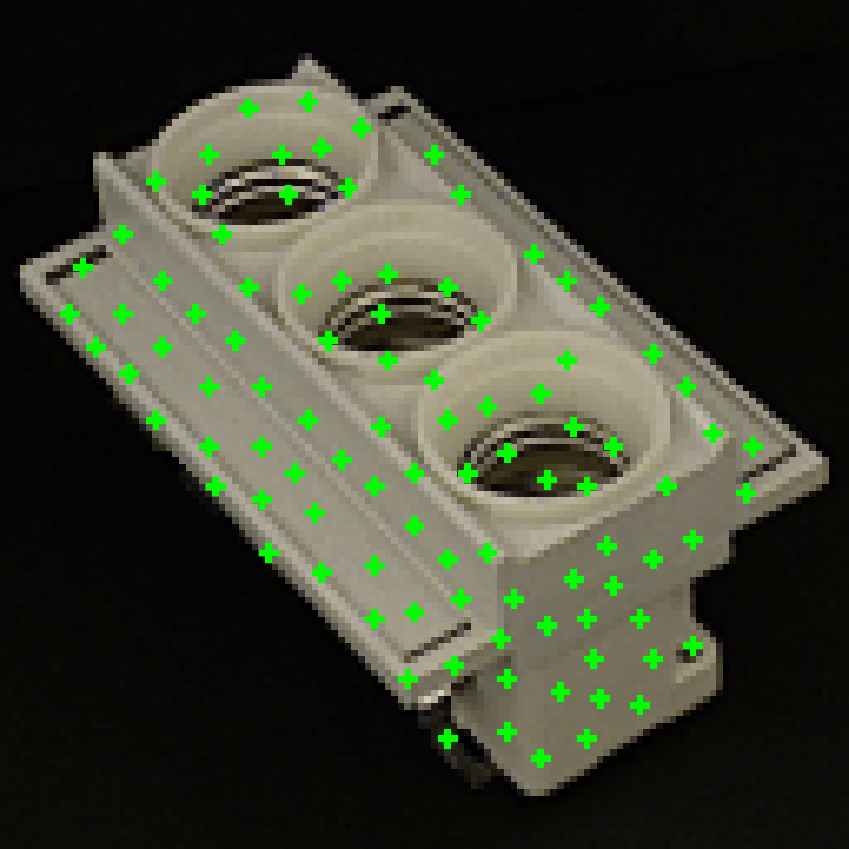}
		\end{tabular}
		\caption{\label{fig:iros15_ver_tests} \textbf{Multimodal hypothesis verification.} The consistency between a sliding window location and a template is evaluated in different modalities at pixels that are selected independently for each template. Left to right: surface normals, image gradients, depth, color.}
	\end{center}

	\endgroup
\end{figure}

\customparagraph{Feature Extraction and Quantization.}
A feature vector $F_h$ associated with a hash table $h$ is defined on a regular grid of $r \times r$ reference points placed over the templates at training time and over the sliding window at test time.
The vector consists of $k-1$ depth differences and $k$ 3D normal vectors measured at $k$ selected reference points (Figure~\ref{fig:iros15_hashing}): $F_h = (d_{2} - d_{1}, d_{3} - d_{1}, \dots, d_{k} - d_{1}, {\bf n}_{1}, \dots, {\bf n}_{k})$.
The depth differences $d_{i} - d_{1}$ are quantized into $b_{d}$ bins, with the quantization boundaries learned from all templates to achieve equal frequency binning, \ie, each bin contains approximately the same number of examples. The normal vectors are quantized as in~\cite{hinterstoisser2012pami} into $b_{n}$ bins according to their orientation.
In the presented experiments, we set $r=12$, $b_d=5$, $b_n=8$, and select $k=3$ reference points as described below.
This setting yields a hash table with $5^2 8^3=12800$ indices.

\customparagraph{Selection of Reference Points.}
To increase the robustness against occlusion and noise, multiple hash tables are constructed and the feature vector $F_h$ is measured at different reference points for each table $h$.
The $k$ reference points associated with a table can be selected randomly. Alternatively,
the selection can be optimized to fill the tables uniformly (for stable detection time) and to cover maximally independent measurements (for robustness). Since the optimal selection is NP-complete, a hybrid
strategy is employed -- a set of $q$, where $q \gg |H|$,
$k$-tuples is generated and the subset with the largest joint entropy of the quantized measurements is retained ($q = 5000$ and $|H| = 100$ in the experiments).

\pagebreak

\customparagraph{}The hashing-based voting procedure is inspired by Cai \etal~\cite{cai2013fast}, who assume a grayscale input image and hash
distances and orientations of
intensity edges instead of depth and surface normals, and similar to the concurrent work of Kehl \etal~\cite{kehl2015hashmod}, who assume an RGB-D image and hash
orientations of image gradients and surface normals.

\subsection{Multimodal Hypothesis Verification} \label{sec:iros15_hypothesis_verification}

Thanks to the preceding hypothesis generation stage, only up to $n$ templates are considered for each window location $w$ that passed the initial objectness filter, which makes the complexity of the verification stage constant in the number of stored templates $|T|$.
The verification proceeds in a sequence of tests evaluating the object size ($\text{T}_1$), surface normals ($\text{T}_2$), image gradients ($\text{T}_3$), surface distance ($\text{T}_4$), and surface color ($\text{T}_5$).
The tests are ordered by
increasing computational cost. If a test fails, the
next ones
are skipped.

Test $\text{T}_1$ verifies that the scale of the detected object $o_t$
(\ie, the level of the image pyramid at which the object is detected)
is in accordance with the depth image channel. The surface of the object $o_t$ as seen in the template $t$ detected at the window location $w$ is expected to be approximately at distance $e = s_w \lVert \mathbf{t}_t \rVert_2 - 0.5d_t$, where $s_w$ is
the image scale associated with the window $w$,
$\lVert \mathbf{t}_t \rVert_2$ is the Euclidean distance between the camera center and the object centroid, and $d_t$ is the object diameter.
The expected distance $e$ is compared with distances calculated at five reference points -- the center of the sliding window and the centers of its four quadrants.\footnote{\label{note:distance}The distance from the camera center to a 3D point $\mathbf{x}_\mathbf{u}$ that projects to the pixel $\mathbf{u}$ can be calculated from the intrinsics and the value at $\mathbf{u}$ in the depth channel, which represents the $Z$ coordinate of $\mathbf{x}_\mathbf{u}$.} The template $t$ passes test $\text{T}_1$ if the distance at one or more of the five reference points is between $e/\sqrt{s}$ and $e\sqrt{s}$, where $s$ is the scale factor between two consecutive levels of the image pyramid.

Tests $\text{T}_2$ and $\text{T}_3$ verify the orientation of the surface normals and of the intensity gradients at $p$ selected pixels ($p=100$ in our experiments; Figure~\ref{fig:iros15_ver_tests}). Similarly to~\cite{hinterstoisser2012accv}, the pixels are selected independently for each template at training time.
Pixels for test $\text{T}_2$ are selected at locations with locally stable orientation of surface normals.
Pixels for test $\text{T}_3$ are selected at locations with large gradient magnitude (\ie, typically on the contour in the case of texture-less objects). The orientation of surface normals and intensity gradients is quantized and compared between the template and the window, which can be done efficiently by bitwise operations using the response maps described in~\cite{hinterstoisser2011linemod}.

The distance test $\text{T}_4$ and the color test $\text{T}_5$ reuse pixels selected for the surface normal test $\text{T}_2$. In test $\text{T}_4$, a difference $\delta$ between the distance measured in the template and the distance measured in the sliding window is calculated at each tested pixel.\footnoteref{note:distance} A tested pixel is matched if $|\delta - \text{med}_\delta| < \alpha d_t$, where $\text{med}_\delta$ is the median value of differences $\delta$ from all tested pixels, $d_t$ is the object diameter, and $\alpha$ is a coefficient set to $0.05$ in our experiments. Finally, pixel colors in test $\text{T}_5$ are compared in the HSV space as in~\cite{hinterstoisser2012accv}.

A template passes tests $\text{T}_2$ to $\text{T}_5$ if at least $v$ pixels in each test have a matching value within a small neighborhood of $g \times g$ pixels around the corresponding location in the sliding window. The matching value is searched in the neighborhood to compensate for the sliding window step and for the discretization of orientations during training. In our experiments, $v$ was set to $0.6p$ to leave a tolerance for partial occlusions, and the extent of the neighborhood $g$ was set to $5\,$px.

A template $t$ that passes all the tests at a
location $w$ is assigned a final score computed as $c_{t,w} = \sum_{i \in \{2 \dots 5\}} c_{t,w,i}$, where $c_{t,w,i}$ is the fraction of matching feature points in test $\text{T}_i$.

\subsection{Non-Maxima Suppression} \label{sec:iros15_nonmaxima_suppression}

The verified templates are collected from all different locations and scales. Since different views of an object are often alike, and since multiple objects may be similar, there may be multiple overlapping detections. Unique detections are identified by repeatedly retaining the candidate with the highest score $r$, and removing all detections that have a large overlap with it. The score is defined as $r = c (a / s)$, where $c$ is the verification score defined above, $s$ is the detection scale, and $a$ is the area of the object silhouette in the template. Weighting the score by the silhouette area favors detections which explain more of the scene (\eg, when a cup is seen from a side, with the handle visible, a template depicting the handle is preferred over other templates which may have the same score $c$ but do not show the handle). The retained detections are passed to the pose refinement stage, together with the approximate 6D poses associated with the templates.

\section{Refinement by Particle Swarm Optimization}
\label{sec:iros15_refinement}

For a template $t$ detected at a window location $w$, the pose refinement stage receives as input a 3D model of the object $o_t$, the object pose $\mathbf{P}_{t} = [\mathbf{R}_{t} \,|\, \mathbf{t}_{t}]$, the depth image channel, and the camera intrinsic parameters. %

The pose $\mathbf{P}_{t}$ associated with the template $t$ is valid when the center of $t$ is aligned with the principal point of the image in the original scale. To reflect the 2D offset $(u_w, v_w)$ and the scale $s_w$ of the window location $w$, at which $t$ was detected, $\mathbf{P}_{t}$ is transformed to: $\mathbf{P}_{t,w} = [\mathbf{R}_a\mathbf{R}_{t} \,|\, \mathbf{R}_a\mathbf{t}_{t}/s_w]$, where the rotation matrix $\mathbf{R}_a$ aligns the optical axis to a vector that originates at the optical center and passes through the pixel $(u_w, v_w)$. %

The pose $\mathbf{P}_{t,w}$ is refined by the Particle Swarm Optimization (PSO)~\cite{poli2007swarm}, which stochastically evolves a population of candidate poses over multiple iterations. A candidate pose is evaluated by rendering the depth image of the 3D object model in that pose and comparing the rendering with the input depth image. As shown in~\cite{zabulis2015object}, PSO is less prone to local minima compared to the commonly used Iterative Closest Point (ICP) algorithm~\cite{besl1992method}. A detailed description of the pose refinement procedure is in~\cite{zabulis2015object,hodan2015detection}.

\section{Experimental Evaluation} \label{sec:iros15_experiments}

This section compares the performance of HashMatch with other methods for 6D object localization (Section~\ref{sec:iros15_localization}) and describes a robotic assembly project in which the method was deployed (Section~\ref{sec:iros15_app}). Results of HashMatch on more datasets are in Section~\ref{sec:bop_challenge_2017}.

\subsection{6D Object Localization} \label{sec:iros15_localization}

\customparagraph{Evaluation Methodology.}
Given a single RGB-D test image and an object identifier, the task is to estimate the 6D pose of the object.
The estimate with the highest confidence $c_{t,w}$ is selected and evaluated in each image.
The estimates are evaluated as in~\cite{hinterstoisser2012accv} using the Average Distance (AD) pose-error function defined in Section~\ref{sec:rel_eval} -- the ADI variant is used to evaluate pose estimates of objects \#3, \#7, \#10, and \#11, and the ADD variant for the other objects. A pose estimate is considered correct if the AD error is below $10\%$ of the object diameter. The accuracy of a method is measured by the recall rate, \ie, the fraction of annotated object instances for which a correct pose is estimated.

\customparagraph{Dataset.}
HashMatch is evaluated on the Linemod (LM) dataset~\cite{hinterstoisser2012accv}, which includes 3D models of 15 texture-less objects and, for each object, a test set consisting of approximately 1200 RGB-D images in VGA resolution. Each image shows one annotated object instance under mild occlusion, heavy 2D and 3D clutter, and large viewpoint variation.

\customparagraph{Compared Methods.}
The proposed method is compared with the Linemod~\cite{hinterstoisser2011linemod} and Linemod++~\cite{hinterstoisser2012pami} methods by Hinterstoisser \etal, and with the methods by Drost \etal~\cite{drost2010model} and Kehl \etal~\cite{kehl2015hashmod}. Linemod follows an exhaustive template matching approach. Linemod++ extends it by two verification steps -- a color check and a depth check by a rough but fast ICP.
A finer ICP is then applied to the best of the remaining pose hypotheses. The essential difference of HashMatch is the addition of the objectness filter and the hypothesis generation stage which avoids the exhaustive search. The Hashmod method by Kehl \etal~\cite{kehl2015hashmod} is similar to HashMatch (they hash orientations of image gradients and surface normals, instead of depth measurements and orientations of surface normals). The method of Drost \etal~\cite{drost2010model} takes a different approach based on matching pairs of oriented 3D points between the point cloud of the test scene and the 3D object model, and aggregating the matches via a voting scheme.

\customparagraph{Templates and Parameters.}
The set of templates $T$ is obtained by rendering the 3D object models from a uniformly sampled upper half of a hemisphere centered at the object centroid (with a step of $10^{\circ}$ in both azimuth and elevation). To achieve invariance to rotation around the optical axis, an in-plane rotation is additionally applied to each template (from $-40^{\circ}$ to $40^{\circ}$ with a step of $10^{\circ}$). In total, an object is represented by 2916 templates of size $108 \times 108\,$px.
Each test image is scanned at $9$ scales ($4$ larger and $4$ smaller scales with a scaling factor of $1.2$) with a scanning step of $5\,$px.

\begin{figure}[t!]
	\begin{center}
		\begingroup
		\renewcommand{\arraystretch}{0.9} %
		\begin{tabular}{ @{}c@{ } @{}c@{ } @{}c@{ } @{}c@{ } @{}c@{ } }
			\includegraphics[width=0.15\linewidth]{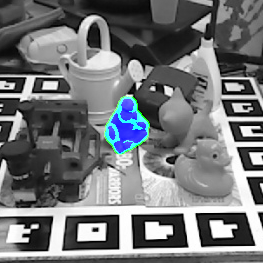} &
			\includegraphics[width=0.15\linewidth]{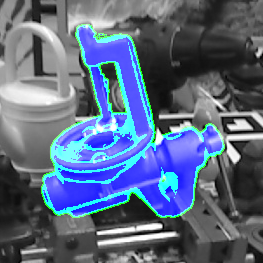} &
			\includegraphics[width=0.15\linewidth]{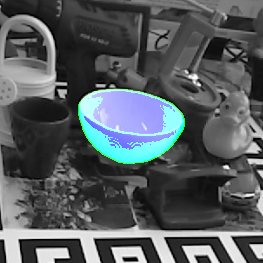} &
			\includegraphics[width=0.15\linewidth]{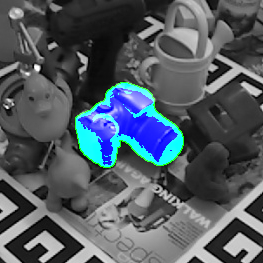} &
			\includegraphics[width=0.15\linewidth]{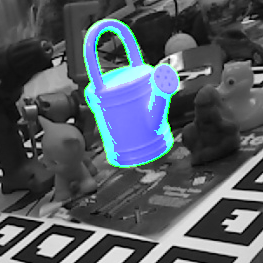} \\
			\includegraphics[width=0.15\linewidth]{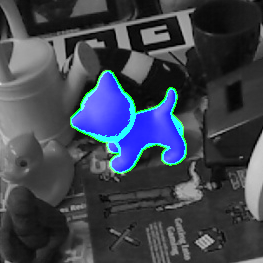} &
			\includegraphics[width=0.15\linewidth]{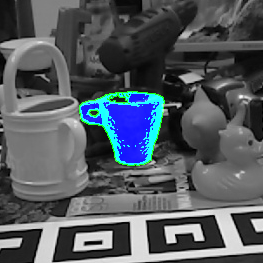} &
			\includegraphics[width=0.15\linewidth]{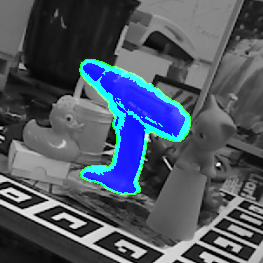} &
			\includegraphics[width=0.15\linewidth]{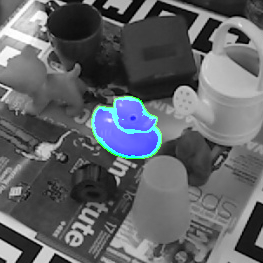} &
			\includegraphics[width=0.15\linewidth]{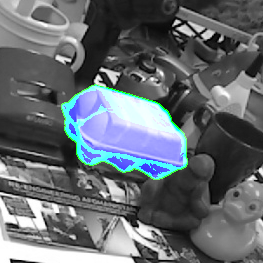} \\
			\includegraphics[width=0.15\linewidth]{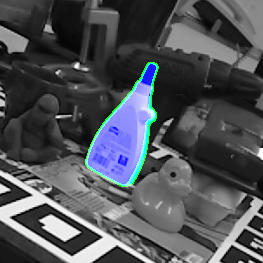} &
			\includegraphics[width=0.15\linewidth]{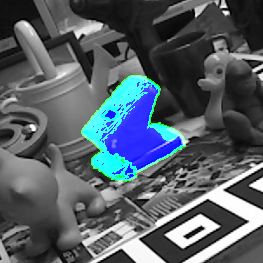} &
			\includegraphics[width=0.15\linewidth]{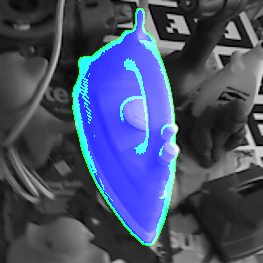} &
			\includegraphics[width=0.15\linewidth]{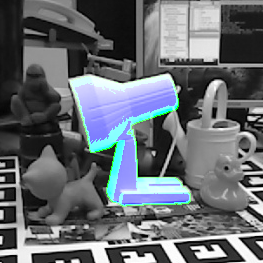} &
			\includegraphics[width=0.15\linewidth]{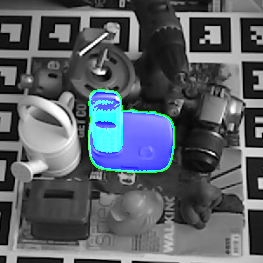}
		\end{tabular}
		\endgroup
		\caption{\label{fig:iros15_res_hinter} \textbf{HashMatch results on the LM dataset~\cite{hinterstoisser2012accv}.} Shown are sample test images which were darkened, cropped and overlaid with 3D object models in the estimated 6D poses.}
	\end{center}

\vspace{-2.0ex}

	\begin{center}
		\centering
		\begingroup
		\footnotesize
		\setlength{\tabcolsep}{2pt}	
		\begin{tabularx}{1.0\textwidth}{l *{16}{Y}}
			\toprule
			Method & 1 & 2 & 3 & 4 & 5 & 6 & 7 & 8 & 9 & 10 & 11 & 12 & 13 & 14 & 15 & Avg. \\
			\midrule
			\Rowcolor{lightgray} HashMatch &
			93.9 & \textbf{99.8} & 98.8 & 95.5 & \textbf{95.9} & 98.2 & \textbf{99.5} & 94.1 & 94.3 & \textbf{100} & \textbf{98.0} & 88.0 & 97.0 & 88.8 & 89.4 & 95.4 \\
			
			Hashmod &
			\textbf{96.1} & 92.8 & 99.3 & \textbf{97.8} & 92.8 & 98.9 & 96.2 & \textbf{98.2} & 94.1 & 99.9 & 96.8 & 95.7 & 96.5 & 98.4 & \textbf{93.3} & 96.5 \\
			
			LM++ &
			95.8 & 98.7 & \textbf{99.9} & 97.5 & 95.4 & \textbf{99.3} & 97.1 & 93.6 & \textbf{95.9} & 99.8 & 91.8 & \textbf{95.9} & \textbf{97.5} & \textbf{97.7} & \textbf{93.3} & \textbf{96.6} \\
			
			LM &
			69.4 & 94.0 & 99.5 & 79.5 & 79.5 & 88.2 & 80.7 & 81.3 & 75.9 & 99.1 & 64.3 & 78.4 & 88.8 & 89.8 & 77.8 & 83.0 \\
			
			Drost &
			86.5 & 70.7 & 95.7 & 78.6 & 80.2 & 85.4 & 68.4 & 87.3 & 46.0 & 97.0 & 57.2 & 77.4 & 84.9 & 93.3 & 80.7 & 79.3 \\
			\bottomrule
		\end{tabularx}
		\captionof{table}{
			\label{tab:iros15_rec_rates} \textbf{Recall rates [\%] on the LM dataset~\cite{hinterstoisser2012accv}.} A pose estimate is considered correct if the AD error is below $10\%$ of the object diameter. The object ID's are as in~\cite{hodan2018bop}.
		}
		\endgroup
	\end{center}

\vspace{-3.5ex}

	\begin{center}
		\begingroup
		\renewcommand{\arraystretch}{0.9} %
		\begin{tabular}{ @{}c@{ } @{}c@{ } }
			\includegraphics[width=0.48\linewidth]{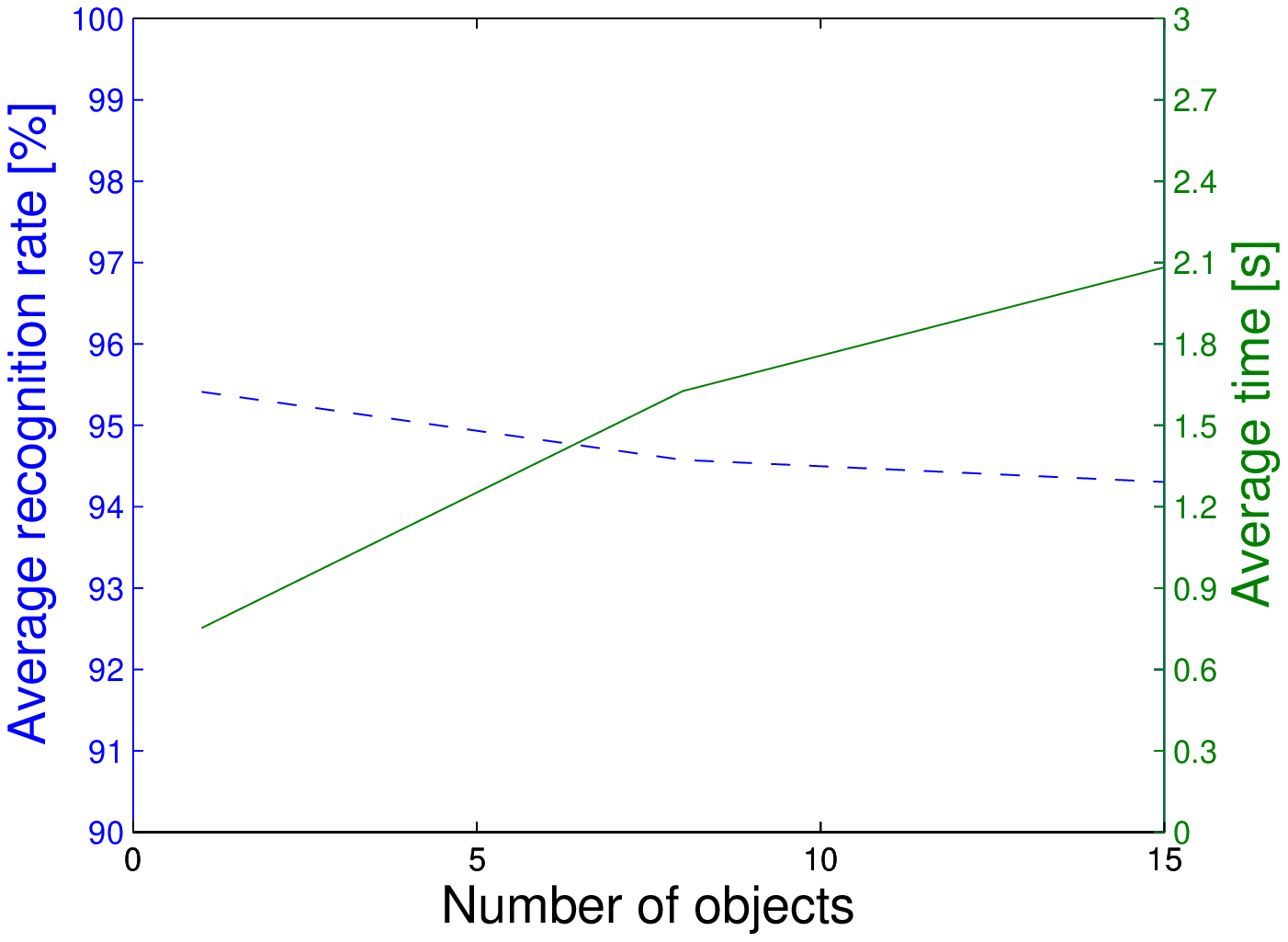}
			&
			\includegraphics[width=0.48\linewidth]{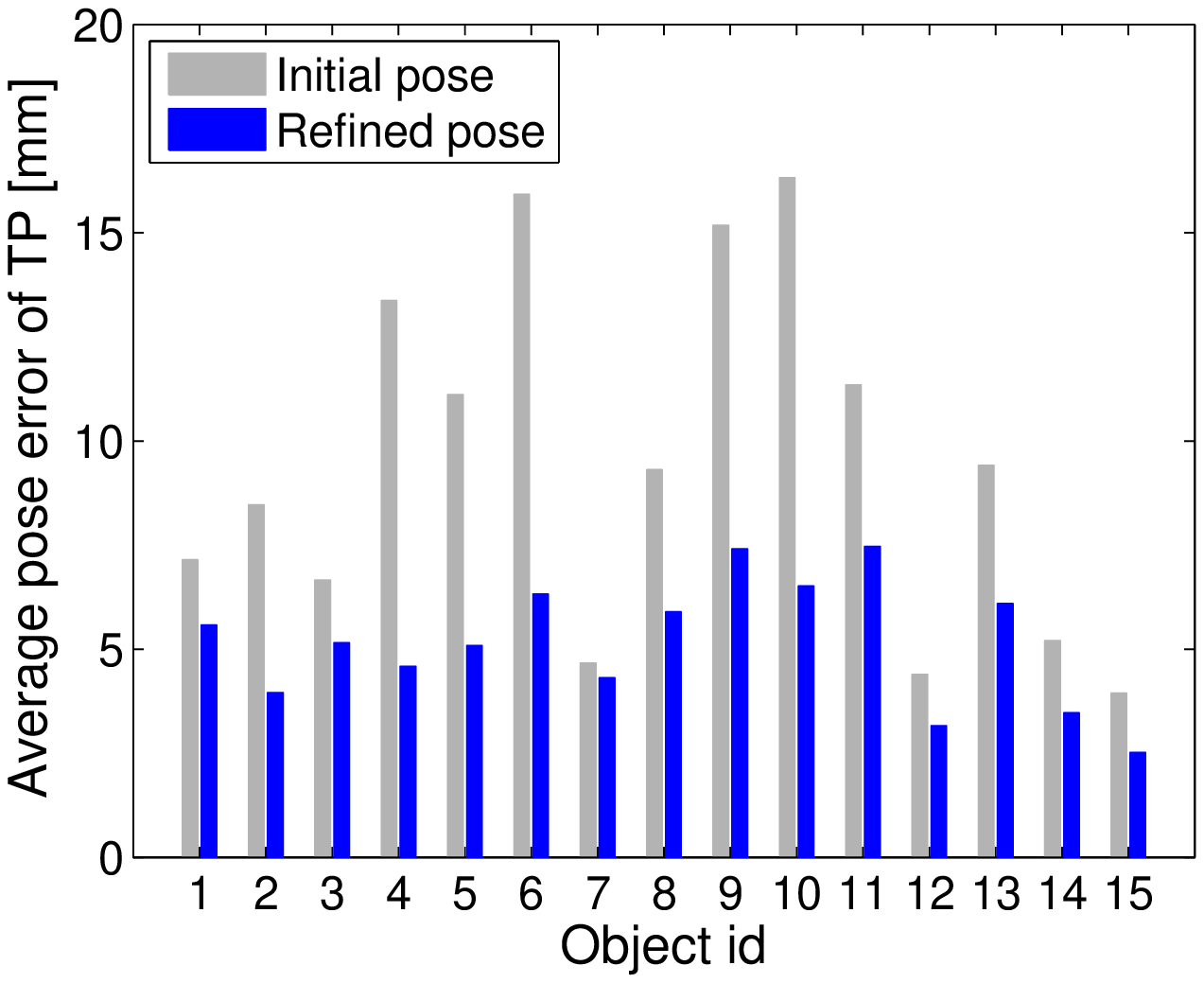}
		\end{tabular}
		\endgroup
		\caption{\label{fig:iros15_eval_graphs}
			\textbf{Left: Sub-linear complexity in the \# of templates.} The average recall rate (dashed line) and the average processing time are plotted \wrt the number of loaded templates (2916 templates per object).
			The processing time is $0.75\,$s when templates of a single object are loaded and $2.08\,$s when templates of $15$ objects are loaded, which is noticeably lower than the expected processing time of an exhaustive template matching approach, \ie, $0.75 \cdot 15 = 11.25\,$s.
			\textbf{Right: Average AD errors of initial/refined pose estimates.}
			Only pose esti\-ma\-tes whose AD error is below $10\%$ of the object diameter after the refinement are considered.
		}
	\end{center}

\vspace{-6.0ex}
\end{figure}

\customparagraph{Results.}
HashMatch achieves the average recall rate of $95.4\%$, \ie, the recall rate averaged over the 15 objects (Table~\ref{tab:iros15_rec_rates}).
HashMatch outperforms Linemod~\cite{hinterstoisser2011linemod} by $12.4\%$, Drost \etal~\cite{drost2010model} by $16.1\%$, and is slightly inferior to Linemod++~\cite{hinterstoisser2012pami} ($-1.2\%$) and Hashmod ($-1.1\%$).
As for HashMatch, the computational complexity of Hashmod is sub-linear in the number of templates, while the complexity of the other compared methods is linear. %

The sub-linear complexity of HashMatch in the number of loaded templates is demonstrated in Figure~\ref{fig:iros15_eval_graphs} (left). The average processing time per image is $0.75\,$s when only $2916$ templates of a single object are loaded and $2.08\,$s when $43740$ templates of $15$ objects are loaded. The latter is noticeably lower than the expected processing time of exhaustive template matching, \ie, $0.75 \cdot 15 = 11.25\,$s. The recall rate drops by only $1\%$ when the templates of all $15$ objects are loaded. The benefit of the refinement stage can be seen in Figure~\ref{fig:iros15_eval_graphs} (right) that compares the average AD errors of the initial and refined estimates.

The reported processing time is achieved with our parallelized C++ implementation on a modern desktop PC equipped with a $16$ core CPU and an NVIDIA GTX 780 GPU (the GPU was employed only for the refinement stage). As reported in~\cite{hinterstoisser2012accv}, the method of Drost \etal takes on average $6.3\,$s and Linemod++ $0.12\,$s. The latter was achieved with a highly optimized implementation using SSE parallelization and a limited scale space (for each object, only a limited set of scales was considered). With a similar level of optimization, HashMatch is expected to run even faster thanks to its sub-liner complexity. %

HashMatch placed fourth out of the 15 participants of the BOP Challenge 2017 (Section~\ref{sec:bop_challenge_2017}).
The leaderboard, comparing the participating methods on seven datasets, are presented in Section~\ref{sec:bop17_results}. Besides the results of the complete HashMatch method, the leaderboard shows also results of HashMatch without the pose refinement stage -- the average recall dropped by $11.8\%$ without the refinement (from $67.2\%$ to $55.4\%$). Visualizations of sample HashMatch results on the LM dataset can be found in Figure~\ref{fig:iros15_res_hinter}.

Compared to modern methods based on neural networks, such as EPOS presented in Chapter~\ref{ch:method_epos}, the advantage of HashMatch is the speed of learning new objects. While retraining a network may take several hours or days, HashMatch only requires to recalculate the hash tables, which typically takes less than a minute. However, as shown in Section~\ref{sec:bop17_results}, EPOS significantly outperforms HashMatch in the recall rate on most datasets (+38\% on TUD-L, +27\% on LM-O, +15\% on T-LESS, -2\% on IC-BIN).
The fact that EPOS uses only RGB channels makes these improvements even more notable.

\subsection{Robotic Application} \label{sec:iros15_app}

\begin{figure}[b!]
	\begin{center}		
		\begingroup
		\renewcommand{\arraystretch}{0.9} %
		\begin{tabular}{ @{}c@{ } @{}c@{ } @{}c@{ } @{}c@{ } }
			\includegraphics[width=0.21\columnwidth]{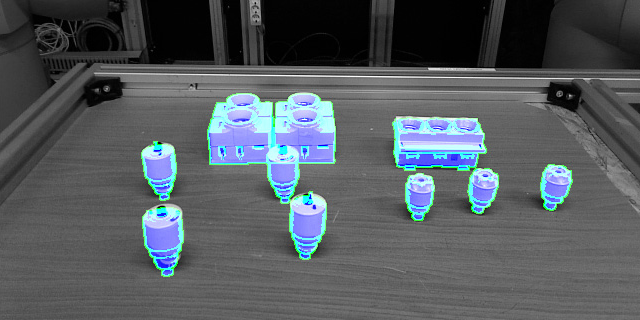} &
			\includegraphics[width=0.21\columnwidth]{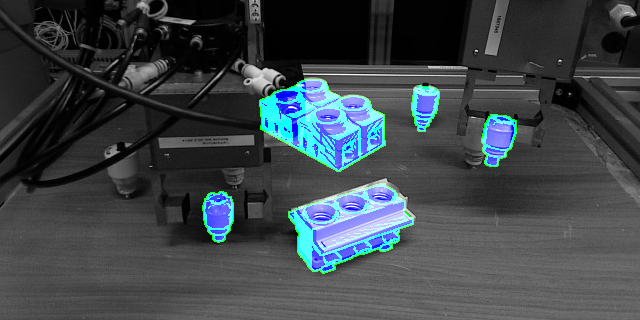} &
			\includegraphics[width=0.21\columnwidth]{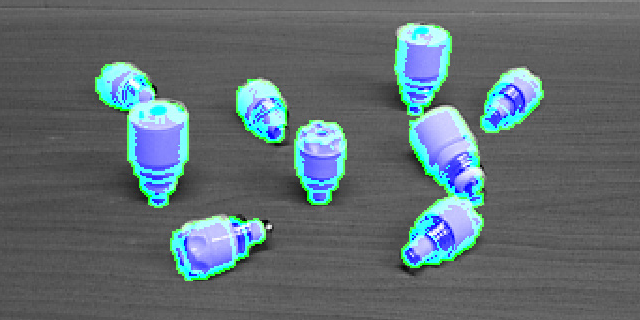} \\
			\includegraphics[width=0.21\linewidth]{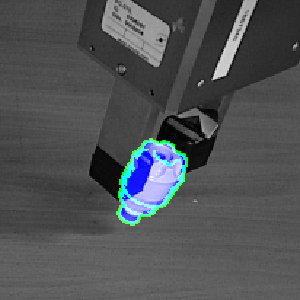} &
			\includegraphics[width=0.21\linewidth]{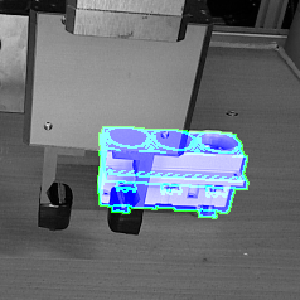} &
			\includegraphics[width=0.21\linewidth]{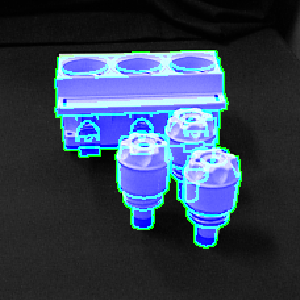}
		\end{tabular}
		\endgroup
		\caption{\label{fig:iros15_app} \textbf{HashMatch supporting robotic manipulation in the DARWIN project.} Industrial robotic arms with compliant grippers were manipulating texture-less electrical fuses and fuse boxes of different types, which were later included in the T-LESS dataset (Chapter~\ref{ch:tless}).
		} %
	\end{center}
\end{figure}

HashMatch was successfully deployed in the DARWIN FP7 EU project~\cite{darwinproject} aiming to develop a dexterous assembler robot working with embodied intelligence.
Figure~\ref{fig:iros15_app} shows the estimated poses in the considered environment.
When a robotic gripper was present in the scene, its posture was accurately provided by motor encoders and used to mask out the corresponding pixels in the image, preventing them from contaminating the pose estimation.
As shown in Figure~\ref{fig:iros15_eval_graphs} (right), HashMatch achieves a sub-centimeter average accuracy of the estimated poses,
which meets the requirements imposed by the compliant grippers of the industrial robotic arms used in the project (St{\"a}ubli RX130 and RX90).

	\chapter[EPOS:\ Estimating 6D Pose of Objects with Symmetries]{EPOS\\ {\Large Estimating 6D Pose of Objects with Symmetries}} \label{ch:method_epos}

\begin{figure}[t!]
	\begin{center}		
		\begingroup
		\renewcommand{\arraystretch}{0}
		\begin{tabular}{ @{}c@{ } @{}c@{ } }
			\includegraphics[width=0.32\linewidth]{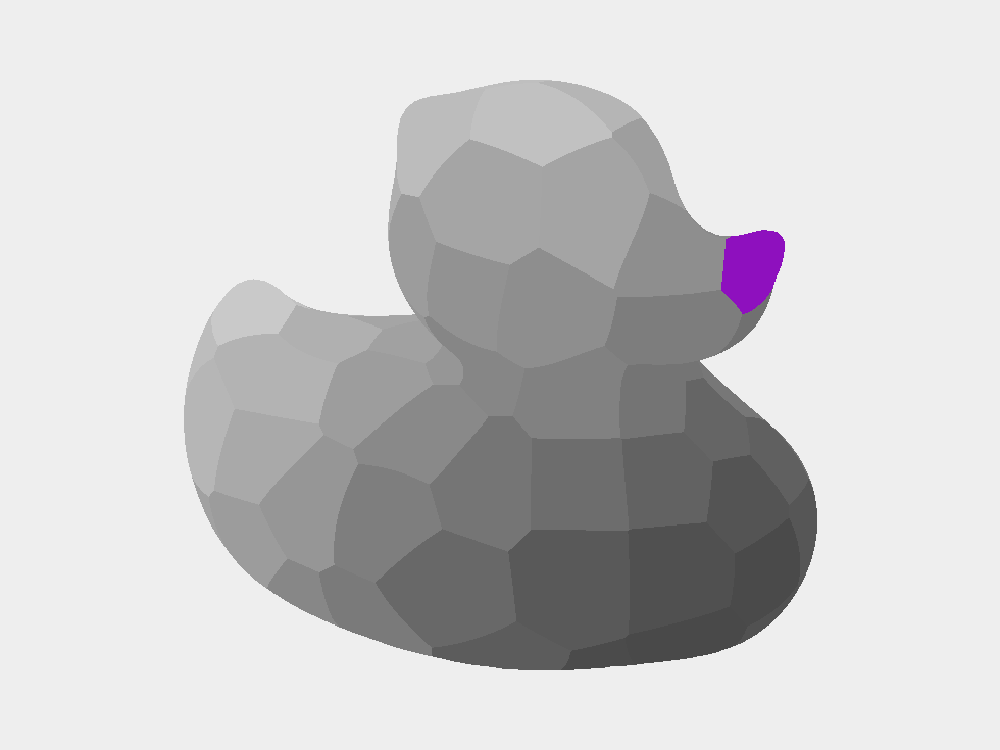} %
			&
			\includegraphics[width=0.32\linewidth]{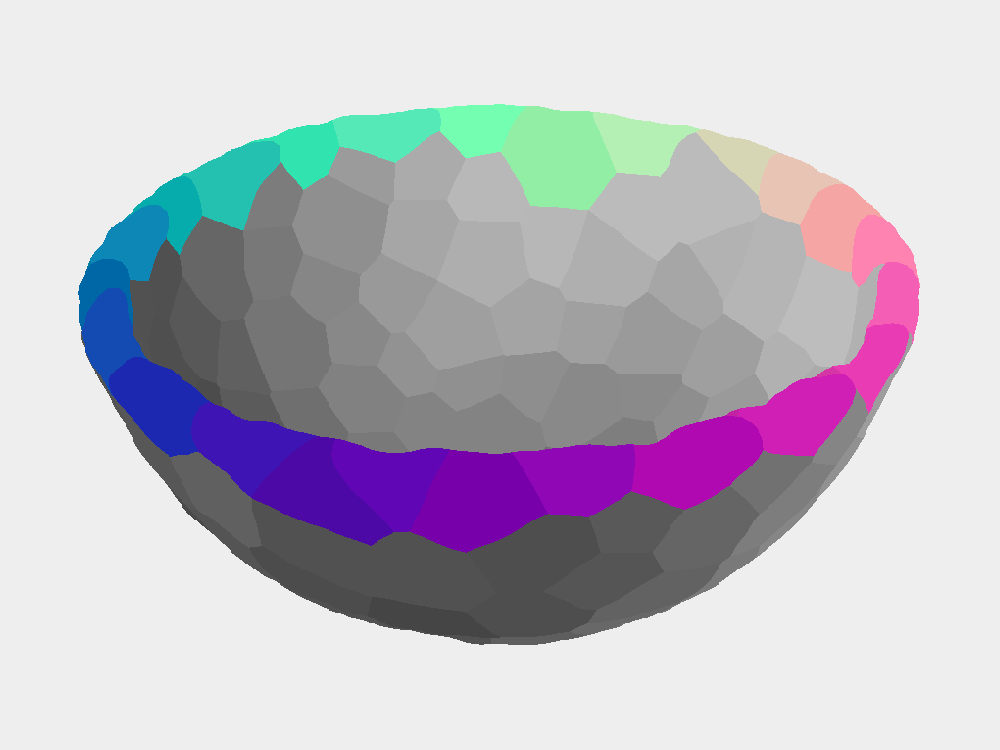} \\
			\includegraphics[width=0.32\linewidth]{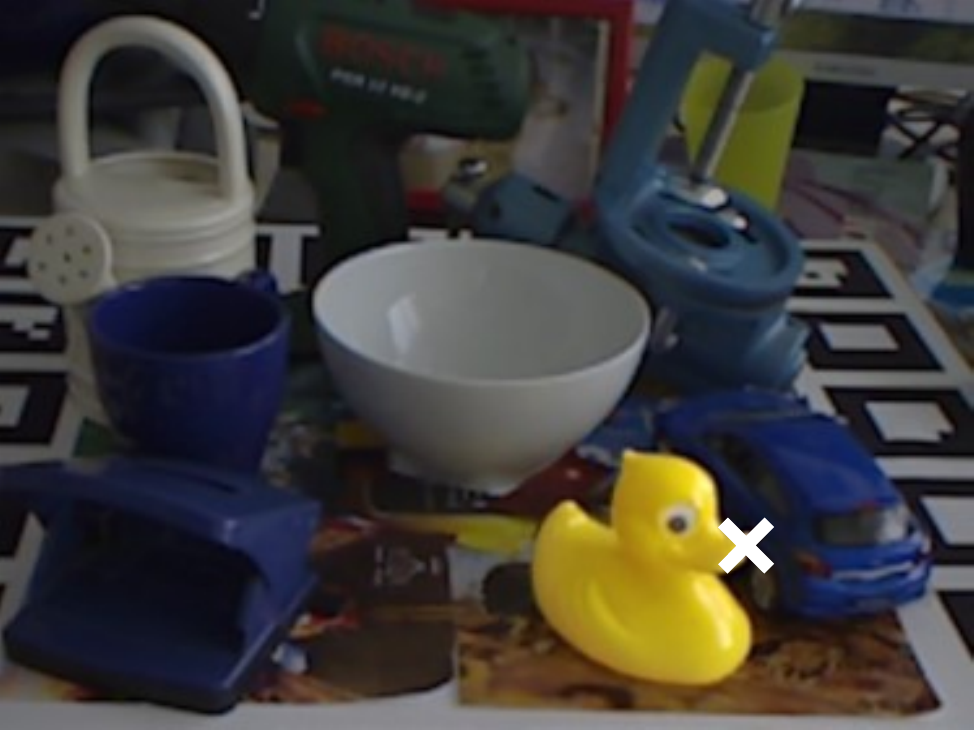} %
			&
			\includegraphics[width=0.32\linewidth]{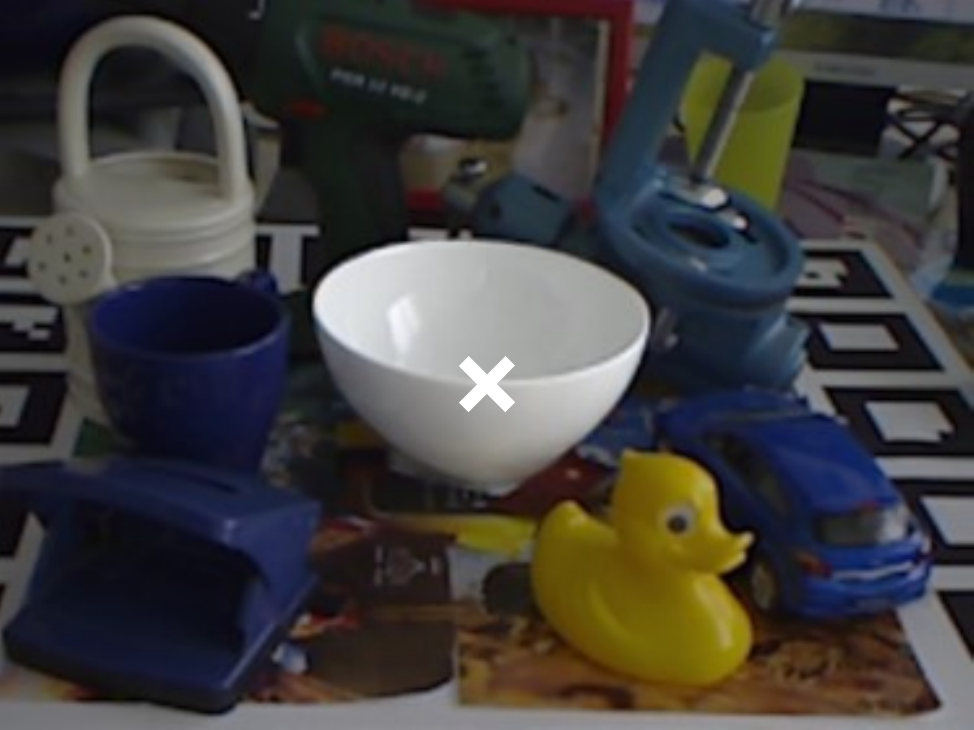} \\
		\end{tabular}
		\vspace{0.7ex}
		\endgroup
		\caption{\label{fig:epos_teaser}
			\textbf{Ambiguity of 2D-3D correspondences.}
			A 2D image location corresponds to a \emph{single} 3D location on the object model in the case of distinct object parts (left), but to one of \emph{multiple} indistinguishable 3D locations in the case of global or partial object symmetries (right). Representing an object by surface fragments allows predicting \emph{possibly multiple} potential correspondences per pixel.
			The potentially corresponding fragments are colored by mapping $(x,y,z)$ coordinates of their centers in the model space to $(R,G,B)$ channels.
		}
	\end{center}
\end{figure}

This chapter presents EPOS, a method for estimating the 6D pose of specific rigid objects with available 3D models from a single RGB input image.
The method is based on establishing 2D-3D correspondences between densely sampled pixels of the input image and 3D locations on the object model. The key idea is to represent an object by a controllable number of compact surface fragments, which allows handling object symmetries
by predicting multiple potential 2D-3D correspondences per pixel (Figure~\ref{fig:epos_teaser}), and ensures a consistent number and uniform coverage of candidate 3D locations on objects of any type.
Thanks to this representation, the method is applicable to a broad range of objects -- besides those with a distinct and non-repeatable shape or texture (a shoe, a box of corn flakes,~\etc \cite{lowe1999object,collet2011moped}), the method can handle texture-less objects and objects with global or partial symmetries (a~bowl, a cup,~\etc \cite{hodan2017tless,drost2017introducing,hinterstoisser2012accv}).

\begin{figure}[t!]
	\begin{center}
		\begin{tabular}{ @{}c@{ } @{}c@{ } }
			\hspace{2ex} & \includegraphics[width=0.9\linewidth]{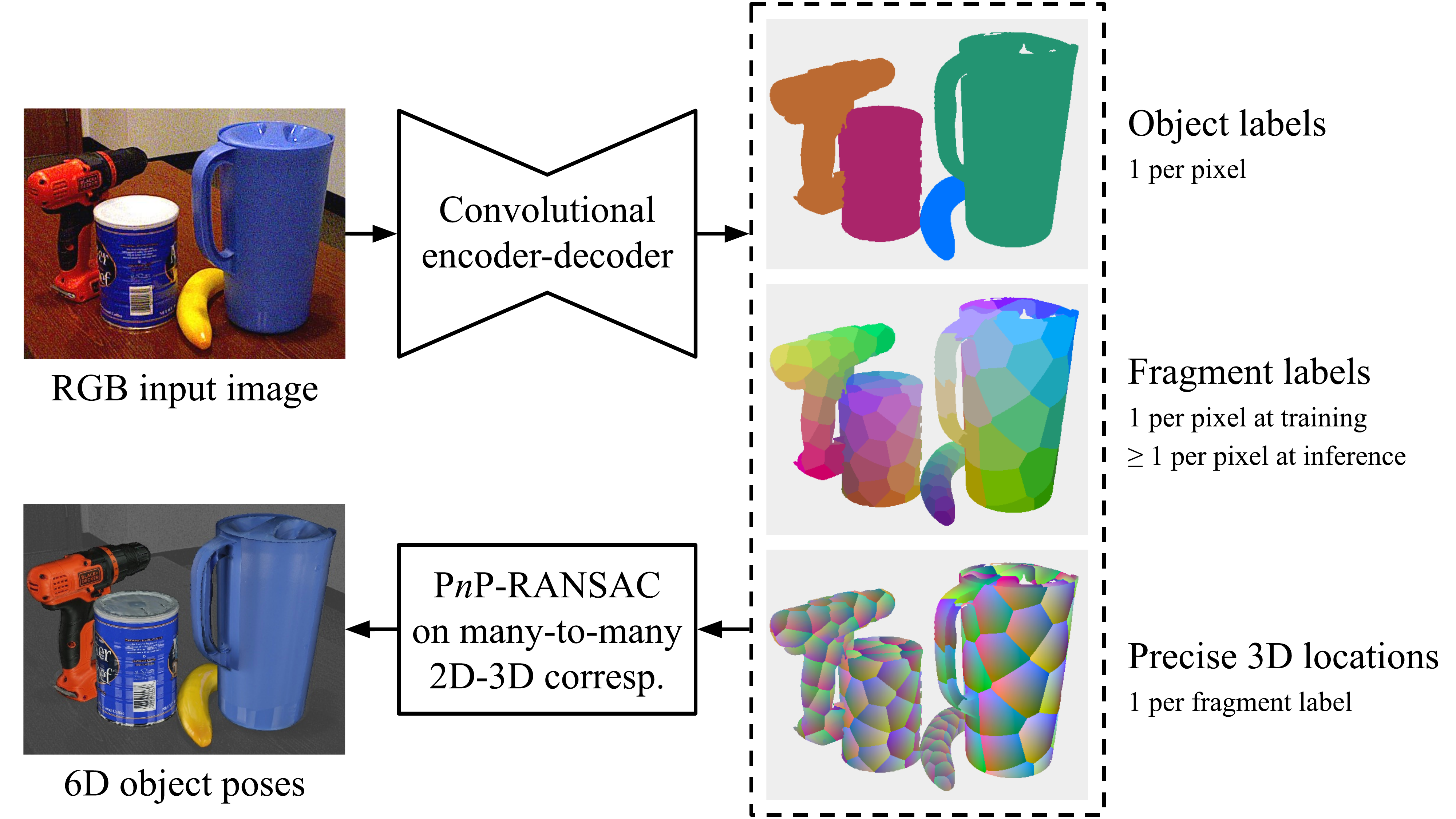}
		\end{tabular}
		\vspace{0.5ex}
		\caption{\label{fig:epos_pipeline} \textbf{EPOS pipeline.}
			At training time, the encoder-decoder network is provided for each pixel an annotation in the form of a single object label (representing an object model), a~single fragment label, and a single 3D location in the local coordinate frame of the fragment.
			At test time, the network predicts at each pixel (i)~a probability distribution over object models, (ii)~a probability distribution over all fragments of the most probable object model, which is used to select a set of potentially corresponding fragments, and (iii)~a precise 3D location in the local coordinate frame of each of the selected fragments.
			Each pixel is then linked with the precise 3D locations on the potentially corresponding fragments, and 6D poses of possibly multiple instances of each object model are estimated
			by an efficient and robust variant of the P\emph{n}P-RANSAC algorithm.
			In the visualizations,
			fragments are colored by mapping $(x,y,z)$ coordinates of their centers in the model space to $(R,G,B)$, and the precise 3D locations are colored by mapping $(x,y,z)$ coordinates in the local fragment space to $(R,G,B)$.
		}
	\end{center}
\end{figure}

Potential correspondences between densely sampled pixels and the
fragments are predicted by an encoder-decoder
network.
At each pixel, the network predicts (i) the probability of each object's presence, (ii) the probability of each fragment given the object's presence, and (iii) a precise 3D location on each fragment (Figure~\ref{fig:epos_pipeline}).
By modeling the probability of fragments conditionally, the uncertainty due to global and partial object symmetries (which are assumed unknown) is decoupled from the uncertainty of the object's presence.
The decoupled uncertainty due to symmetries is then used to guide the selection of possibly multiple
potential correspondences at each pixel.
The resulting set of potential correspondences collected at
all pixels forms a many-to-many relationship, \ie, a pixel may correspond to multiple 3D locations on the model surface and vice versa.

Poses of possibly multiple instances of every object model are estimated from the potential correspondences by an efficient and robust
variant of the P\emph{n}P-RANSAC algorithm
integrated in the Progressive-X scheme~\cite{barath2019progx}. The efficiency is achieved by the PROSAC sampler~\cite{chum2005matching} that prioritizes potential correspondences with a high predicted probability, while the robustness is achieved by a novel function for measuring the quality of a pose hypothesis. This function considers only the most accurate potential correspondence at each pixel as only up to one may be compatible with the hypothesis; the other potential correspondences provide alternative explanations and do not influence the quality.

Recent correspondence-based methods predict only a single 3D location per pixel and therefore cannot reliably handle object symmetries (Section~\ref{sec:related_object_coordinates}).
On the other hand, classical correspondence-based methods based on matching local features (Sections \ref{sec:rel_methods_2d_features} and \ref{sec:rel_methods_3d_features}) can establish multiple correspondences per pixel, but perform poorly on texture-less objects because the feature detectors often fail to provide a sufficient number of reliable locations and the descriptors are no longer discriminative enough~\cite{tombari2013bold,hodan2015detection}.

Other part-based object representations have appeared in the literature, mainly with the motivation to increase the robustness of methods against occlusion, but have not been utilized to handle object symmetries.
For example, Crivellaro \etal~\cite{crivellaro2017robust} represent an object by a set of manually selected distriminative parts and estimate the 6D pose of each part by predicting the 2D projections of pre-selected 3D keypoints.
Brachmann \etal~\cite{brachmann2014learning} and Nigam \etal~\cite{nigam2018detect} split the 3D bounding box of the object model into uniform bins and predict up to one corresponding bin per pixel.
However, they represent each bin with its center which yields correspondences with limited precision.
For human pose estimation, G\"{u}ler \etal~\cite{alp2018densepose} segment the 3D surface of the human body into semantically-defined parts.
At each pixel, they predict a single label of the corresponding part and the UV texture coordinates defining the precise 3D location on the part.
In contrast, to effectively capture the partial object symmetries, we represent an object by a set of compact surface fragments of similar
size and predict possibly multiple labels of the corresponding fragments per pixel.
Besides, we regress the precise location in local 3D coordinates of the fragment instead of the UV coordinates.
Using the UV coordinates requires a well-defined topology of the mesh model, which may need manual intervention, and is problematic for objects with a complicated surface such as a coil or an engine~\cite{drost2017introducing}.

Compared with the participants of the BOP Challenge 2019~\cite{bop19challenge,hodan2018bop}, EPOS outperformed all RGB methods and most RGB-D and D methods on the T-LESS~\cite{hodan2017tless} and LM-O~\cite{brachmann2014learning} datasets, which include texture-less and symmetric objects captured in cluttered scenes under various levels of occlusion.
On the YCB-V~\cite{xiang2017posecnn} dataset, which includes both textured and texture-less objects, the method was superior to all competitors, with a significant $27\%$ absolute improvement over the second-best RGB method.
These results are achieved without any post-refinement of the estimated poses, such as~\cite{manhardt2018deep,li2018deepim,rad2017bb8}.
As shown later in Section~\ref{sec:bop17_results}, EPOS significantly outperforms also the HashMatch method proposed in Chapter~\ref{ch:method_template}, even though HashMatch relies on RGB-D image channels.

EPOS was published in~\cite{hodan2020epos}, is described in detail in Sections~\ref{sec:epos_fragments}--\ref{sec:epos_fitting} and evaluated in Section~\ref{sec:epos_experiments}. The project website with a video presentation, a real-world demonstration, the source code, and pre-trained models is available at: \texttt{\href{http://cmp.felk.cvut.cz/epos/}{cmp.felk.cvut.cz/epos}}.

\section{Surface Fragments} \label{sec:epos_fragments}

A mesh model $M_o = (V_o, T_o)$, defined by a set of 3D vertices $V_o$ and a set of triangular faces $T_o$, is assumed available for each object, where objects are referred to by labels $o \in O = \{1, \dots, m\}$.
The set of all 3D points on the model surface, denoted as $S_o$, is split into $n$ fragments referred to by labels $f \in F = \{1, \dots, n\}$.
A surface fragment $f$ of an object $o$ is defined as $S_{o,f} = \{\mathbf{x} \in S_o \; | \; d(\mathbf{x}, \mathbf{g}_{o,f}) < d(\mathbf{x}, \mathbf{g}_{o,i}), \forall i \in F, i \neq f\}$, where $d(\cdot)$ is the Euclidean distance of two 3D points and $\{\mathbf{g}_{o,f}\}_{f=1}^n$ is a set of fragment centers.
The centers are found by the farthest point sampling algorithm which iteratively selects the vertex from $V_o$ that is farthest from the already selected vertices. The algorithm starts with the centroid of the model which is then removed from the final set of centers.

Note that the number of fragments was fixed across all objects for the experiments presented in Section~\ref{sec:epos_experiments}. A study of object-specific numbers of fragments, which may improve performance and may depend on factors such as the physical object size, shape, or the range of distances of the object from the camera, is left for future work.

\section{Predicting Potential 2D-3D Correspondences} \label{sec:epos_corrs}
\label{sec:epos_coord_pred}

This section describes the prediction of potential 2D-3D correspondences between densely sampled pixels of the input image and precise 3D locations on the object models. Each object is represented by a set of surface fragments which allow predicting multiple potential correspondences per pixel.

\subsection{Decoupling Uncertainty Due to Symmetries}
The probability that a fragment $f$ of an object $o$ is visible at a pixel with image coordinates $\mathbf{u} = (u, v) \in \{1, \dots, w \} \times \{1, \dots, h\}$, where $(w,h)$ is the image size,
is modeled~as:
\begin{align*}
	\Pr(Y\myeq f, X\myeq o\,|\, \mathbf{u}) = \Pr(Y\myeq f\,|\,X\myeq o, \mathbf{u})\Pr(X\myeq o\,|\, \mathbf{u})
\end{align*}
where $X$ and $Y$ are random variables representing the object and the fragment respectively.
The probability $\Pr(Y\myeq f,X\myeq o\,|\,\mathbf{u})$ may be low because (1) the object $o$ is not present at the pixel $\mathbf{u}$ or the object identity is ambiguous, or (2) the identity of the fragment visible at $\mathbf{u}$ is ambiguous, which may be caused by global or partial symmetries of the object.
To disentangle these two ambiguities, we predict $a_{o}(\mathbf{u})$ $=$ $\Pr(X\myeq o\,|\,\mathbf{u})$ and $b_{o,f}(\mathbf{u})$ $=$ $\Pr(Y\myeq f\,|\,X\myeq o,\mathbf{u})$ separately, instead of directly predicting $\Pr(Y\myeq f,X\myeq o\,|\,\mathbf{u})$. Expressing the probability distribution $b_{o,f}$ separately enables selecting a data-dependent number of potentially corresponding fragments per pixel (described in Section~\ref{sec:epos_establish_coords}).

\subsection{Regressing Precise 3D Locations}
A surface fragment~$f$ of an object~$o$ is associated with a regressor $\mathbf{r}_{o,f}: \mathbb{R}^2 \rightarrow \mathbb{R}^3$, which at a pixel~$\mathbf{u}$ predicts the corresponding 3D location: $\mathbf{r}_{o,f}(\mathbf{u}) = (\mathbf{x} - \mathbf{g}_{o,f}) / h_{o,f}$.
The regressor produces
\emph{3D~fragment coordinates}, which are defined in a 3D coordinate frame with the origin at the fragment center $\mathbf{g}_{o,f}$. The 3D point $\mathbf{x}$ is expressed in the 3D coordinate frame of the model, and the scalar $h_{o,f}$ normalizes the regression range and is defined as the length of the longest side of the 3D bounding box of the fragment.

\subsection{Dense Prediction}
A single deep neural network with an encoder-decoder structure, \eg, DeepLabv3+~\cite{chen2018encoder}, is adopted to densely predict $a_{i}(\mathbf{u})$, $b_{o,f}(\mathbf{u})$ and $\mathbf{r}_{o,f}(\mathbf{u})$, $\forall o \in O$, $\forall f \in F$, $\forall i \in O \cup \{0\}$, where $0$ is reserved for the background class.
For $m$ objects, each represented by $n$ surface fragments, the network has $4mn$$+$$m$$+$$1$ output channels ($m$$+$$1$ for probabilities of the objects and the background, $mn$ for probabilities of the surface fragments, and $3mn$ for the 3D fragment coordinates).

\subsection{Network Training} \label{sec:epos_network_training}
The network is trained by minimizing the following loss averaged over all pixels $\mathbf{u}$:
\begin{align}\label{eq:epos_loss}
	\begin{split}
		L(\mathbf{u}) = \; &
		E\big(\bar{\mathbf{a}}{(\mathbf{u})}, \mathbf{a}{(\mathbf{u})}\big) \\
		&+ \lambda_1 \sum\nolimits_{o\in O} \bar{a}_{o}{(\mathbf{u})} E\big(\bar{\mathbf{b}}_{o}{(\mathbf{u})}, \mathbf{b}_{o}{(\mathbf{u})}\big) \\
		&+ \lambda_2 \sum\nolimits_{o\in O} \bar{a}_{o}{(\mathbf{u})} \sum\nolimits_{f \in F}  \bar{b}_{o,f}{(\mathbf{u})} H\big(\bar{\mathbf{r}}_{o,f}{(\mathbf{u})}, \mathbf{r}_{o,f}{(\mathbf{u})}\big)
	\end{split}
\end{align}
where $E$ is the softmax cross-entropy loss, $H$ is the Huber loss~\cite{huber1992robust}, $\mathbf{a}(\mathbf{u}) = [a_0(\mathbf{u}),$ $\dots,$ $a_m(\mathbf{u})]$, and $\mathbf{b}_{o}(\mathbf{u}) = [b_{o,1}(\mathbf{u}),$ $\dots,$ $b_{o,n}(\mathbf{u})]$.
The ground-truth one-hot vectors $\bar{\mathbf{a}}{(\mathbf{u})}$ and $\bar{\mathbf{b}}_{o}{(\mathbf{u})}$ indicate which object (or the background) and which fragment is visible at the pixel~$\mathbf{u}$. Elements of these ground-truth vectors are denoted as $\bar{a}_{o}{(\mathbf{u})}$ and $\bar{b}_{o,f}{(\mathbf{u})}$. The vector $\bar{\mathbf{b}}_{o}{(\mathbf{u})}$ is defined only if the object $o$ is present at $\mathbf{u}$.
The ground-truth 3D fragment coordinates are denoted as $\bar{\mathbf{r}}_{o,f}(\mathbf{u})$. Weights $\lambda_1$ and $\lambda_2$ are used to balance the loss terms.

\begin{figure}[t!]
	\begin{center}
		\begingroup
		\small
		\begin{tabular}{ @{}c@{ } @{}c@{ } @{}c@{ } @{}c@{ } }
			RGB image & Object labels & Fragment labels & 3D fragment coord. \vspace{0.5ex} \\
			
			\includegraphics[width=0.245\linewidth]{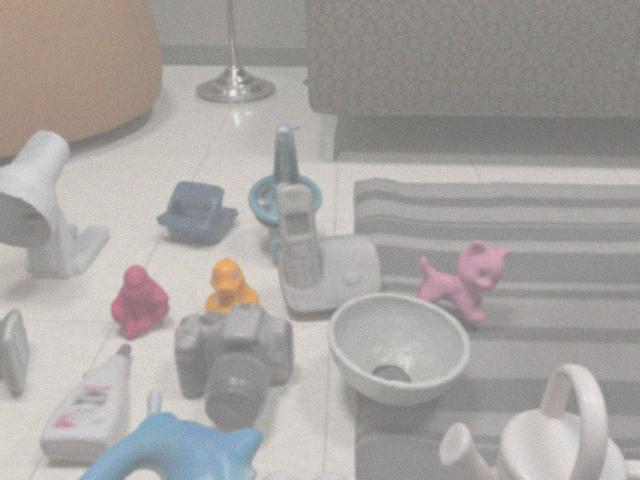} &
			\includegraphics[width=0.245\linewidth]{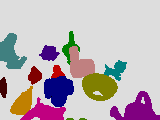} &
			\includegraphics[width=0.245\linewidth]{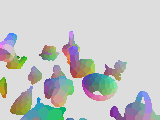} &
			\includegraphics[width=0.245\linewidth]{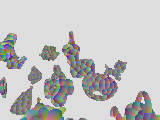} \\
			
			\includegraphics[width=0.245\linewidth]{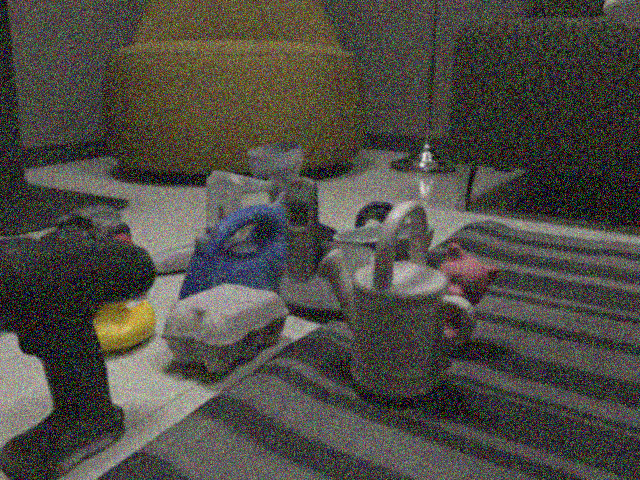} &
			\includegraphics[width=0.245\linewidth]{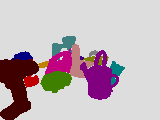} &
			\includegraphics[width=0.245\linewidth]{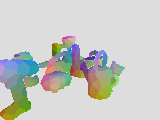} &
			\includegraphics[width=0.245\linewidth]{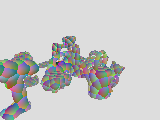} \\
			
			\includegraphics[width=0.245\linewidth]{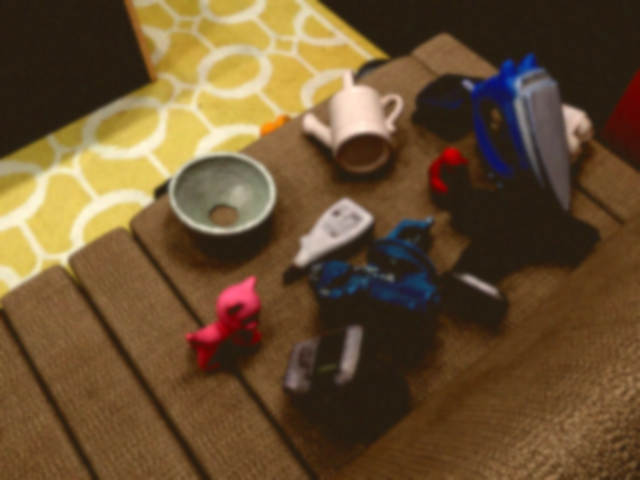} &
			\includegraphics[width=0.245\linewidth]{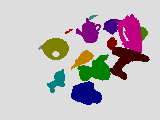} &
			\includegraphics[width=0.245\linewidth]{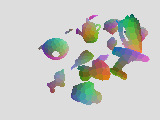} &
			\includegraphics[width=0.245\linewidth]{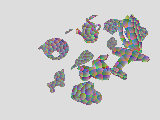} \\
			
		\end{tabular}
		\endgroup
		\caption{\label{fig:epos_training_data}
			\textbf{Training data.} The left column shows sample synthetic images from~\cite{hodan2019photorealistic} augmented by randomly adjusting brightness, contrast, hue, and saturation, and by applying random Gaussian noise and blur, similarly to~\cite{hinterstoisser2017pre}. The right three columns show the ground-truth
			generated by a custom OpenGL shader
			(labels are converted to one-hot vectors for training).
		}
	\end{center}
\end{figure}

The network is trained on images annotated with ground-truth 6D object poses.
Vectors $\bar{\mathbf{a}}(\mathbf{u})$, $\bar{\mathbf{b}}_{o}(\mathbf{u})$, and $\bar{\mathbf{r}}_{o,f}(\mathbf{u})$ are obtained by rendering the 3D object models in the ground-truth poses with a custom OpenGL shader.
Pixels outside the visibility masks of the objects, which are calculated as in~\cite{hodan2016evaluation}, are considered to be the background.
The training images are augmented by randomly adjusting brightness, contrast, hue, and saturation, and by applying random Gaussian noise and blur, similarly to~\cite{hinterstoisser2017pre} (Figure~\ref{fig:epos_training_data}).

\subsection{Learning in the Presence of Object Symmetries}

In the case of global or partial object symmetries,
a pixel may have multiple potentially corresponding 3D locations that are indistinguishable in the given image (Figure~\ref{fig:epos_teaser}).
One could try to find the indistinguishable 3D locations and train the network to predict a high probability for each of the surface fragments on which these locations lie.
Finding 3D locations that are indistinguishable due to global object symmetries is relatively easy as the global symmetries can be detected from the shape and texture of the object models (Section~\ref{sec:symmetries}).
On the other hand, finding 3D locations that are indistinguishable due to partial object symmetries is difficult as one would need to identify the visible object parts in each training image and find their fits to the object models.

\begin{figure}[t!]
	\begin{center}
		\begingroup
		\begin{tabular}{ @{}c@{ } @{}c@{ } @{}c@{ } @{}c@{ } }
			\includegraphics[width=0.245\linewidth]{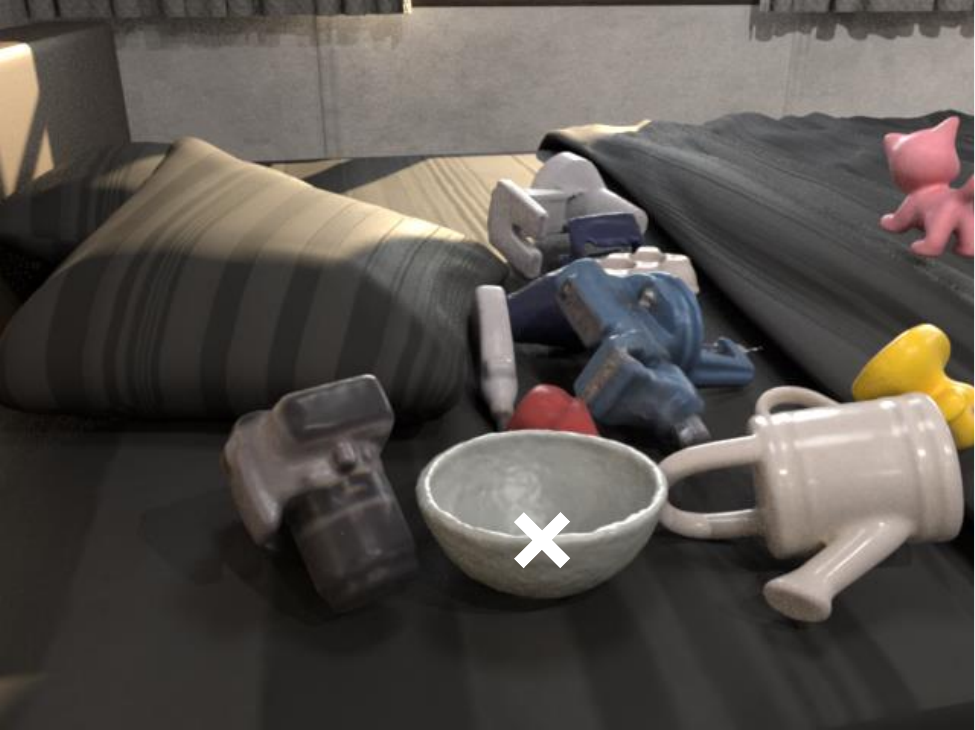} &
			\includegraphics[width=0.245\linewidth]{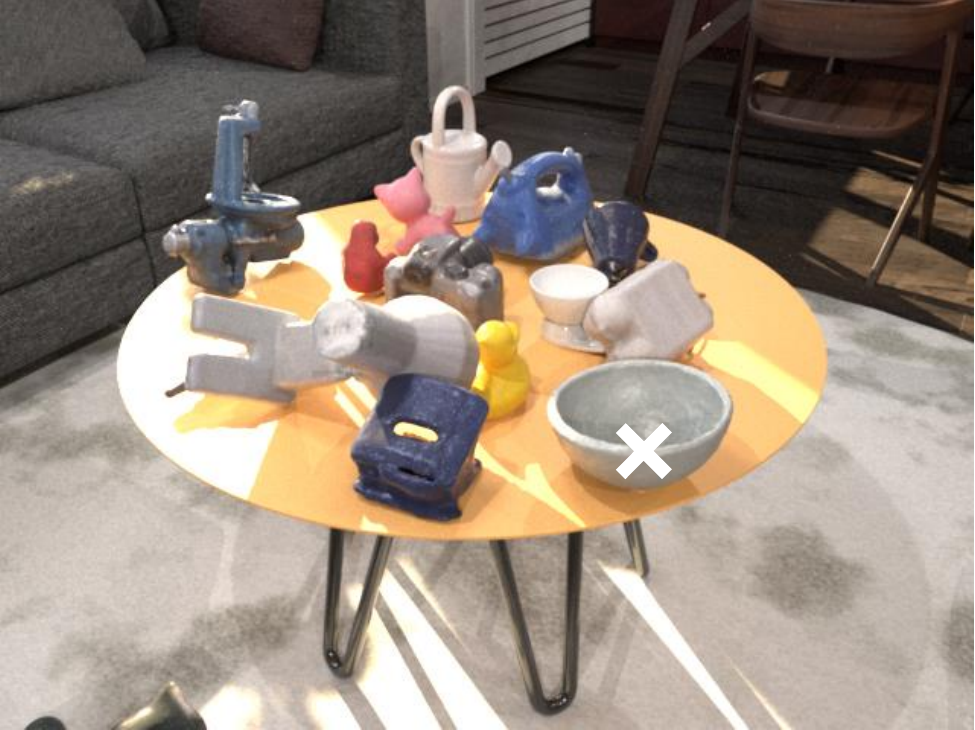} &
			\includegraphics[width=0.245\linewidth]{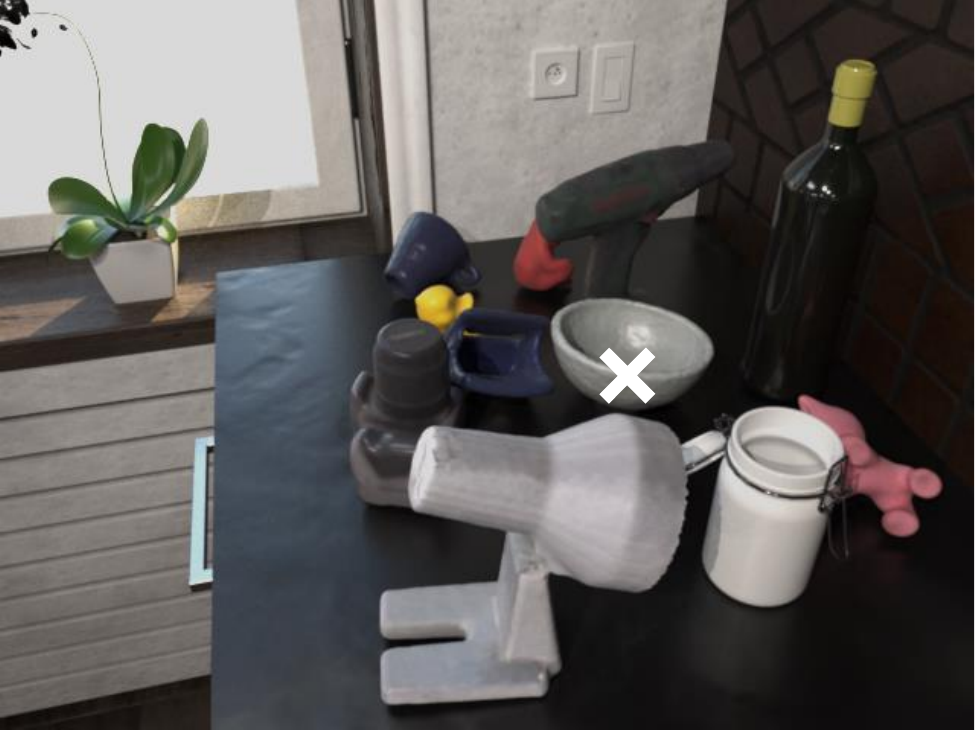} &
			\includegraphics[width=0.245\linewidth]{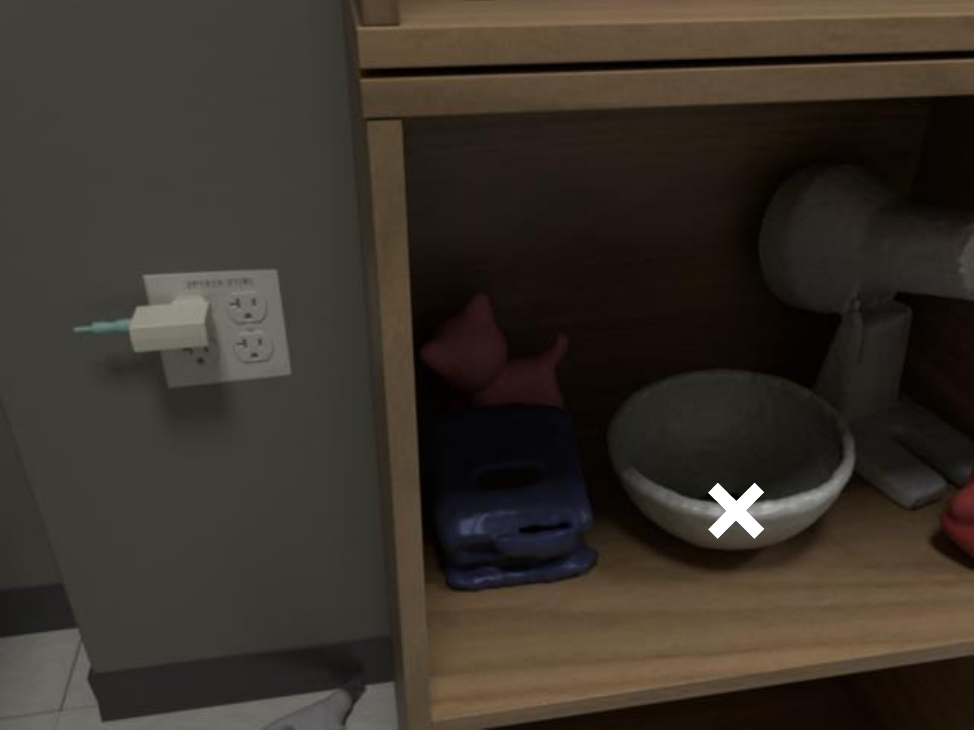} \\
			
			\includegraphics[width=0.245\linewidth]{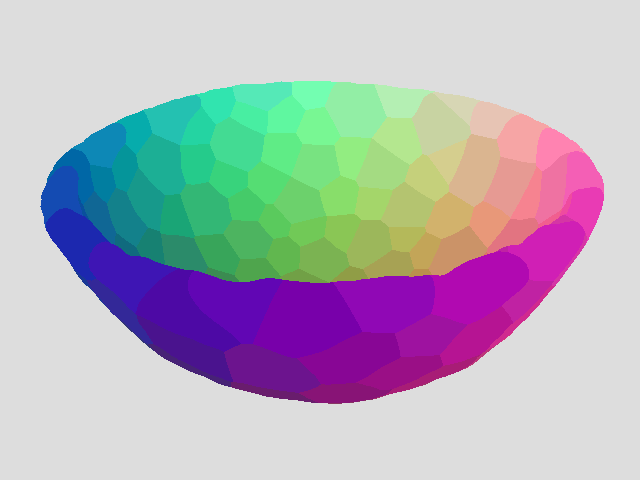} &
			\includegraphics[width=0.245\linewidth]{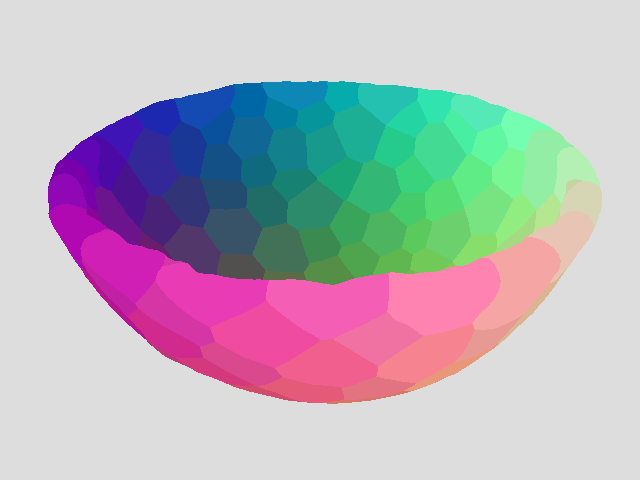} &
			\includegraphics[width=0.245\linewidth]{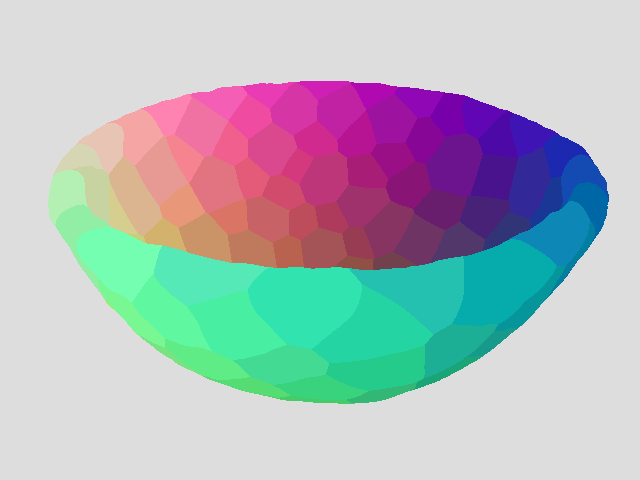} &
			\includegraphics[width=0.245\linewidth]{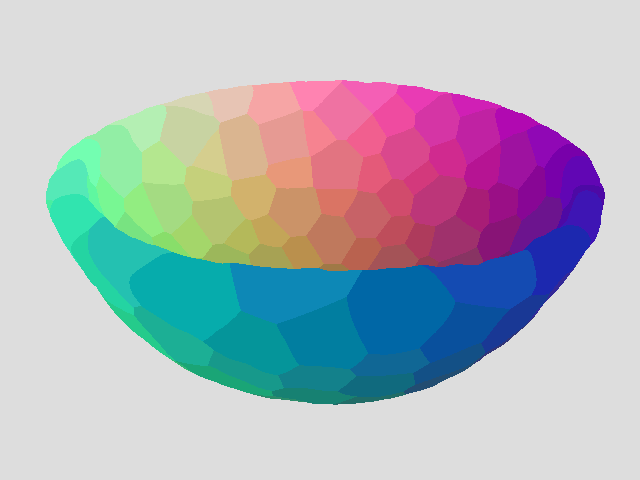} \\
			
			\includegraphics[width=0.245\linewidth]{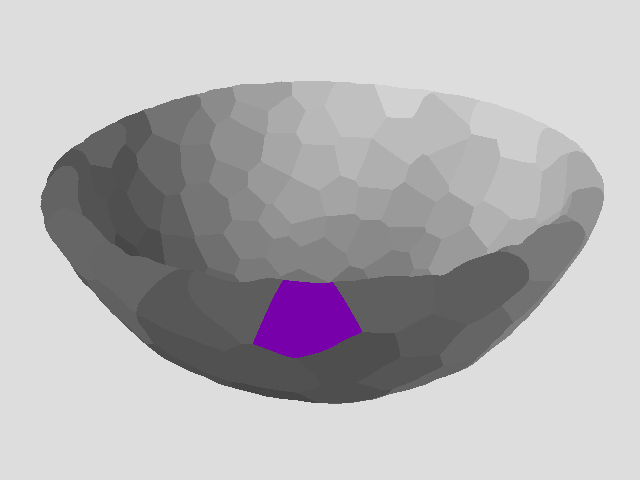} &
			\includegraphics[width=0.245\linewidth]{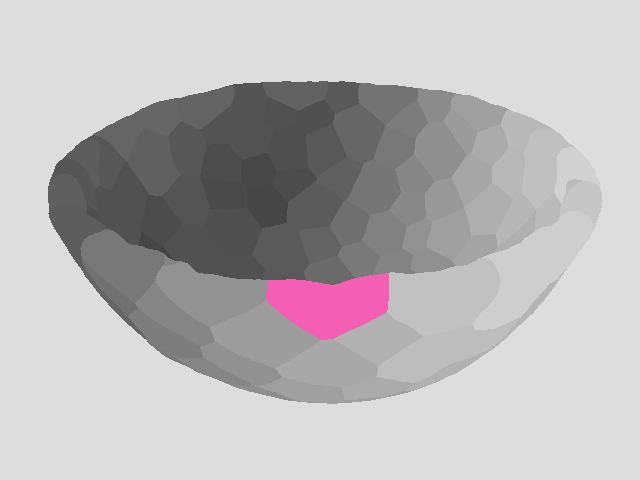} &
			\includegraphics[width=0.245\linewidth]{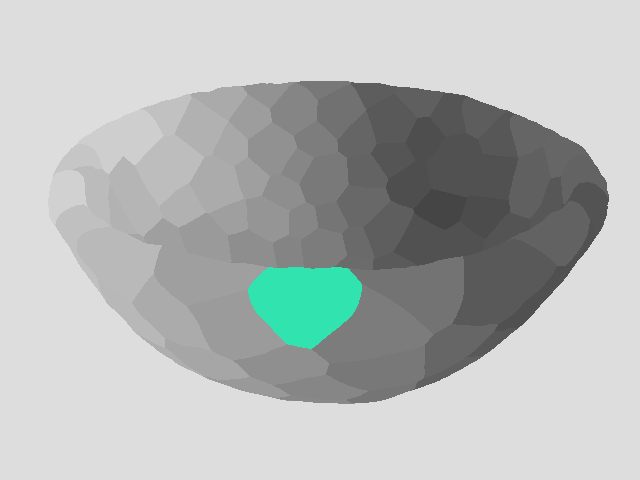} &
			\includegraphics[width=0.245\linewidth]{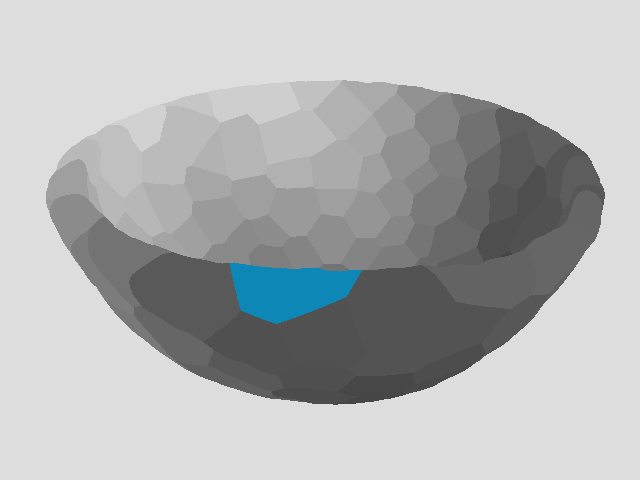} \\
		\end{tabular}
		\endgroup
		\caption{\label{fig:epos_learning_symmetries}
			\textbf{Learning in the presence of object symmetries.}
			The first row shows sample synthetic training images, the second row visualizes the surface fragments of the bowl oriented around the vertical axis as in the images, and the third row highlights the ground-truth fragments assigned to the pixels marked in the first row.
			Instead of explicitly identifying all indistinguishable (\ie, potentially corresponding) fragments at each pixel of the training images (which may be not trivial if the object pose is ambiguous due to occlusions), each pixel is assigned a single ground-truth fragment. By providing the network with multiple training images showing the objects in uniformly distributed poses, the network is expected to learn a uniform distribution over the indistinguishable fragments. The color-coding of fragments is explained in Figure~\ref{fig:epos_teaser}.
		}
	\end{center}
\end{figure}

Instead of explicitly listing all indistinguishable 3D locations, we provide the network with only a single corresponding 3D location per pixel during training (see Equation~\ref{eq:epos_loss}) and let the network learn all of the potential correspondences implicitly. This is achieved by minimizing the softmax cross-entropy loss $E\left(\bar{\mathbf{b}}_o(\mathbf{u}), \mathbf{b}_o(\mathbf{u})\right)$, which corresponds exactly to minimizing the Kullback-Leibler divergence of distributions $\bar{\mathbf{b}}_o(\mathbf{u})$ and $\mathbf{b}_o(\mathbf{u})$~\cite{goodfellow2016deep}, and by ensuring that object poses in the training set are sampled uniformly \wrt the three rotation angles and the distance from the camera.
Such sampling of object poses is easy to achieve with synthetic training images and ensures that each of the indistinguishable fragments is indicated by the ground-truth one-hot distribution $\bar{\mathbf{b}}_o(\mathbf{u})$ at a comparable number of pixels $\mathbf{u}$.
The network is then expected to learn at these pixels a high value of the probability $b_{o,f}(\mathbf{u})$ for all of the indistinguishable fragments.
For example, the probability distribution on a sphere is expected to be uniform, \ie, $b_{o,f}(\mathbf{u})=1/n, \forall f \in F$.
On a bowl, the probability is expected to be constant around the axis of symmetry (Figure~\ref{fig:epos_learning_symmetries}).

\subsection{Establishing Potential 2D-3D Correspondences} \label{sec:epos_establish_coords}
A pixel~$\mathbf{u}$ is linked with a 3D location $\mathbf{x}_{o,f}(\mathbf{u}) = h_{o,f} \mathbf{r}_{o,f}(\mathbf{u}) + \mathbf{g}_{o,f}$ on all fragments for which $a_o(\mathbf{u}) > \tau_a$ and $b_{o,f}(\mathbf{u})/\max_{i=1}^n\!\left(b_{o,i}(\mathbf{u})\right) > \tau_b$.
The threshold $\tau_b$ is relative to the maximum to collect locations from all indistinguishable fragments that are expected to have a similarly high probability~$b_{o,f}(\mathbf{u})$.
The set of potential 2D-3D correspondences established for object~$o$, which may cover multiple instances of this object, is denoted as $C_o = \left\{\left(\mathbf{u}, \mathbf{x}_{o,f}(\mathbf{u}), s_{o,f}(\mathbf{u})\right)\right\}$, where $s_{o,f}(\mathbf{u}) = a_o(\mathbf{u})b_{o,f}(\mathbf{u})$ is a confidence.

\section{Robust and Efficient 6D Pose Fitting} \label{sec:epos_fitting}

The next step of the pipeline is to estimate the 6D poses of possibly multiple instances of each object $o$ from the set of potential 2D-3D correspondences $C_o$. A 6D pose is defined by a rigid transformation from the 3D model frame to the 3D camera frame, and the correspondences define links between the 3D model frame and the 2D image frame.

If only a single instance of an object $o$ was visible in the image, and if only up to one correspondence was established per pixel, the pose could be effectively estimated by solving the Perspective-\emph{n}-Point (P\emph{n}P) problem~\cite{lepetit2009epnp}.
Popular algorithms to solve this problem include P3P~\cite{kneip2011novel} for $n=3$ correspondences, EP\emph{n}P~\cite{lepetit2009epnp} for $n \geq 4$, and DLS-P\emph{n}P~\cite{hesch2011direct} for $n \geq 3$. For robust estimation on real-world data, these algorithms are often combined with the RANSAC fitting scheme~\cite{fischler1981random}.
In this scheme, the P\emph{n}P problem is solved repeatedly on a randomly sampled subset of correspondences and the final output is defined by the pose hypothesis with the highest quality, which is typically defined by the number of inlier correspondences~\cite{fischler1981random} or their likelihood~\cite{torr2002bayesian}.
A correspondence $(\mathbf{u}, \mathbf{x})$ is considered an inlier \wrt a pose estimate $\hat{\textbf{P}} = [\hat{\mathbf{R}}\,|\, \hat{\mathbf{t}}]$, if the re-projection error:
\begin{equation}\label{eq:reproj_error}
	e\big(\mathbf{u}, \mathbf{x}, \hat{\textbf{R}}, \hat{\textbf{t}}\big) = \big\Vert \mathbf{u} - \text{proj}\big(\hat{\mathbf{R}}\textbf{x} + \hat{\mathbf{t}}\big)\big\Vert_2
\end{equation}

\noindent is below a
threshold $\tau_r$. Function $\text{proj}(\cdot)$ is the 2D projection with values in pixels. If the re-projection error is larger or equal to $\tau_r$, the correspondence is considered an outlier.

In our case, multiple instances of an object $o$ may be visible in the image, their number may be unknown (depending on the 6D object pose estimation problem; Section~\ref{sec:problem_formulation}) and there may be multiple potential correspondences established at each pixel. Specifically, with respect to a single pose hypothesis, the set $C_o$ of potential correspondences may include three types of outliers.
First, it may include outliers due to erroneous prediction of the 3D locations. Second, for each 2D/3D location there may be up to one correspondence which is compatible with the pose hypothesis; the other correspondences act as outliers. Third, correspondences originating from different instances of the object $o$ are also incompatible with the pose hypothesis.
The set $C_o$ may be therefore contaminated with a high proportion of outliers and a robust and efficient estimator able to estimate poses of multiple object instances is needed. %

\subsection{Single-Instance Fitting} \label{sec:single_instance_fitting}

The 6D pose of a single instance of an object $o$ is estimated from the set of correspondences~$C_o$ by GC-RANSAC~\cite{barath2018gcransac}, a variant of RANSAC which locally optimizes every new so-far-the-best hypothesis by alternating between two steps: (i)~selecting the inliers by
a graph-cut optimization, and (ii)~refining the hypothesis from the currently selected inliers.
The inlier selection approach is motivated by the fact that both inliers and outliers are typically spatially coherent, which is the case also for 2D-3D correspondences -- a correspondence near an inlier or outlier (in 2D and 3D)
is more likely to be an inlier or outlier respectively.
A neighborhood graph of correspondences for the graph-cut optimization is constructed by describing each correspondence with a 5D vector consisting of the 2D and 3D coordinates (in pixels and centimeters) and linking two 5D descriptors if their Euclidean distance is below a threshold $\tau_d$.\footnote{Optimal units of the 2D and 3D coordinates forming the 5D space may depend on factors such as the object size, its distance from the camera or intrinsic camera parameters. Representing the 2D and 3D coordinates in pixels and centimeters respectively worked well on all datasets used in our experiments.}
The unary and pair-wise energy terms to optimize are defined using a kernel function of the re-projection error truncated at $\tau_r$.

A pose hypothesis is calculated by the P3P solver~\cite{kneip2011novel} from a triplet of correspondences sampled by PROSAC~\cite{chum2005matching}, which first focuses on correspondences with high confidence~$s_{o,f}$ (defined in Section~\ref{sec:epos_establish_coords}) and progressively blends to a uniform sampling. Triplets which form 2D triangles with the area below a threshold $\tau_t$ or have collinear 3D locations are rejected, and pose hypotheses behind the camera or with the determinant of the rotation matrix equal to $-1$ (\ie, an improper rotation matrix~\cite{haber2011three}) are discarded.
The hypothesis that passes these early-rejection tests is refined from all inliers by the EP\emph{n}P solver~\cite{lepetit2009epnp} followed by the Levenberg-Marquardt (L-M) optimization~\cite{more1978levenberg}. EP\emph{n}P and L-M are used also for refinement in the local optimization which is applied when a new so-far-the-best hypothesis is found.

The quality of a pose hypothesis $\hat{\textbf{P}} = [\hat{\mathbf{R}}\,|\, \hat{\mathbf{t}}]$ is defined as:
\begin{align} \label{eq:pose_quality_single}
	q\big(\hat{\textbf{R}}, \hat{\textbf{t}}, U_o, C_{o,\mathbf{u}}, \tau_r\big) = \frac{1}{|U_o|} \sum_{\mathbf{u} \in U_o} \max_{(\mathbf{u}, \mathbf{x}, \cdot) \in C_{o,\mathbf{u}}} \max\left(0, \, 1 - \frac{e\big(\mathbf{u}, \mathbf{x}, \hat{\textbf{R}}, \hat{\textbf{t}} \big)^2}{\tau_r^2}\right)
\end{align}
where $U_o$ is a set of pixels at which the correspondences $C_o$ are established, $C_{o,\mathbf{u}} \subset C_o$ are correspondences established at the pixel $\mathbf{u}$, $e(\cdot)$ is the re-projection (Equation~\ref{eq:reproj_error}), and $\tau_r$ is the inlier-outlier threshold. The use of the truncated quadratic function is inspired by MSAC~\cite{torr1998robust,lebeda2012fixing}.
At each pixel, the quality $q$ considers only the most accurate potential correspondence as only up to one may be compatible with the hypothesis; the
others
provide alternative explanations and do not influence the quality.

GC-RANSAC runs for up to $\tau_i$ iterations until the quality $q$ of a pose hypothesis reaches a threshold $\tau_q$. The hypothesis with the highest $q$ defines the final outcome.

\subsection{Multi-Instance Fitting} \label{sec:epos_multi_instance_fitting}

The 6D poses of multiple instances of an object $o$ are estimated from correspondences $C_o$ by the Progressive-X fitting scheme~\cite{barath2019progx}. In this scheme, pose hypotheses are proposed one by one and maintained in a set $H$.
The hypotheses are proposed by the single-instance fitting method (described in Section~\ref{sec:single_instance_fitting}) with the quality function modified to encourage exploring correspondences that have not been explained by any hypothesis from $H$:
\begin{align} \label{eq:pose_quality_multi}
	\begin{split}
	q\big(\hat{\textbf{R}},\,& \hat{\textbf{t}}, H, U_o, C_{o,\mathbf{u}}, \tau_r\big) \\
	& = \frac{1}{|U_o|} \sum_{\mathbf{u} \in U_o} \max_{(\mathbf{u}, \mathbf{x}, \cdot) \in C_{o,\mathbf{u}}} \max\left(0, \, \min\left( 1 - \frac{e\big(\mathbf{u}, \mathbf{x}, \hat{\textbf{R}}, \hat{\textbf{t}} \big)^2}{\tau_r^2}, \, \frac{e'\big(\mathbf{u}, \mathbf{x}, H \big)^2}{\tau_r^2} \right)\right)
	\end{split}
\end{align}

\noindent where the function $e'$ is defined by the minimum re-projection error over the set $H$:
\begin{align}
	e'\big(\mathbf{u}, \mathbf{x}, H \big) = \min\limits_{[\textbf{R} \, | \, \textbf{t}] \in H} e\big(\mathbf{u}, \mathbf{x}, \textbf{R}, \textbf{t} \big)
\end{align}

A proposed pose hypothesis is added to $H$ if the Jaccard similarity~\cite{toldo2008robust}, calculated between the set of inliers of the proposed hypothesis and the set of inliers of all hypotheses from $H$, is below a threshold $\tau_j$.
After adding a hypothesis, the set $H$ is consolidated by the PEARL convex optimization~\cite{isack2012energy}, which minimizes an energy balancing geometric errors and regularity of inlier clusters by alternating between two steps:
(i) selecting inliers for each hypothesis from $H$ by the alpha-expansion optimization, and (ii)~refining all hypotheses from the selected inliers.
A hypothesis is removed from $H$ when no inliers are assigned to it in the first step.
PEARL utilizes the spatial coherence of correspondences using the same neighborhood graph as GC-RANSAC and runs until convergence.

Progressive-X terminates if the probability of finding another instance~\cite{barath2019progx} falls below a threshold $\tau_p$ or if the specified number of object poses is estimated (the latter condition applies only in the case of the 6D object localization problem).

\begin{figure}[t!]
	\begin{center}
	    \begingroup
		\begin{tabular}{ @{}c@{ } @{}c@{ } @{}c@{ } }
			\includegraphics[width=0.325\linewidth]{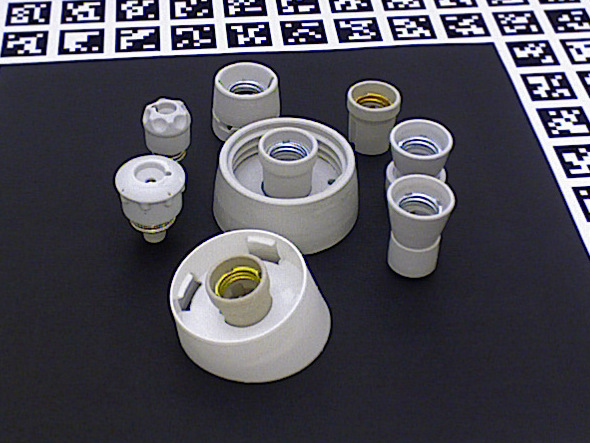} &
			\includegraphics[width=0.325\linewidth]{images/epos/qualitative/ycbv/59_001070_rgb.jpg} &
			\includegraphics[width=0.325\linewidth]{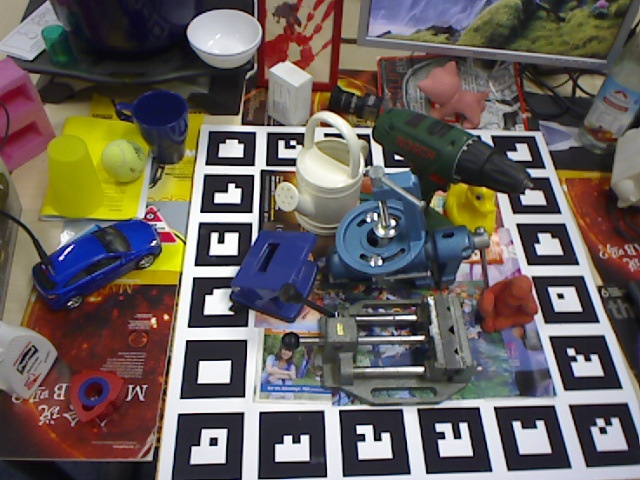} \\
			
			\includegraphics[width=0.325\linewidth]{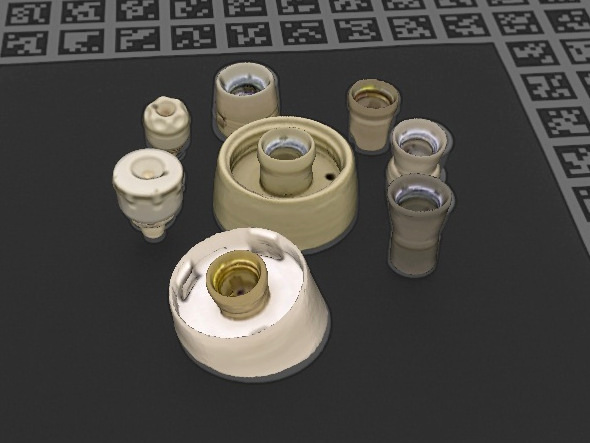} &
			\includegraphics[width=0.325\linewidth]{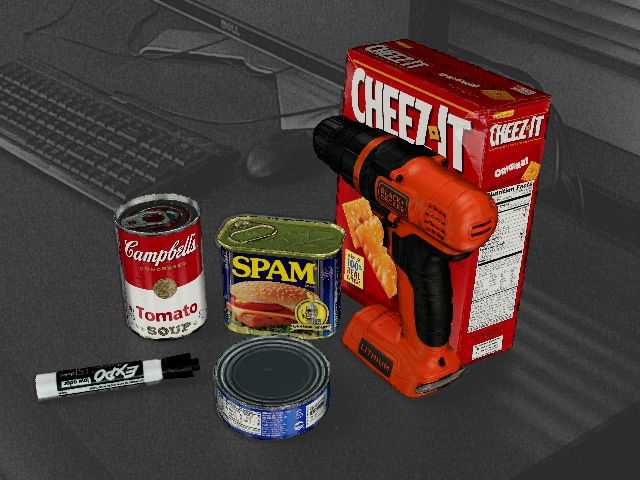} &
			\includegraphics[width=0.325\linewidth]{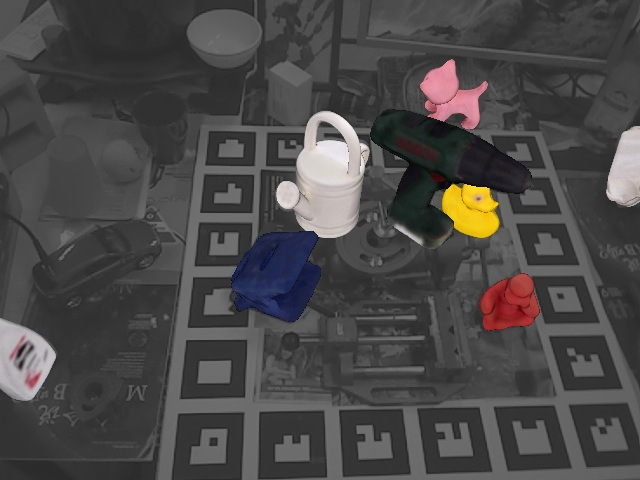} \\
		\end{tabular}
		\endgroup
		\caption{\label{fig:epos_qualitative}
			\textbf{Example EPOS results} on T-LESS (left), YCB-V (middle) and LM-O (right).
			At the top are renderings of the 3D object models in poses estimated from the RGB images at the bottom.
			All eight LM-O objects, including two truncated ones, are detected in the bottom example.
			More results, including a real-world demo, can be found on the project website.
		}
	\end{center}
\end{figure}

\section{Experimental Evaluation} \label{sec:epos_experiments}

This section compares the performance of EPOS with other 6D object pose estimation methods and presents ablation experiments demonstrating the benefit of representing an object by surface fragments, of regressing the precise 3D locations on the fragments, and of fitting the pose by a modern robust estimator.

\subsection{Experimental Setup}

\noindent
\textbf{Evaluation Methodology.}
We follow the evaluation methodology of the BOP Challenge 2019 described in Section~\ref{sec:bop_methodology}. In short, a method is evaluated on the 6D object localization problem, and the error of an estimated pose $\hat{\textbf{P}}$ \wrt the ground-truth pose $\bar{\textbf{P}}$ is calculated by three pose-error functions: Visible Surface Discrepancy (VSD), which treats indistinguishable poses as equivalent by considering only the visible object part; Maximum Symmetry-Aware Surface Distance (MSSD), which considers a set of pre-identified global object symmetries and measures the surface deviation in 3D; and Maximum Symmetry-Aware Projection Distance (MSPD), which considers the object symmetries and measures the perceivable deviation. MSPD is suitable for the evaluation of RGB methods, for which estimating the $Z$ translational component (along the optical axis) is more challenging. An estimated pose is considered correct \wrt a pose-error function~$e$, if $e < \theta_e$, where $e \in \{\text{VSD}, \text{MSSD}, \text{MSPD}\}$ and $\theta_e$ is the threshold of correctness.
The fraction of annotated object instances, for which a correct pose is estimated, is referred to as Recall.
The Average Recall \wrt a function~$e$, denoted as $\text{AR}_e$, is defined as the average of the Recall rates calculated for multiple settings of the threshold $\theta_e$ and also for multiple settings of the misalignment tolerance $\tau$ in the case of $\text{VSD}$.
The accuracy of a method is measured by the Average Recall: $\text{AR} = (\text{AR}_{\text{VSD}} + \text{AR}_{\text{MSSD}} + \text{AR}_{\text{MSPD}}) \, / \, 3$.
As EPOS uses only the RGB image channels, besides $\text{AR}$ we report $\text{AR}_{\text{MSPD}}$.

\customparagraph{Datasets.}
The experiments are conducted on datasets T-LESS~\cite{hodan2017tless}, YCB-V~\cite{xiang2017posecnn}, and LM-O~\cite{brachmann2014learning}.
The datasets include 3D object models and RGB-D images of VGA resolution with ground-truth 6D object poses (EPOS uses only the RGB channels). 
LM-O contains eight, mostly texture-less objects and 200 test images
of the objects in a clutered scene under various levels of occlusion.
YCB-V includes 21 objects, which are both textured and texture-less, and 900 test images showing the objects with occasional occlusion and limited clutter.
T-LESS contains 30 objects with no significant texture
and with symmetries and mutual similarities.
It includes 1000 test images from 20 scenes with varying complexity, including challenging scenes with multiple instances of several objects and with a high amount of clutter and occlusion.
See Section~\ref{sec:bop_datasets} for more details.

\customparagraph{Training Images.}
The network is trained on several types of
images.
For T-LESS, we use all 30K physically-based rendered (PBR) images from SyntheT-LESS~\cite{pitteri2019object}, 50K images of objects rendered by OpenGL on random photographs from NYU Depth V2~\cite{silberman2012indoor}, which we generated similarly to~\cite{hinterstoisser2017pre}, and all 38K real images from~\cite{hodan2017tless} showing objects on black background, where we replaced the background with random photographs.
For YCB-V, we use the provided 113K real and 80K synthetic images.
For LM-O, we use 67K PBR images from~\cite{hodan2019photorealistic} (scenes 1 and 2) and 50K images of objects rendered by OpenGL on random photographs. No real images of the objects are used for training on LM-O.

\customparagraph{Optimization.}
We use the DeepLabv3+ encoder-decoder network~\cite{chen2018encoder} with Xception-65~\cite{chollet2017xception} as the backbone.
The network is pre-trained on the Microsoft COCO dataset~\cite{lin2014microsoft} and fine-tuned on the training images detailed above for 2M iterations.
The training images are augmented as described in Section~\ref{sec:epos_network_training}, the batch size is set to $1$, the initial learning rate to $0.0001$, parameters of batch normalization are not fine-tuned and other hyper-parameters are set as in~\cite{chen2018encoder}.
To overcome the domain gap between the synthetic training and real test images, we apply the simple technique from~\cite{hinterstoisser2017pre} and freeze the ``early flow'' part of Xception-65.
For LM-O, we additionally freeze the ``middle flow'' since there are no real training images in this dataset.

\customparagraph{Method Parameters.}
The rates of atrous spatial pyramid pooling in DeepLabv3+ are set to $12$, $24$, and $36$, and the output stride to $8\,\text{px}$.
The spatial resolution of the output channels is doubled by bilinear interpolation, \ie, locations $\mathbf{u}$ for which the predictions are made are at the centers of $4\times4\,\text{px}$ regions of the input image.
A single network is trained for all objects in a dataset, an object is represented by $n$ = $64$ surface fragments (unless stated otherwise), and the other parameters are set as follows: $\lambda_1$~=~$1$, $\lambda_2$~=~$100$, $\tau_a$~=~$0.1$, $\tau_b$~=~$0.5$, $\tau_d$~=~$20$, $\tau_r$~=~$4\,\text{px}$, $\tau_i=400$, $\tau_q$~=~$0.5$, $\tau_t$~=~$100\,\text{px}$, $\tau_j$~=~$0.8$, $\tau_p$~=~$0.5$.

\setlength{\tabcolsep}{3pt}
\begin{table*}[t!]
	\begin{center}
		\begingroup
		\footnotesize
		\begin{tabularx}{\linewidth}{l l Y Y Y Y Y Y Y}
			\toprule
			
			\multirow{2}{*}{\vspace{-1.0ex}6D object pose estimation method} &
			\multirow{2}{*}{\vspace{-1.0ex}Image} &
			\multicolumn{2}{c}{T-LESS~\cite{hodan2017tless}} &
			\multicolumn{2}{c}{YCB-V~\cite{xiang2017posecnn}} &
			\multicolumn{2}{c}{LM-O~\cite{brachmann2014learning}} &
			\multirow{2}{*}{\vspace{-1.0ex}Time} \\
			
			\cmidrule(lr){3-4} \cmidrule(lr){5-6} \cmidrule(lr){7-8}
			
			& &
			$\text{AR}$ &
			$\text{AR}_\text{$\ast$}$ &
			$\text{AR}$ &
			$\text{AR}_\text{$\ast$}$ &
			$\text{AR}$ &
			$\text{AR}_\text{$\ast$}$ & \\
			
			\midrule

			\Rowcolor{lightgray} EPOS & RGB & \wincat{47.6} & \win{63.5} & \win{69.6} & \win{78.3} & \wincat{44.3} & \win{65.9} & $\phantom{00}$0.75 \\
			
			Zhigang-CDPN-ICCV19~\cite{li2019cdpn} & RGB & 12.4 & 17.0 & 42.2 & 51.2 & 37.4 & 55.8 & $\phantom{00}$0.67 \\
			Sundermeyer-IJCV19~\cite{sundermeyer2019augmented} & RGB & 30.4 & 50.4 & 37.7 & 41.0 & 14.6 & 25.4 & $\phantom{00}$0.19 \\
			Pix2Pose-BOP-ICCV19~\cite{park2019pix2pose} & RGB & 27.5 & 40.3 & 29.0 & 40.7 & $\phantom{0}$7.7 & 16.5 & $\phantom{00}$0.81 \\
			DPOD-ICCV19 (synthetic)~\cite{zakharov2019dpod} & RGB & $\phantom{0}$8.1 & 13.9 & 22.2 & 25.6 & 16.9 & 27.8 & $\phantom{00}$0.24 \\
			
			\cmidrule{1-9}
			
			Pix2Pose-BOP\_w/ICP-ICCV19~\cite{park2019pix2pose} & RGB-D & $\phantom{00.}$-- & $\phantom{00.}$-- & \wincat{67.5} & \wincat{63.0} & $\phantom{00.}$-- & $\phantom{00.}$-- & $\phantom{000.0}$-- \\
			Drost-CVPR10-Edges~\cite{drost2010model} & RGB-D & \wincat{50.0} & \wincat{51.8} & 37.5 & 27.5 & \wincat{51.5} & \wincat{56.9} & 144.10 \\
			F{\'e}lix\&Neves-ICRA17-IET19~\cite{rodrigues2019deep,raposo2017using} & RGB-D & 21.2 & 21.3 & 51.0 & 38.4 & 39.4 & 43.0 & $\phantom{0}$52.97 \\
			Sundermeyer-IJCV19+ICP~\cite{sundermeyer2019augmented} & RGB-D & 48.7 & 51.4 & 50.5 & 47.5 & 23.7 & 28.5 & $\phantom{00}$1.10 \\
			
			\cmidrule{1-9}
			
			Vidal-Sensors18~\cite{vidal2018method} & D & \win{53.8} & \wincat{57.4} & \wincat{45.0} & \wincat{34.7} & \win{58.2} & \wincat{64.7} & $\phantom{00}$4.93 \\
			Drost-CVPR10-3D-Only~\cite{drost2010model} & D & 44.4 & 48.0 & 34.4 & 26.3 & 52.7 & 58.1 & $\phantom{0}$10.47 \\
			Drost-CVPR10-3D-Only-Faster~\cite{drost2010model} & D & 40.5 & 43.6 & 33.0 & 24.4 & 49.2 & 54.2 & $\phantom{00}$2.20 \\

			\bottomrule
		\end{tabularx}
		\endgroup
		\caption{\label{tab:epos_bop_results} \textbf{BOP Challenge 2019}~\cite{bop19challenge,hodan2018bop} results on the T-LESS, YCB-V and LM-O datasets. Objects are represented by $64$ surface fragments in EPOS.
		Top scores among methods using the same image channels are \wincat{bold}, the best overall scores are \win{blue}. $\text{AR}_\text{$\ast$}$ represents $\text{AR}_\text{MSPD}$.
		The time (in seconds) is the average image processing time averaged over the three datasets.
		} \vspace{-0.26em}
	\end{center}
\end{table*}

\subsection{Main Results}

\noindent
\textbf{Accuracy.} Table~\ref{tab:epos_bop_results} compares the performance of EPOS with the participants of the BOP Challenge 2019~\cite{bop19challenge,hodan2018bop}.
EPOS outperforms all RGB methods on all three datasets by a large margin in both $\text{AR}$ and $\text{AR}_{\text{MSPD}}$ scores.
On the YCB-V dataset, it achieves $27\%$ absolute improvement in both scores over the second-best RGB method and also outperforms all RGB-D and D methods.
On the T-LESS and LM-O datasets, which include symmetric and texture-less objects, EPOS achieves the overall best $\text{AR}_{\text{MSPD}}$ score.
As the BOP rules require the method parameters to be fixed across datasets, Table~\ref{tab:epos_bop_results} reports scores achieved with objects from all datasets represented by $64$ fragments.
As reported in Table\ \ref{tab:epos_fragments}, increasing the number of fragments from $64$ to $256$ yields in some cases additional improvements but around double image processing time.
Note that we do not perform any post-refinement of the estimated poses, such as \cite{manhardt2018deep,li2018deepim,zakharov2019dpod,rad2017bb8}.

Results of EPOS on the latest BOP Challenge 2020~\cite{hodan2020bop} can be found in Table~\ref{tab:bop_2020_results}. EPOS was trained only on the synthetic training images provided for the 2020 edition of the challenge and all parameters were set as described above. Overall, EPOS placed 16th out of 26 methods. However, the better performing methods applied an RGB-based or depth-based post-refinement of the estimated poses, used also real training images, or trained one neural network per object instead of one network per dataset.
As other methods, EPOS would likely benefit from these design choices as well. For example, the ICP post-refinement~\cite{rusinkiewicz2001efficient} improved the accuracy of the Pix2Pose method~\cite{park2019pix2pose} by absolute $24.9\%$ and teleported this method from the 22nd to the 4th place. The importance of a post-refinement stage was demonstrated also by other methods -- top nine methods applied ICP or an RGB-based refiner, similar to DeepIM~\cite{li2018deepim}. Moreover, adding real training images improved the accuracy on datasets T-LESS, TUD-L, and YCB-V\footnote{The other datasets included in the challenge do not provide real training images.} on average by $15.8\%$ for the CosyPose method~\cite{labbe2020cosypose} and by $13.2\%$ for the CDPN method~\cite{li2019cdpn} (compare rows \#3 and \#5 and rows \#10 and \#14 in Table~\ref{tab:bop_2020_results}).
The top-performing CosyPose method identified strong augmentation of training images as one of the key components which could be used in EPOS as well.
Experiments with a post-refinement stage, real training images and the strong augmentation are left for future work.

\customparagraph{Speed.}
EPOS takes $0.75\,\text{s}$ per image on average on datasets T-LESS, YCB-V, and LM-O (Table~\ref{tab:epos_bop_results}).\footnote{With a 6-core Intel i7-8700K CPU, 64GB RAM, and Nvidia P100 GPU.}
As the other RGB methods, which are all based on convolutional neural networks, EPOS is noticeably faster than the RGB-D and D methods, which are slower typically due to an ICP post-refinement stage.
The RGB methods of~\cite{sundermeyer2019augmented,zakharov2019dpod} are $3$--$4$ times faster but significantly less accurate than EPOS.
On the seven core datasets from the BOP Challenge 2020, EPOS takes $1.87\,\text{s}$ per image on average (Table~\ref{tab:bop_2020_results}). This processing time is higher mainly because of the ITODD and IC-BIN datasets which include images with multiple object instances (the average and the maximum number of instances per image are $4.2$ and $81$ for ITODD and $12.3$ and $19$ for IC-BIN).
Depending on the application requirements, the trade-off between the accuracy and speed of EPOS can be controlled by, \eg, the number of
fragments, the network size, the image resolution, the density of pixels at which the correspondences are predicted, or the maximum
number of GC-RANSAC iterations. Besides, our implementation could be further optimized.

\subsection{Ablation Experiments}

\noindent
\textbf{Surface Fragments.}
The accuracy scores of EPOS for different numbers of surface fragments are shown in the upper half of Table~\ref{tab:epos_fragments}.
With a single fragment, the method performs direct regression of the so-called 3D object coordinates~\cite{brachmann2014learning}, similarly to~\cite{jafari2018ipose,park2019pix2pose,li2019cdpn}.
The accuracy increases with the number of fragments and reaches the peak at $64$ or $256$ fragments.
On all three datasets, the peaks of both $\text{AR}$ and $\text{AR}_\text{MSPD}$ scores are $18$--$33\%$ higher than the scores achieved with the direct regression of the 3D object coordinates.
This significant improvement demonstrates the effectiveness of fragments on various types of objects, including symmetric, texture-less, and textured objects.

On T-LESS, the accuracy drops when the number of fragments is increased from $64$ to $256$.
We suspect this is because the fragments become too small (T-LESS includes smaller objects) and training 
becomes challenging due to fewer examples per fragment.

The average number of potential correspondences increases with the number of fragments, \ie, each pixel is linked with more fragments (columns Corr.\ in Table~\ref{tab:epos_fragments}).
At the same time, the average number of fitting iterations tends to decrease (columns Iter.), \ie, a pose hypothesis with the quality~$q$ %
above threshold $\tau_q$ is found earlier.
This shows that the fitting method can benefit from knowing multiple potential correspondences per pixel.
However, although the average number of iterations decreases, the average image processing time tends to increase (with a higher number of fragments) due to a higher computational cost of the network inference and of each fitting iteration.
Setting the number of fragments to $64$ provides a practical trade-off between the speed and accuracy.

\setlength{\tabcolsep}{0pt}
\begin{table*}[!t]
	\begin{center}
		\begingroup
		\footnotesize
		\begin{tabularx}{\textwidth}{Y Y Y Y Y Y Y Y Y Y Y Y Y Y Y Y}

		    \toprule
			
			\multirow{2}{*}{\vspace{-1.0ex}{$n$}} &
			\multicolumn{5}{c}{{T-LESS~\cite{hodan2017tless}}} &
			\multicolumn{5}{c}{{YCB-V~\cite{xiang2017posecnn}}} &
			\multicolumn{5}{c}{{LM-O~\cite{brachmann2014learning}}} \\
			
			\cmidrule(lr){2-6} \cmidrule(lr){7-11} \cmidrule(lr){12-16}
			
			&
			$\text{AR}$ &
			$\text{AR}_\text{$\ast$}$ &
			$\text{Corr.}$ &
			$\text{Iter.}$ &
			$\text{Time}$ &
			$\text{AR}$ &
			$\text{AR}_\text{$\ast$}$ &
			$\text{Corr.}$ &
			$\text{Iter.}$ &
			$\text{Time}$ &
			$\text{AR}$ &
			$\text{AR}_\text{$\ast$}$ &
			$\text{Corr.}$ &
			$\text{Iter.}$ &
			$\text{Time}$ \\
			
		    \cmidrule{1-16}
			\multicolumn{16}{c}{{\hspace{10.5mm}\textit{With regression of 3D fragment coordinates}}} \\
			\cmidrule{1-16}

			\leavevmode\hphantom{00}1 & 17.2 & 30.7 & \leavevmode\hphantom{0}911 & 347 & 0.97 & 41.7 & 52.6 & 1079 & 183 & 0.56 & 26.8 & 47.5 & 237 & 111 & 0.53 \\
			\leavevmode\hphantom{00}4 & 39.5 & 57.1 & 1196 & 273 & 0.95 & 54.4 & 66.1 & 1129 & 110 & 0.52 & 33.5 & 56.0 & 267 & \leavevmode\hphantom{0}58 & 0.51 \\
			\leavevmode\hphantom{0}16 & 45.4 & 62.7 & 1301 & 246 & 0.96 & 63.2 & 72.7 & 1174 & \leavevmode\hphantom{0}71 & 0.51 & 39.3 & 61.3 & 275 & \leavevmode\hphantom{0}54 & 0.50 \\
			\leavevmode\hphantom{0}64 & \wincat{47.6} & \wincat{63.5} & 1612 & 236 & 1.18 & 69.6 & 78.3 & 1266 & \leavevmode\hphantom{0}56 & 0.57 & 44.3 & \wincat{65.9} & 330 & \leavevmode\hphantom{0}53 & 0.49 \\
			256 & 45.6 & 59.7 & 3382 & 230 & 2.99 & \wincat{71.4} & \wincat{79.8} & 1497 & \leavevmode\hphantom{0}56 & 0.94 & \wincat{46.0} & 65.4 & 457 & \leavevmode\hphantom{0}70 & 0.60 \\
			
			\cmidrule{1-16}
			\multicolumn{16}{c}{{\hspace{10.5mm}\textit{Without regression of 3D fragment coordinates}}} \\
			\cmidrule{1-16}
			
			\leavevmode\hphantom{00}1 & \leavevmode\hphantom{0}0.0 & \leavevmode\hphantom{0}0.0 & \leavevmode\hphantom{0}911 & 400 & 0.23 & \leavevmode\hphantom{0}0.0 & \leavevmode\hphantom{0}0.0 & 1079 & 400 & 0.17 & \leavevmode\hphantom{0}0.0 & \leavevmode\hphantom{0}0.0 & 237 & 400 & 0.24 \\
			\leavevmode\hphantom{00}4 & \leavevmode\hphantom{0}3.2 & \leavevmode\hphantom{0}8.8 & 1196 & 399 & 0.89 & \leavevmode\hphantom{0}3.0 & \leavevmode\hphantom{0}7.4 & 1129 & 400 & 0.53 & \leavevmode\hphantom{0}5.2 & 15.2 & 267 & 390 & 0.50 \\
			\leavevmode\hphantom{0}16 & 13.9 & 37.5 & 1301 & 396 & 1.02 & 16.1 & 36.4 & 1174 & 400 & 0.61 & 17.1 & 47.7 & 275 & 359 & 0.55 \\
			\leavevmode\hphantom{0}64 & 29.4 & 55.0 & 1612 & 380 & 1.35 & 41.5 & 66.6 & 1266 & 383 & 0.73 & 31.0 & 62.3 & 330 & 171 & 0.55 \\
			256 & 43.0 & 58.2 & 3382 & 299 & 2.95 & 64.5 & 77.7 & 1497 & 206 & 0.88 & 43.2 & 64.9 & 457 & \leavevmode\hphantom{0}72 & 0.58 \\

		    \bottomrule
		\end{tabularx}
		\endgroup
		\caption{\label{tab:epos_fragments} \textbf{Number of fragments and regression.}
		The table reports the Average Recall scores for different numbers of surface fragments ($n$) with and without regression of the 3D fragment coordinates (the fragment centers are used in the case of no regression). $\text{AR}_\text{$\ast$}$ represents $\text{AR}_\text{MSPD}$.
		The table also reports the average number of correspondences established per object model in an image, the average number of GC-RANSAC iterations to fit a single pose (both are rounded to integers), and the average image processing time (in seconds).
		}
	\end{center}
\end{table*}

\setlength{\tabcolsep}{3pt}
\begin{table*}[t!]
	\begin{center}
		\begingroup
		\footnotesize
		\begin{tabularx}{\linewidth}{l l Y Y Y Y Y Y Y}
			\toprule
			
			\multirow{2}{*}{\vspace{-1.0ex}RANSAC variant} &
			\multirow{2}{*}{\vspace{-1.0ex}Non-minimal solver} &
			\multicolumn{2}{c}{T-LESS~\cite{hodan2017tless}} &
			\multicolumn{2}{c}{YCB-V~\cite{xiang2017posecnn}} &
			\multicolumn{2}{c}{LM-O~\cite{brachmann2014learning}} &
			\multirow{2}{*}{\vspace{-1.0ex}Time} \\
			
			\cmidrule(lr){3-4} \cmidrule(lr){5-6} \cmidrule(lr){7-8}
		
			& &
			$\text{AR}$ &
			$\text{AR}_\text{$\ast$}$ &
			$\text{AR}$ &
			$\text{AR}_\text{$\ast$}$ &
			$\text{AR}$ &
			$\text{AR}_\text{$\ast$}$ & \\
			
			\midrule
			
			OpenCV RANSAC & EP\emph{n}P~\cite{lepetit2009epnp} & 35.5 & 47.9 & 67.2 & 76.6 & 41.2 & 63.5 & 0.16 \\
			MSAC~\cite{torr2002bayesian} & EP\emph{n}P~\cite{lepetit2009epnp} + L-M~\cite{more1978levenberg} & 44.3 & 61.0 & 63.8 & 73.7 & 39.7 & 61.7 & 0.49 \\
			GC-RANSAC~\cite{barath2018gcransac} & DLS-P\emph{n}P~\cite{hesch2011direct} & 44.3 & 59.5 & 67.5 & 76.1 & 35.6 & 53.9 & 0.53 \\
			GC-RANSAC~\cite{barath2018gcransac} & EP\emph{n}P~\cite{lepetit2009epnp} &  46.9 & 62.6 & 69.2 & 77.9 & 42.6 & 63.6 & 0.39 \\
			GC-RANSAC~\cite{barath2018gcransac} & EP\emph{n}P~\cite{lepetit2009epnp} + L-M~\cite{more1978levenberg} & \wincat{47.6} & \wincat{63.5} & \wincat{69.6} & \wincat{78.3} & \wincat{44.3} & \wincat{65.9} & 0.52 \\
			\bottomrule
		\end{tabularx}
		\endgroup
		\caption{\label{tab:epos_ransac_variants} 
		\textbf{RANSAC variants and non-minimal solvers.}
		The P3P solver~\cite{kneip2011novel} is used to estimate the pose from a minimal sample of 2D-3D correspondences.
		The non-minimal solvers are applied when estimating the pose from a larger-than-minimal sample.
		The reported time~(s) is the average time to fit poses of all object instances in an image averaged over the datasets.
		}
	\end{center}
\end{table*}

\customparagraph{Regression of 3D Fragment Coordinates.}
The upper half of Table~\ref{tab:epos_fragments} shows scores achieved with regressing the precise 3D locations, while the lower half shows scores achieved with the same network models but using the fragment centers (Section~\ref{sec:epos_fragments}) instead of the regressed locations.
Without the regression, the scores increase with the number of fragments as the deviation of the fragment centers from the true corresponding 3D locations decreases.
However, the accuracy is often noticeably lower than with the regression.
With a single fragment and without the regression, all pixels are linked to the same fragment center and all samples of three correspondences are immediately rejected because they fail the non-collinearity test, hence the low processing time.

Even though the regressed locations are not guaranteed to lie on the model surface, their average distance from the surface was measured to be less than $1\,\text{mm}$ (with $64$ and $256$ fragments), which is negligible compared to the object sizes.
No improvement was observed when the regressed locations were replaced by the closest points on the surface.

\customparagraph{Robust Pose Fitting.}
Table\ \ref{tab:epos_ransac_variants} evaluates several methods for robust pose estimation from 2D-3D correspondences: RANSAC~\cite{fischler1981random} from OpenCV (function \texttt{solvePnPRansac}), MSAC~\cite{torr2000mlesac}, and GC-RANSAC~\cite{barath2018gcransac}.
The methods were evaluated within the Progressive-X scheme (Section~\ref{sec:epos_multi_instance_fitting}), with the P3P solver~\cite{kneip2011novel} to estimate the pose from a minimal sample, \ie, three correspondences, and with several solvers to estimate the pose from a non-minimal sample.
In OpenCV RANSAC and MSAC, the non-minimal solver refines the pose from all inliers.
In GC-RANSAC, it is additionally used in the graph-cut-based local optimization which is applied when a new so-far-the-best pose is found.
We tested OpenCV RANSAC with all available non-minimal solvers and achieved the best scores with EP\emph{n}P~\cite{lepetit2009epnp}.
The top-performing
method on all datasets is GC-RANSAC with EP\emph{n}P followed by the Levenberg-Marquardt optimization~\cite{more1978levenberg} as the non-minimal solver.
Note the improvement in accuracy, especially on T-LESS, over OpenCV RANSAC which is used by many recent correspondence-based methods, \eg, \cite{park2019pix2pose,zakharov2019dpod,rad2017bb8,tekin2018real}.

	\chapter[ObjectSynth:\ Synthesis of Photorealistic Training Images]{ObjectSynth\\{\Large Synthesis of Photorealistic Training Images}} \label{ch:synthesis}

\begin{figure*}[t!]
	\begin{center}
		\begin{tabular}{ @{}c@{ } @{}c@{ } @{}c@{ } @{}c@{ } }
			\includegraphics[width=0.2435\linewidth]{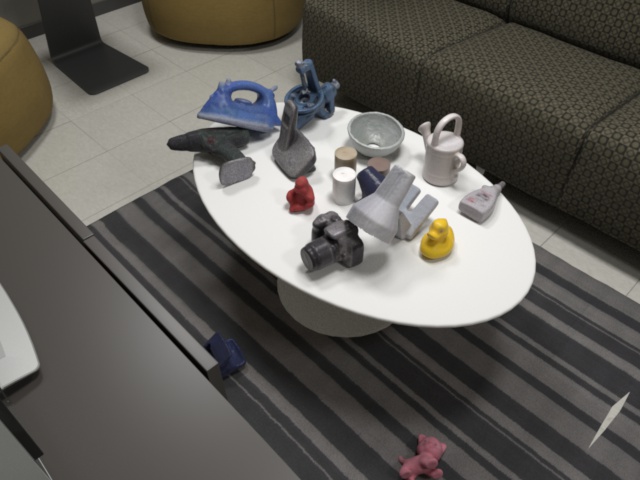} &
			\includegraphics[width=0.2435\linewidth]{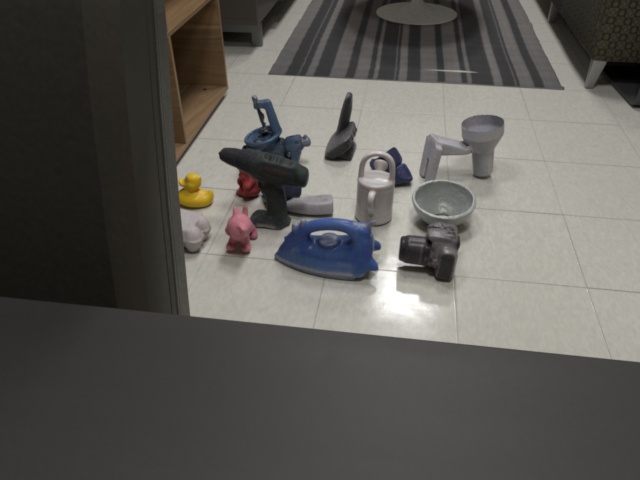} &
			\includegraphics[width=0.2435\linewidth]{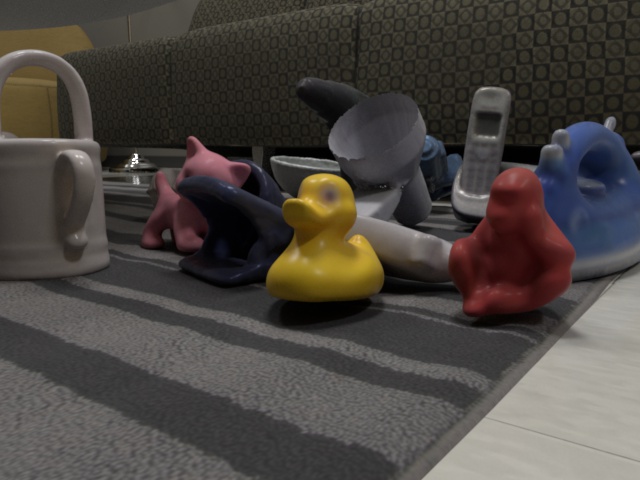} &
			\includegraphics[width=0.2435\linewidth]{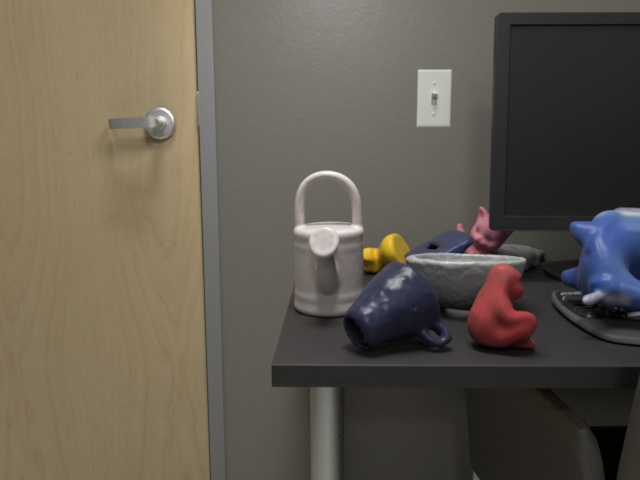} \\
			\includegraphics[width=0.2435\linewidth]{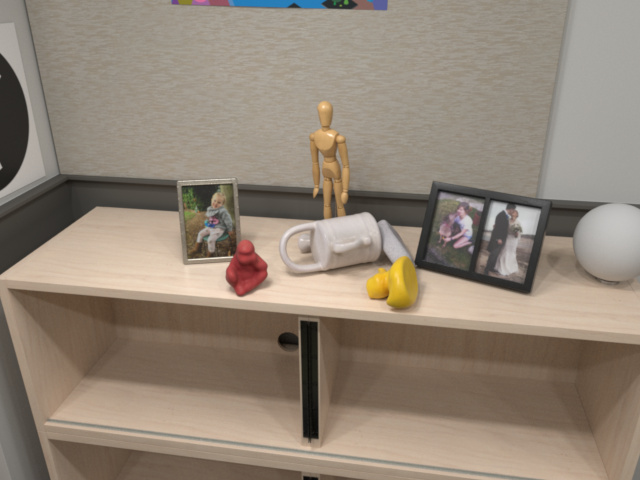} &
			\includegraphics[width=0.2435\linewidth]{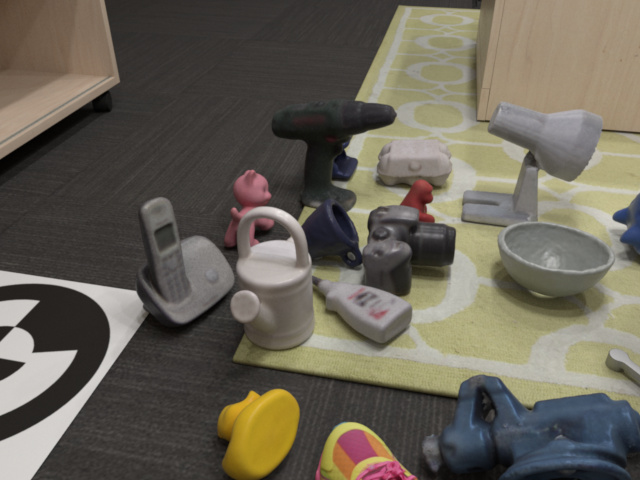} &
			\includegraphics[width=0.2435\linewidth]{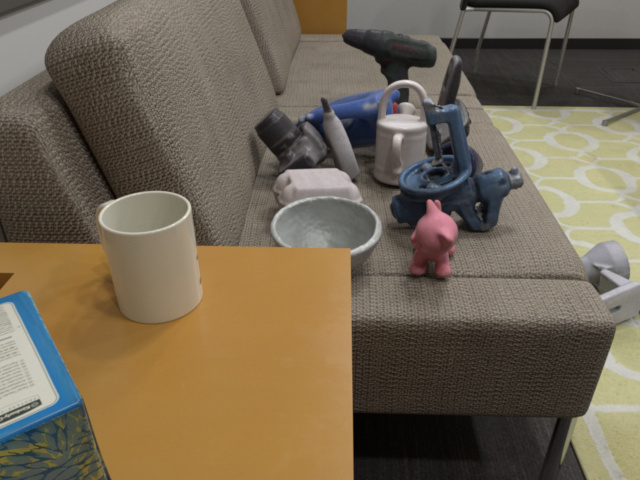} &
			\includegraphics[width=0.2435\linewidth]{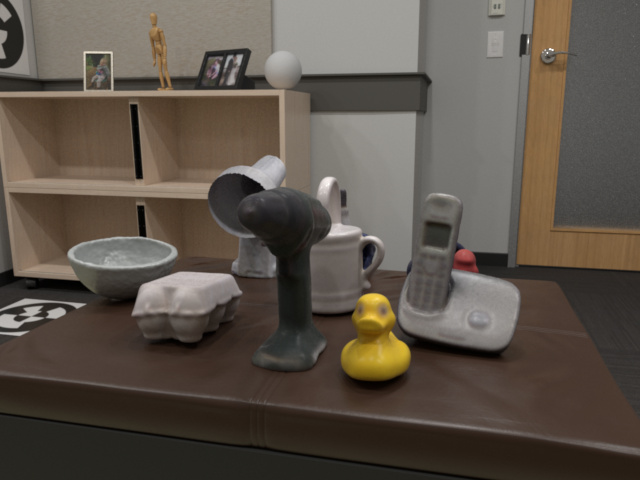} \\
			\includegraphics[width=0.2435\linewidth]{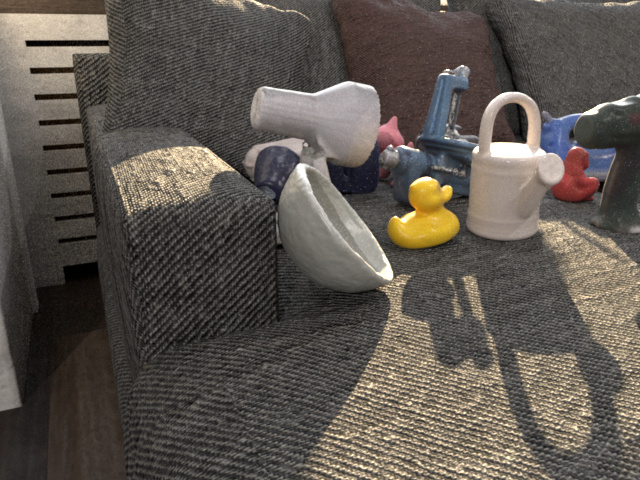} &
			\includegraphics[width=0.2435\linewidth]{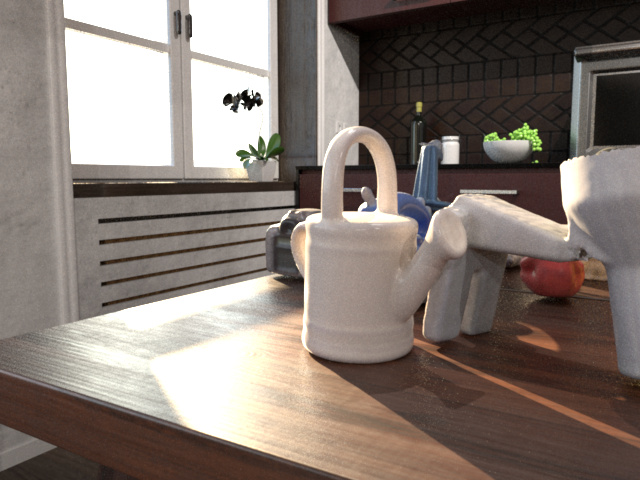} &
			\includegraphics[width=0.2435\linewidth]{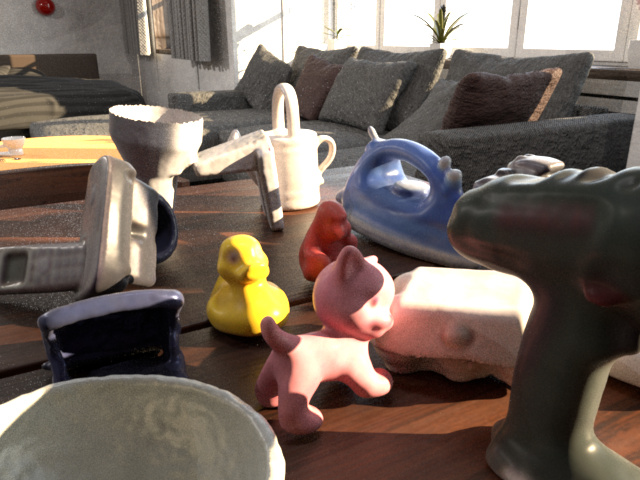} &
			\includegraphics[width=0.2435\linewidth]{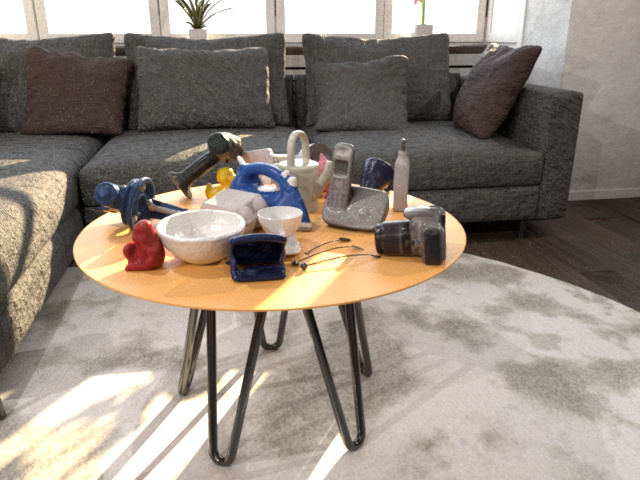} \\
			\includegraphics[width=0.2435\linewidth]{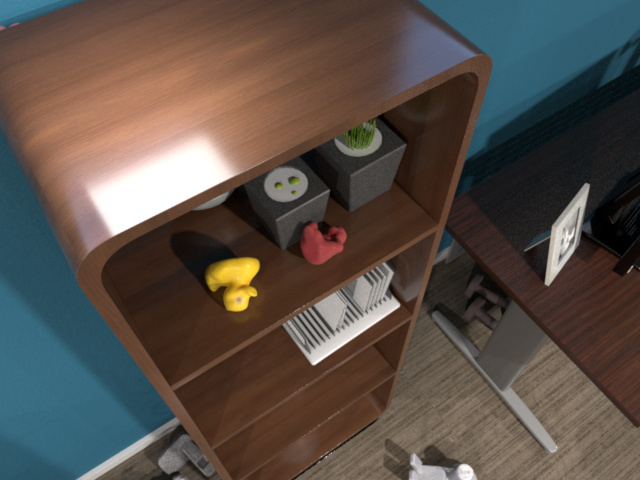} &
			\includegraphics[width=0.2435\linewidth]{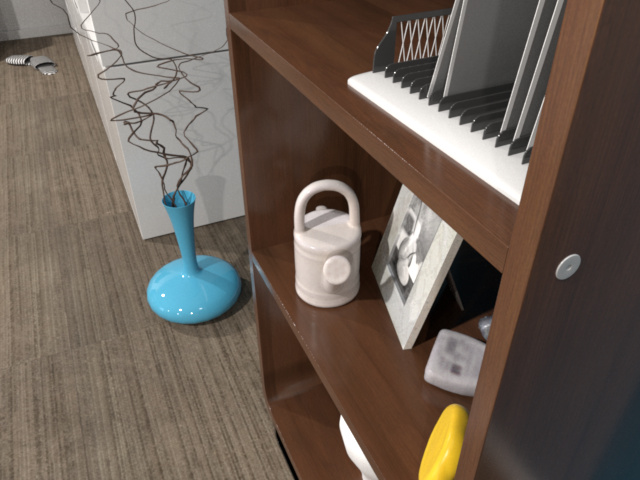} &
			\includegraphics[width=0.2435\linewidth]{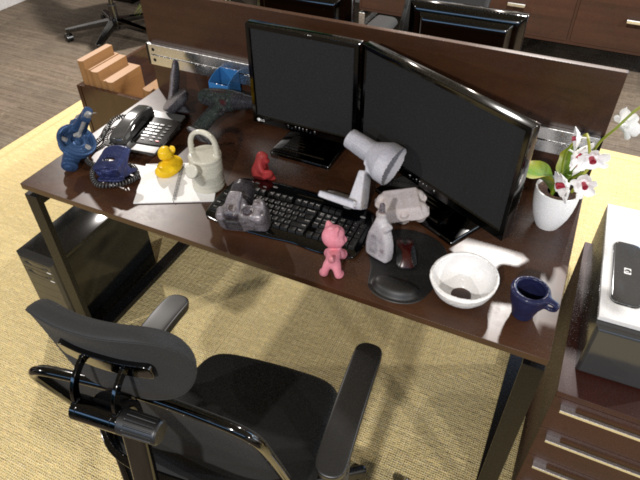} &
			\includegraphics[width=0.2435\linewidth]{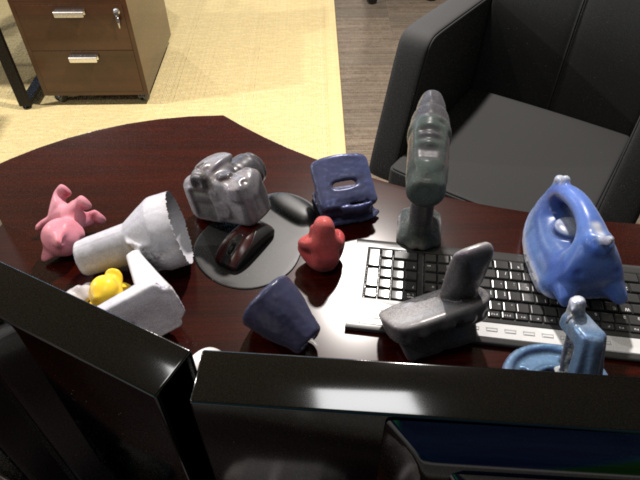} \\
			\includegraphics[width=0.2435\linewidth]{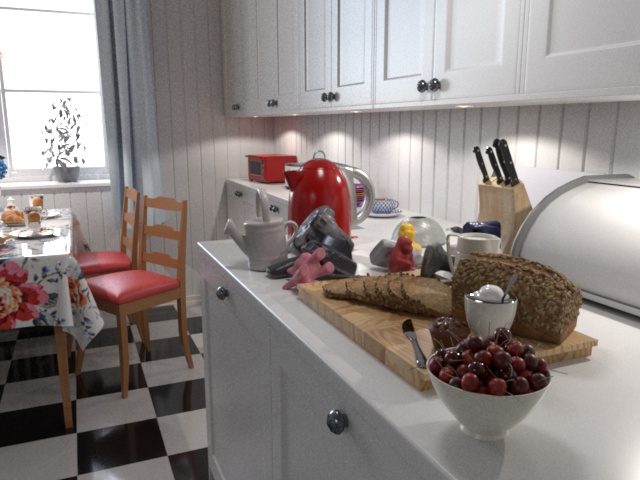} &
			\includegraphics[width=0.2435\linewidth]{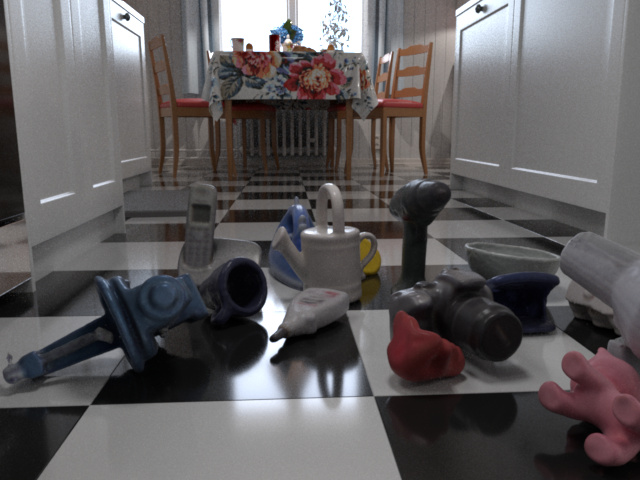} &
			\includegraphics[width=0.2435\linewidth]{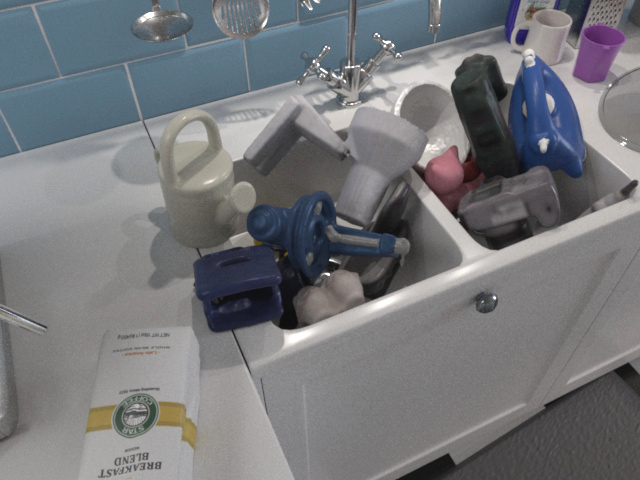} &
			\includegraphics[width=0.2435\linewidth]{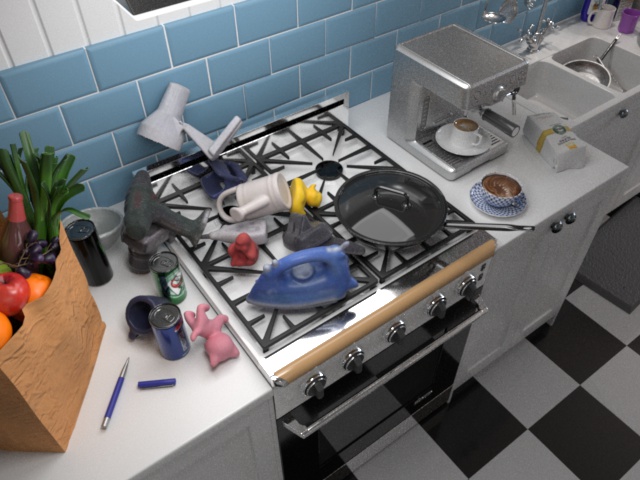} \\
			
			\includegraphics[width=0.2435\linewidth]{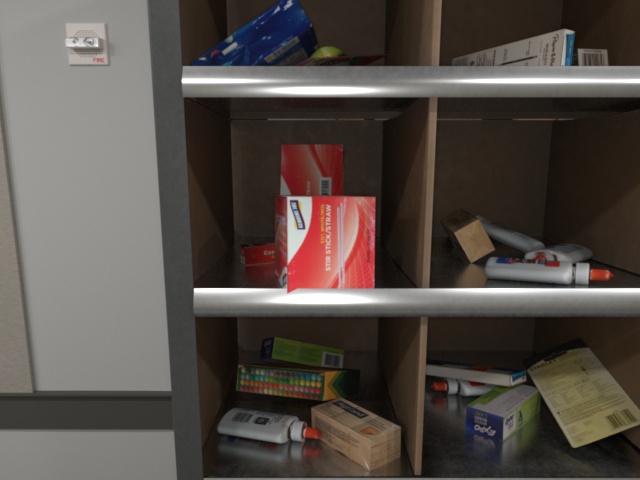} &
			\includegraphics[width=0.2435\linewidth]{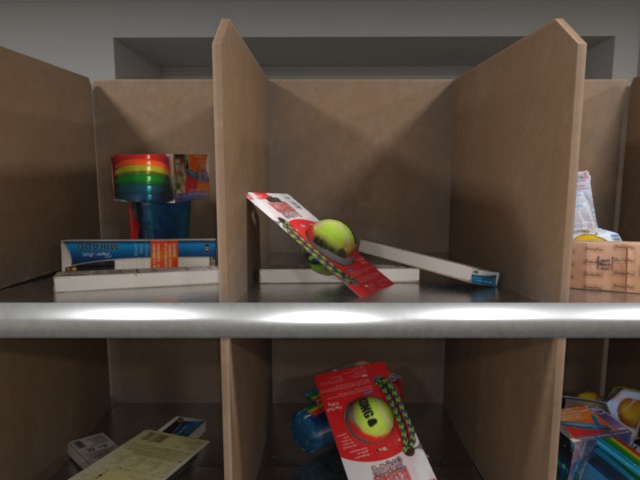} &
			\includegraphics[width=0.2435\linewidth]{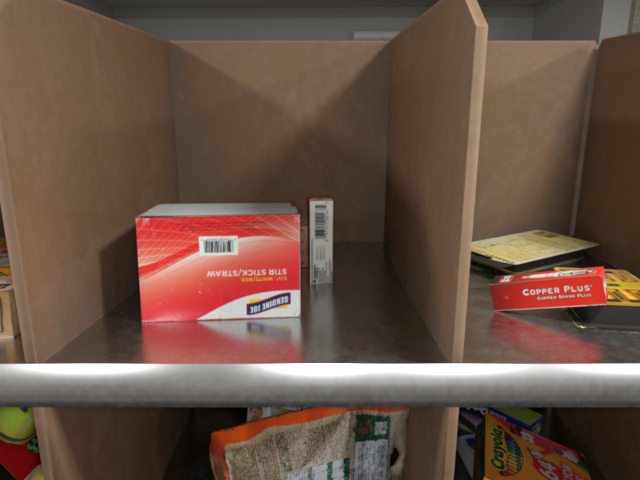} &
			\includegraphics[width=0.2435\linewidth]{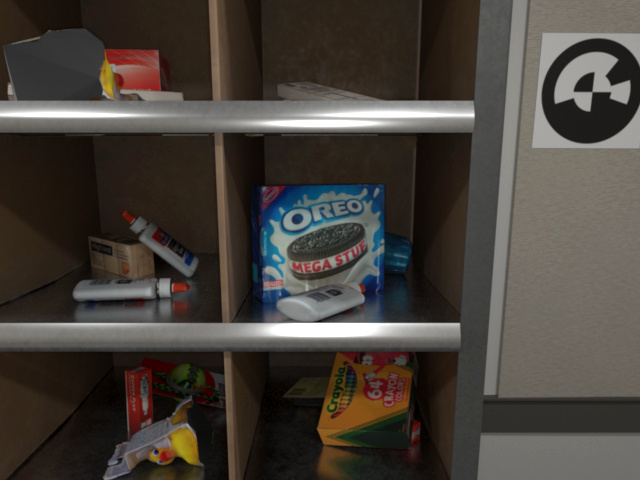} \\
		\end{tabular}
		\caption{\label{fig:synth_pbr_examples}
			\textbf{Example images synthesized by ObjectSynth.}
			The top five rows show images of the LM objects~\cite{hinterstoisser2012accv}, the bottom row shows images of the RU-APC objects~\cite{rennie2016dataset}.
			The images
			are annotated with 2D bounding boxes, masks and 6D poses of the objects. %
		} \vspace{-2.0ex}
	\end{center}
\end{figure*}

This chapter presents an approach to synthesize highly photorealistic images of 3D object models, which are experimentally shown effective for training deep neural networks for %
detecting the objects and estimating their pose from real test images.
The proposed approach has three key ingredients.
First, 3D models of objects are arranged in 3D models of complete indoor scenes with realistic materials and lighting.
Second, plausible geometric configurations of objects and cameras in the scenes are generated by physics simulation. And, third, a high degree of visual realism is achieved by physically-based rendering (PBR).
Example images synthesized by the proposed approach are in Figure~\ref{fig:synth_pbr_examples}.

The experiments show that the Faster R-CNN object detector~\cite{ren2017faster} trained on the synthesized images achieves a 24\% absolute improvement in mAP@.75IoU on real test images from the RU-APC dataset~\cite{rennie2016dataset}, and 11\% on real test images from the LM-O dataset~\cite{brachmann2014learning,hinterstoisser2012accv}.
The improvements are relative to a baseline where the training images are synthesized by rendering object models on top of random photographs -- similar images are commonly used for training methods for
tasks such as object instance detection~\cite{dwibedi2017cut}, object instance segmentation~\cite{hinterstoisser2017pre}, and 6D object pose estimation~\cite{kehl2017ssd,rad2017bb8}.

Modeling of objects and scenes is described in Section~\ref{sec:synth_modeling}, arranging the object models in the scene models and generating camera poses in Section~\ref{sec:synth_composition}, and rendering in Section~\ref{sec:synth_pbr}. The effectiveness of the photorealistic training images is evaluated in Section~\ref{sec:synth_experiments}. A refined version of the proposed approach, which was used to synthesize training images for participants of the BOP Challenge 2020~(Chapter~\ref{ch:bop}), is described in Section~\ref{sec:synth_bop20}.

Tom{\'a}{\v{s}} Hoda{\v{n}} was working on ObjectSynth during his internship at Microsoft Research in Redmond in 2018. The work was published in~\cite{hodan2019photorealistic}.
The project website with a dataset of 400K photorealistic images is available at: \texttt{\href{https://thodan.github.io/objectsynth/}{thodan.github.io/objectsynth}}, and a dataset of 350K images synthesized for the BOP Challenge 2020 at: \texttt{\href{https://bop.felk.cvut.cz/}{bop.felk.cvut.cz}}.

\section{Scene and Object Modeling} \label{sec:synth_modeling}

\begin{figure}[t!]
	\begin{center}
		\includegraphics[width=0.8\columnwidth]{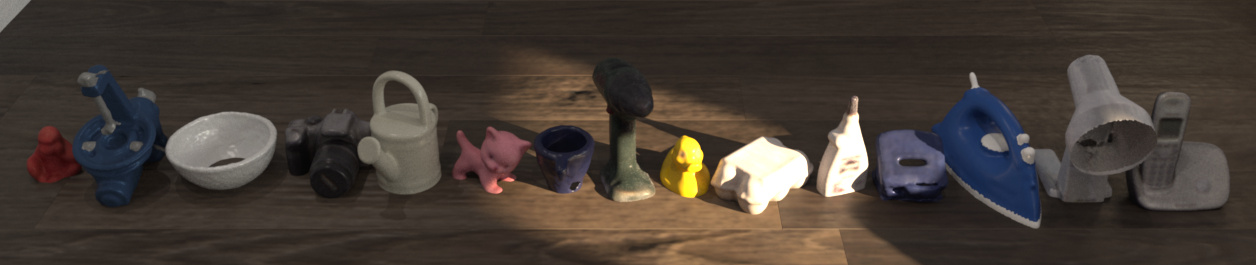} \\
		\vspace{0.5ex}
		\includegraphics[width=0.8\columnwidth]{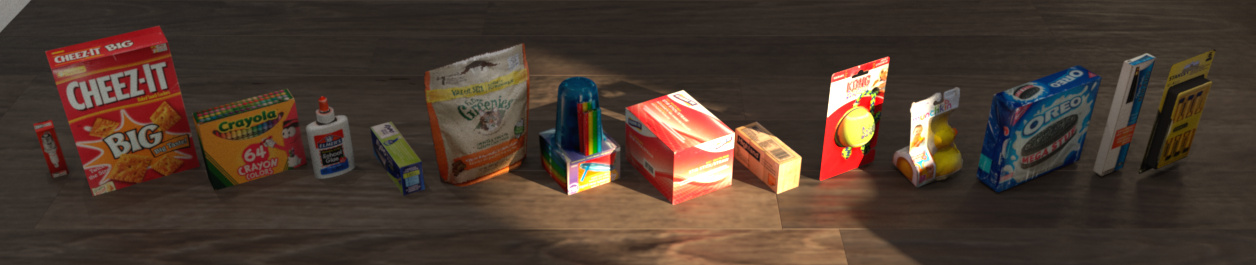} \\
		\vspace{0.7ex}
		\begin{tabular}{ @{}c@{ } @{}c@{ } @{}c@{ } }
			\small{Scene 1} & \small{Scene 2} & \small{Scene 3} \\
			\includegraphics[width=0.3265\columnwidth]{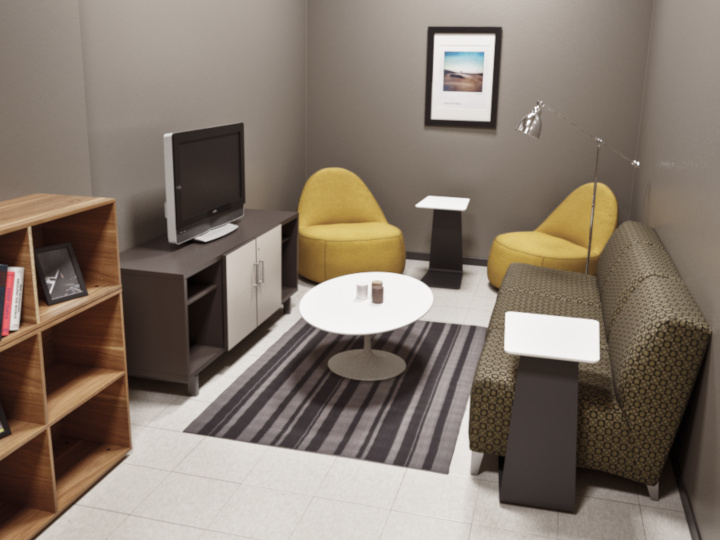} &
			\includegraphics[width=0.3265\columnwidth]{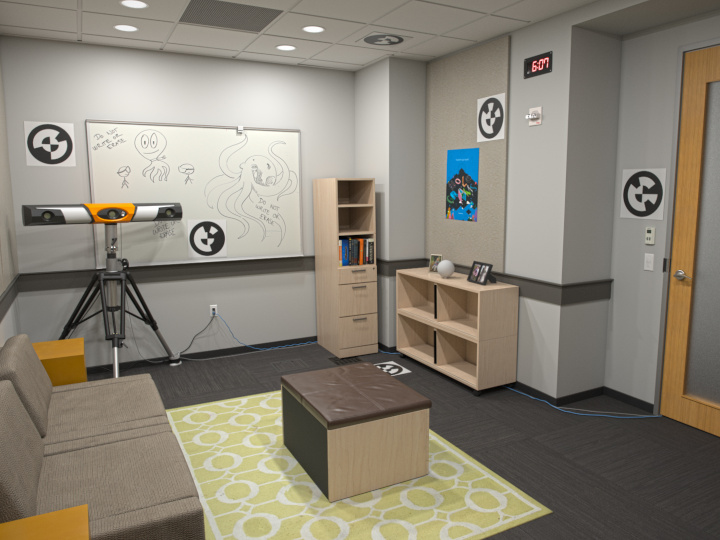} &
			\includegraphics[width=0.3265\columnwidth]{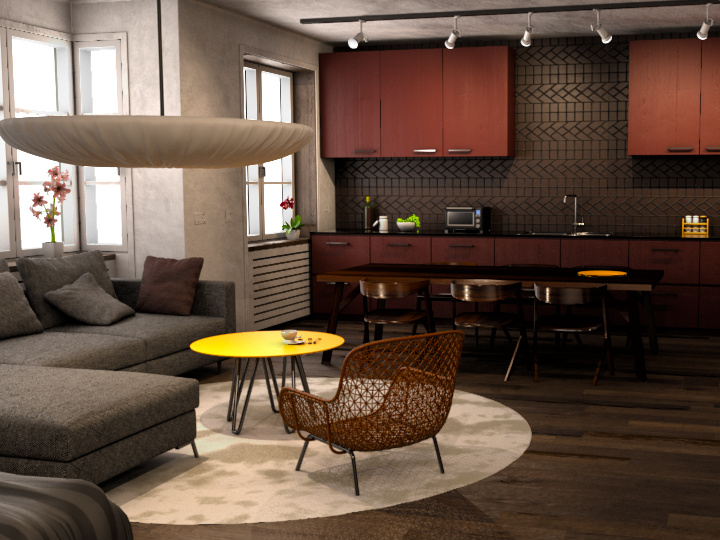} \\
			\small{Scene 4} & \small{Scene 5} & \small{Scene 6} \\
			\includegraphics[width=0.3265\columnwidth]{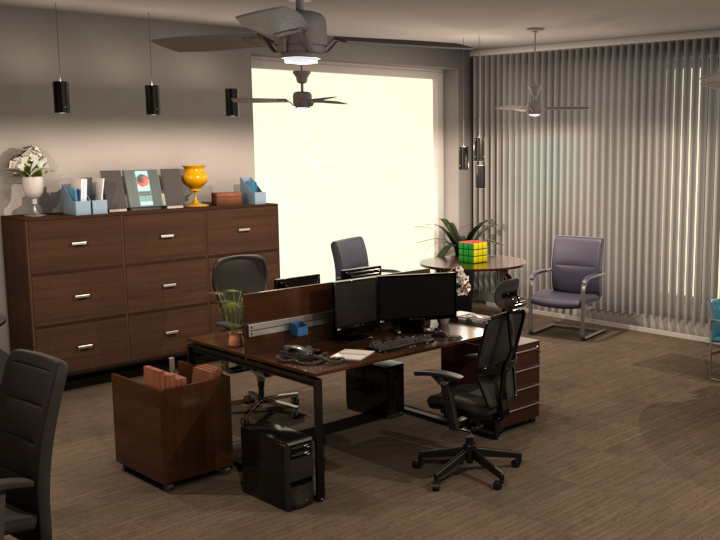} &
			\includegraphics[width=0.3265\columnwidth]{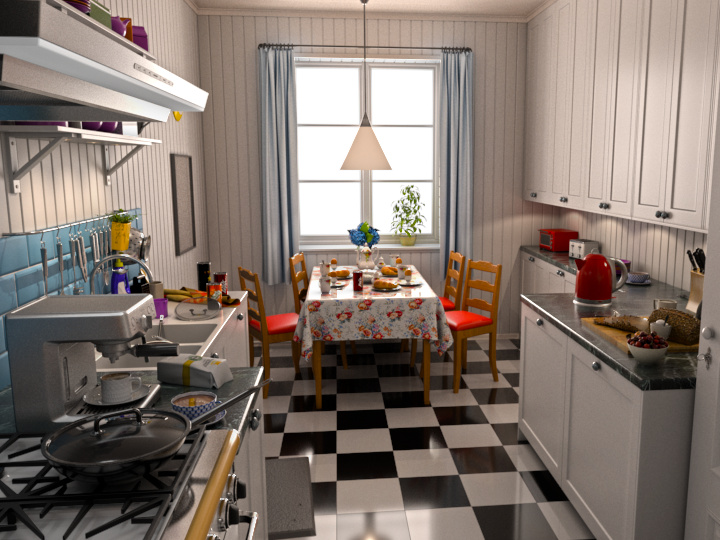} &
			\includegraphics[width=0.3265\columnwidth]{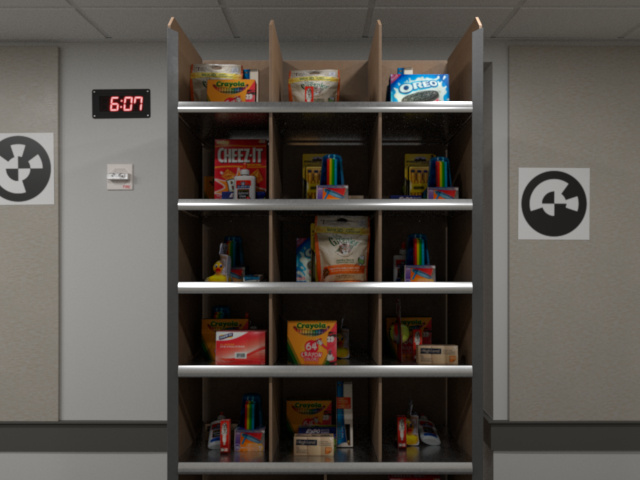}
		\end{tabular}
		\caption{\label{fig:synt_models} \textbf{3D models of objects and scenes.} Objects from the LM dataset~\cite{hinterstoisser2012accv} (1st row) were rendered in Scenes 1--5 and objects from RU-APC~\cite{rennie2016dataset} (2nd row) in Scenes 3 and 6.
		}
	\end{center}
\end{figure}

Achieving a high degree of visual realism requires 3D models that are accurate in terms of geometry, texture, and material.
We worked with 3D models of 15 objects from the Linemod (LM) dataset~\cite{hinterstoisser2012accv} and 14 objects from the Rutgers APC (RU-APC) dataset~\cite{rennie2016dataset} (Figure~\ref{fig:synt_models}, top).
We started with the refined versions of the object models from BOP~\cite{hodan2018bop}, provided as color meshes with surface normals, and manually assigned PBR materials to the models which match the appearance of the objects in real images. The material properties were controlled by the roughness, specular, and metallic
parameters.\footnote{The roughness parameter controls the range of angles in which the reflected light is scattered, which influences the sharpness of the reflection. The specular parameter controls the facing (along normal) reflectivity. The metallic parameter controls the blending factor of the metallic and non-metallic material models.
In the case of a metallic material, the base color controls the color of the specular reflection and most light is reflected as specular reflections. Non-metallic materials have specular reflections that are the same color as the incoming light and barely reflects when looking at the surface face-on~\cite{unrealenginepbr}.} Since objects from both datasets are mostly texture-less and of a homogeneous material, the same material properties were assigned to the whole model surface. A Lambertian material was assigned to the RU-APC models as the objects are mostly made of cardboard.

The 3D object models were arranged
in 3D models of six furnished scenes (Figure~\ref{fig:synt_models}, bottom).
Scenes 1--5 represent the office and household environments and include fine details and typical objects, \eg, the kitchen (Scene~5) contains dishes in a sink or a bowl of cherries. Scene~6 contains a shelf from the Amazon Picking Challenge 2015~\cite{yu2016summary}.

The scene models were created using standard 3D tools, primarily Autodesk Maya, by artists at Microsoft Research.
Scenes 1 and 2 are reconstructions of real-world environments obtained using LiDAR and photogrammetry 3D scans which served as a
guide for artists.
Materials were recreated using photographic reference, PBR material scanning~\cite{muravision}, and color swatch samples~\cite{nixsensor}.
Scenes 3--5 were purchased online~\cite{evermotion}, their geometry and materials were refined, and clutter and chaos was added to mimic a real environment.
A 3D geometry model of the shelf in Scene~6 was provided in the Amazon Picking Challenge 2015~\cite{yu2016summary}.
Reference imagery of the shelf was used to create textures and materials that
match its appearance.
Exterior light coming to the scene through windows was modeled with Arnold Physical Sky~\cite{georgiev2018arnold}
which can accurately depict atmospheric effects and time-of-day variation. Interior lights
were modeled with standard light sources such as area and point lights.

\section{Scene and Object Composition} \label{sec:synth_composition}

Multiple stages to arrange the objects on were manually selected in each scene. A stage is defined by a polygon and is typically located on tables, chairs and other places with a distinct structure, texture or illumination. Placing objects on such locations maximizes the diversity of the rendered images.
One stage per shelf bin was created in Scene~6.

An arrangement of a set of objects was generated in two steps: (1) poses of the 3D object models were initialized above a randomly selected stage, and (2) a physically plausible arrangement was reached by
simulating the objects falling on the stage under gravity and undergoing mutual collisions.
The poses were initialized by FLARE~\cite{gal2014flare}, a rule-based system for generation of object layouts for augmented reality applications, and physics was simulated by NVIDIA PhysX.
The initial height of the objects above the stage was randomly sampled from $5$ to $50\,$cm. For the LM objects, one instance per object model was added to the stage and initialized with the canonical orientation, \ie, the cup was up-right, the cat was standing on her legs, \etc. For the RU-APC objects, up to five instances per object model were added to the stage and initialized with random orientations.

Multiple cameras were positioned around each object arrangement.
Instead of fitting all objects within the camera frustum, the camera was pointed at a randomly selected object. This allowed for better control of the scale of the rendered objects.
The azimuth angles, elevation angles, and distances of the cameras were uniformly sampled from ranges determined by the ground-truth 6D object poses from the test images.
A mask of the focused object and a mask of its visible part were rendered before rendering the RGB image.
The RGB image was rendered only if at least $30\%$ of the object was visible.

\setlength{\tabcolsep}{3pt}
\begin{figure*}[t!]
	\begin{center}
		\begingroup
		\footnotesize
		\begin{tabularx}{0.85\linewidth}{l Y Y Y Y Y Y}
			\toprule
			Quality & AA & D rays & S rays & D depth & S depth & Max~depth \\
			\midrule
			Low & 1 & 1 & 1 & 1 & 1 & 2\\
			Medium & 9 & 4 & 4 & 3 & 2 &3 \\
			High & 25 & 9 & 4 & 3 & 3 & 4 \\
			\midrule
			BlenderProc4BOP & 50 & 1 & 1 & 3 & 0 & 3 \\
			\bottomrule
		\end{tabularx}
		\endgroup
		\captionof{table}{\label{tab:synth_arnoldsampling}
			\textbf{Ray-tracing parameters~\cite{georgiev2018arnold} for different quality settings.} AA is the number of rays sampled per image pixel. D/S rays is the number of indirect diffuse/specular rays sampled
			when a ray from the camera hits a diffuse/specular surface.
			After the first reflection, only a single indirect diffuse/specular ray is sampled.
			D/S depth is the maximum number of diffuse/specular reflections, and Max depth is the maximum number of reflections in total. Besides indirect rays that reflect from other scene elements, the color of a surface is determined by direct rays from light sources. BlenderProc4BOP is discussed later in Section~\ref{sec:synth_bop20}.
		}
	\end{center}
	
	\begin{center}
		\begin{tabular}{ @{}c@{ } @{}c@{ } @{}c@{ } }
			\includegraphics[width=0.3265\linewidth]{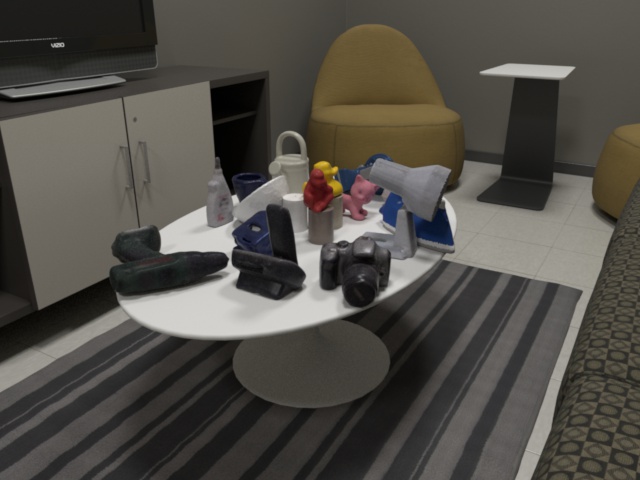} &
			\includegraphics[width=0.3265\linewidth]{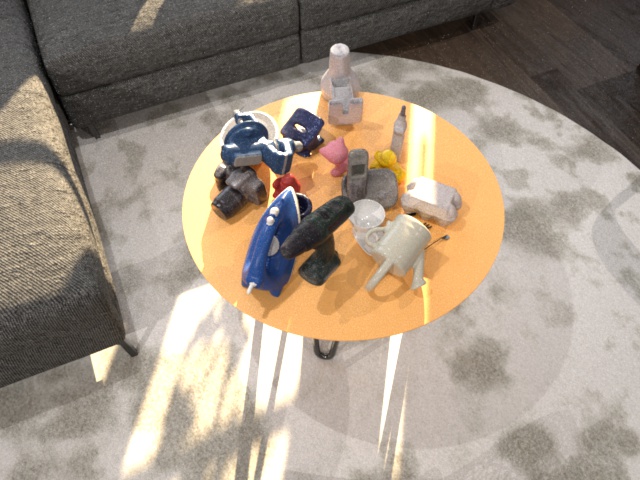} &
			\includegraphics[width=0.3265\linewidth]{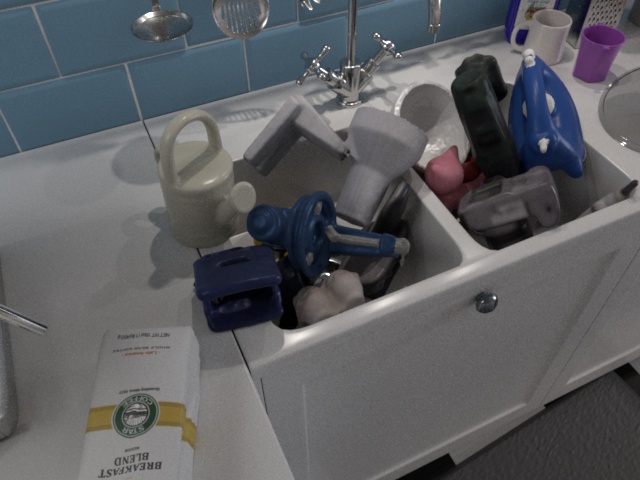} \\
			\includegraphics[width=0.3265\linewidth]{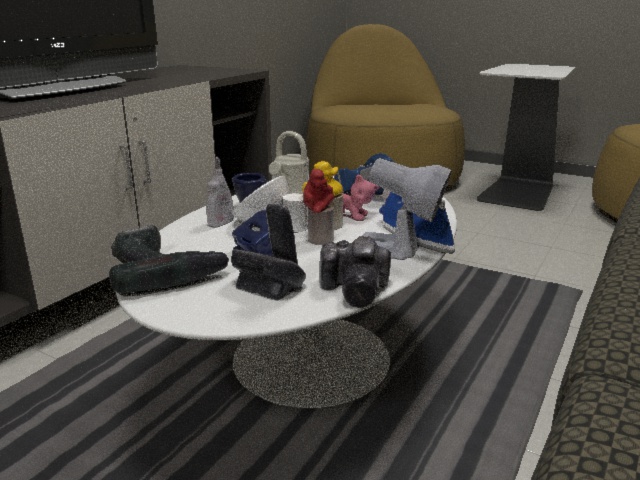} &
			\includegraphics[width=0.3265\linewidth]{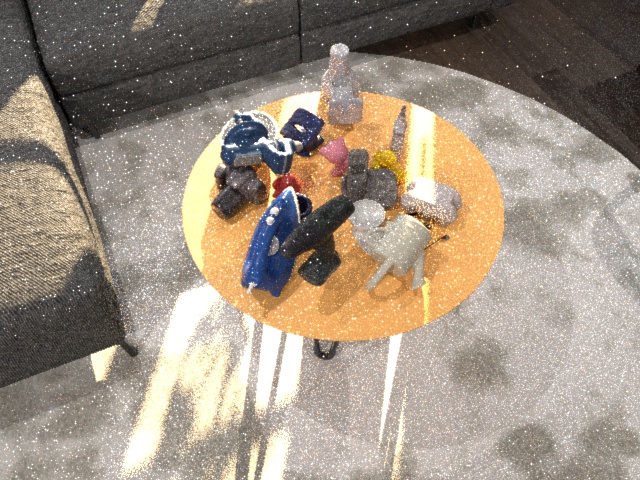} &
			\includegraphics[width=0.3265\linewidth]{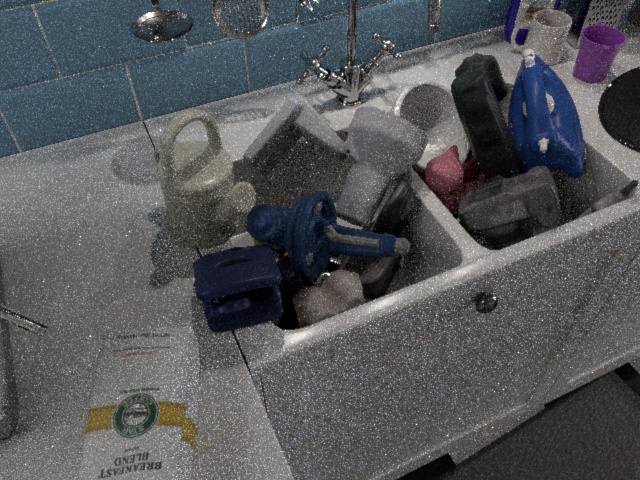} \\
		\end{tabular}
		\caption{\label{fig:synth_pbr_quality}
			\textbf{High \vs low rendering quality.}
			The same images rendered at the high (top) and the low (bottom) quality settings of the Arnold physically-based renderer~\cite{georgiev2018arnold}. Notice the noise due to insufficient illumination sampling in the low-quality images.
		}
	\end{center}
\end{figure*}

\section{Physically-Based Rendering} \label{sec:synth_pbr}

Besides accurate 3D models, achieving a high degree of visual realism requires accurate simulation of lighting, including soft shadows, reflections, refractions and indirect light bounces. %
For the experiments in Section~\ref{sec:synth_experiments}, the images were synthesized by Arnold~\cite{georgiev2018arnold}, a physically-based renderer
simulating the flow of light energy in the scene by ray~tracing.

Three images were rendered
per camera in a low, medium and high quality settings.
The mapping of the settings to Arnold parameters is in Table~\ref{tab:synth_arnoldsampling}.
Increasing the number of rays sampled per pixel (AA) reduces aliasing artifacts due to insufficient sampling of geometry and noise due to insufficient sampling of illumination. Increasing the number of indirect diffuse rays sampled when a ray from the camera hits a diffuse surface (D rays) and
and from a specular surface (S rays) reduces illumination noise. Increasing the maximum number of diffuse and specular reflections
(D/S depth)
improves the accuracy of the illumination
by gathering more of the light energy bounced around in the scene.

The images were rendered on 16-core Intel Xeon 2.3GHz processors with 112GB RAM. GPU's were not used. The average rendering time per image in VGA resolution was $15\,$s in low, $120\,$s in medium, and $720\,$s in high quality.
A CPU cluster with 400 nodes allowed us to render 2.3M images in low, 288K in medium, and 48K in high quality within a day.

A set of 200K images in low and 200K images in high quality is
available on the project website. 
LM objects are rendered in Scenes 1--5 and RU-APC objects in Scenes 3 and 6.
Each instance is annotated with the 2D bounding box, segmentation mask and 6D pose.

\section{Experiments} \label{sec:synth_experiments}

The experiments presented in this section evaluate the effectiveness of PBR images for training the Faster R-CNN object detector~\cite{ren2017faster}.
Specifically, the experiments focus on three aspects:
(1)~the importance of PBR images over the commonly used images of objects rendered on top of random photographs, (2)~the importance of the high PBR quality, and (3)~the importance of accurately modeling the scene context.

\subsection{Experimental Setup}

\begin{figure}[t!]
	
	\begin{center}
		\begingroup
		\footnotesize
		\begin{tabularx}{0.63\linewidth}{l Y Y Y Y Y Y Y Y Y}
			\toprule									
			& 1 & 5 & 6 & 8 & 9 & 10 & 11 & 12 & mAP \\
			\midrule		
			\multicolumn{10}{c}{Inception-ResNet-v2} \\
			\midrule
			PBR-h & \textbf{60.3} & \textbf{44.5} & \textbf{56.7} & \textbf{53.4} & \textbf{81.8} & \textbf{48.6} & 9.6 & \textbf{92.3} & \textbf{55.9} \\
			PBR-l & 57.3	& 35.8	& 53.3	& 52.6	& 77.8	& 23.8	& 3.1	& 94.5 & 49.8 \\
			R\&P & 30.7 & 45.4	& 42.5	& 32.4	& 77.1	& 33.4 & \textbf{19.6}	& 76.7	& 44.7 \\
			\midrule								
			\multicolumn{10}{c}{ResNet-101} \\
			\midrule	
			PBR-h & \textbf{46.3} & \textbf{40.3} & \textbf{48.5} & \textbf{58.0}	& \textbf{76.4} & \textbf{39.5} & 4.7	& \textbf{85.5}	& \textbf{49.9} \\
			PBR-l & 44.1	& 26.6	& 41.6	& 53.7	& 73.7	& 24.5	& 1.1	& 91.6	& 44.6 \\
			R\&P & 35.5 & 45.3 & 37.1 & 44.6 & 75.0 & 33.6 & \textbf{12.7} & 76.8 &	45.1 \\
			\bottomrule
		\end{tabularx}
		\endgroup
		\captionof{table}{\label{tab:synth_scores_lmo} \textbf{Object detection scores on LM-O:}  Per-object average precision (AP@.75IoU) and mean average precision (mAP@.75IoU) of~Faster R-CNN trained on high/low PBR images of objects in Scenes 1--5 (PBR-h, PBR-l), and baseline R\&P images.
		Object ID's as in~\cite{hodan2018bop}.
		} \vspace{0.1ex}
	\end{center}
	
	\begin{center}
		\begingroup
		\footnotesize
		\begin{tabularx}{\linewidth}{l Y Y Y Y Y Y Y Y Y Y Y Y Y Y Y}
			\toprule
			& 1 & 2 & 3 & 4 & 5 & 6 & 7 & 8 & 9 & 10 & 11 & 12 & 13 & 14 & mAP \\
			\midrule
			\multicolumn{16}{c}{Inception-ResNet-v2} \\
			\midrule
			PBR-h & 57.1 & 93.3 & \textbf{88.0}	& 61.2 & 80.4 & \textbf{62.5} & \textbf{99.0} & 98.1 & \textbf{73.2}	& \textbf{44.8} & 65.4 & \textbf{70.9} & 86.8 & \textbf{26.4} &	71.9 \\
			PBR-l & \textbf{57.6} & \textbf{96.3} & 84.6 & \textbf{62.2} & \textbf{81.3} & 60.5 & 98.7 & \textbf{98.6} & 73.1 & \textbf{44.8} & \textbf{79.3} & 67.2 & \textbf{90.5} & 25.6 & \textbf{72.9} \\
			PBR-ho & 46.2 & 61.9 & 56.0	& 55.8 & 54.4 & 69.8 & 89.0 & 89.3 & 81.6 & 21.5 & 72.7 &	58.3 & 43.3 & 21.7 & 58.7 \\
			R\&P & 33.5 & 47.5 & 71.5 & 32.7 & 42.4 & 13.5 & 44.9	& 73.0 & 57.4	& 44.5 & 47.6 & 35.8 & 87.6	& 40.6	& 48.0 \\
			\midrule
			\multicolumn{16}{c}{ResNet-101} \\
			\midrule
			PBR-h & 30.6 & \textbf{93.7} & \textbf{91.6} & \textbf{68.2} & 72.1 & 56.7 & 93.4 & \textbf{93.6} & \textbf{75.2} & \textbf{42.6} & \textbf{84.5} & 60.9 & 73.2 & 21.4 & \textbf{68.4} \\
			PBR-l & 26.8 & 87.6 & 87.3 & 64.0 & \textbf{79.8} & 27.8 & \textbf{95.2} & 90.4 & 66.2 & 37.5 & 83.1 & \textbf{61.4} & \textbf{79.3} & \textbf{25.3} & 65.1 \\
			PBR-ho & \textbf{35.2} & 64.4 & 58.4 & 52.9 & 46.7 & 53.0 & 71.5	& 73.8 & 69.3 & 32.2 & 66.2 & 51.8 & 28.3 &	19.3 & 51.6 \\
			R\&P & 29.1 & 38.5 & 82.0 & 59.2 & 52.4 & \textbf{59.1} & 79.5 & 75.0 & 36.4 & 36.8 & 75.1 & 50.6 & 48.5 & 14.8 & 52.7 \\
			\bottomrule
		\end{tabularx}
		\endgroup
		\captionof{table}{\label{tab:synth_scores_ruapc} \textbf{Object detection scores on RU-APC:} Per-object average precision (AP@.75IoU) and mean average precision (mAP@.75IoU) of~Faster R-CNN trained on high/low-quality PBR images of in-context objects in Scene 6 (PBR-h, PBR-l), high-quality PBR images of out-of-context objects in Scene 3 (PBR-ho), and baseline images (R\&P). Object ID's as in BOP~\cite{hodan2018bop}.
		} \vspace{-1.5ex}
	\end{center}

	\begin{center}
		\begin{tabular}{ @{}c@{ } @{}c@{ } @{}c@{ } @{}c@{ } }
			\includegraphics[width=0.243\linewidth]{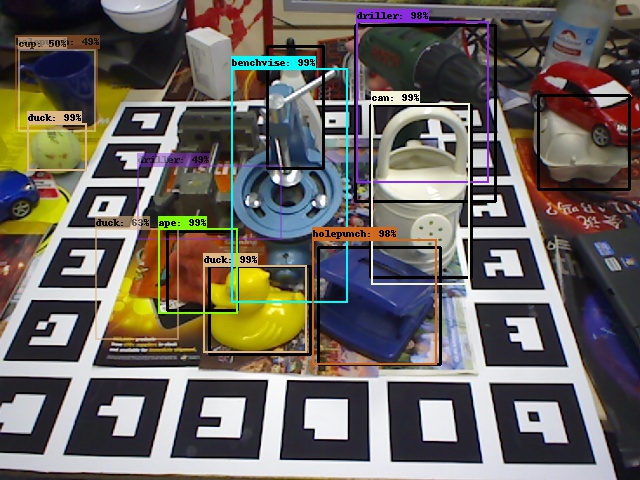} &
			\includegraphics[width=0.243\linewidth]{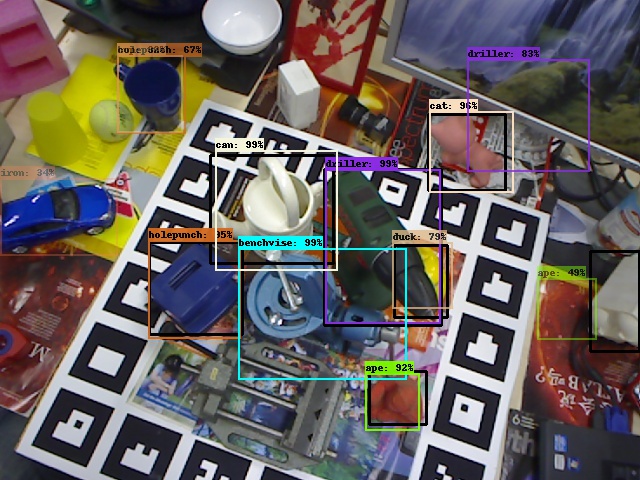} &
			\includegraphics[width=0.243\linewidth]{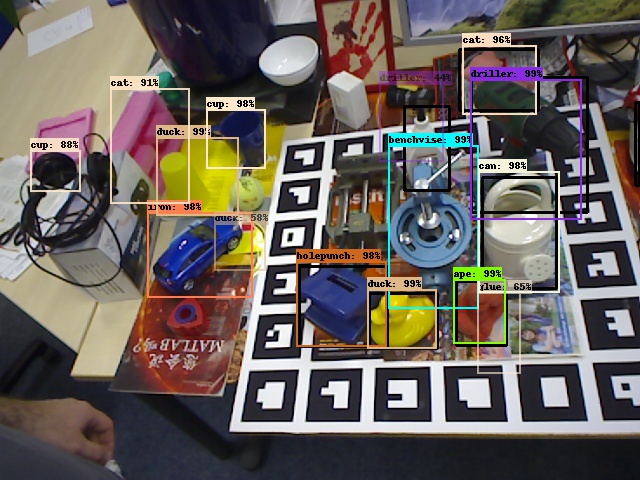} &
			\includegraphics[width=0.243\linewidth]{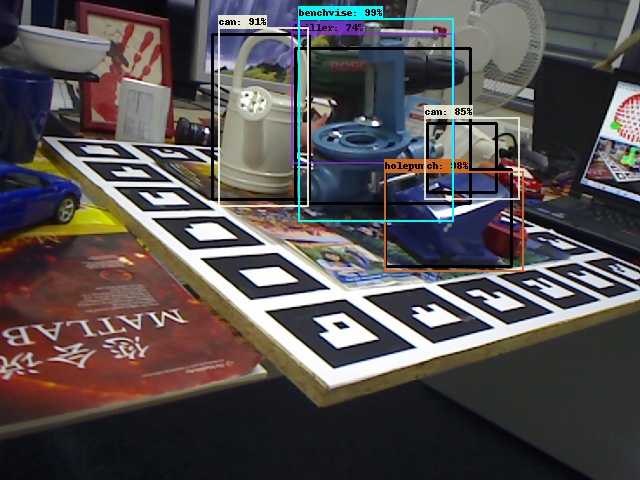} \\
			\includegraphics[width=0.243\linewidth]{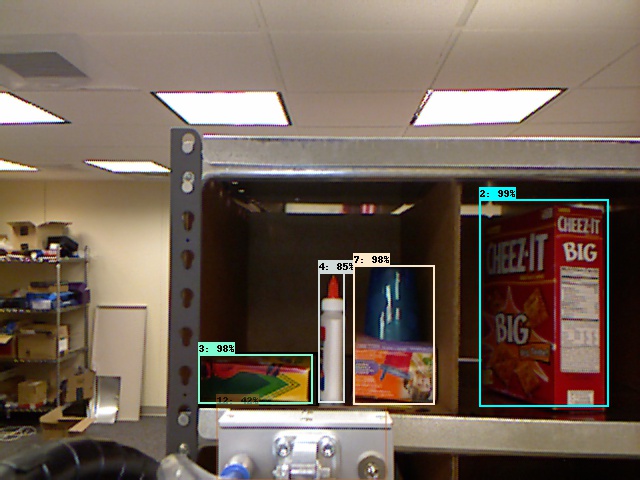} &
			\includegraphics[width=0.243\linewidth]{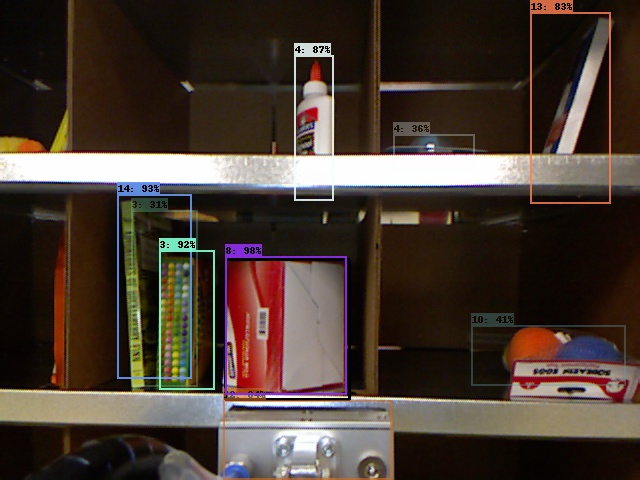} &
			\includegraphics[width=0.243\linewidth]{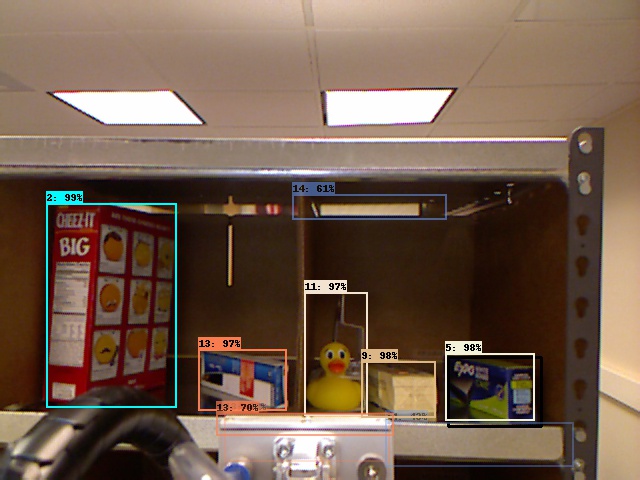} &
			\includegraphics[width=0.243\linewidth]{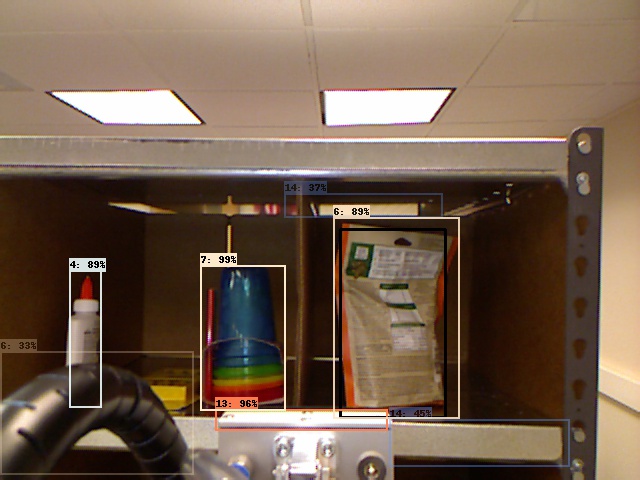} \\
		\end{tabular}
		\caption{\label{fig:synth_frcnn_results}
			\textbf{Object detections.}
			The Faster R-CNN object detector was trained on high-quality PBR images and evaluated on real test images from LM-O (top) and RU-APC (bottom).
		} \vspace{-3.0ex}
	\end{center}
\end{figure}

\customparagraph{Datasets.}
The experiments were conducted on the reduced versions~\cite{hodan2018bop} of datasets Linemod-Occluded (LM-O)~\cite{brachmann2014learning,hinterstoisser2012accv} and Rutgers APC
(RU-APC)~\cite{rennie2016dataset}.
The datasets include 3D object models and real RGB-D test images of VGA resolution (only RGB channels were used).
The ground-truth 2D bounding boxes for the evaluation of 2D object detection were calculated from the provided ground-truth 6D object poses.
LM-O contains 200 images with ground-truth annotations for 8 LM objects captured with various levels of occlusion. RU-APC contains 14 object models and 1380 images which show the objects in a cluttered warehouse shelf. Example test images are in Figure~\ref{fig:synth_frcnn_results}.

\begin{figure}[t!]
	\begin{center}
		\begin{tabular}{ @{}c@{ } @{}c@{ } @{}c@{ } }
			\includegraphics[width=0.3265\columnwidth]{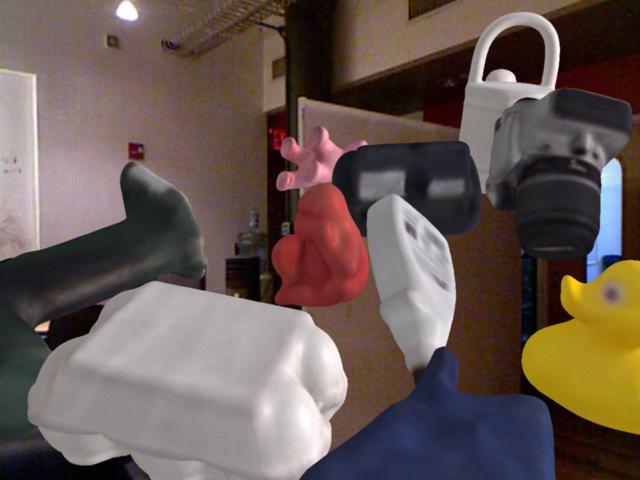} &
			\includegraphics[width=0.3265\columnwidth]{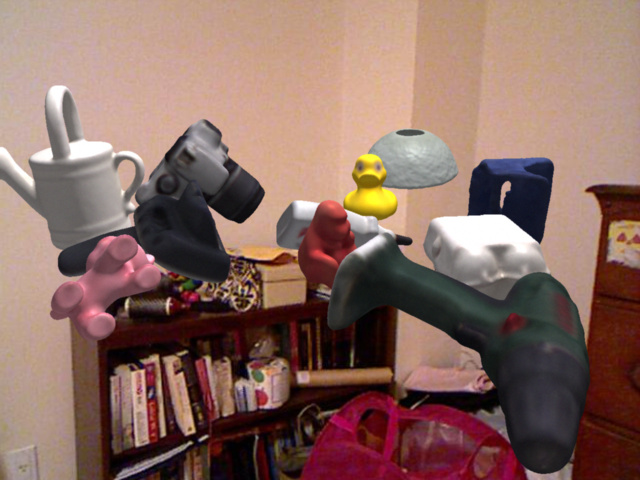} &
			\includegraphics[width=0.3265\columnwidth]{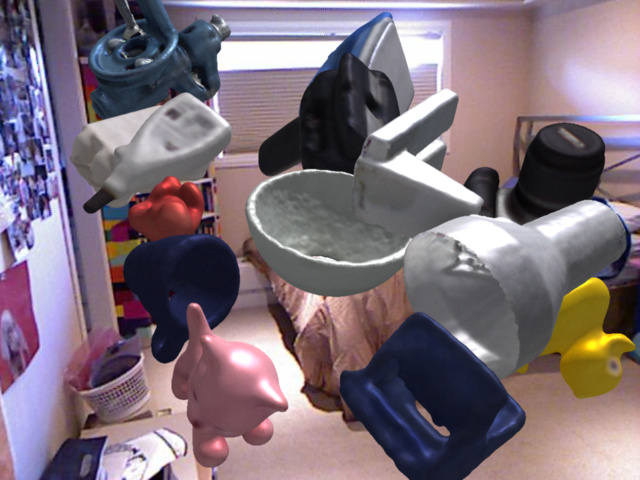} \\
			\includegraphics[width=0.3265\columnwidth]{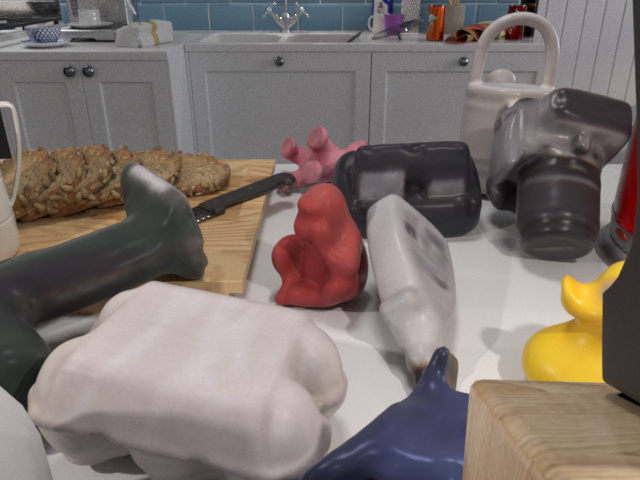} &
			\includegraphics[width=0.3265\columnwidth]{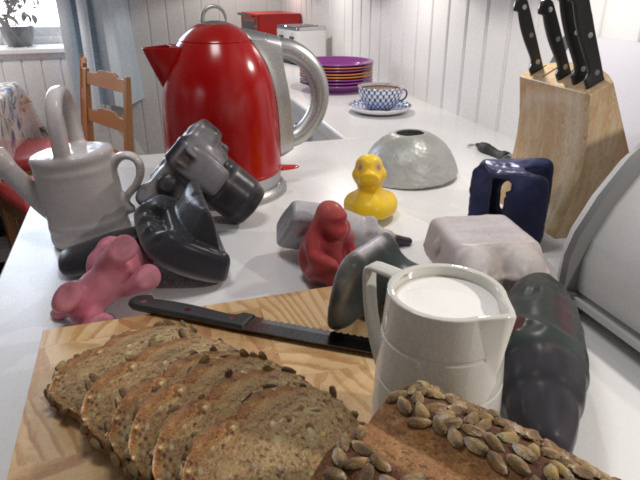} &
			\includegraphics[width=0.3265\columnwidth]{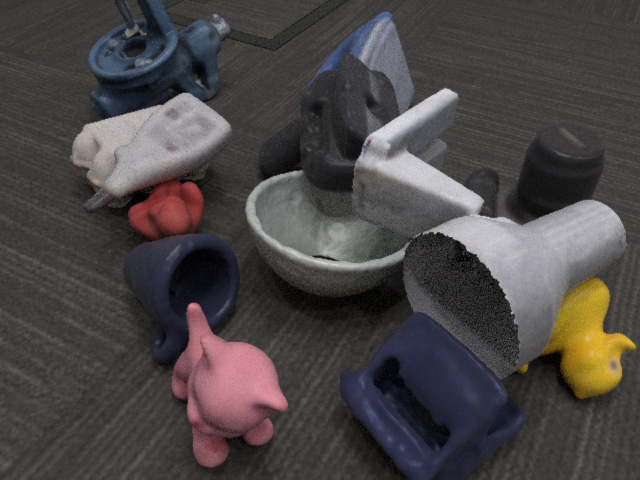} \\
		\end{tabular}
		\caption{\label{fig:synth_random_bg}
			\textbf{Types of training images.} The baseline ``render \& paste'' training images (top) were generated by rendering the 3D object models in the same poses as in the PBR images (bottom) and pasting the renderings on top of random photographs.
		}
	\end{center}
\end{figure}

\customparagraph{``Render\;\&\;Paste'' Training Images (R\&P).}
The baseline images were generated by a ``render\;\&\;paste'' pipeline similar to~\cite{hinterstoisser2017pre} -- the 3D object models were rendered by OpenGL on top of random real photographs pulled from NYU Depth Dataset V2~\cite{silberman2012indoor}.
For a more direct comparison, the R\&P images were generated by imitating the PBR images -- the 3D object models were rendered in the same poses as in the PBR images. %
Generating one R\&P image took $3\,$s on average. Examples are in Figure~\ref{fig:synth_random_bg}.

\customparagraph{Object Instance Detection.}
The experiments were conducted with two network architectures of Faster R-CNN:
ResNet-101~\cite{he2016deep} and Inception-ResNet-v2~\cite{szegedy2017inception}.
The networks were pre-trained on the Microsoft COCO dataset~\cite{lin2014microsoft} and fine-tuned on synthetic images for 100K iterations.
The learning rate was set to 0.001 and multiplied by 0.96 every 1K iterations.
To virtually increase diversity of the training set, the images were augmented by randomly adjusting brightness, contrast, hue, and saturation, and by applying random Gaussian noise and blur, similarly to~\cite{hinterstoisser2017pre}.
The implementation of Faster R-CNN from Tensorflow Object Detection API~\cite{huang2017speed} was used.
The detection accuracy was measured by mAP@.75IoU, \ie, the mean average precision with a strict IoU threshold of 0.75~\cite{lin2014microsoft}.
For each test image, detections of object models annotated in the image were considered.

\setlength{\tabcolsep}{3pt}

\subsection{Importance of PBR Images for Training}
On RU-APC, Faster R-CNN with Inception-ResNet-v2 trained on the high-quality PBR images achieves a significant 24$\%$ absolute improvement of mAP@.75IoU over the same method trained on R\&P images (Table~\ref{tab:synth_scores_ruapc}, PBR-h \vs R\&P).
It is noteworthy that PBR images yield almost 35$\%$ or higher absolute improvement on five objects, and overall achieve a better performance on 12 out of 14 objects.
This is achieved when the objects are rendered in Scene 6 that accurately imitates the scene visible in test images.
When the objects are rendered in Scene 3 (PBR-ho \vs R\&P), the accuracy score still increases by 11$\%$
(the importance of scene context is discussed later in Section~\ref{sec:context}).
Improvements, although not so dramatic, can also be observed with the ResNet-101 network.
On LM-O, PBR images yield almost 11$\%$ absolute improvement, with a large gain on 7 out of 8 objects (Table~\ref{tab:synth_scores_lmo}, PBR-h \vs R\&P).

\subsection{Importance of PBR Quality}

On LM-O, Faster R-CNN with Inception-ResNet-v2 trained on the high-quality PBR images achieves an improvement of almost 6$\%$ over the low-quality PBR images (Table~\ref{tab:synth_scores_lmo}, PBR-h \vs PBR-l).
A similar improvement is not observed on RU-APC when training on PBR images rendered in Scene~6.
The illumination in Scene 6 is simpler, there is no incoming outdoor light and the materials are mainly Lambertian. There are therefore no complex reflections and the low-quality PBR images from this scene are cleaner than, \eg, when rendered in Scenes 3--5. This suggests that the low quality is sufficient for scenes with simpler illumination and materials.

\subsection{Importance of Scene Context} \label{sec:context}

\begin{figure}[h!]
	\begin{center}
		\small
		\begin{tabular}{ @{}c@{ } @{}c@{ } @{}c@{ } @{}c@{ } @{}c@{ } }
			\multicolumn{2}{c}{In context} & & \multicolumn{2}{c}{Out of context} \vspace{0.5ex} \\			
			\includegraphics[width=0.239\linewidth]{images/icip19/supp/pbr/ruapc_03.jpg} &
			\includegraphics[width=0.239\linewidth]{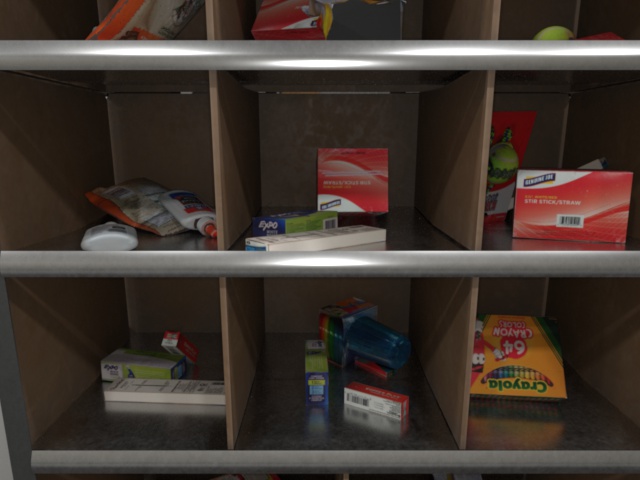} & \hspace{1.0ex} &
			\includegraphics[width=0.239\columnwidth]{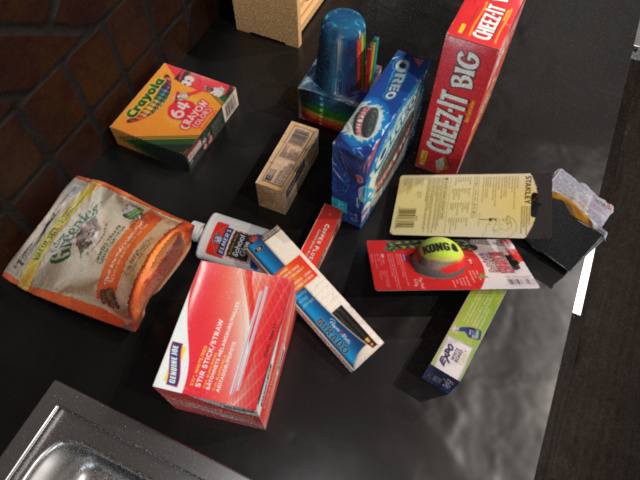} &
			\includegraphics[width=0.239\columnwidth]{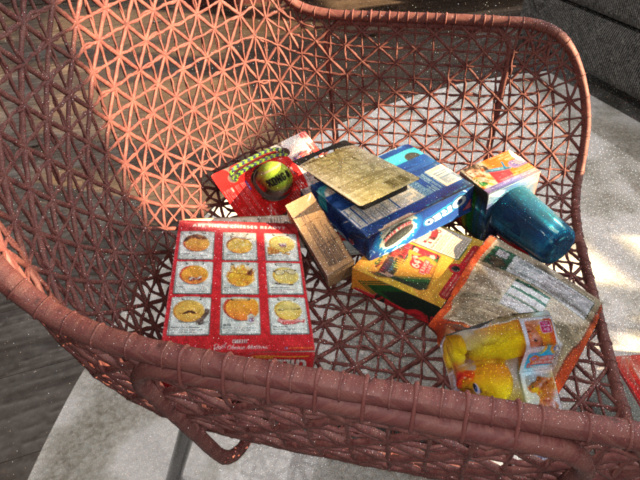} \\
		\end{tabular}
		\caption{\label{fig:synth_context}
			\textbf{Images of in-context and out-of-context RU-APC objects.} The in-context images are rendered in Scene 6 which accurately models the scene in test images.
		} \vspace{-3.5ex}
	\end{center}	
\end{figure}

To analyze the importance of accurately modeling the scene context, the RU-APC objects were rendered in two setups:
(1)~``in context'' in
Scene~6, and (2)~``out of context'' in Scene~3 (Figure~\ref{fig:synth_context}).
Following the taxonomy of contextual information from~\cite{divvala2009empirical},
the in-context setup faithfully imitates the gist, geometric, semantic, and illumination aspects of the context of the test scene. The out-of-context setup exhibits discrepancies in all of these aspects.
Training images of in-context objects yield an absolute improvement of 13$\%$ with Inception-ResNet-v2 and 16$\%$ with ResNet-101 over the images of out-of-context objects (Table~\ref{tab:synth_scores_ruapc}, PBR-h \vs PBR-ho). This demonstrates the benefit of accurately modeling the context of the test scene.

\begin{figure}[t!]
	\begin{center}
		
		\begingroup
		\small
		\renewcommand{\arraystretch}{0.9}
		
		\begin{tabular}{ @{}c@{ } @{}c@{ } @{}c@{ } @{}c@{ } }
			
			\multicolumn{4}{c}{Commonly used ``render\;\&\;paste'' synthetic training images} \vspace{0.5ex} \\
			
			\includegraphics[width=0.243\columnwidth]{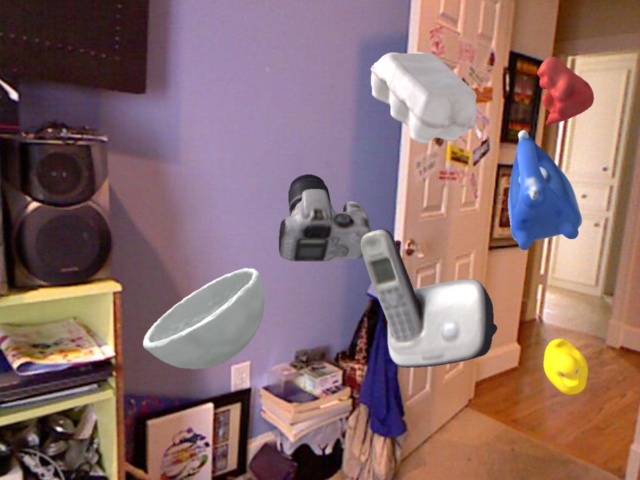} &
			\includegraphics[width=0.243\columnwidth]{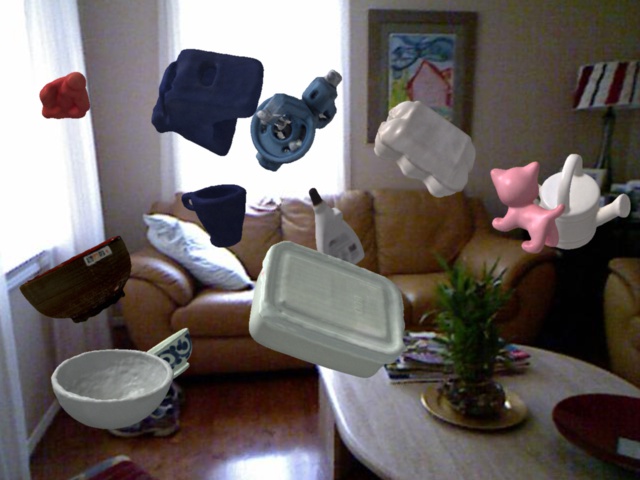} &
			\includegraphics[width=0.243\columnwidth]{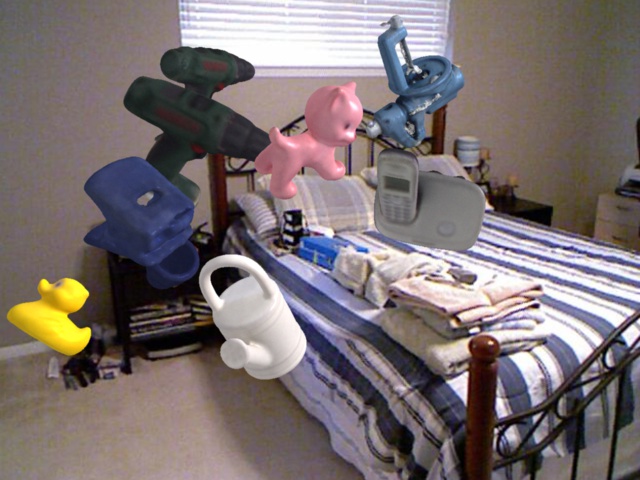} &
			\includegraphics[width=0.243\columnwidth]{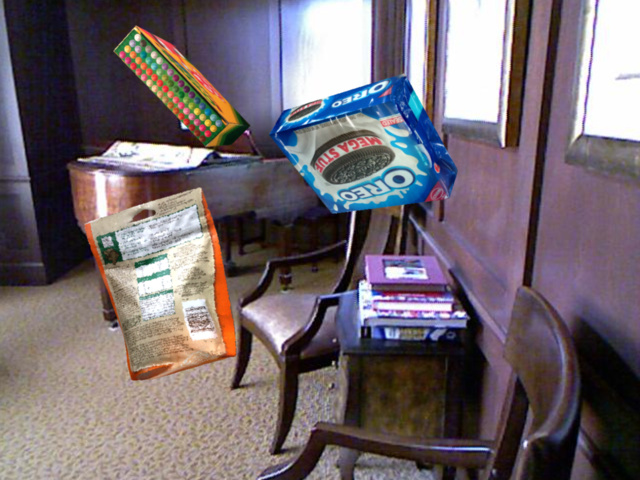} \\
			
			\multicolumn{4}{c}{}\vspace{-2.0ex} \\
			\multicolumn{4}{c}{Photorealistic training images rendered by BlenderProc4BOP~\cite{denninger2019blenderproc,denninger2020blenderproc}} \vspace{0.5ex} \\
			
			\includegraphics[width=0.243\columnwidth]{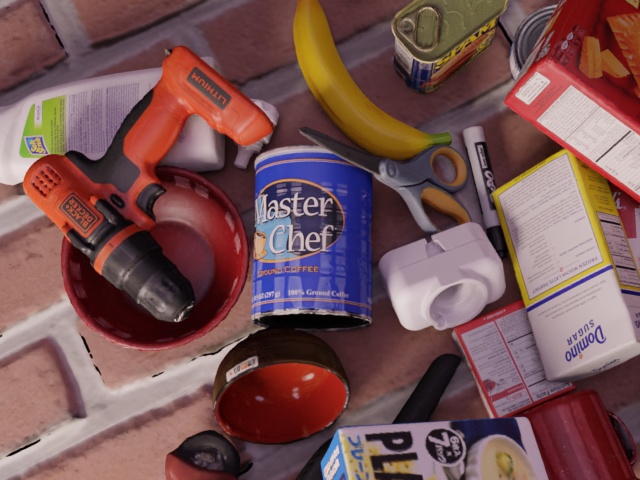} &
			\includegraphics[width=0.243\columnwidth]{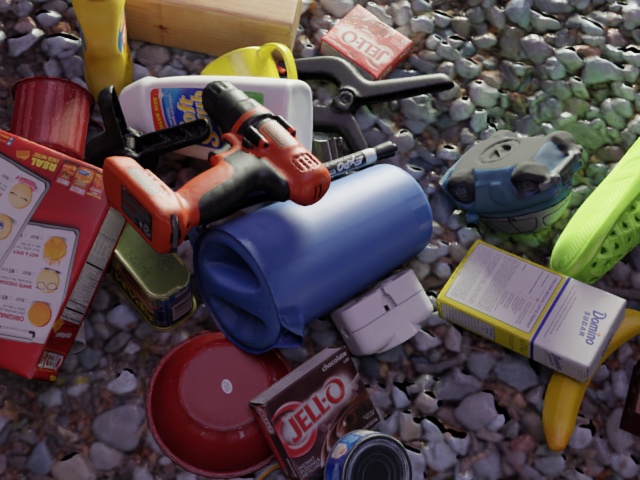} &
			\includegraphics[width=0.243\columnwidth]{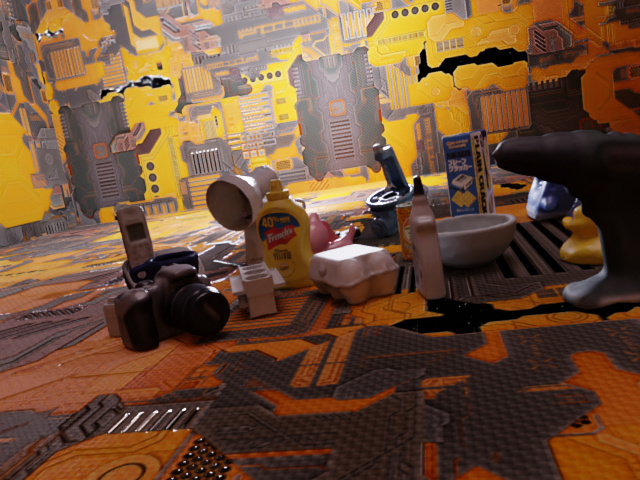} &
			\includegraphics[width=0.243\columnwidth]{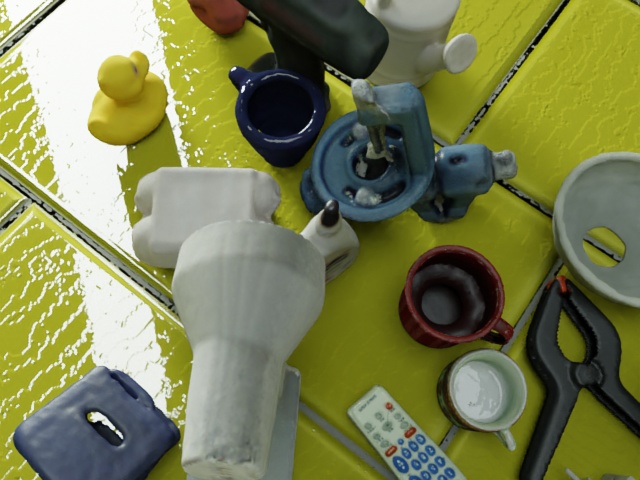} \\
			
			\includegraphics[width=0.243\columnwidth]{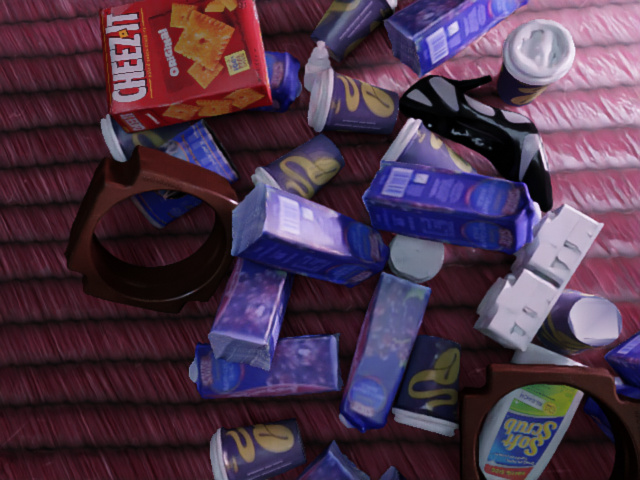} &
			\includegraphics[width=0.243\columnwidth]{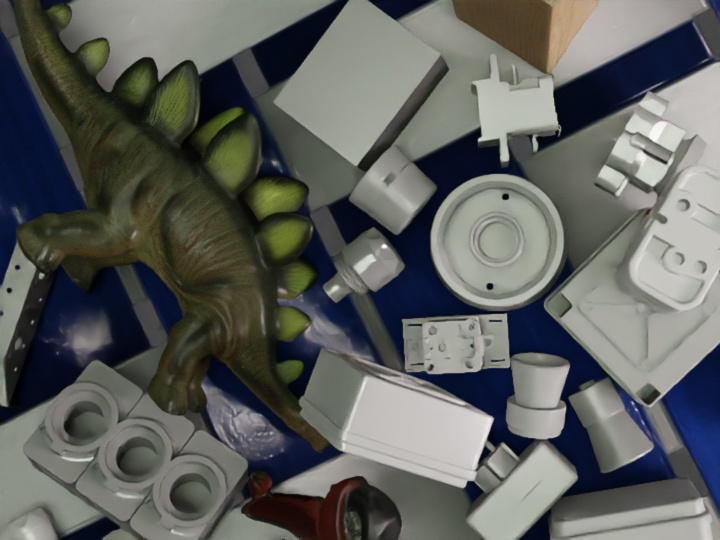} &
			\includegraphics[width=0.243\columnwidth]{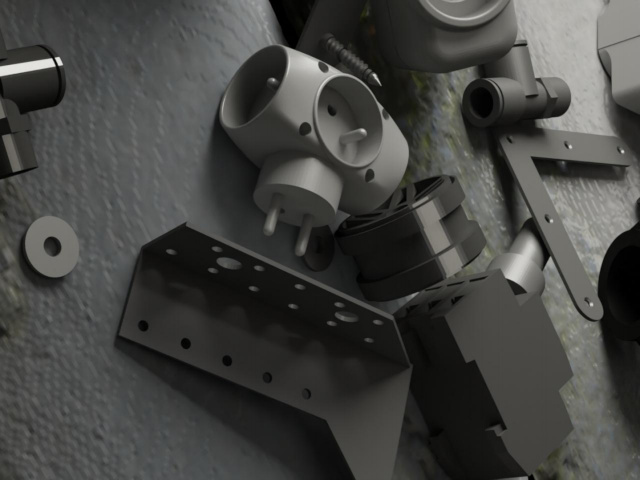} &
			\includegraphics[width=0.243\columnwidth]{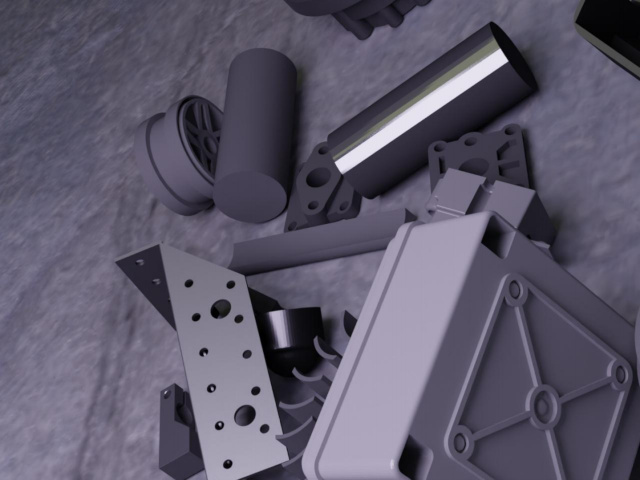} \\
			
		\end{tabular}
		\endgroup
		
		\caption{\label{fig:synth_pbr_bop20} \textbf{Synthetic training images.} Methods for 6D object pose estimation have been commonly trained on ``render\;\&\;paste'' images synthesized by OpenGL rendering of 3D object models randomly positioned on top of random backgrounds. Instead, participants of the BOP Challenge 2020 were provided 350K PBR training images rendered by ray tracing and showing the 3D object models in physically plausible poses inside a cube with a random PBR material.}
		
	\end{center}
\end{figure}

\section{Training Images for BOP Challenge 2020} \label{sec:synth_bop20}

In the BOP Challenge 2020 (Section~\ref{sec:bop_challenge_2020}), the participants were provided with 350K PBR training images
generated and automatically annotated by BlenderProc4BOP~\cite{denninger2019blenderproc,denninger2020blenderproc}, an open-source and light-weight physically-based renderer
which implements a refined version of the synthesis approach proposed above.

To improve efficiency, BlenderProc4BOP renders objects not in 3D models of 
indoor scenes but inside an empty cube, with objects from other datasets serving as distractors.
To achieve a rich spectrum of generated images, a random PBR material from the CC0 Textures library~\cite{demes2020textures} is assigned to the walls of the cube, and light with a random intensity and color is emitted from the room ceiling and from a randomly positioned point source.
As shown in Table~\ref{fig:synth_pbr_quality}, the ray-tracing parameters are set economically, with $50$ rays sampled per pixel but only a single indirect ray sampled when a ray from the camera hits a surface. Noise in the rendered images is reduced by the Intel Open Image Denoiser~\cite{inteldenoise}.
This setup keeps the computational cost low -- the full generation of one RGB-D image in VGA resolution takes \textbf{only 1--3 seconds} on a standard desktop computer with a modern GPU. A set of 50K images can be therefore rendered on 5 GPU's overnight.

Instead of trying to accurately model the object materials, the specular, metallic, and roughness properties are randomized.
Realistic object poses are achieved by dropping the 3D object models on the ground plane of the cube using the PyBullet physics engine integrated in Blender~\cite{blender}. This allows creating dense but shallow piles of objects that introduce various levels of occlusion.
Since test images from the LM dataset show the objects always standing upright, the objects from LM are not dropped but instead densely placed on the ground plane in upright poses using automated collision checks.
Each object arrangement is rendered from $25$ random camera poses. The azimuth angles, elevation angles, and distances of the cameras are uniformly sampled from ranges determined by the ground-truth 6D object poses from the test images, and the in-plane rotation angles are generated randomly.
Examples of the generated images are in Figure~\ref{fig:synth_pbr_bop20}.

The data generated by BlenderProc4BOP (RGB and depth images, ground-truth 6D object poses, camera intrinsics) is saved in the BOP format, allowing to interface with utilities from the BOP toolkit~\cite{boptoolkit}. 
The source code and configuration files to reproduce or modify the generation process are publicly available~\cite{blenderproc4bop}. Users are encouraged to build on top of the open-source code and release their extensions.

In the BOP Challenge 2020, methods achieved noticeably higher accuracy scores when trained on the PBR training images generated by BlenderProc4BOP than when trained on the ``render\;\&\;paste'' images. Although adding real training images yielded even higher scores, competitive results were achieved with PBR images only -- out of the 26 evaluated methods, the fifth was trained only on the PBR images. See Section~\ref{sec:bop_challenge_2020} for details.

	\chapter[T-LESS:\ An RGB-D Dataset with Texture-Less Objects]{T-LESS\\{\Large An RGB-D Dataset with Texture-less Objects}} \label{ch:tless}

\begin{figure}[!t]
	\begin{center}
		\begingroup
		\begin{tabular}{ @{}c@{ } @{}c@{ } @{}c@{ } }
			\includegraphics[width=0.3265\columnwidth]{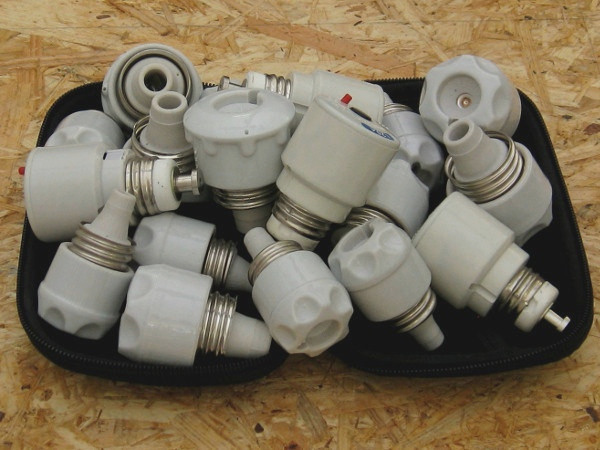} &
			\includegraphics[width=0.3265\columnwidth]{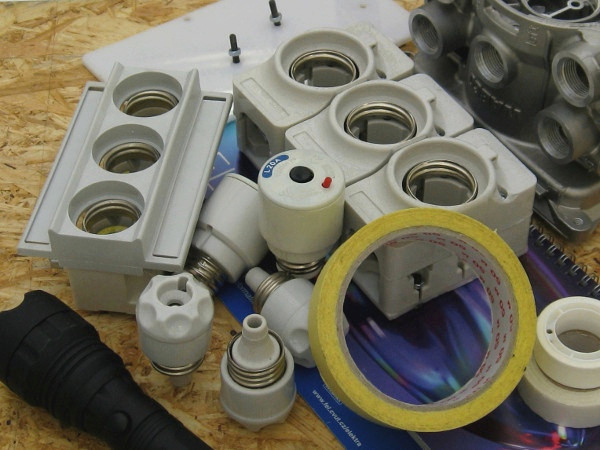} &
			\includegraphics[width=0.3265\columnwidth]{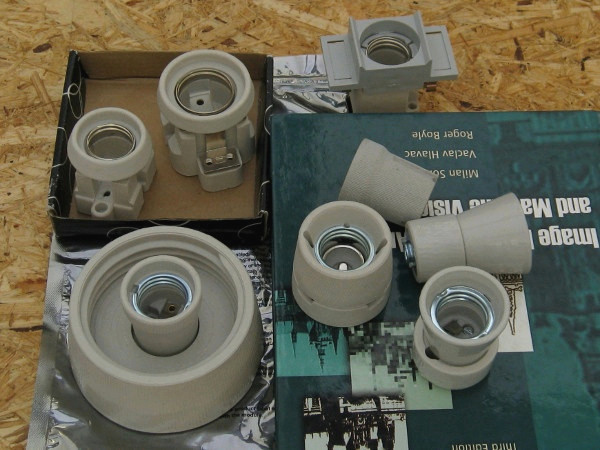} \\
			
			\includegraphics[width=0.3265\columnwidth]{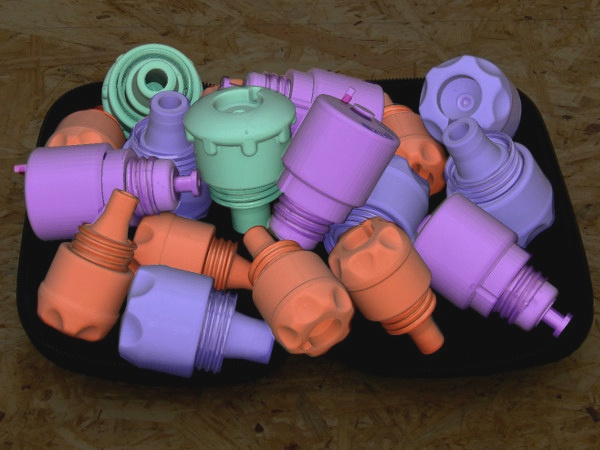} &
			\includegraphics[width=0.3265\columnwidth]{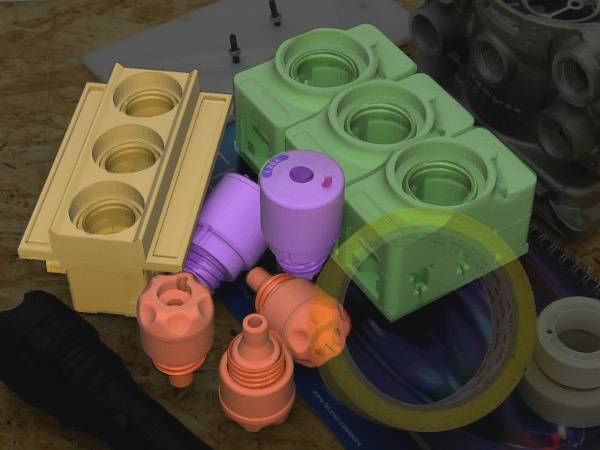} &
			\includegraphics[width=0.3265\columnwidth]{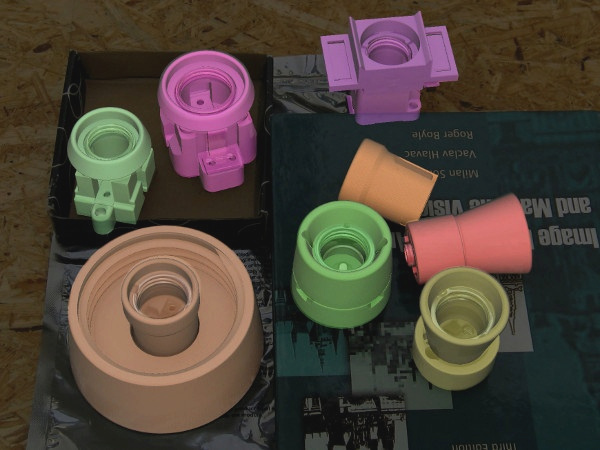} \\
		\end{tabular}
		\endgroup
		\caption{\label{fig:tless_challenges} \textbf{Challenges in T-LESS.}
		Recognizing and estimating the 6D pose of T-LESS objects is challenging because the objects are texture-less, without a discriminative color, exhibit symmetries and similarities in shape and size, and are often seen under heavy occlusion and clutter.
		Cropped
		test images (top) are shown overlaid with colored 3D object models in the ground-truth 6D poses (bottom). Instances of the same object have the same color.}
	\end{center}
\end{figure}

\begin{figure}[t!]
\begin{center}
	
\begingroup
\setlength{\tabcolsep}{0.0pt} %
\renewcommand{\arraystretch}{0.0} %

\begin{tabular}{ c c c c c c c c c c }
\rowcolor{black}
\begin{overpic}[width=0.1\columnwidth]{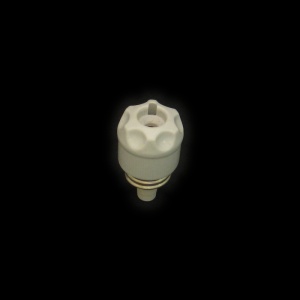}
\put(3,82){\scriptsize$\displaystyle\textbf{\light{1}}$}\end{overpic} &
\begin{overpic}[width=0.1\columnwidth]{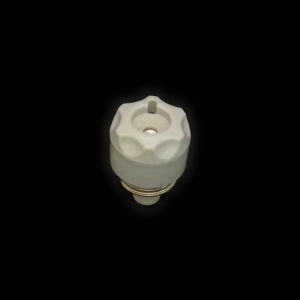}
\put(3,82){\scriptsize$\displaystyle\textbf{\light{2}}$}\end{overpic} &
\begin{overpic}[width=0.1\columnwidth]{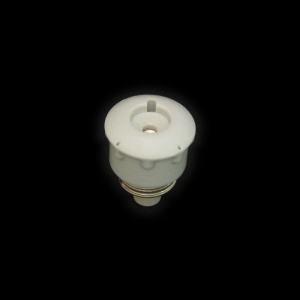}
\put(3,82){\scriptsize$\displaystyle\textbf{\light{3}}$}\end{overpic} &
\begin{overpic}[width=0.1\columnwidth]{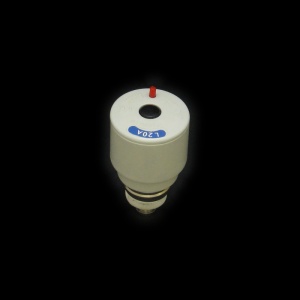}
\put(3,82){\scriptsize$\displaystyle\textbf{\light{4}}$}\end{overpic} &
\begin{overpic}[width=0.1\columnwidth]{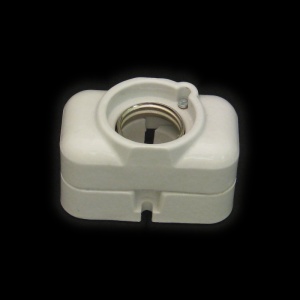}
\put(3,82){\scriptsize$\displaystyle\textbf{\light{5}}$}\end{overpic} &
\begin{overpic}[width=0.1\columnwidth]{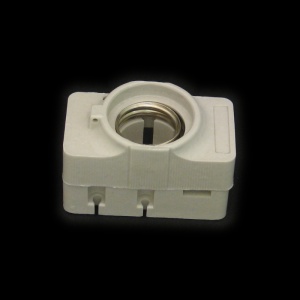}
\put(3,82){\scriptsize$\displaystyle\textbf{\light{6}}$}\end{overpic} &
\begin{overpic}[width=0.1\columnwidth]{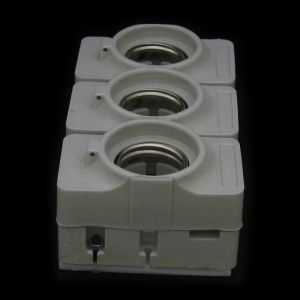}
\put(3,82){\scriptsize$\displaystyle\textbf{\light{7}}$}\end{overpic} &
\begin{overpic}[width=0.1\columnwidth]{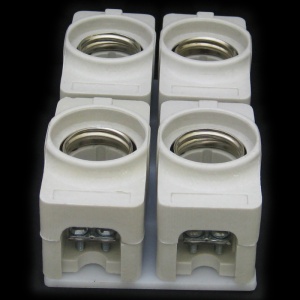}
\put(3,82){\scriptsize$\displaystyle\textbf{\light{8}}$}\end{overpic} &
\begin{overpic}[width=0.1\columnwidth]{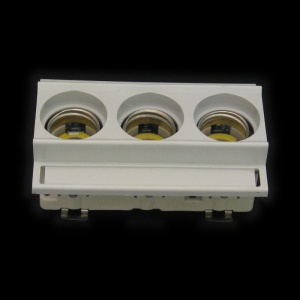}
\put(3,82){\scriptsize$\displaystyle\textbf{\light{9}}$}\end{overpic} &
\begin{overpic}[width=0.1\columnwidth]{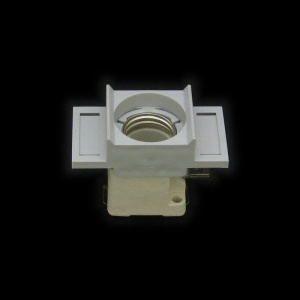}
\put(3,82){\scriptsize$\displaystyle\textbf{\light{10}}$}\end{overpic} \\

\rowcolor{black}
\begin{overpic}[width=0.1\columnwidth]{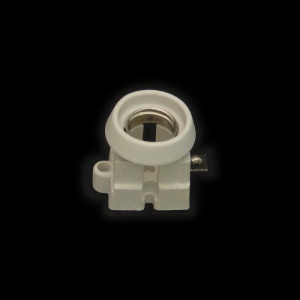}
\put(3,82){\scriptsize$\displaystyle\textbf{\light{11}}$}\end{overpic} &
\begin{overpic}[width=0.1\columnwidth]{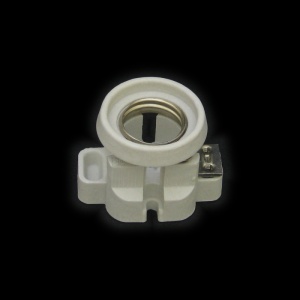}
\put(3,82){\scriptsize$\displaystyle\textbf{\light{12}}$}\end{overpic} &
\begin{overpic}[width=0.1\columnwidth]{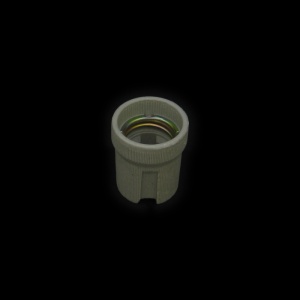}
\put(3,82){\scriptsize$\displaystyle\textbf{\light{13}}$}\end{overpic} &
\begin{overpic}[width=0.1\columnwidth]{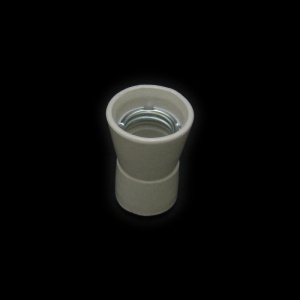}
\put(3,82){\scriptsize$\displaystyle\textbf{\light{14}}$}\end{overpic} &
\begin{overpic}[width=0.1\columnwidth]{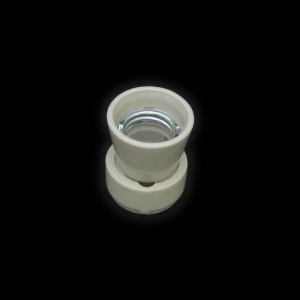}
\put(3,82){\scriptsize$\displaystyle\textbf{\light{15}}$}\end{overpic} &
\begin{overpic}[width=0.1\columnwidth]{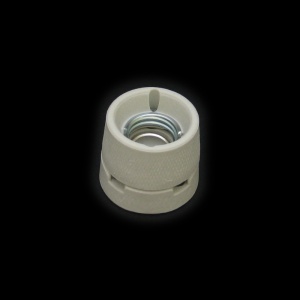}
\put(3,82){\scriptsize$\displaystyle\textbf{\light{16}}$}\end{overpic} &
\begin{overpic}[width=0.1\columnwidth]{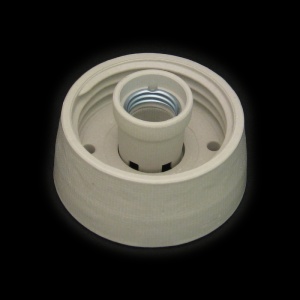}
\put(3,82){\scriptsize$\displaystyle\textbf{\light{17}}$}\end{overpic} &
\begin{overpic}[width=0.1\columnwidth]{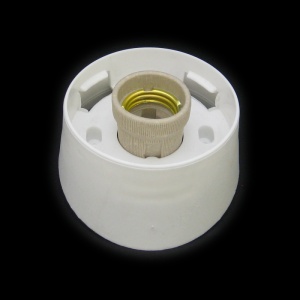}
\put(3,82){\scriptsize$\displaystyle\textbf{\light{18}}$}\end{overpic} &
\begin{overpic}[width=0.1\columnwidth]{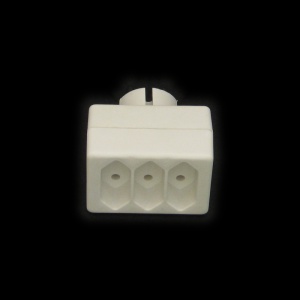}
\put(3,82){\scriptsize$\displaystyle\textbf{\light{19}}$}\end{overpic} &
\begin{overpic}[width=0.1\columnwidth]{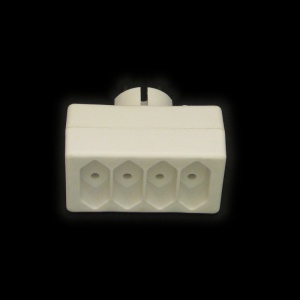}
\put(3,82){\scriptsize$\displaystyle\textbf{\light{20}}$}\end{overpic} \\

\rowcolor{black}
\begin{overpic}[width=0.1\columnwidth]{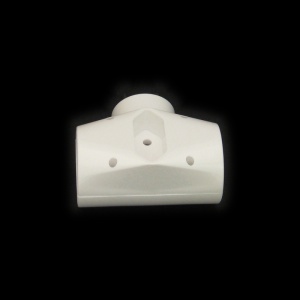}
\put(3,82){\scriptsize$\displaystyle\textbf{\light{21}}$}\end{overpic} &
\begin{overpic}[width=0.1\columnwidth]{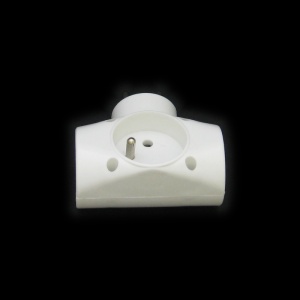}
\put(3,82){\scriptsize$\displaystyle\textbf{\light{22}}$}\end{overpic} &
\begin{overpic}[width=0.1\columnwidth]{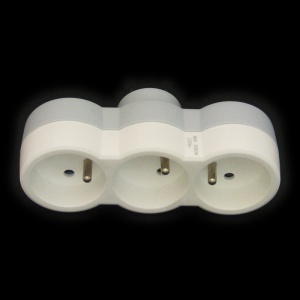}
\put(3,82){\scriptsize$\displaystyle\textbf{\light{23}}$}\end{overpic} &
\begin{overpic}[width=0.1\columnwidth]{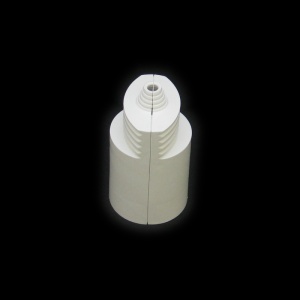}
\put(3,82){\scriptsize$\displaystyle\textbf{\light{24}}$}\end{overpic} &
\begin{overpic}[width=0.1\columnwidth]{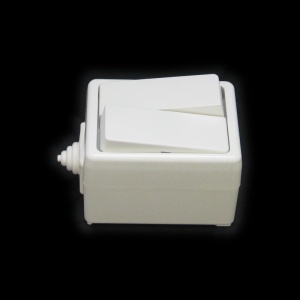}
\put(3,82){\scriptsize$\displaystyle\textbf{\light{25}}$}\end{overpic} &
\begin{overpic}[width=0.1\columnwidth]{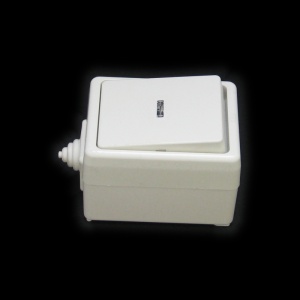}
\put(3,82){\scriptsize$\displaystyle\textbf{\light{26}}$}\end{overpic} &
\begin{overpic}[width=0.1\columnwidth]{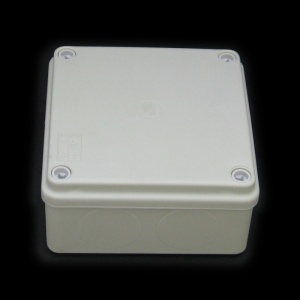}
\put(1,82){\scriptsize$\displaystyle\textbf{\light{27}}$}\end{overpic} &
\begin{overpic}[width=0.1\columnwidth]{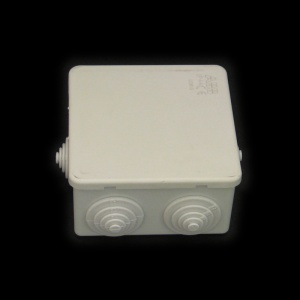}
\put(3,82){\scriptsize$\displaystyle\textbf{\light{28}}$}\end{overpic} &
\begin{overpic}[width=0.1\columnwidth]{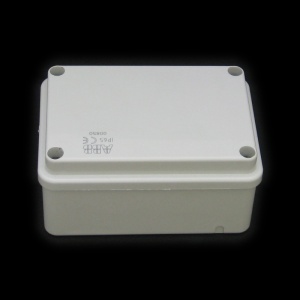}
\put(3,82){\scriptsize$\displaystyle\textbf{\light{29}}$}\end{overpic} &
\begin{overpic}[width=0.1\columnwidth]{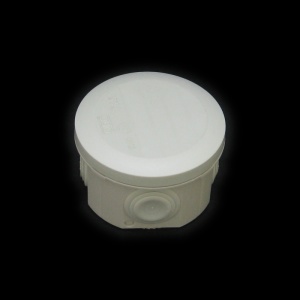}
\put(3,82){\scriptsize$\displaystyle\textbf{\light{30}}$}\end{overpic}

\end{tabular}

\vspace{1.5ex}

\begin{tabular}{ c c c c c }

\rowcolor{black}
\begin{overpic}[width=0.2\columnwidth]{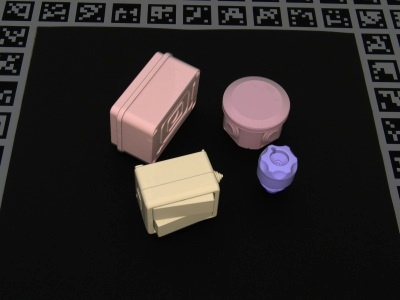}
\put(2,65){\scriptsize$\displaystyle\textbf{\light{1}}$}\end{overpic} &
\begin{overpic}[width=0.2\columnwidth]{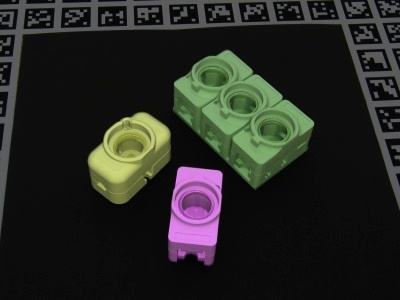}
\put(2,65){\scriptsize$\displaystyle\textbf{\light{2}}$}\end{overpic} &
\begin{overpic}[width=0.2\columnwidth]{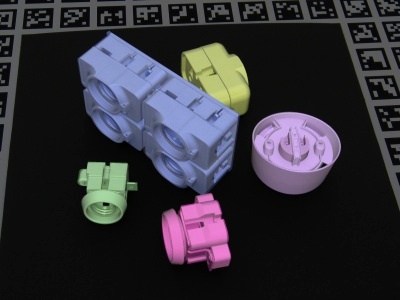}
\put(2,65){\scriptsize$\displaystyle\textbf{\light{3}}$}\end{overpic} &
\begin{overpic}[width=0.2\columnwidth]{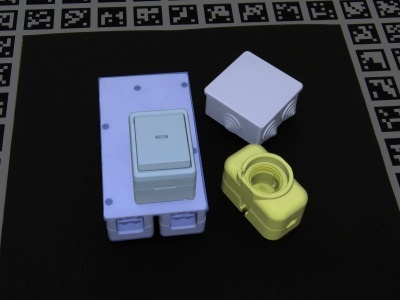}
\put(2,65){\scriptsize$\displaystyle\textbf{\light{4}}$}\end{overpic} &
\begin{overpic}[width=0.2\columnwidth]{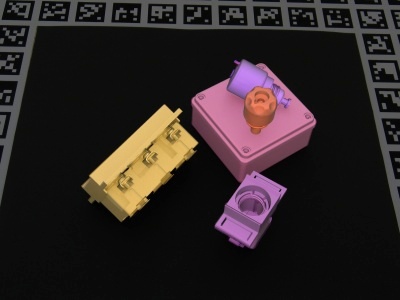}
\put(2,65){\scriptsize$\displaystyle\textbf{\light{5}}$}\end{overpic} \\

\rowcolor{black}
\begin{overpic}[width=0.2\columnwidth]{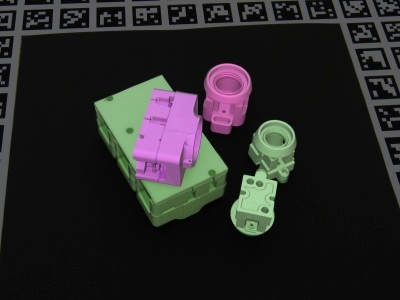}
\put(2,65){\scriptsize$\displaystyle\textbf{\light{6}}$}\end{overpic} &
\begin{overpic}[width=0.2\columnwidth]{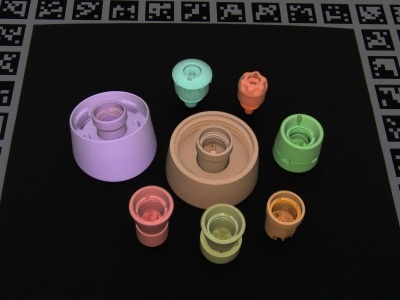}
\put(2,65){\scriptsize$\displaystyle\textbf{\light{7}}$}\end{overpic} &
\begin{overpic}[width=0.2\columnwidth]{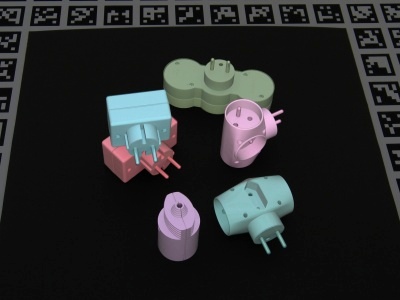}
\put(2,65){\scriptsize$\displaystyle\textbf{\light{8}}$}\end{overpic} &
\begin{overpic}[width=0.2\columnwidth]{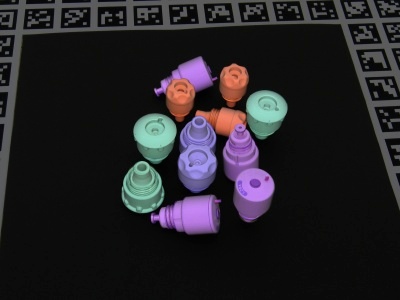}
\put(2,65){\scriptsize$\displaystyle\textbf{\light{9}}$}\end{overpic} &
\begin{overpic}[width=0.2\columnwidth]{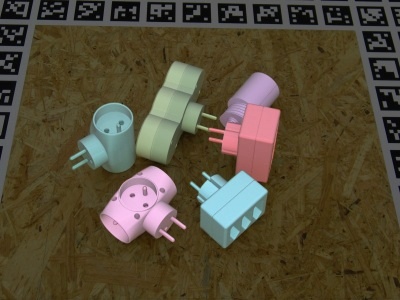}
\put(2,65){\scriptsize$\displaystyle\textbf{\light{10}}$}\end{overpic} \\

\rowcolor{black}
\begin{overpic}[width=0.2\columnwidth]{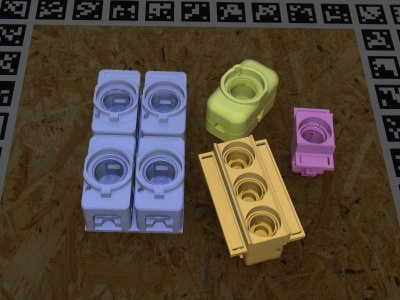}
\put(2,65){\scriptsize$\displaystyle\textbf{\light{11}}$}\end{overpic} &
\begin{overpic}[width=0.2\columnwidth]{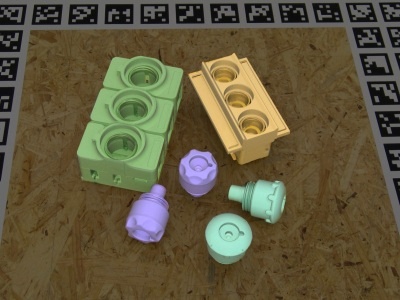}
\put(2,65){\scriptsize$\displaystyle\textbf{\light{12}}$}\end{overpic} &
\begin{overpic}[width=0.2\columnwidth]{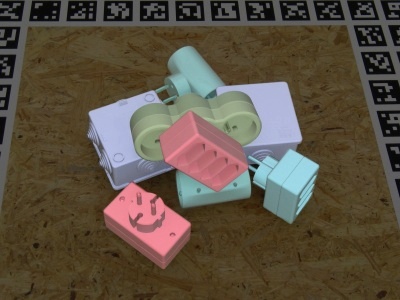}
\put(2,65){\scriptsize$\displaystyle\textbf{\light{13}}$}\end{overpic} &
\begin{overpic}[width=0.2\columnwidth]{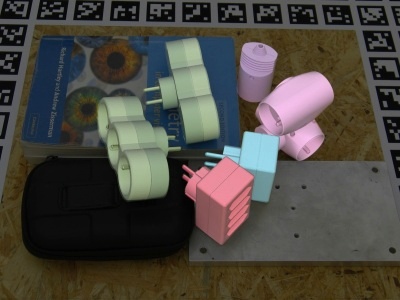}
\put(2,65){\scriptsize$\displaystyle\textbf{\light{14}}$}\end{overpic} &
\begin{overpic}[width=0.2\columnwidth]{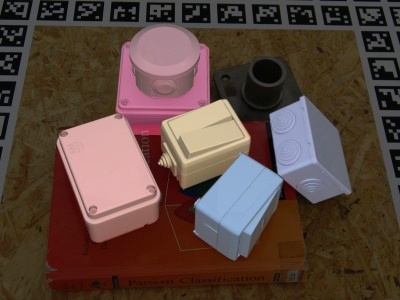}
\put(2,65){\scriptsize$\displaystyle\textbf{\light{15}}$}\end{overpic} \\

\rowcolor{black}
\begin{overpic}[width=0.2\columnwidth]{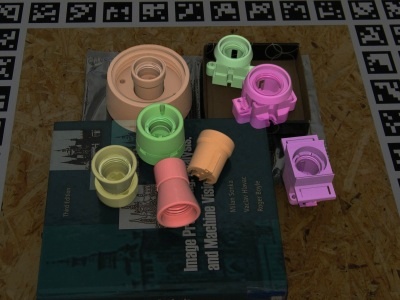}
\put(2,65){\scriptsize$\displaystyle\textbf{\light{16}}$}\end{overpic} &
\begin{overpic}[width=0.2\columnwidth]{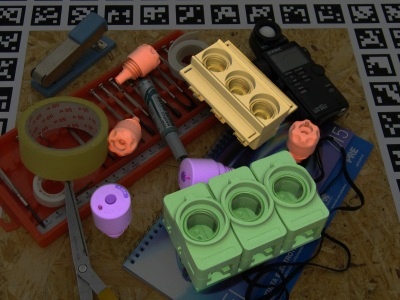}
\put(2,65){\scriptsize$\displaystyle\textbf{\light{17}}$}\end{overpic} &
\begin{overpic}[width=0.2\columnwidth]{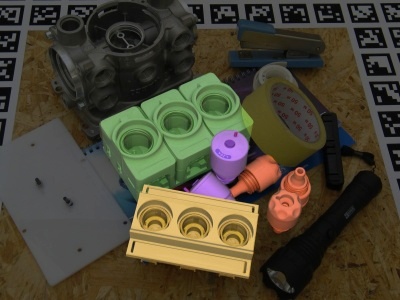}
\put(2,65){\scriptsize$\displaystyle\textbf{\light{18}}$}\end{overpic} &
\begin{overpic}[width=0.2\columnwidth]{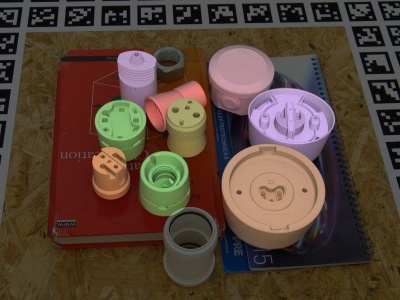}
\put(2,65){\scriptsize$\displaystyle\textbf{\light{19}}$}\end{overpic} &
\begin{overpic}[width=0.2\columnwidth]{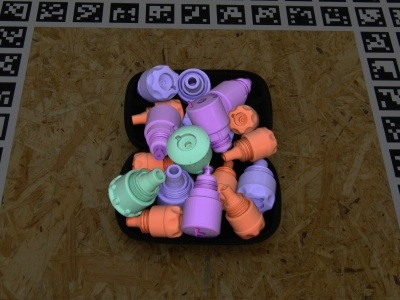}
\put(2,65){\scriptsize$\displaystyle\textbf{\light{20}}$}\end{overpic}
\end{tabular}
\endgroup

\vspace{0.9ex}

\caption{\label{fig:tless_overview}
\textbf{T-LESS objects and scenes.}
The dataset includes 3D models and training images of 30 objects (top) and test images of 20 scenes (bottom) -- shown overlaid with colored 3D object models in the ground truth poses.
The images were captured from a systematically sampled view sphere around an object/scene and are annotated with accurate ground-truth 6D poses of all instances of the modeled objects.
Some objects are a composition of others. For example, objects 7 and 8 are built up from object 6, object 9 is made of three copies of object 10,
and the center part of objects 17 and 18 is nearly identical to object 13.
}
\end{center}
\end{figure}

This chapter introduces T-LESS, a publicly available dataset for 6D object pose estimation that includes 3D models and training and test RGB-D images of thirty
commodity electrical parts. Recognizing and estimating the 6D pose of these objects is challenging because the objects have no significant texture or discriminative color, exhibit symmetries and similarities in shape and size, and some objects are a composition of others (Figures~\ref{fig:tless_challenges} and \ref{fig:tless_overview}).
Objects exhibiting similar properties are common in industrial environments and T-LESS is the first dataset to include such objects.

The training and test images were captured with a triplet of sensors, \ie, a~structured light RGB-D sensor (Primesense Carmine 1.09), %
a time-of-flight RGB-D sensor (Microsoft Kinect v2), and an RGB sensor (Canon IXUS 950 IS). The sensors were time-synchronized and had similar perspectives. The images were captured with an automated acquisition setup
which systematically sampled images from a view sphere, resulting in \mytilde39K training and \mytilde10K test images from each sensor.
The training images show objects in isolation on a black background, while the test images originate from twenty table-top scenes with arbitrarily arranged objects. Complexity of the test scenes varies from those with several isolated objects and a clean background to very challenging ones with multiple instances of several objects and with a high amount of occlusion and clutter. Additionally, the dataset contains two types of 3D mesh models for each object -- one manually created in CAD software and one semi-automatically reconstructed from the training RGB-D images. All instances of the modeled objects in the training and test images are annotated with accurate ground-truth 6D poses -- see the visualizations in Figure~\ref{fig:tless_challenges} for a qualitative and Section~\ref{sec:tless_gt_eval} for a quantitative evaluation of the ground-truth poses.

The dataset is intended for evaluating various flavors of the 6D object pose estimation problem and other related problems such as 2D object detection~\cite{tombari2013bold,hodan2015efficient} and segmentation~\cite{tombari2011online,georgakis2016multiview}. Since images from three types of sensors are available, the dataset allows to study the importance of different input modalities for the aforementioned problems.
Another possibility is to use the training images for evaluating 3D object reconstruction methods~\cite{steinbrucker2014fastfusion} where the provided CAD models can serve as the ground truth.

The objectives in designing T-LESS were to provide a dataset of substantial but manageable size, with a rigorous and complete ground-truth annotation that is accurate to the level of the sensor resolution, and with graded complexity
that makes the dataset reasonably future-proof, \ie, solvable but not solved by the current state-of-the-art methods.
The difficulty of T-LESS is demonstrated in Section~\ref{sec:tless_6d_loc_eval} on the performance of the HashMatch method, which achieves a relatively low accuracy on T-LESS compared to the accuracy on the well-established LM dataset (Section~\ref{sec:iros15_localization}). T-LESS was selected as one of the core datasets in the BOP benchmark (Chapter~\ref{ch:bop}) and, as of 2020, it is still one of the more difficult datasets in the benchmark.

The applied approach to capture the training and test images, annotate the images with ground-truth 6D object poses, and reconstruct the 3D object models is described in Sections~\ref{sec:tless_acquisition_setup}--\ref{sec:tless_gt_poses}. The accuracy of the ground-truth poses and the difficulty of the dataset is assessed in Section~\ref{sec:tless_exp}.
The T-LESS dataset was published in~\cite{hodan2017tless} and is available on the project website: \texttt{\href{http://cmp.felk.cvut.cz/t-less/}{cmp.felk.cvut.cz/t-less}}.

\begin{figure}[t!]
	\begin{center}
		\begin{overpic}[width=0.6\columnwidth]{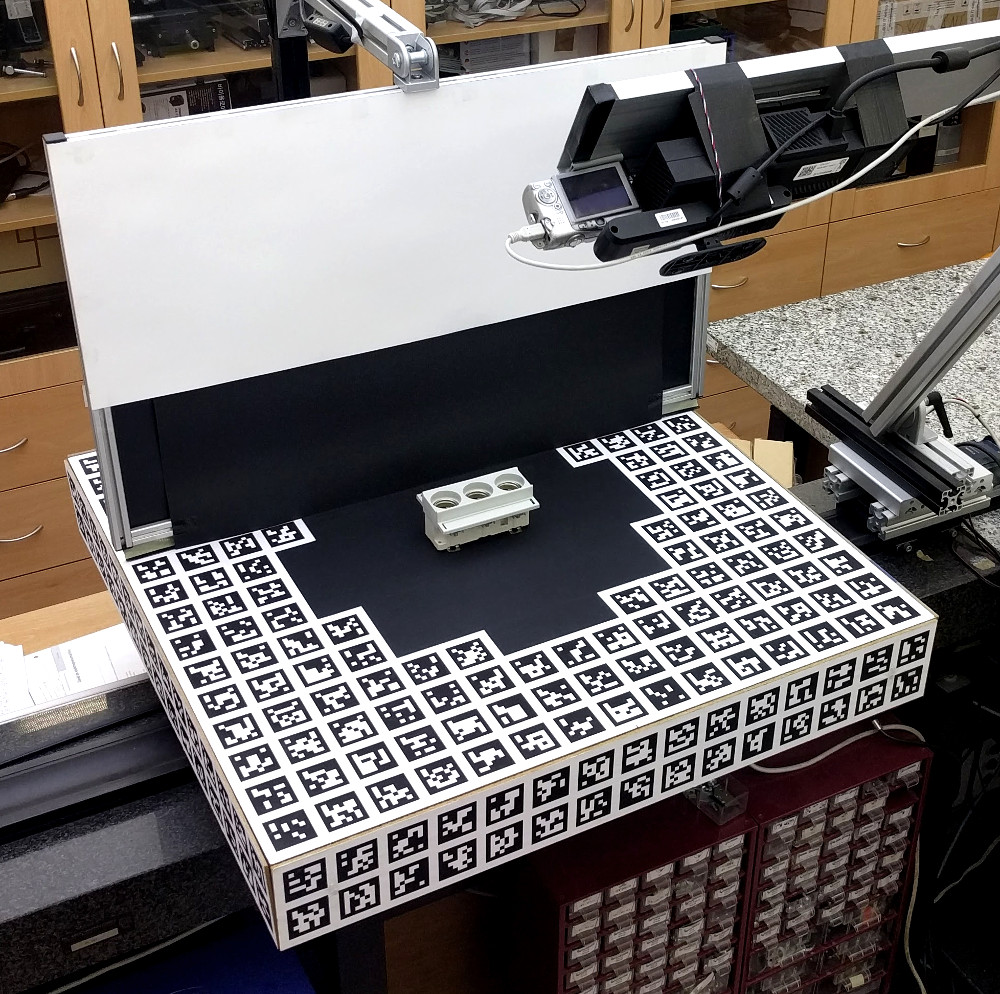}
			\put(80.5,81){\color{white}\circle*{10}}
			\put(78.9,79.25){\large$\displaystyle\textbf{3}$}
			\put(33,25){\color{white}\circle*{10}}
			\put(31.4,23.25){\large$\displaystyle\textbf{1}$}
			\put(34,55){\color{white}\circle*{10}}
			\put(32.4,53.25){\large$\displaystyle\textbf{2}$}
		\end{overpic}
		\vspace{0.9ex}
		\caption{\label{fig:tless_acquisition_setup}
			\textbf{Acquisition setup:} (1)~a turntable with a marker field, (2)~a screen ensuring a black background in training images (removed when capturing test images), (3)~a triplet of sensors attached to a jig with adjustable tilt.}
	\end{center}
\end{figure}

\section{Acquisition Setup}
\label{sec:tless_acquisition_setup}

The training and test images were captured with the aid of a setup shown in Figure~\ref{fig:tless_acquisition_setup}.
The objects were placed on a turntable and the sensors were attached to a jig with adjustable tilt.
Poses of the sensors were estimated using a marker field that was affixed to the turntable.
The marker field was extended vertically to the turntable sides to facilitate estimating the poses at lower elevations.
When capturing training images, the objects were placed in the middle of the turntable and in the front of a black screen that ensured a uniform background from all elevations. To introduce a non-uniform background in test images, a sheet of plywood with markers at its edges was placed on 
the turntable.
In some scenes, the objects were placed on other objects such as books to
invalidate a ground plane assumption that might be made by some methods.
The depth of object surfaces in the training and test images is in the range of $0.53$--$0.92\,$m, which is within the sensing ranges of the
RGB-D sensors:\ $0.35$--$1.4\,$m for Carmine and $0.5$--$4.5\,$m for Kinect.

\section{Calibration of Sensors}
\label{sec:tless_calibration}

The intrinsic and distortion parameters of the sensors were estimated with the standard checkerboard-based procedure from OpenCV~\cite{bradski2008learning}.
High calibration accuracy is affirmed by low values of the root-mean-square re-projection error calculated at the checkerboard corners: $0.51\,\text{px}$ for Carmine, $0.35\,\text{px}$ for Kinect, and $0.43\,\text{px}$ for Canon. For the RGB-D sensors, the calibration was performed with the RGB images. The depth images were aligned to the RGB images by the factory depth-to-color registration from the official SDK's (OpenNI 2.2 and Kinect for Windows SDK 2.0). The color and aligned depth images, which are included in the dataset, are already processed to remove radial distortion. %

All sensors were synchronized and extrinsically calibrated with respect to the turntable to enable registration of any pair of images.
Since the images were taken while the turntable was moving, precise synchronization was essential to ensure that the images from the three sensors are taken from similar viewpoints. %
Poses of the sensors were estimated using fiducial BCH-code markers from ARToolKitPlus~\cite{wagner2007artoolkitplus}. Specifically, the 2D detections of particular markers in an image combined with the knowledge of their physical location on the turntable provided a set of 2D-3D correspondences. The sensor pose was then estimated from the correspondences by the P\emph{n}P-RANSAC algorithm~\cite{fischler1981random,lepetit2009epnp} and refined by a non-linear minimization of the cumulative re-projection error~\cite{lourakis13model}.
The root-mean-square re-projection error, which was calculated at the marker corners in all test images, is $1.27\,$px for Carmine, $1.37\,$px for Kinect, and $1.50\,$px for Canon.
This measure combines errors in the intrinsic calibration, marker field detection and pose estimation, and is therefore larger than the aforementioned error of the intrinsic calibration.

\section{Training and Test Images}
\label{sec:tless_train_test_images}

The dataset includes training images of every object in isolation captured from a full view sphere with the acquisition setup described in Section~\ref{sec:tless_acquisition_setup}.
The training images were captured from elevation angles $-85^{\circ}$ to $85^{\circ}$ with a $10^{\circ}$ step and from the complete range of azimuth angles with a $5^{\circ}$ step.
Views from the upper and lower hemispheres were taken separately, turning the object upside down in between.
In total, the dataset includes $18 \cdot 72 = 1296$ training images per object from each sensor.
Exceptions are objects 19 and 20 which were captured only from the upper hemisphere ($648$ images from elevation $85^{\circ}$ to $5^{\circ}$). These objects are horizontally symmetric and the views from the upper hemisphere are therefore sufficient.
The test scenes were captured from elevation angles $15^{\circ}$ to $75^{\circ}$ with a $10^{\circ}$ step and from the complete range of azimuth angles with a $5^{\circ}$ step. In total, the dataset includes $7 \cdot 72 = 504$ test images per test scene from each sensor.

The captured images were cropped to remove regions at the image borders, which mostly showed the fiducial markers and were irrelevant for 6D object pose estimation.
To hide markers that were still visible in the training images after the cropping, and therefore to ensure a black background everywhere around the objects, the training images were gradually darkened from the object mask towards the image borders.

\begin{figure}[t!]
	\begin{center}
		\begingroup
		\footnotesize
		\begin{tabular}{ @{}c@{ } @{}c@{ } @{}c@{ } }
			Primesense Carmine 1.09, RGB-D & Microsoft Kinect v2, RGB-D & Canon IXUS 950 IS, RGB \vspace{0.5ex} \\
			Training: $400\times400\,$px & Training: $400\times400\,$px & Training: $1900\times1900\,$px \vspace{0.5ex} \\
			\includegraphics[width=0.3265\columnwidth]{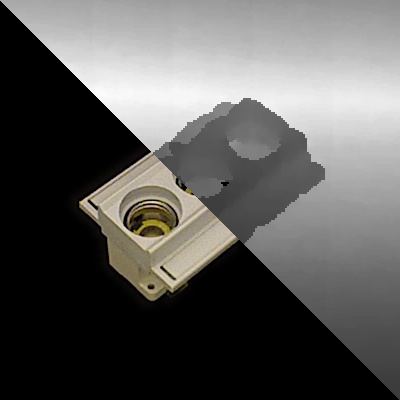} &
			\includegraphics[width=0.3265\columnwidth]{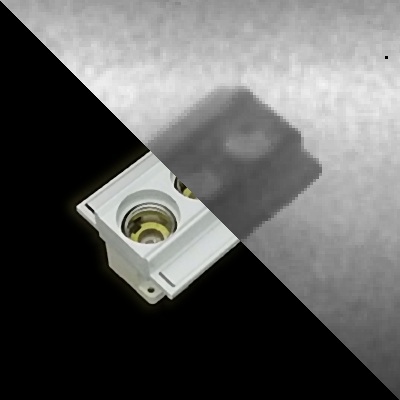} &
			\includegraphics[width=0.3265\columnwidth]{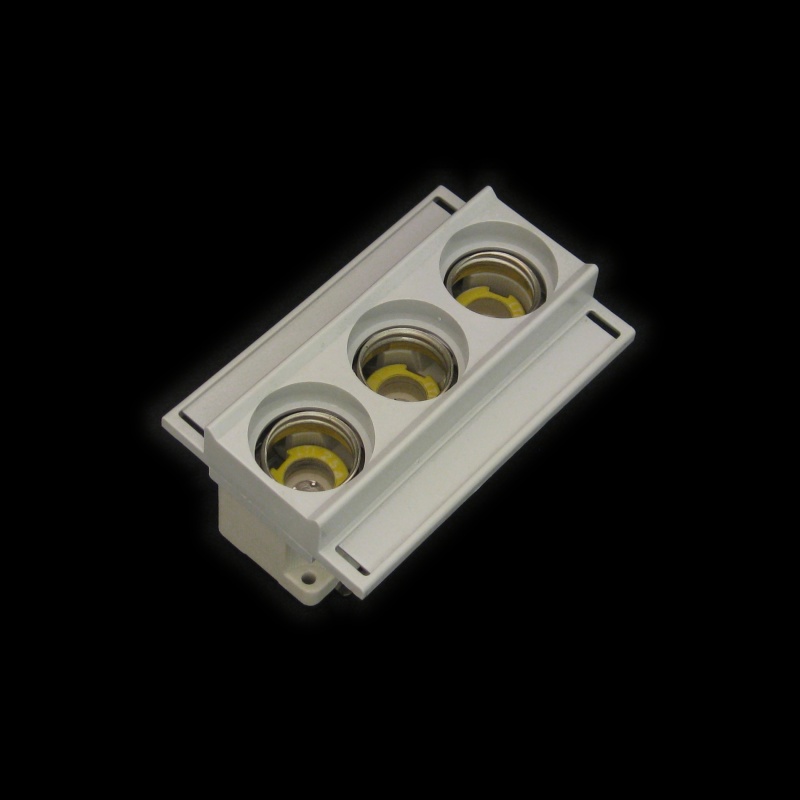} \\
			Test: $720\times540\,$px & Test: $720\times540\,$px & Test: $2560\times1920\,$px \vspace{0.5ex} \\
			\includegraphics[width=0.3265\columnwidth]{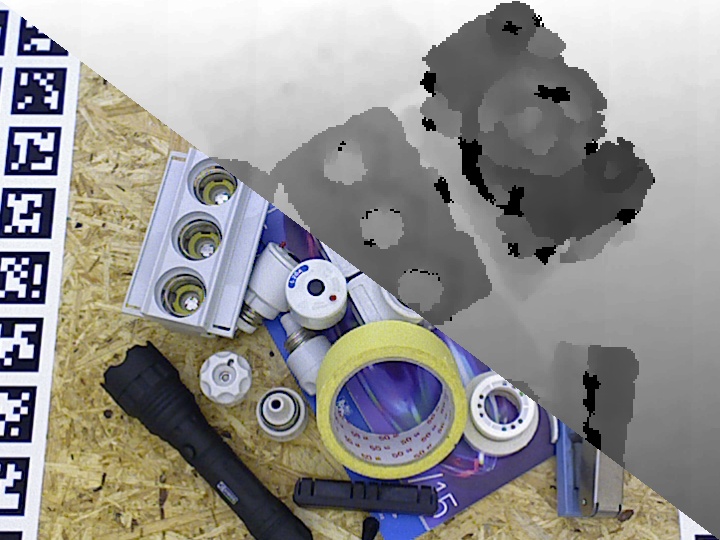} &
			\includegraphics[width=0.3265\columnwidth]{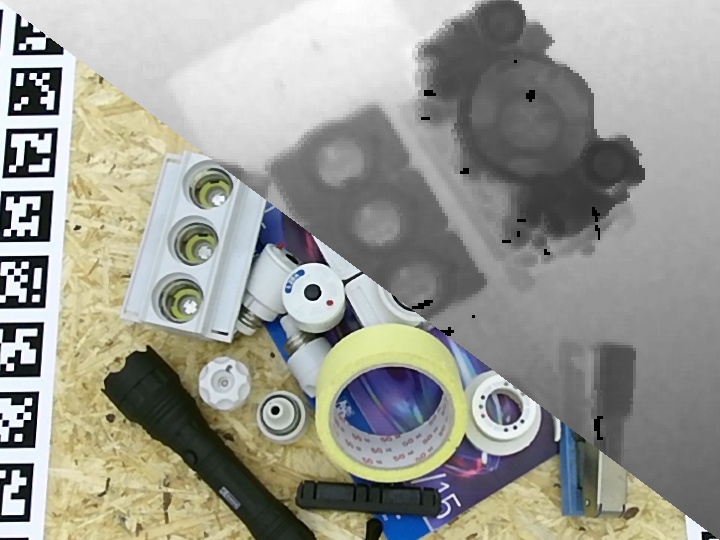} &
			\includegraphics[width=0.3265\columnwidth]{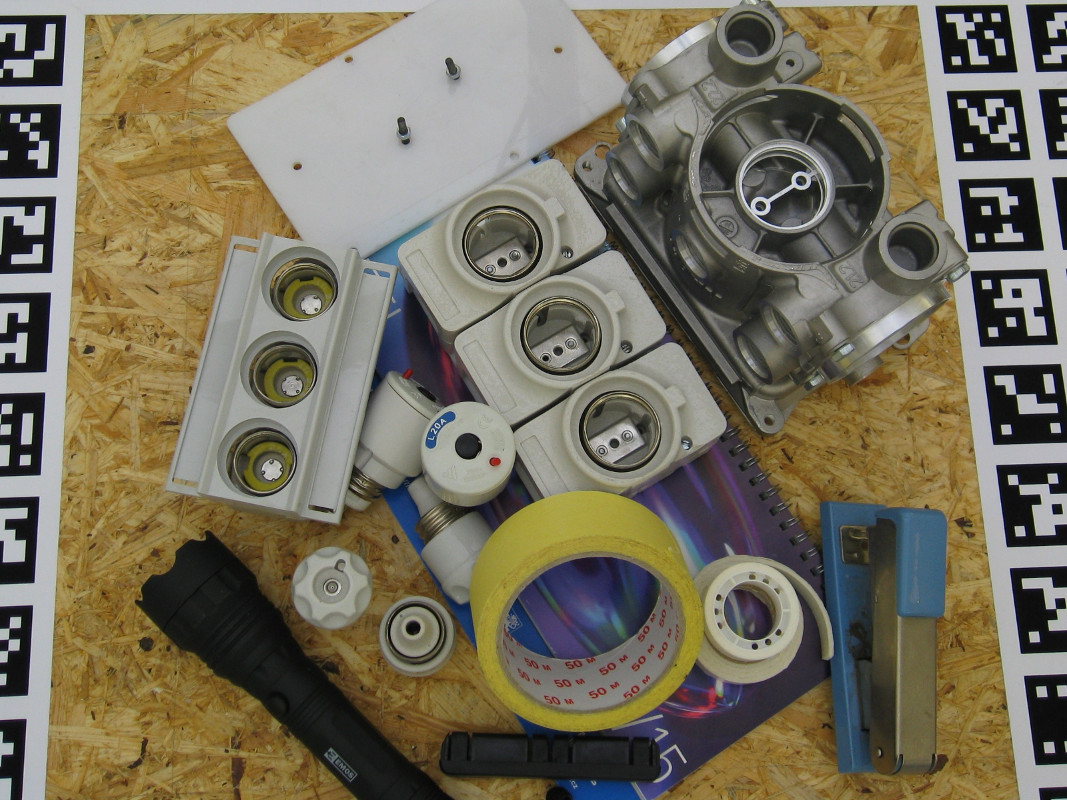}
		\end{tabular}
		\endgroup
		\caption{\label{fig:tless_rgbd_views}
			\textbf{Training and test images captured by three sensors.}
			For the RGB-D images, the left halves show the RGB channels and the right halves show the depth channel.
		}
	\end{center}
\end{figure}

The provided training RGB-D images from Carmine and Kinect are of resolution $400 \times 400\,\text{px}$, the training RGB images from Canon of $1900 \times 1900\,\text{px}$,
the test RGB-D images from Carmine and Kinect of $720 \times 540\,\text{px}$,
and the test RGB images from Canon of $2560 \times 1920\,\text{px}$.
Example images from the three sensors are in Figure~\ref{fig:tless_rgbd_views}.

\section{Depth Correction}
\label{sec:tless_depth_correction}

Similarly to~\cite{foix2011lock,sturm2012benchmark}, the depth measurements from the RGB-D sensors were observed to exhibit a systematic error.
To eliminate the error, the depth measurements were collected at image locations of the marker corners together with their expected values calculated from the known marker coordinates.
The measurements were collected from the range of $0.53$\,--\,$0.92\,$m in which the objects appear in the training and test images.
The following
correction models were then found by least-squares fitting: $d_c = 1.0247 \cdot d - 5.19$ for Carmine, and $d_c = 1.0266 \cdot d - 26.88$ for Kinect, where $d$ and $d_c$ are the measured and corrected depth values in millimeters.
The correction reduced the mean absolute difference from the expected depth from $12.4$ to $2.8\,\text{mm}$ for Carmine and from $7.0$ to $3.6\,\text{mm}$ for Kinect.
The correction is already applied to all depth images included in the dataset.

\section{3D Object Models}
\label{sec:tless_obj_models}

\begin{figure}[t!]
	\begin{center}
		\begingroup
		\begin{tabular}{ @{}c@{ } }
			\includegraphics[width=0.8\columnwidth]{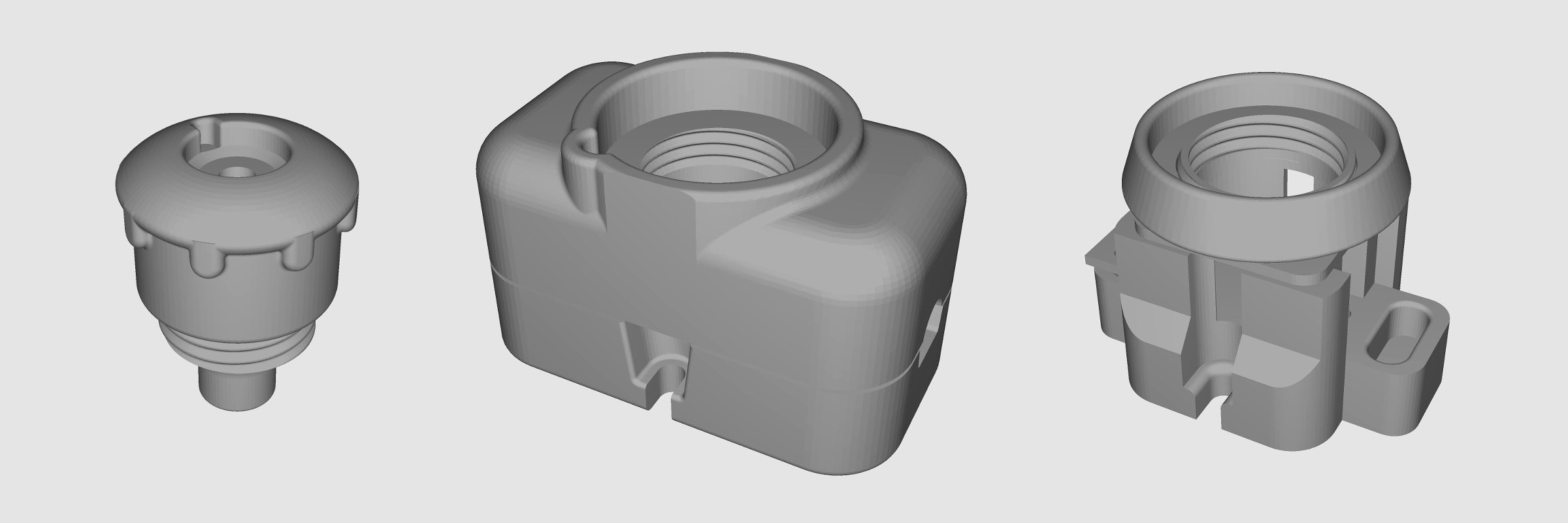} \\
			\includegraphics[width=0.8\columnwidth]{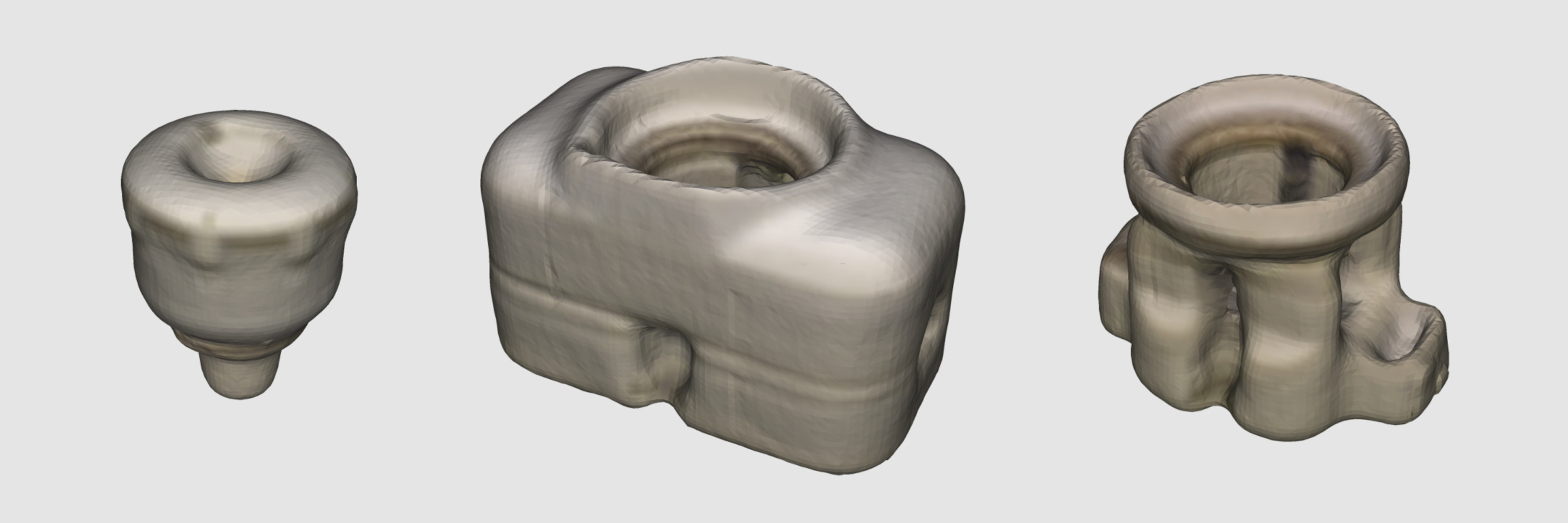}

		\end{tabular}
		\endgroup
		\caption{\label{fig:tless_obj_models} \textbf{Examples of 3D object models.} Top: Manually created CAD models. Bottom: Semi-automatically reconstructed models which include also surface color. Surface normals at model vertices are included in both model types.}
	\end{center}
\end{figure}

A manually-created CAD model and a semi-automatically reconstructed model are available for each object (Figure~\ref{fig:tless_obj_models}).
Models of both types are provided in the form of 3D meshes with surface normals.
The surface normal at a mesh vertex was calculated by MeshLab~\cite{cignoni2008meshlab} as the angle-weighted sum of face normals incident to the vertex~\cite{thurrner1998computing}. The surface color, stored at the mesh vertices, is available only in the reconstructed models. %

The reconstructed models were obtained using the volumetric 3D mapping method by Steinbr\"ucker \etal~\cite{steinbrucker2014fastfusion}.
The input to the method was the RGB-D training images from Carmine and the camera poses estimated from the fiducial markers (Section~\ref{sec:tless_calibration}). For each object, two partial models were reconstructed first, one from the upper and one from the lower view hemisphere. The partial models were aligned by the Iterative Closest Point (ICP) algorithm applied to the model vertices, followed by manual refinement to ensure correct registration of surface details that are visible only in color. 
The camera poses were then brought into a common reference frame by the found alignment transformation and the full object model was reconstructed using images from all cameras.
The models contained some minor artifacts, \eg, small spikes, which were removed manually.
Some of the objects have small shiny metal parts, such as the plug poles, which are not reconstructed because their depth was not reliably captured by the depth sensor.

The reconstructed models were aligned to the CAD models by the ICP algorithm followed by manual refinement.
Models of both types are therefore defined in the same coordinate system and share the same ground-truth 6D poses.
The origin of the model coordinate system coincides with the center of the 3D bounding box of the CAD model.

The geometrical similarity of the two model types was assessed by calculating the average distance from vertices of the reconstructed models to the closest points on the surface of the CAD models. The average distance over all object models is $1.01\,$mm, which is very low compared to the size of the objects that ranges from $58.13\,$mm for object 13 to $217.16\,$mm for object 8.
The Metro software~\cite{cignoni1998metro}
was used to measure the distance.

\section{Ground-Truth 6D Object Poses}
\label{sec:tless_gt_poses}

To collect the ground-truth 6D object poses for images of a test scene, a dense 3D model of the scene was first reconstructed from all images of the scene by the method of Steinbr\"ucker et al.~\cite{steinbrucker2014fastfusion}.
The CAD object models were then manually aligned to the scene model, the object models were rendered into several high-resolution images from the Canon camera using known camera-to-turntable
transformations, misalignments were identified in the renderings, and the poses were manually refined accordingly.
The refinement was repeated until a satisfactory alignment of the renderings with the images of the scene was achieved.

\section{Design Validation and Experiments}
\label{sec:tless_exp}

This section evaluates the accuracy of the ground-truth 6D object poses and assesses the difficulty of the 6D object localization task on the T-LESS dataset.

\subsection{Accuracy of the Ground-Truth 6D Object Poses}
\label{sec:tless_gt_eval}

\begin{figure}[!t]
	\begin{center}
		\footnotesize
		\begin{tabularx}{0.69\linewidth}{l Y Y Y Y}
			\toprule
			Sensor, model type & $\mu_\delta$ & $\sigma_\delta$ & $\mu_{|\delta|}$ & $\text{med}_{|\delta|}$ \\
			\midrule
			Carmine, CAD & -0.60 & 8.12 & 4.53 & 2.57 \\
			Carmine, reconstructed & -0.79 & 7.72 & 4.28 & 2.46 \\
			Kinect, CAD & 4.46 & 11.76 & 8.76 & 5.67 \\
			Kinect, reconstructed & 4.08 & 11.36 & 8.40 & 5.45 \\
			\bottomrule
		\end{tabularx}
		\captionof{table}{\label{tab:tless_gt_error}
			\textbf{Quality of the ground-truth 6D object poses.}
			Statistics of differences between the depth of object models in the ground-truth poses and the captured depth (in millimeters) -- $\mu_\delta$ and $\sigma_\delta$ are the mean and the standard deviation of the differences, $\mu_{|\delta|}$ and $\text{med}_{|\delta|}$ are the mean and the median of the absolute differences.}
	\end{center}

	\begin{center}
		\begingroup
		\begin{tabular}{c}
			\includegraphics[width=0.69\columnwidth]{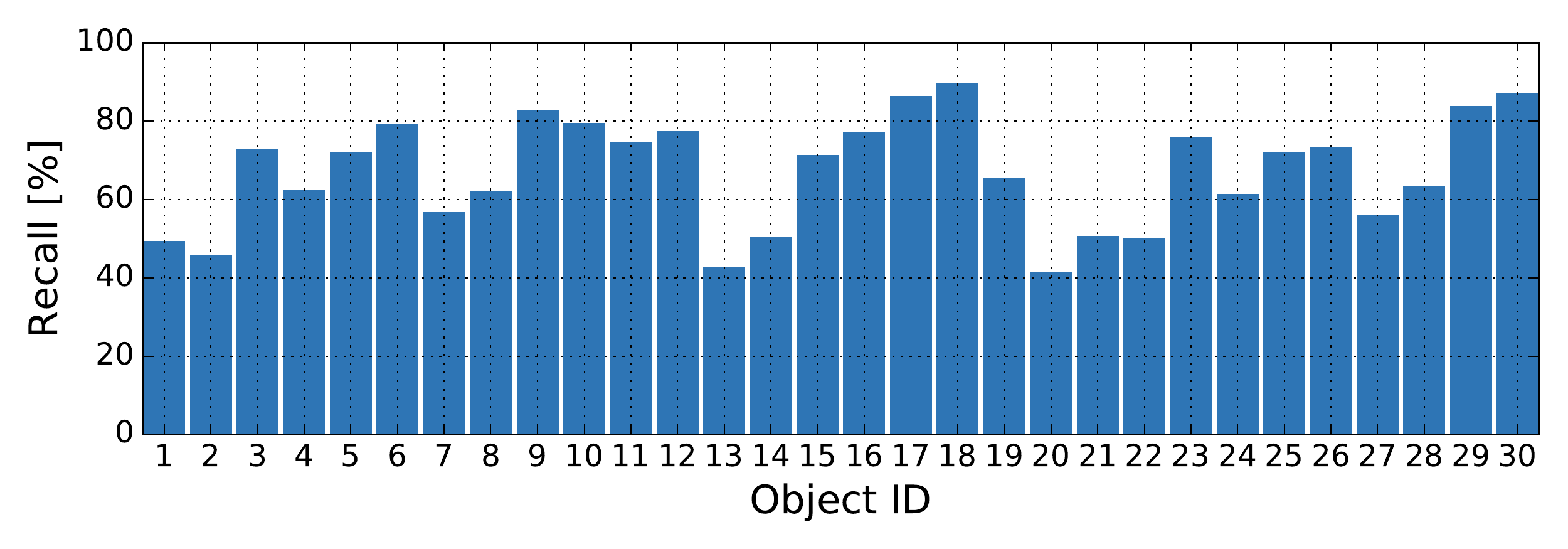} \\
			\includegraphics[width=0.69\columnwidth]{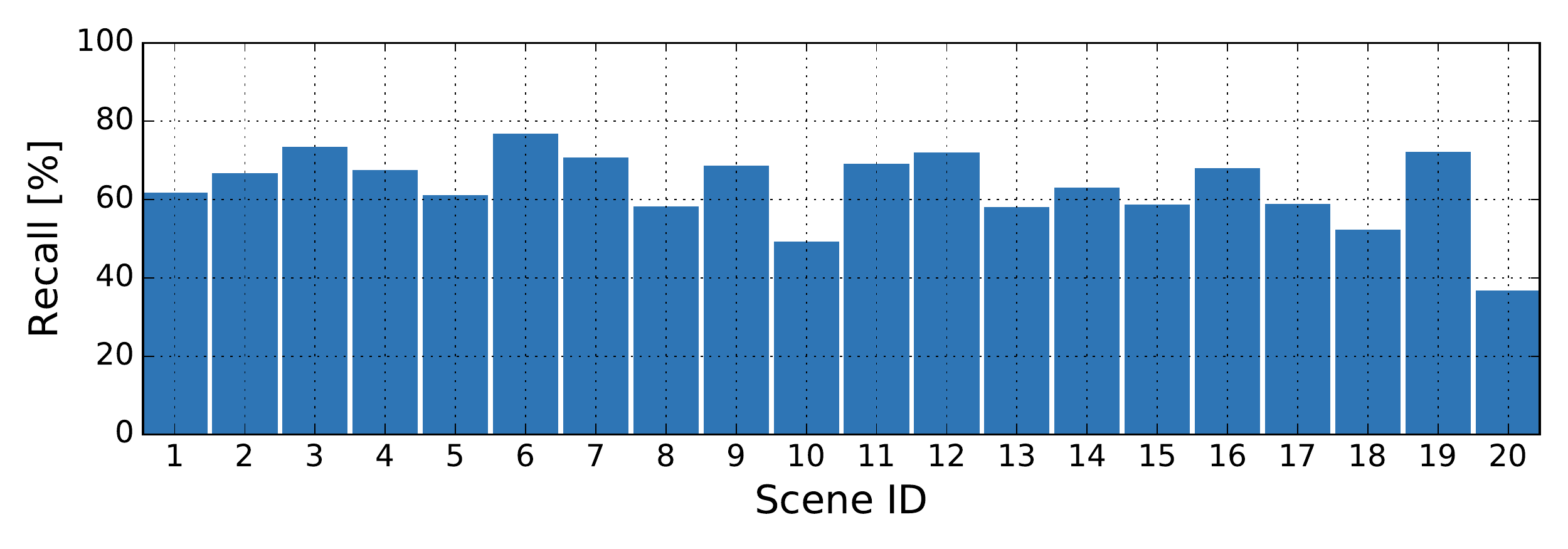} \\ 
			\includegraphics[width=0.69\columnwidth]{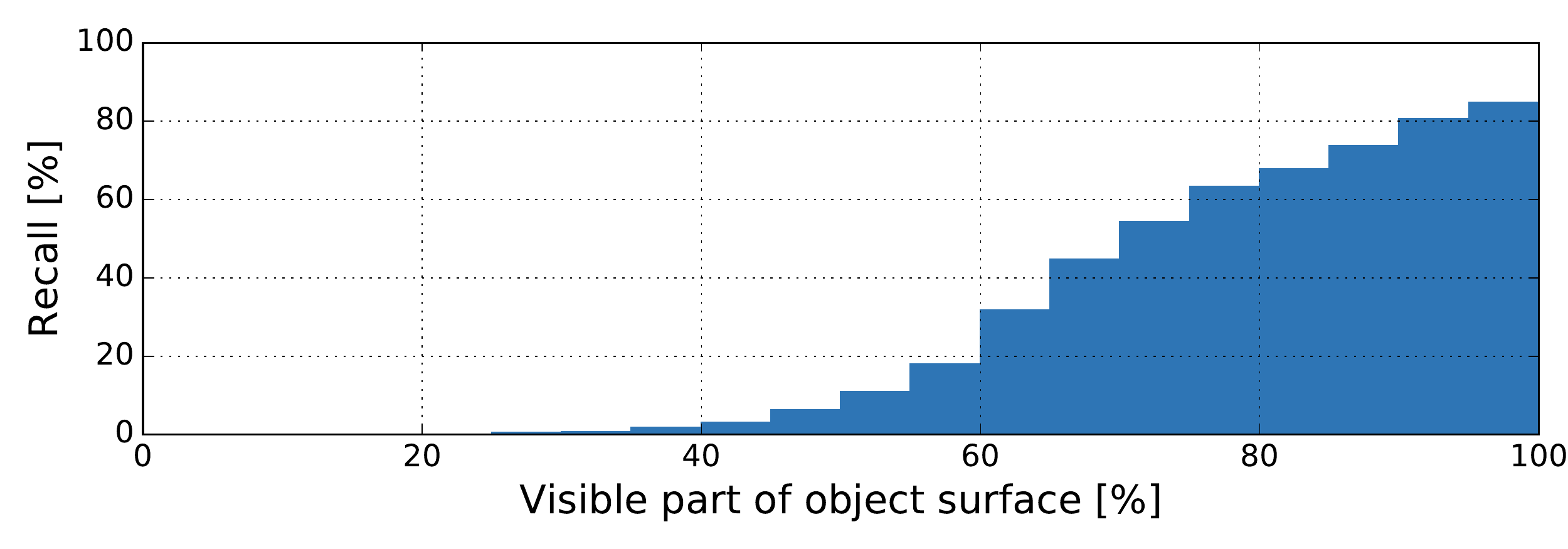}
		\end{tabular}
		\endgroup
		\caption{\label{fig:tless_loc_results}
			\textbf{HashMatch results on T-LESS.}
			Shown are the recall rates
			per object (top), per scene (middle), and \wrt the visible fraction of the object surface (bottom).
		}
	\end{center}
\end{figure}

The depth images, corrected as described in Section~\ref{sec:tless_depth_correction}, were compared with depth images obtained by graphically rendering the 3D object models in the ground-truth poses.
Table~\ref{tab:tless_gt_error} shows the statistics of the difference $\delta$ between depth values in the two images, which was calculated at pixels with a valid depth value in both images and
aggregated over all training and test images.
Differences exceeding $5\,$cm were considered to be outliers and pruned before calculating the statistics. The outliers represent $\mytilde2.5\,\%$ of the differences and are typically caused by erroneous depth measurements or by occlusion induced by distractor objects in the case of test images.

In the case of Carmine, the captured depth images align well with the rendered depth images as indicated by the mean difference $\mu_{\delta}$ being close to zero.
In the case of Kinect, the RGB and depth images were observed to be slightly misregistered, which is the cause of the positive bias in $\mu_{\delta}$.
The average absolute difference $\mu_{|\delta|}$ is less than $5\,\text{mm}$ for Carmine and $9\,\text{mm}$ for Kinect, which is near the accuracy of the sensors~\cite{khoshelham2012accuracy} and is relatively small compared to the size of the T-LESS objects.
The error statistics are slightly favorable for the reconstructed object models
which were created from the captured depth images and exhibit similar characteristics and artifacts as the images. For example, the plug poles are not visible in the depth images and are therefore missing in the reconstructed models, but are present in the CAD models.

\subsection{6D Object Localization} \label{sec:tless_6d_loc_eval}

The difficulty of 6D object localization on the T-LESS dataset is assessed with HashMatch, the template-matching method described in Chapter~\ref{ch:method_template}. The method is evaluated on all test RGB-D images from Carmine, the templates are generated from the RGB-D training images from Carmine, and the CAD models are used in the pose refinement stage.
The accuracy of the method is measured by the recall rate, \ie, the fraction of annotated
object instances for which a correct pose is estimated. A pose estimate is considered correct if the ADI error defined in Section~\ref{sec:rel_eval} is below $10\%$ of the object diameter.

Figure~\ref{fig:tless_loc_results} presents the recall rates per object (top) and per scene (middle).
The lowest recall rates are achieved on objects that are similar to other objects.
For example, object 1 is often confused with object 2, as are objects 20, 21 and 22.
Likewise, test scenes containing similar objects are more difficult, with the hardest one being scene 20 that contains multiple instances of several similar objects.
Besides the mutual similarity of objects, occlusion is another challenge presented in T-LESS -- as shown at the bottom of Figure~\ref{fig:tless_loc_results}, the recall rate decreases proportionally with the visible fraction of objects.

The recall rate of HashMatch averaged over all T-LESS objects is $67.2\%$, which is noticeably lower than the recall rate of $95.4\%$ which this method achieves on the well-established and rather saturated LM dataset~\cite{hinterstoisser2012accv}. A margin for improvement, indicating new challenges brought by the T-LESS dataset,
is evident also from the results of the recent BOP Challenge 2020 described in Section~\ref{sec:bop_challenge_2020}.

	\chapter[BOP:\ The Benchmark for 6D Object Pose Estimation]{BOP\\{\Large The Benchmark for 6D Object Pose Estimation}} \label{ch:bop}

The progress of research in computer vision has been strongly influenced by challenges and benchmarks, which enable to evaluate and compare methods and better understand their limitations. The Middlebury benchmark~\cite{scharstein2002taxonomy,scharstein2007learning} for depth from stereo and optical flow estimation was one of the first that gained large attention. The PASCAL VOC challenge~\cite{everingham2010pascal}, based on a photo collection from the internet, was the first to standardize the evaluation of object detection and image classification. It was followed by 
the ImageNet challenge~\cite{russakovsky2015imagenet}
which has pushed image classification methods to new levels of accuracy. The key was a large-scale dataset that enabled training of deep neural networks, which then quickly became a game-changer for many other tasks~\cite{krizhevsky2012imagenet}.
With increasing maturity of computer vision methods, recent benchmarks moved to real-world scenarios. A great example is the KITTI benchmark \cite{geiger2012kitti} focusing on problems related to autonomous driving. It showed that methods ranking high on established benchmarks, such as the Middlebury, may perform below average when moved outside the laboratory conditions.

Many methods for 6D object pose estimation have been published recently, \eg, \cite{tejani2014latent,krull2015learning,hodan2015detection,brachmann2016uncertainty,wohlhart2015learning,kehl2017ssd,rad2017bb8,michel2017global}, but it was unclear which methods perform well and in which scenarios.
The most commonly used LM dataset by Hinterstoisser \etal \cite{hinterstoisser2012accv} was not intended as a general benchmark and has several limitations: the lighting conditions are constant and the objects are easy to distinguish, unoccluded, and located around the image center. Some of these limitations were later addressed -- Brachmann \etal \cite{brachmann2014learning} created the LM-O dataset by adding ground-truth annotation for occluded objects in the LM dataset, the T-LESS dataset described in Chapter~\ref{ch:tless} introduced industry-relevant objects with symmetries and similarities, and Drost \etal \cite{drost2017introducing} created the ITODD dataset with objects from reflective materials. However, the datasets were provided in different formats and there was no standard evaluation methodology.
New methods were usually compared with only a few competitors on a small subset of datasets.

\begin{figure}[!t]
	\begin{center}
		
		\vspace{-1.0ex}
		
		\begingroup
		\setlength{\tabcolsep}{1pt} %
		\renewcommand{\arraystretch}{1} %
		\tiny
		
		\begin{tabu}{*{30}{c}}
		\includegraphics[scale=0.21]{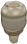} &
		\includegraphics[scale=0.21]{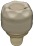} &
		\includegraphics[scale=0.21]{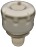} &
		\includegraphics[scale=0.21]{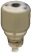} &
		\includegraphics[scale=0.21]{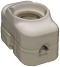} &
		\includegraphics[scale=0.21]{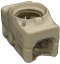} &
		\includegraphics[scale=0.21]{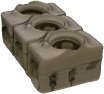} &
		\includegraphics[scale=0.21]{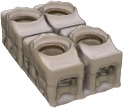} &
		\includegraphics[scale=0.21]{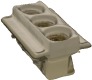} &
		\includegraphics[scale=0.21]{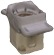} &
		\includegraphics[scale=0.21]{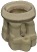} &
		\includegraphics[scale=0.21]{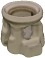} &
		\includegraphics[scale=0.21]{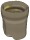} &
		\includegraphics[scale=0.21]{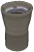} &
		\includegraphics[scale=0.21]{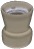} &
		\includegraphics[scale=0.21]{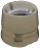} &
		\includegraphics[scale=0.21]{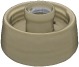} &
		\includegraphics[scale=0.21]{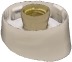} &
		\includegraphics[scale=0.21]{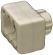} &
		\includegraphics[scale=0.21]{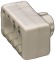} &
		\includegraphics[scale=0.21]{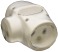} &
		\includegraphics[scale=0.21]{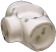} &
		\includegraphics[scale=0.21]{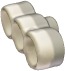} &
		\includegraphics[scale=0.21]{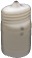} &
		\includegraphics[scale=0.21]{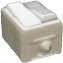} &
		\includegraphics[scale=0.21]{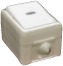} &
		\includegraphics[scale=0.21]{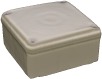} &
		\includegraphics[scale=0.21]{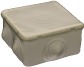} &
		\includegraphics[scale=0.21]{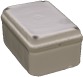} &
		\includegraphics[scale=0.21]{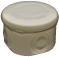}
		\\

		\rowfont{\color{black!75}}1 & 2 & 3 & 4 & 5 & 6 & 7 & 8 & 9 & 10 & 11 & 12 & 13 & 14 & 15 & 16 & 17 & 18 & 19 & 20 & 21 & 22 & 23 & 24 & 25 & 26 & 27 & 28 & 29 & 30 \\
		\multicolumn{30}{c}{\vspace{-1.5ex}} \\
		\multicolumn{30}{c}{{\scriptsize T-LESS}}
		\end{tabu} \vspace{0.0ex} \\
	
		\begin{tabu}{*{28}{c}}
		\includegraphics[scale=0.21]{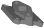} &
		\includegraphics[scale=0.21]{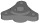} &
		\includegraphics[scale=0.21]{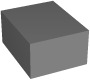} &
		\includegraphics[scale=0.21]{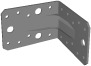} &
		\includegraphics[scale=0.21]{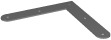} &
		\includegraphics[scale=0.21]{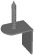} &
		\includegraphics[scale=0.21]{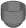} &
		\includegraphics[scale=0.21]{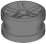} &
		\includegraphics[scale=0.21]{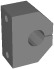} &
		\includegraphics[scale=0.21]{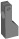} &
		\includegraphics[scale=0.21]{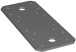} &
		\includegraphics[scale=0.21]{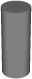} &
		\includegraphics[scale=0.21]{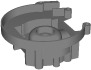} &
		\includegraphics[scale=0.21]{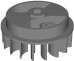} &
		\includegraphics[scale=0.21]{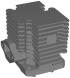} &
		\includegraphics[scale=0.21]{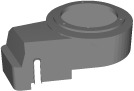} &
		\includegraphics[scale=0.21]{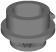} &
		\includegraphics[scale=0.21]{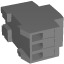} &
		\includegraphics[scale=0.21]{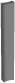} &
		\includegraphics[scale=0.21]{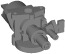} &
		\includegraphics[scale=0.21]{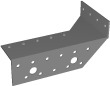} &
		\includegraphics[scale=0.21]{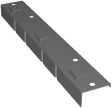} &
		\includegraphics[scale=0.21]{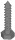} &
		\includegraphics[scale=0.21]{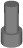} &
		\includegraphics[scale=0.21]{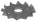} &
		\includegraphics[scale=0.21]{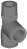} &
		\includegraphics[scale=0.21]{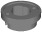} &
		\includegraphics[scale=0.21]{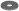} \\
		
		\rowfont{\color{black!75}}1 & 2 & 3 & 4 & 5 & 6 & 7 & 8 & 9 & 10 & 11 & 12 & 13 & 14 & 15 & 16 & 17 & 18 & 19 & 20 & 21 & 22 & 23 & 24 & 25 & 26 & 27 & 28 \\
		\multicolumn{28}{c}{\vspace{-1.5ex}} \\
		\multicolumn{28}{c}{{\scriptsize ITODD}}
		\end{tabu} \vspace{-0.5ex} \\
	
		\begin{tabu}{*{22}{c}}
		\includegraphics[scale=0.21]{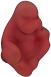} &
		\includegraphics[scale=0.21]{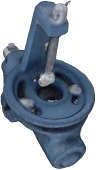} &
		\includegraphics[scale=0.21]{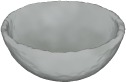} &
		\includegraphics[scale=0.21]{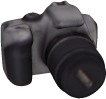} &
		\includegraphics[scale=0.21]{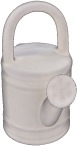} &
		\includegraphics[scale=0.21]{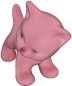} &
		\includegraphics[scale=0.21]{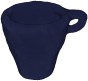} &
		\includegraphics[scale=0.21]{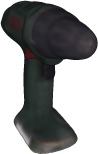} &
		\includegraphics[scale=0.21]{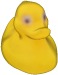} &
		\includegraphics[scale=0.21]{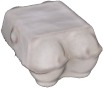} &
		\includegraphics[scale=0.21]{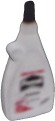} &
		\includegraphics[scale=0.21]{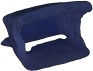} &
		\includegraphics[scale=0.21]{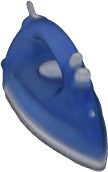} &
		\includegraphics[scale=0.21]{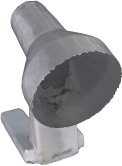} &
		\includegraphics[scale=0.21]{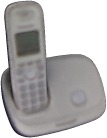} &
		&
		\includegraphics[scale=0.21]{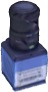} &
		\includegraphics[scale=0.21]{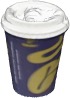} &
		\includegraphics[scale=0.21]{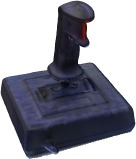} &
		\includegraphics[scale=0.21]{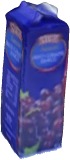} &
		\includegraphics[scale=0.21]{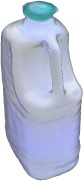} &
		\includegraphics[scale=0.21]{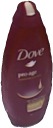}
		\\
		
		\rowfont{\color{black!75}}1 & 2 & 3 & 4 & 5 & 6 & 7 & 8 & 9 & 10 & 11 & 12 & 13 & 14 & 15 & \hspace{2em} & 1 & 2 & 3 & 4 & 5 & 6 \\
		\multicolumn{22}{c}{\vspace{-1.5ex}} \\
		\multicolumn{15}{c}{{\scriptsize LM/LM-O}} & & \multicolumn{6}{c}{{\scriptsize IC-MI/IC-BIN}}
		\end{tabu} \vspace{0.0ex} \\
	
		\begin{tabu}{*{17}{c}}
		\includegraphics[scale=0.21]{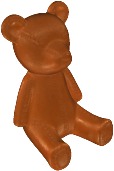} &
		\includegraphics[scale=0.21]{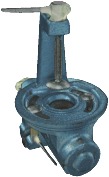} &
		\includegraphics[scale=0.21]{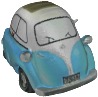} &
		\includegraphics[scale=0.21]{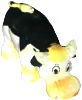} &
		\includegraphics[scale=0.21]{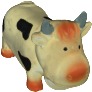} &
		\includegraphics[scale=0.21]{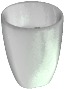} &
		\includegraphics[scale=0.21]{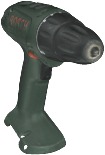} &
		\includegraphics[scale=0.21]{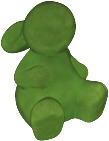} &
		\includegraphics[scale=0.21]{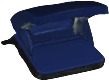} &
		\includegraphics[scale=0.21]{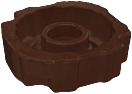} &
		\includegraphics[scale=0.21]{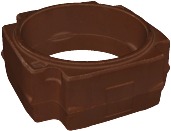} &
		\includegraphics[scale=0.21]{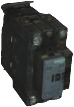} &
		\includegraphics[scale=0.21]{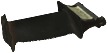} &
		\includegraphics[scale=0.21]{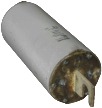} &
		\includegraphics[scale=0.21]{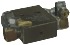} &
		\includegraphics[scale=0.21]{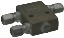} &
		\includegraphics[scale=0.21]{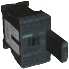} \\
		\rowfont{\color{black!75}}1 & 2 & 3 & 4 & 5 & 6 & 7 & 8 & 9 & 10 & 11 & 12 & 13 & 14 & 15 & 16 & 17 \\
		\end{tabu} \\
		
		\begin{tabu}{*{16}{c}}
		\includegraphics[scale=0.21]{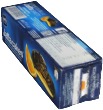} &
		\includegraphics[scale=0.21]{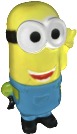} &
		\includegraphics[scale=0.21]{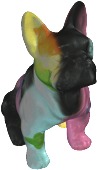} &
		\includegraphics[scale=0.21]{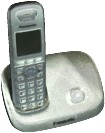} &
		\includegraphics[scale=0.21]{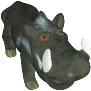} &
		\includegraphics[scale=0.21]{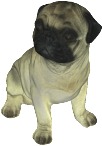} &
		\includegraphics[scale=0.21]{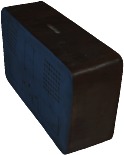} &
		\includegraphics[scale=0.21]{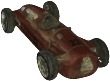} &
		\includegraphics[scale=0.21]{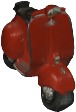} &
		\includegraphics[scale=0.21]{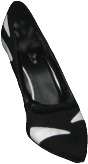} &
		\includegraphics[scale=0.21]{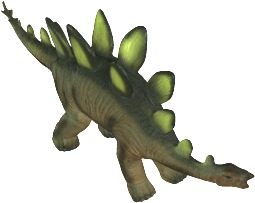} &
		\includegraphics[scale=0.21]{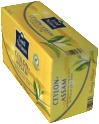} &
		\includegraphics[scale=0.21]{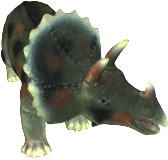} &
		\includegraphics[scale=0.21]{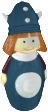} &
		\includegraphics[scale=0.21]{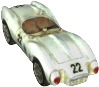} &
		\includegraphics[scale=0.21]{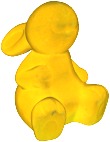} \\
		
		\rowfont{\color{black!75}}18 & 19 & 20 & 21 & 22 & 23 & 24 & 25 & 26 & 27 & 28 & 29 & 30 & 31 & 32 & 33 \\
		\multicolumn{16}{c}{\vspace{-1.5ex}} \\
		\multicolumn{16}{c}{{\scriptsize HB}}
		\end{tabu} \vspace{-0.5ex} \\
	
		\begin{tabu}{*{18}{c}}
		\includegraphics[scale=0.21]{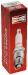} &
		\includegraphics[scale=0.21]{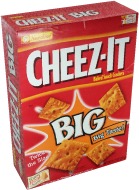} &
		\includegraphics[scale=0.21]{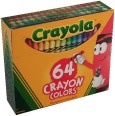} &
		\includegraphics[scale=0.21]{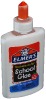} &
		\includegraphics[scale=0.21]{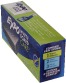} &
		\includegraphics[scale=0.21]{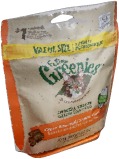} &
		\includegraphics[scale=0.21]{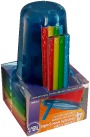} &
		\includegraphics[scale=0.21]{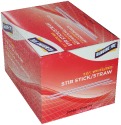} &
		\includegraphics[scale=0.21]{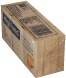} &
		\includegraphics[scale=0.21]{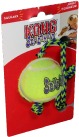} &
		\includegraphics[scale=0.21]{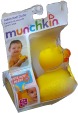} &
		\includegraphics[scale=0.21]{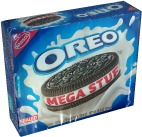} &
		\includegraphics[scale=0.21]{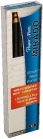} &
		\includegraphics[scale=0.21]{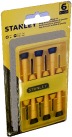} &
		&
		\includegraphics[scale=0.21]{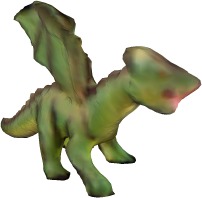} &
		\includegraphics[scale=0.21]{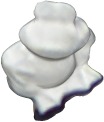} &
		\includegraphics[scale=0.21]{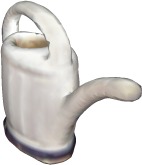} \\
		\rowfont{\color{black!75}}1 & 2 & 3 & 4 & 5 & 6 & 7 & 8 & 9 & 10 & 11 & 12 & 13 & 14 & \hspace{2em} & 1 & 2 & 3 \\
		\multicolumn{3}{c}{\vspace{-1.5ex}} \\
		\multicolumn{14}{c}{{\scriptsize RU-APC}} & & \multicolumn{3}{c}{{\scriptsize TUD-L}}
		\end{tabu} \vspace{0.0ex} \\
	
		\begin{tabu}{*{21}{c}}
		\includegraphics[scale=0.21]{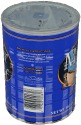} &
		\includegraphics[scale=0.21]{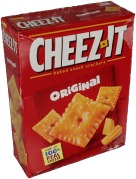} &
		\includegraphics[scale=0.21]{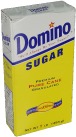} &
		\includegraphics[scale=0.21]{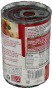} &
		\includegraphics[scale=0.21]{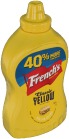} &
		\includegraphics[scale=0.21]{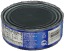} &
		\includegraphics[scale=0.21]{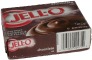} &
		\includegraphics[scale=0.21]{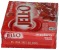} &
		\includegraphics[scale=0.21]{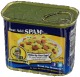} &
		\includegraphics[scale=0.21]{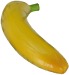} &
		\includegraphics[scale=0.21]{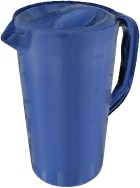} &
		\includegraphics[scale=0.21]{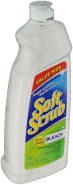} &
		\includegraphics[scale=0.21]{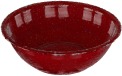} &
		\includegraphics[scale=0.21]{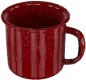} &
		\includegraphics[scale=0.21]{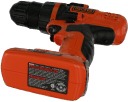} &
		\includegraphics[scale=0.21]{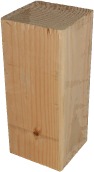} &
		\includegraphics[scale=0.21]{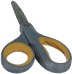} &
		\includegraphics[scale=0.21]{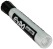} &
		\includegraphics[scale=0.21]{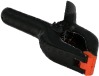} &
		\includegraphics[scale=0.21]{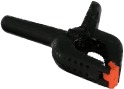} &
		\includegraphics[scale=0.21]{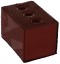} \\
		\rowfont{\color{black!75}}1 & 2 & 3 & 4 & 5 & 6 & 7 & 8 & 9 & 10 & 11 & 12 & 13 & 14 & 15 & 16 & 17 & 18 & 19 & 20 & 21 \\
		\multicolumn{21}{c}{\vspace{-1.5ex}} \\
		\multicolumn{21}{c}{{\scriptsize YCB-V}}
		\end{tabu} \vspace{-2.5ex} \\

		\begin{tabu}{*{21}{c}}
		\includegraphics[scale=0.21]{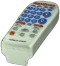} &
		\includegraphics[scale=0.21]{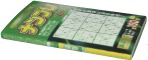} &
		\includegraphics[scale=0.21]{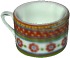} &
		\includegraphics[scale=0.21]{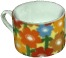} &
		\includegraphics[scale=0.21]{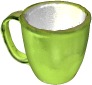} &
		\includegraphics[scale=0.21]{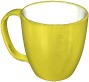} &
		\includegraphics[scale=0.21]{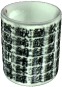} &
		\includegraphics[scale=0.21]{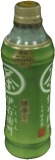} &
		\includegraphics[scale=0.21]{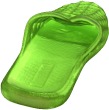} &
		\includegraphics[scale=0.21]{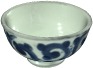} &
		\includegraphics[scale=0.21]{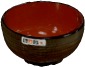} &
		\includegraphics[scale=0.21]{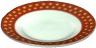} &
		\includegraphics[scale=0.21]{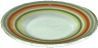} &
		\includegraphics[scale=0.21]{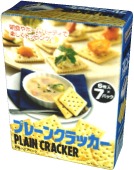} &
		\includegraphics[scale=0.21]{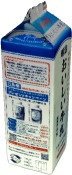} &
		\includegraphics[scale=0.21]{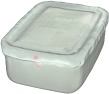} &
		\includegraphics[scale=0.21]{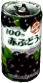} &
		\includegraphics[scale=0.21]{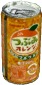} &
		\includegraphics[scale=0.21]{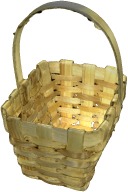} &
		\includegraphics[scale=0.21]{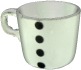} &
		\includegraphics[scale=0.21]{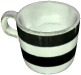}
		\\
		\rowfont{\color{black!75}}1 & 2 & 3 & 4 & 5 & 6 & 7 & 8 & 9 & 10 & 11 & 12 & 13 & 14 & 15 & 16 & 17 & 18 & 19 & 20 & 21 \\
		\multicolumn{21}{c}{\vspace{-1.5ex}} \\
		\multicolumn{21}{c}{{\scriptsize TYO-L}}
		\end{tabu} \vspace{0.5ex} \\

		\endgroup
		
		\caption{\label{fig:bop_overview_objects} \textbf{3D object models available in the BOP datasets.}
		The models are provided as 3D meshes with surface normal vectors and surface color. The color is saved per mesh vertex or as an UV texture.
		The models were created manually in CAD software or by KinectFusion-like systems for 3D surface reconstruction~\cite{newcombe2011kinectfusion}.
		T-LESS includes both CAD and reconstructed models with surface color available only for the reconstructed models shown above. ITODD includes only CAD models with no color. The figure shows models rendered at the same distance.
		}
		\vspace{-1.25ex}
		
	\end{center}
\end{figure}

\begin{figure}[!t]
	\begin{center}
		
		\vspace{-1.5ex}
		
		\begingroup
		\setlength{\tabcolsep}{1.3pt} %
		\renewcommand{\arraystretch}{0.6} %
		\begin{tabular}{ c c c c c }	
			{\scriptsize \mlcellc[b]{LM/LM-O \cite{hinterstoisser2012accv,brachmann2014learning}}} &
			{\scriptsize \mlcellc[b]{T-LESS \cite{hodan2017tless}}} &
			{\scriptsize \mlcellc[b]{ITODD} \cite{drost2017introducing}} &
			{\scriptsize \mlcellc[b]{HB} \cite{kaskman2019homebreweddb}} &
			{\scriptsize \mlcellc[b]{YCB-V} \cite{xiang2017posecnn}} \vspace{0.5ex} \\
			
			\includegraphics[width=0.195\columnwidth]{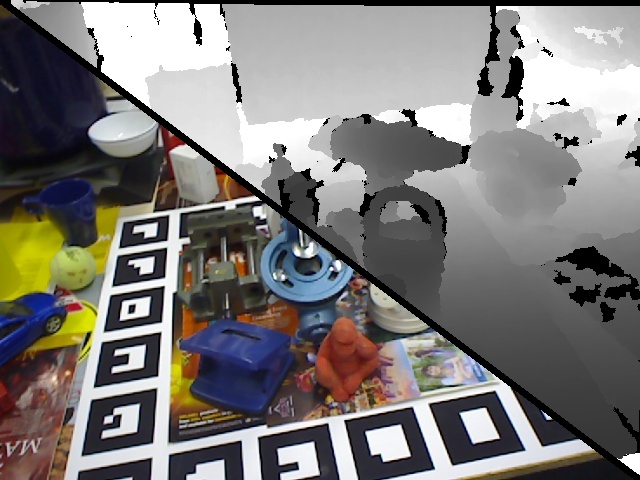} &
			\includegraphics[width=0.195\columnwidth]{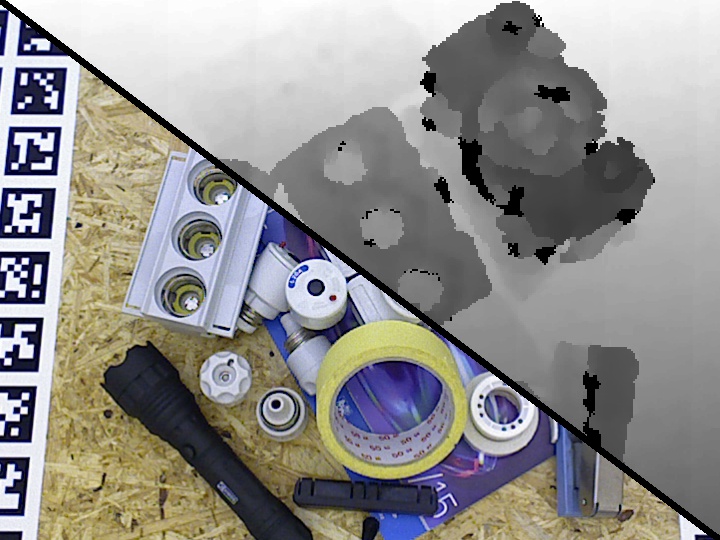} &
			\includegraphics[width=0.195\columnwidth]{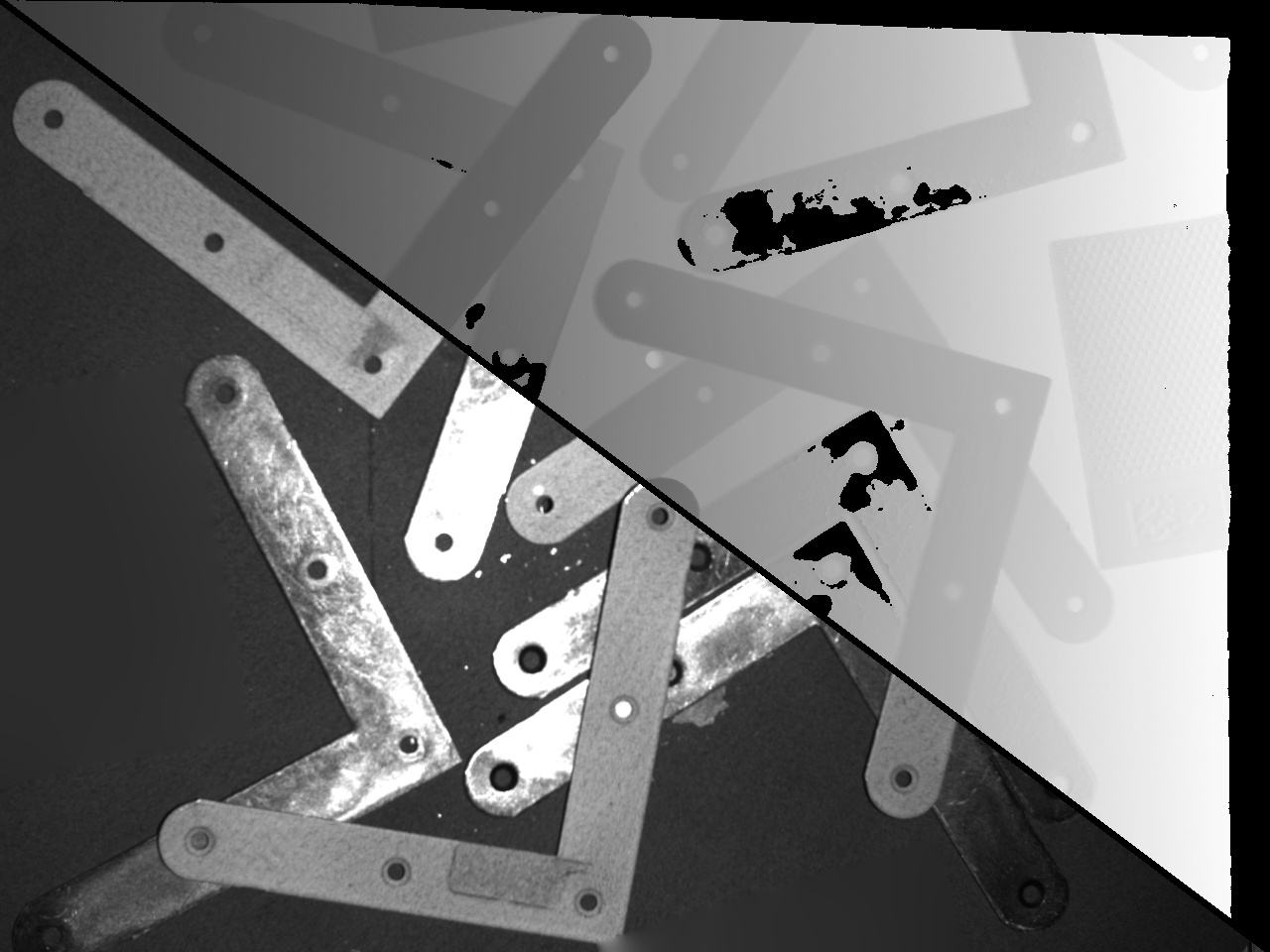} &
			\includegraphics[width=0.195\columnwidth]{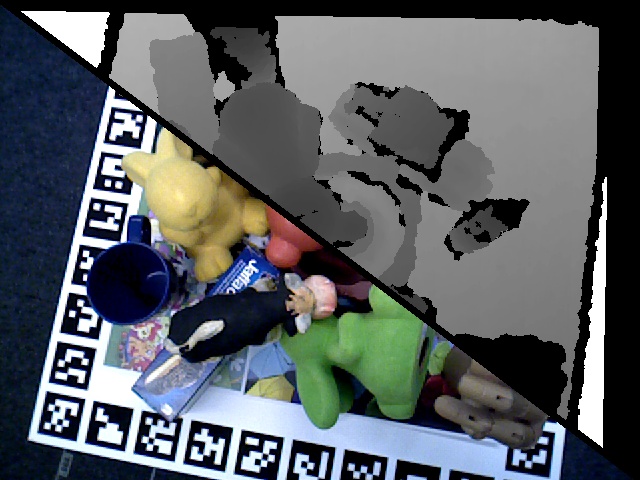} &
			\includegraphics[width=0.195\columnwidth]{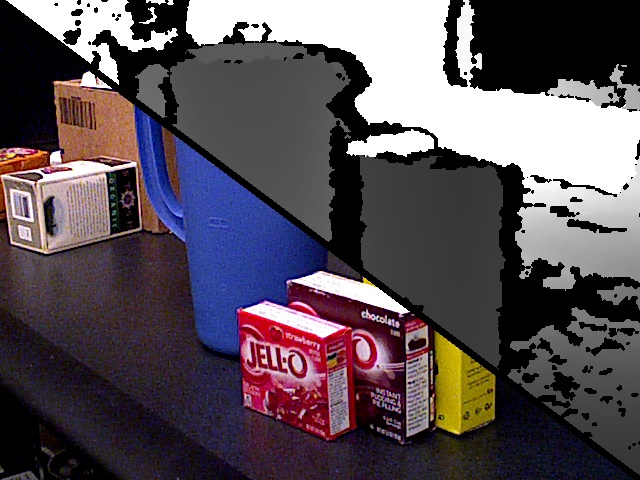} \\
			
			\includegraphics[width=0.195\columnwidth]{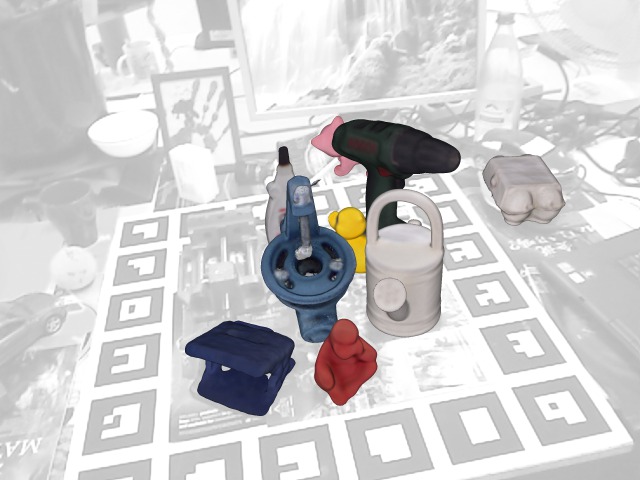} &
			\includegraphics[width=0.195\columnwidth]{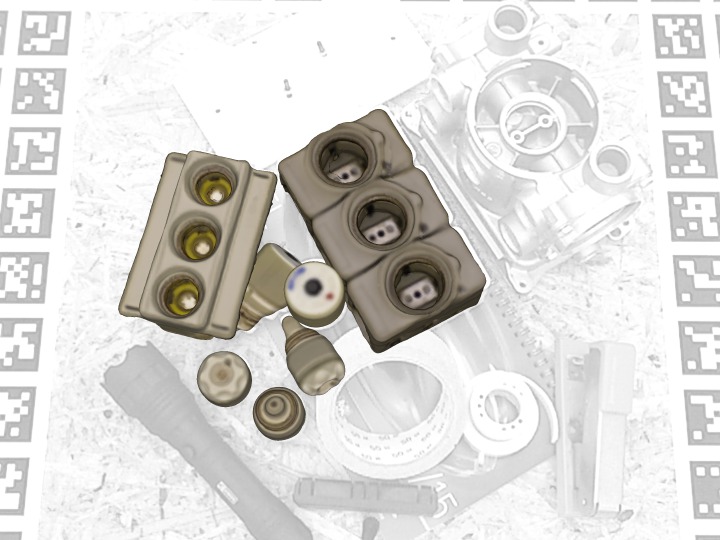} &
			\includegraphics[width=0.195\columnwidth]{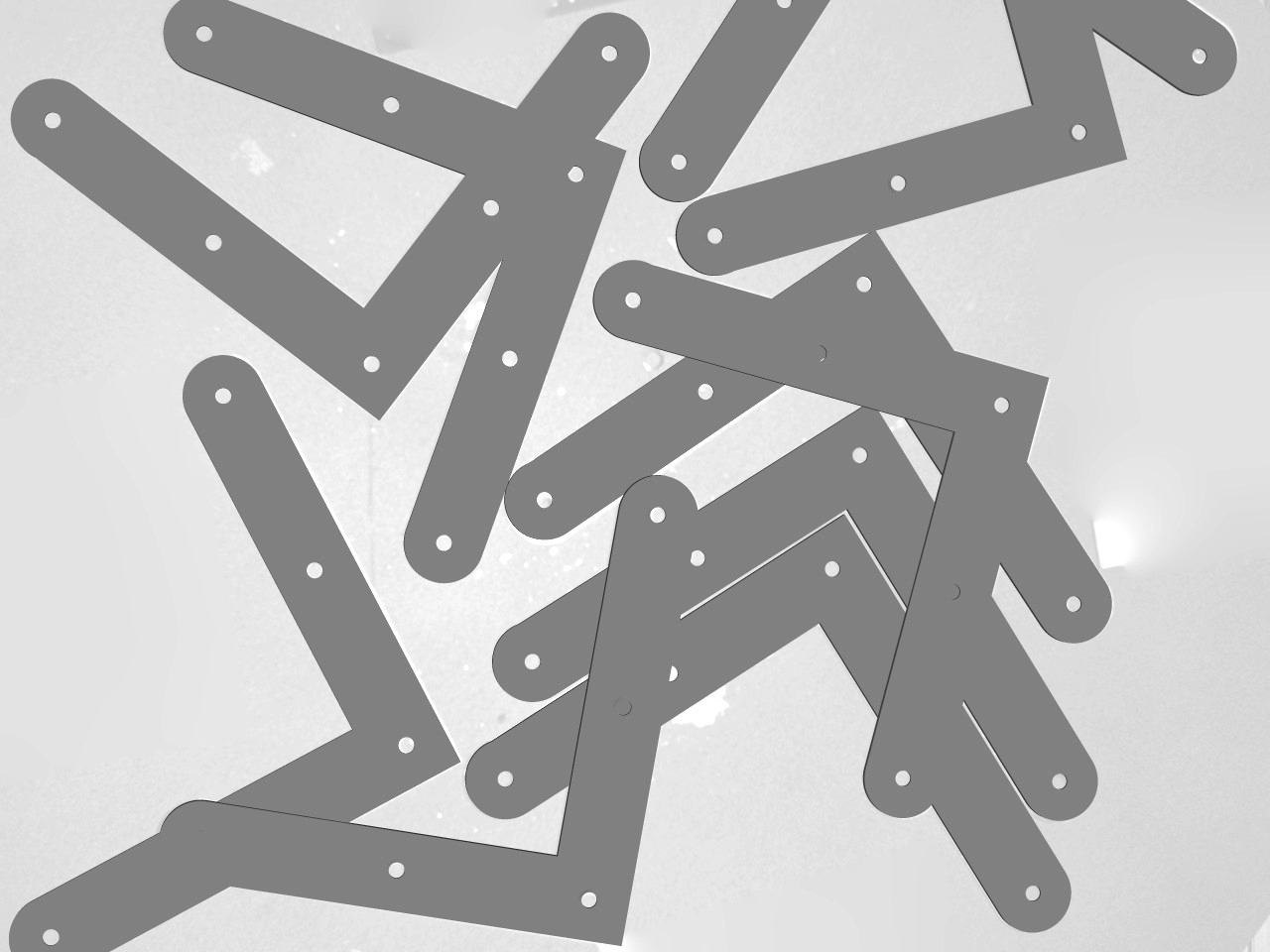} &
			\includegraphics[width=0.195\columnwidth]{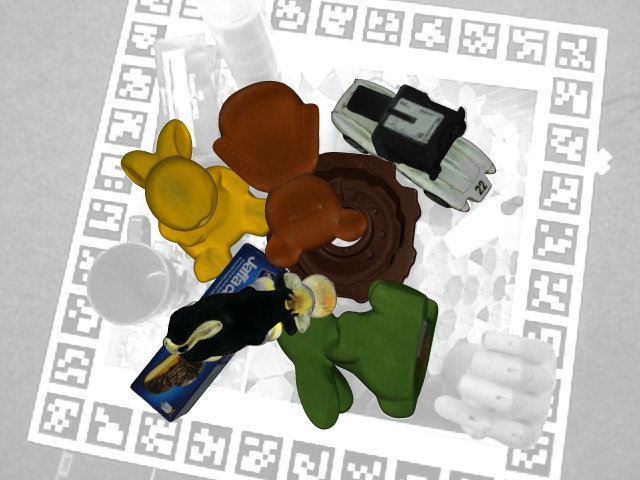} &
			\includegraphics[width=0.195\columnwidth]{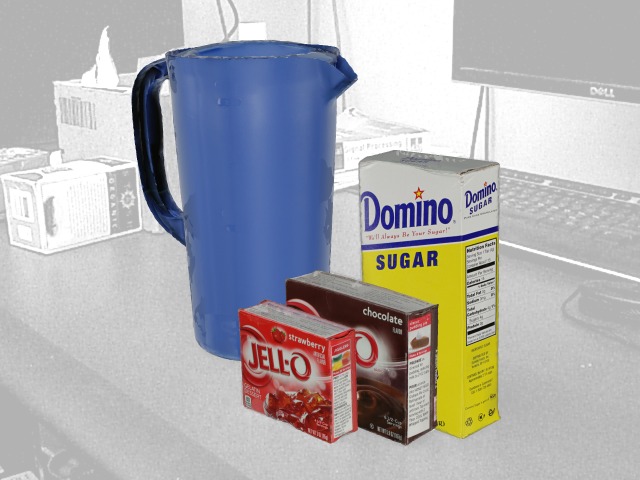} \\
			
			& & & & \\

			{\scriptsize \mlcellc[b]{RU-APC \cite{rennie2016dataset}}} &
			{\scriptsize \mlcellc[b]{IC-BIN} \cite{doumanoglou2016recovering}} &
			{\scriptsize \mlcellc[b]{IC-MI} \cite{tejani2014latent}} &
			{\scriptsize \mlcellc[b]{TUD-L \cite{hodan2018bop}}} &
			{\scriptsize \mlcellc[b]{TYO-L \cite{hodan2018bop}}} \vspace{0.5ex} \\
			
			\includegraphics[width=0.195\columnwidth]{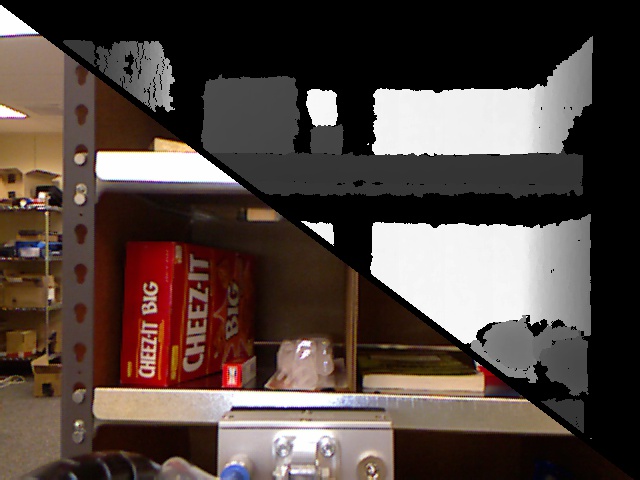} &
			\includegraphics[width=0.195\columnwidth]{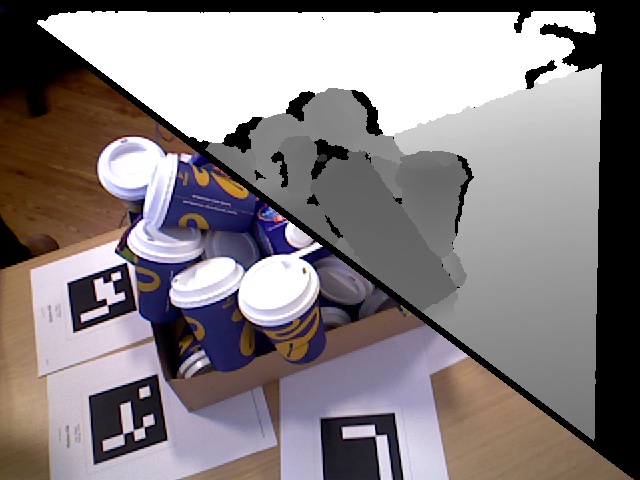} &
			\includegraphics[width=0.195\columnwidth]{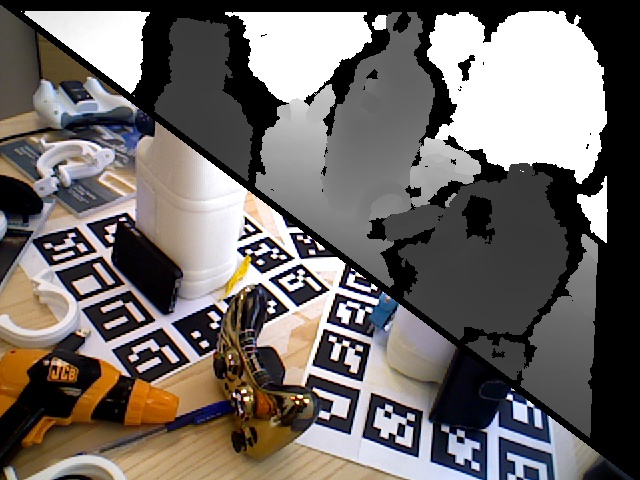} &
			\includegraphics[width=0.195\columnwidth]{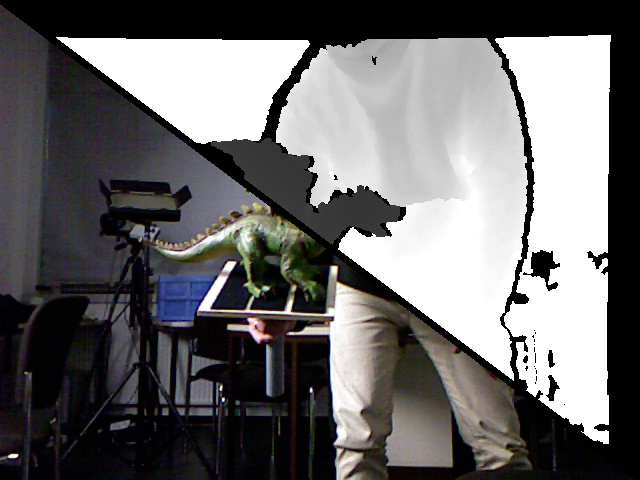} &
			\includegraphics[width=0.195\columnwidth]{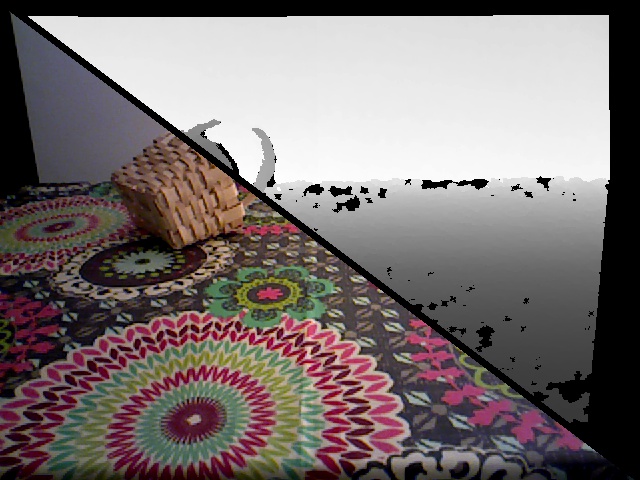} \\
			
			\includegraphics[width=0.195\columnwidth]{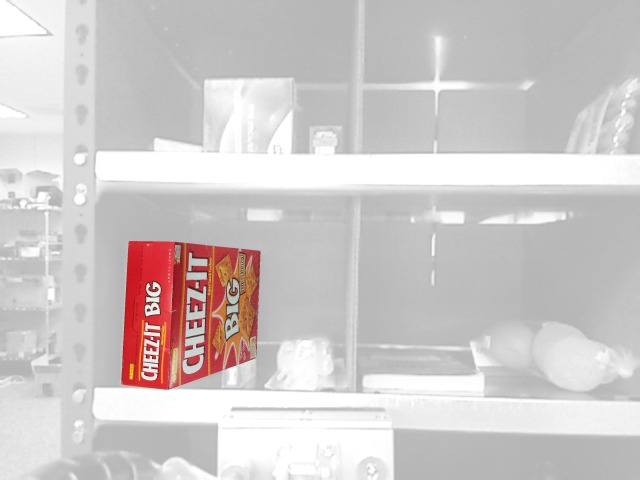} &
			\includegraphics[width=0.195\columnwidth]{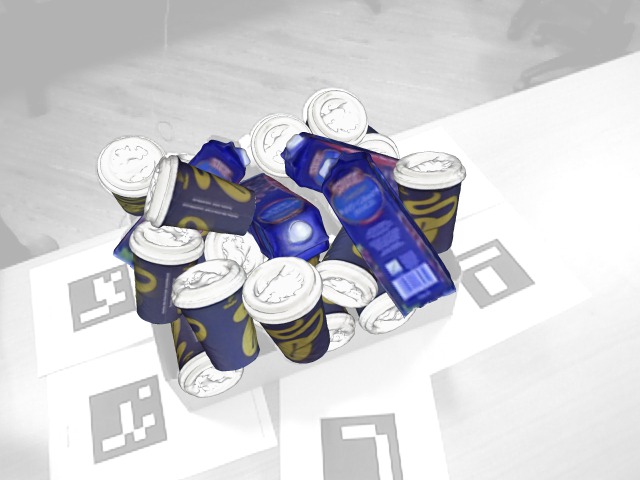} &
			\includegraphics[width=0.195\columnwidth]{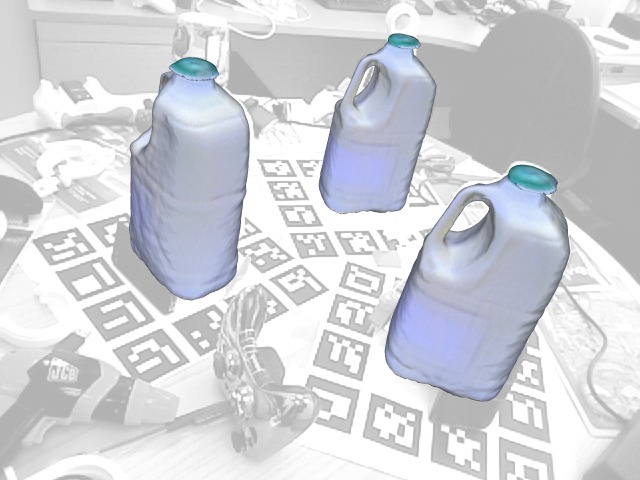} &
			\includegraphics[width=0.195\columnwidth]{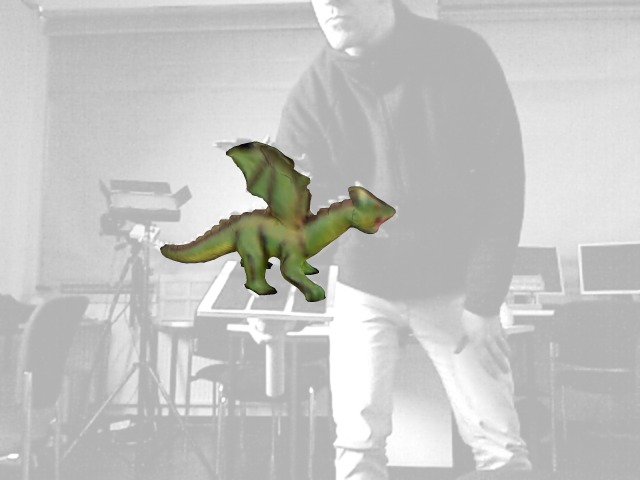} &
			\includegraphics[width=0.195\columnwidth]{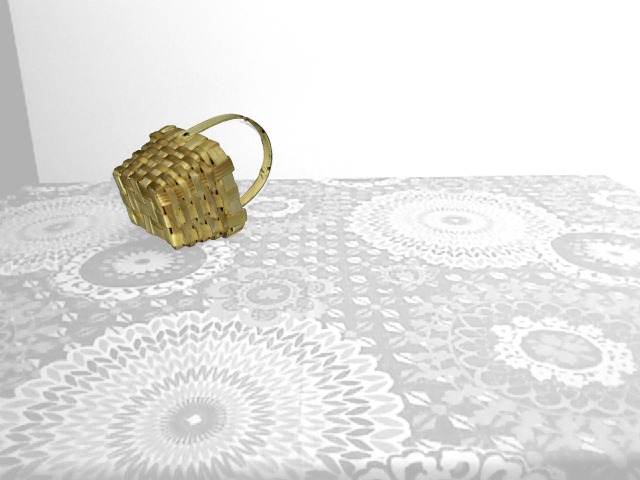} \\
		\end{tabular}
		\endgroup
		
		\caption{\label{fig:bop_overview_images} \textbf{Example test RGB-D images from the BOP datasets}.
			The upper rows show split views of the RGB and D image channels.
			The lower rows show the images overlaid with 3D object models in the ground-truth 6D poses.}
		\vspace{-1.25ex}
	\end{center}
\end{figure}

The BOP benchmark was created both to capture the status quo in the field and to systematically measure the progress in the future. Unlike the PASCAL VOC and ImageNet challenges, the task of 6D object pose estimation requires a specific set of calibrated modalities that cannot be easily acquired from the internet. In contrast to KITTY, it was not necessary to record large amounts of new data. We covered many practical scenarios by combining existing datasets. Additionally, we created two datasets with varying lighting conditions, which is an aspect that was not covered by the existing datasets.

As of 2020, the BOP benchmark comprises of:
(1)~eleven datasets in a unified format which cover various types of objects and practical scenarios -- see Figures~\ref{fig:bop_overview_objects} and \ref{fig:bop_overview_images},
(2)~an evaluation methodology with three new pose-error functions which address limitations of the previously used functions,
(3)~an online evaluation system
which is open for continuous submission of new results and reports the current state of the art,
and (4)~public challenges held at the International Workshops on Recovering 6D Object Pose~\cite{hodan2020r6d} organized annually at the ICCV and ECCV
conferences. Tom{\'a}{\v{s}} Hoda{\v{n}} has been leading the organization of the BOP benchmark together with Frank Michel (up to 2018) and later together with Martin Sundermeyer.

The evaluation methodology is defined in Section~\ref{sec:bop_methodology},
the datasets are introduced in Section~\ref{sec:bop_datasets}, and the public challenges are described and their results analyzed in Sections~\ref{sec:bop_challenge_2017} and \ref{sec:bop_challenge_2020}.
The initial version of the BOP benchmark together with the results of the first public challenge from 2017 were published in~\cite{hodan2018bop} and the results of the following two public challenges from 2019 and 2020 in~\cite{hodan2020bop}. The datasets, evaluation system, and up-to-date leaderbords are available on the project website: \texttt{\href{http://bop.felk.cvut.cz/}{bop.felk.cvut.cz}}. Evaluation scripts are available in the BOP Toolkit~\cite{boptoolkit}.

\section{Evaluation Methodology} \label{sec:bop_methodology}

The evaluation methodology detailed in this section defines the 6D object localization task on which the methods are evaluated (Section~\ref{sec:bop_6d_object_localization}), functions to measure the error of 6D pose estimates which can (fully or partially) deal with pose ambiguities (Sections~\ref{sec:bop_vsd}--\ref{sec:bop_mspd}), and the accuracy score used to rank the methods (Section~\ref{sec:bop_accuracy_score}).

\subsection{6D Object Localization} \label{sec:bop_6d_object_localization}

Methods are evaluated on the task of 6D localization of a \textbf{v}arying number of \textbf{i}nstances of a \textbf{v}arying number of \textbf{o}bjects from a single RGB-D image. This variant of the 6D object localization task is referred to as ViVo and defined as:

\customparagraph{Training input:} For each object, with a label $o \in \{1, \dots, k\}$, a method is given a~3D mesh model $M_o$ (typically with a color texture) and
synthetic or real RGB-D images~show\-ing instances of the object in known 6D poses. Any of the image channels may be used.

\customparagraph{Test input:} The method is provided with an image $I$ and a list $L = [(o_1, n_1),$ $\dots,$ $(o_m, n_m)]$, where $n_i$ is the number of instances of the object $o_i$ present in $I$.

\customparagraph{Test output:} The method produces a list $E = [E_1, \dots, E_m]$, where $E_i$ is a list of $n_i$ pose estimates for instances of the object $o_i$. An estimate is given by a $3\times3$ rotation matrix~$\mathbf{R}$, a $3\times1$ translation vector $\mathbf{t}$, and a confidence
$s$. The matrix $\textbf{P} = [\mathbf{R} \,|\, \mathbf{t}]$ defines a transformation from the 3D model coordinates to the 3D camera coordinates.

\customparagraph{}Note that in the first challenge from 2017, methods were evaluated on a simpler variant of the 6D object localization task with the goal to estimate the 6D pose of a \textbf{s}ingle \textbf{i}nstance of a \textbf{s}ingle \textbf{o}bject. This variant is referred to as SiSo and detailed in Section~\ref{sec:bop17_setup}.

In BOP, methods have been
evaluated on variants of the 6D object localization task for two reasons. First, the accuracy scores on this simpler task are still far from being saturated (Section~\ref{sec:bop_challenge_2020}). Second, the 6D object detection task requires computationally expensive evaluation as many more hypotheses need to be evaluated to calculate the precision/recall curve, which is typically used for evaluating detection. Calculating the 6D pose errors is more expensive than, for example, calculating the intersection over union of 2D bounding boxes used to evaluate 2D object detection.

\subsection{VSD: Visible Surface Discrepancy} \label{sec:bop_vsd}

To calculate the VSD error of an estimated pose~$\hat{\mathbf{P}}$ \wrt the ground-truth pose~$\bar{\mathbf{P}}$ in an image $I$, an object model~$M$ is first rendered in the two poses.
The result of the~rendering is two distance maps $\hat{D}$ and~$\bar{D}$.\footnote{A distance map stores at a pixel~$\mathbf{u}$ the distance from the camera center to a 3D point $\mathbf{x}_\mathbf{u}$ that projects to $\mathbf{u}$. The distance map can be readily computed from the depth map which stores at $\mathbf{u}$ the $Z$ coordinate of $\mathbf{x}_\mathbf{u}$ and which is a typical output of Kinect-like sensors.} As described below, the distance maps are compared with the distance map $D_I$ of the test image~$I$ to obtain the visibility masks $\hat{V}$ and $\bar{V}$, \ie, the sets of pixels where the model $M$ is visible in the image $I$.
Given a misalignment tolerance~$\tau$, the error is calculated as follows ($[\cdot]$ is the Iverson bracket; see also Figure~\ref{fig:bop_vsd_components}):
\begin{equation}
	e_\mathrm{VSD}\big(\hat{D}, \bar{D}, \hat{V}, \bar{V}, \tau\big) =
	\frac{1}{|\hat{V} \cup \bar{V}|} \sum_{\mathbf{u} \in \hat{V} \cup \bar{V}}
	\big[\mathbf{u} \notin \hat{V} \cap \bar{V} \, \vee \, \big|\hat{D}(\mathbf{u}) -
	\bar{D}(\mathbf{u})\big| \geq \tau \big]
\end{equation}

\def\pboxc{\relax\ifvmode\centering\fi}

\begin{figure}[!t]
    \begin{center}
    	
    	\vspace{-1.0ex}

        \begingroup
        \setlength{\tabcolsep}{1.5pt} %
        \renewcommand{\arraystretch}{0.9} %

        \begin{tabular}{ c c c c c c c c }
        	\pbox{\textwidth}{\pboxc{}\footnotesize{$RGB_I$} \\ \vspace{0.5ex}} &
        	\pbox{\textwidth}{\pboxc{}\footnotesize{$D_I$} \\ \vspace{0.5ex}} &
        	\hspace{1em} &
        	\pbox{\textwidth}{\pboxc{}\footnotesize{$\hat{D}$} \\ \vspace{0.5ex}} &
        	\pbox{\textwidth}{\pboxc{}\footnotesize{$\hat{V}$} \\ \vspace{0.5ex}} &
        	\pbox{\textwidth}{\pboxc{}\footnotesize{$\bar{D}$} \\ \vspace{0.5ex}} &
        	\pbox{\textwidth}{\pboxc{}\footnotesize{$\bar{V}$} \\ \vspace{0.5ex}} &
        	\pbox{\textwidth}{\pboxc{}\footnotesize{$D_\mathrm{\Delta}$} \\ \vspace{0.5ex}} \\
        	
	        \includegraphics[width=0.1185\columnwidth]{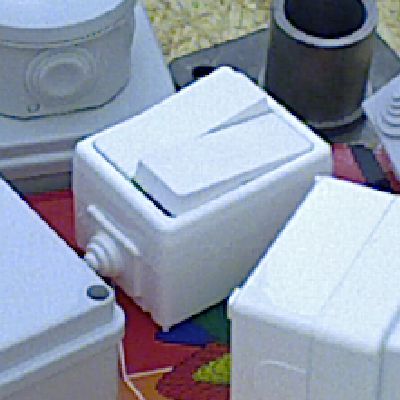} &
        	\includegraphics[width=0.1185\columnwidth]{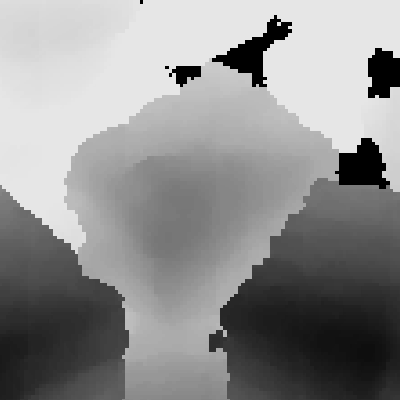} &
            &
        	\includegraphics[width=0.1185\columnwidth]{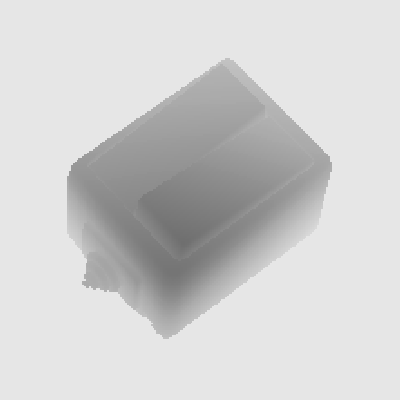} &
        	\includegraphics[width=0.1185\columnwidth]{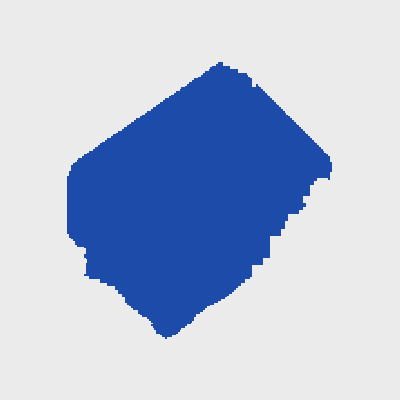} &
        	\includegraphics[width=0.1185\columnwidth]{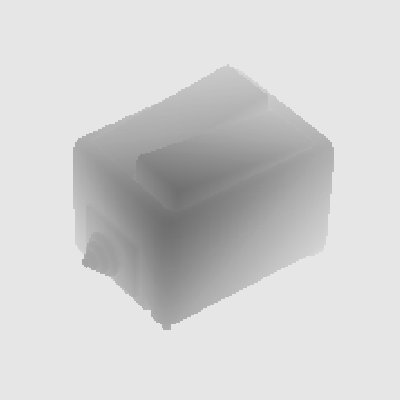} &
        	\includegraphics[width=0.1185\columnwidth]{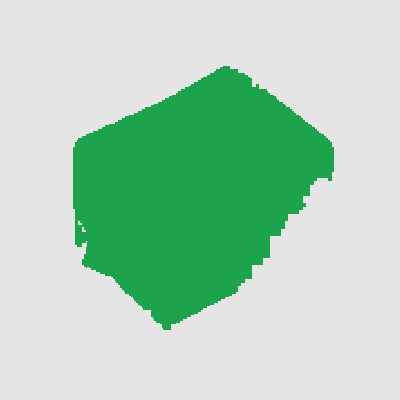} &
        	\includegraphics[width=0.1185\columnwidth]{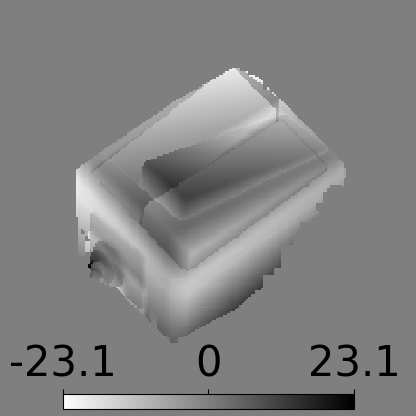} \\
        \end{tabular}
        \caption{\label{fig:bop_vsd_components} \textbf{Quantities used in the calculation of VSD.} %
        Left: The color channels $RGB_I$ (only for illustration) and the distance map $D_I$ of a test image~$I$.
        Right: The distance maps $\hat{D}$ and $\bar{D}$ are obtained by rendering the object model $M$ in the estimated pose $\hat{\mathbf{P}}$ and the ground-truth pose $\bar{\mathbf{P}}$ respectively. The visibility masks $\hat{V}$ and $\bar{V}$ of the model surface that is visible in~$I$ are obtained by comparing $\hat{D}$ and $\bar{D}$ with $D_I$. The distance differences $D_\mathrm{\Delta}(\mathbf{u}) = \hat{D}(\mathbf{u}) - \bar{D}(\mathbf{u})$, $\forall \mathbf{u} \in \hat{V} \cap \bar{V}$, are used for the pixel-wise evaluation of the surface alignment.}
		\vspace{-1.25ex}
        \endgroup
    \end{center}
\end{figure}

\noindent\textbf{Visibility Masks.}
Given the rendered distance image $\bar{D}$, with $\bar{D}(\mathbf{u})$~=~$0$ for pixels outside the object mask, and given the distance image $D_I$ of the test scene, with $D_I(\mathbf{u})$~=~$0$ for pixels with missing measurements,
the visibility mask $\bar{V}$ is defined as a set of pixels where the surface of the model $M$ in the ground-truth pose~$\bar{\mathbf{P}}$ is
at most by a tolerance~$\delta$ behind the scene surface, or where the distance of the scene surface is unknown:
\begin{equation} \label{eq:visibility_gt}
	\bar{V} = \big\{\mathbf{u} \, | \, \bar{D}(\mathbf{u}) > 0 \, \wedge \, \big(\bar{D}(\mathbf{u}) - D_I(\mathbf{u}) \leq \delta \, \vee \, D_I(\mathbf{u}) = 0\big)\big\}
\end{equation}

An object is considered visible at pixels with unknown distance
to enable evaluating poses of glossy objects, \eg, objects from the ITODD dataset~\cite{drost2017introducing}, whose surface is not always captured in the distance image. Note that an object was considered invisible at such pixels in the initial version of the evaluation methodology~\cite{hodan2018bop,hodan2016evaluation}.

For the visibility mask $\hat{V}$ of the model $M$ in the estimated pose~$\hat{\mathbf{P}}$, the definition (\ref{eq:visibility_gt}) is extended by including all pixel from the visibility mask $\bar{V}$ regardless the distance of the scene surface at these pixels. This is to ensure that the surface of the object captured in $D_I$ does not occlude the object model in the estimated pose. The
mask $\hat{V}$ is defined as:
\begin{equation}
	\hat{V} = \big\{\mathbf{u} \, | \, \hat{D}(\mathbf{u}) > 0 \, \wedge \, \big(\hat{D}(\mathbf{u}) - D_I(\mathbf{u}) \leq \delta  \, \vee \, D_I(\mathbf{u}) = 0 \, \vee \, \mathbf{u} \in \bar{V} \big) \big\}
\end{equation}

\noindent\textbf{Properties of VSD.}
The VSD error is calculated only over the visible part of the model
and thus indistinguishable poses are treated as equivalent.
This is a desirable property not provided by the common pose-error functions, including ADD and ADI~\cite{hinterstoisser2012accv,hodan2016evaluation}.

Note that VSD evaluates the alignment of the object shape but not of its color. This is because most of the
models currently included in BOP have baked shadows and reflections in their textures, which makes it difficult to robustly evaluate the color alignment.

\def\pboxc{\relax\ifvmode\centering\fi}

\begin{figure}[!t]
    \begin{center}
    	
    	\vspace{-1.0ex}

        \begingroup
        \footnotesize
        \setlength{\tabcolsep}{1.5pt} %

        \begin{tabular}{ c c c c c c c c }
        	
            \pbox{\textwidth}{\pboxc{}\footnotesize{a: \textbf{0.04}} \\ \footnotesize{3.7/15.2} \\ \vspace{0.5ex}} &
            \pbox{\textwidth}{\pboxc{}\footnotesize{b: \textbf{0.08}} \\ \footnotesize{3.6/10.9} \\ \vspace{0.5ex}} &
            \pbox{\textwidth}{\pboxc{}\footnotesize{c: \textbf{0.11}} \\ \footnotesize{3.2/13.4} \\ \vspace{0.5ex}} &
            \pbox{\textwidth}{\pboxc{}\footnotesize{d: \textbf{0.19}} \\ \footnotesize{1.0/6.4} \\ \vspace{0.5ex}} &
            \pbox{\textwidth}{\pboxc{}\footnotesize{e: \textbf{0.28}} \\ \footnotesize{1.4/7.7} \\ \vspace{0.5ex}} &
            \pbox{\textwidth}{\pboxc{}\footnotesize{f: \textbf{0.34}} \\ \footnotesize{2.1/6.4} \\ \vspace{0.5ex}} &
            \pbox{\textwidth}{\pboxc{}\footnotesize{g: \textbf{0.40}} \\ \footnotesize{2.1/8.6} \\ \vspace{0.5ex}} &
            \pbox{\textwidth}{\pboxc{}\footnotesize{h: \textbf{0.44}} \\ \footnotesize{4.8/21.7} \\ \vspace{0.5ex}} \\
            
            \includegraphics[width=0.1185\columnwidth]{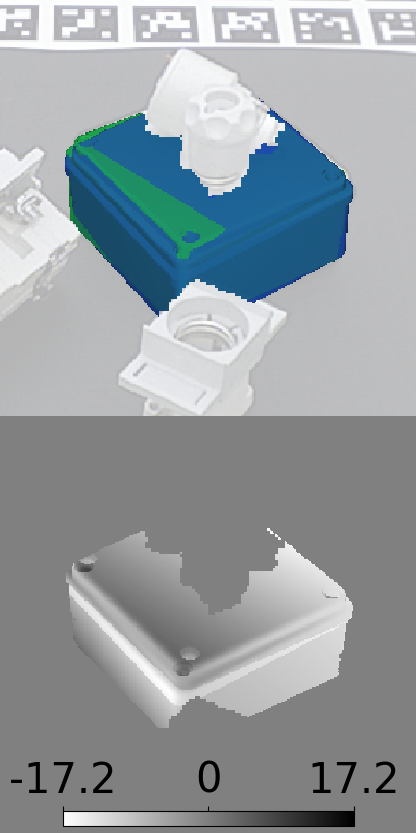} &
            \includegraphics[width=0.1185\columnwidth]{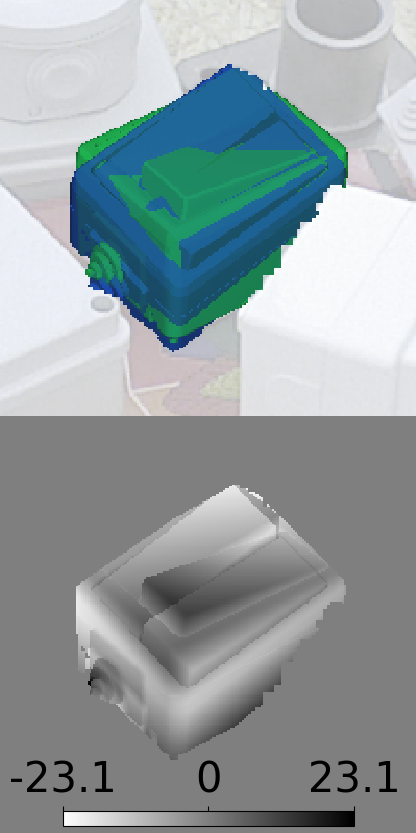} &
            \includegraphics[width=0.1185\columnwidth]{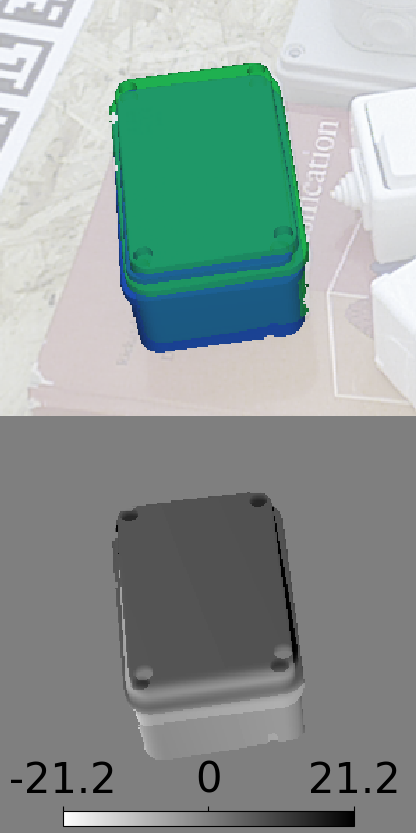} &
            \includegraphics[width=0.1185\columnwidth]{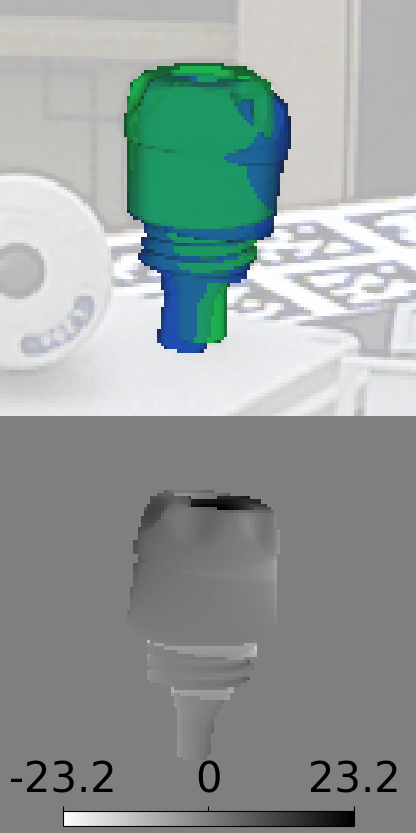} &
            \includegraphics[width=0.1185\columnwidth]{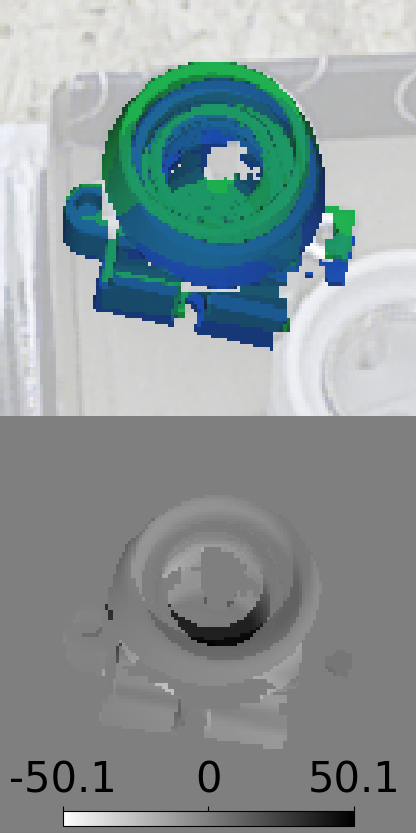} &
            \includegraphics[width=0.1185\columnwidth]{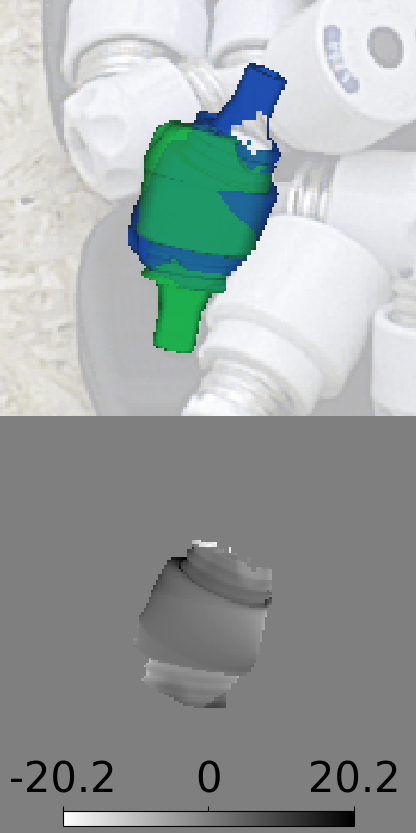} &
            \includegraphics[width=0.1185\columnwidth]{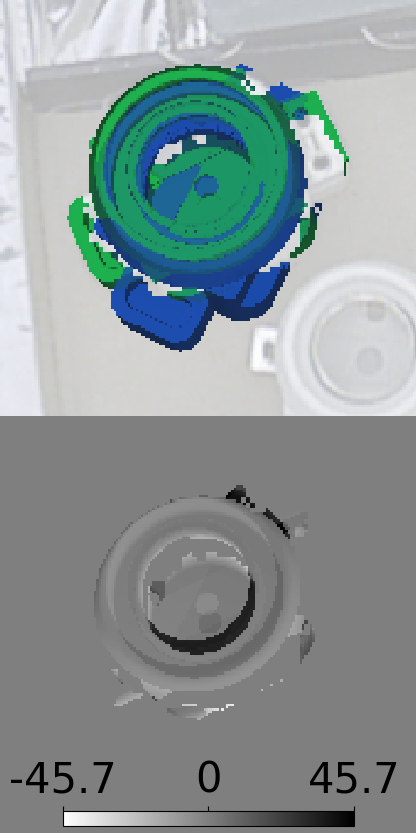} &
            \includegraphics[width=0.1185\columnwidth]{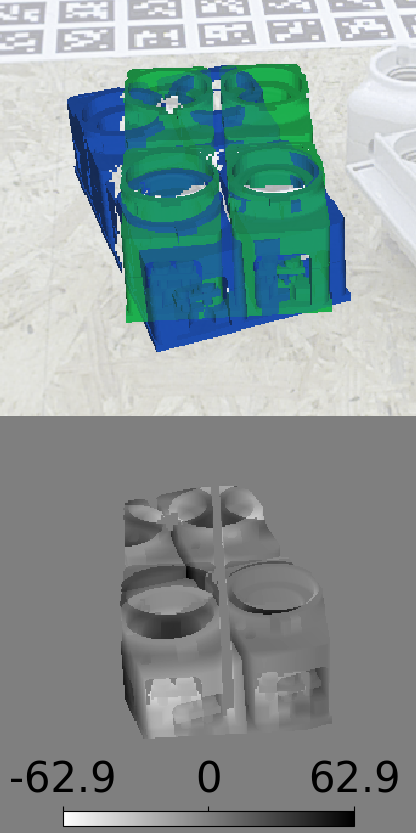} \\
            
            \pbox{\textwidth}{\pboxc{}\vspace{0.7ex}\footnotesize{i: \textbf{0.47}} \\ \footnotesize{4.8/9.2} \\ \vspace{0.5ex}} &
            \pbox{\textwidth}{\pboxc{}\vspace{0.7ex}\footnotesize{j: \textbf{0.54}} \\ \footnotesize{6.9/10.8} \\ \vspace{0.5ex}} &
            \pbox{\textwidth}{\pboxc{}\vspace{0.7ex}\footnotesize{k: \textbf{0.57}} \\ \footnotesize{6.9/8.9} \\ \vspace{0.5ex}} &
            \pbox{\textwidth}{\pboxc{}\vspace{0.7ex}\footnotesize{l: \textbf{0.64}} \\ \footnotesize{21.0/21.7} \\ \vspace{0.5ex}} &
            \pbox{\textwidth}{\pboxc{}\vspace{0.7ex}\footnotesize{m: \textbf{0.66}} \\ \footnotesize{4.4/6.5} \\ \vspace{0.5ex}} &
            \pbox{\textwidth}{\pboxc{}\vspace{0.7ex}\footnotesize{n: \textbf{0.76}} \\ \footnotesize{8.8/9.9} \\ \vspace{0.5ex}} &
            \pbox{\textwidth}{\pboxc{}\vspace{0.7ex}\footnotesize{o: \textbf{0.89}} \\ \footnotesize{49.4/11.1} \\ \vspace{0.5ex}} &
            \pbox{\textwidth}{\pboxc{}\vspace{0.7ex}\footnotesize{p: \textbf{0.95}} \\ \footnotesize{32.8/10.8} \\ \vspace{0.5ex}} \\
            
            \includegraphics[width=0.1185\columnwidth]{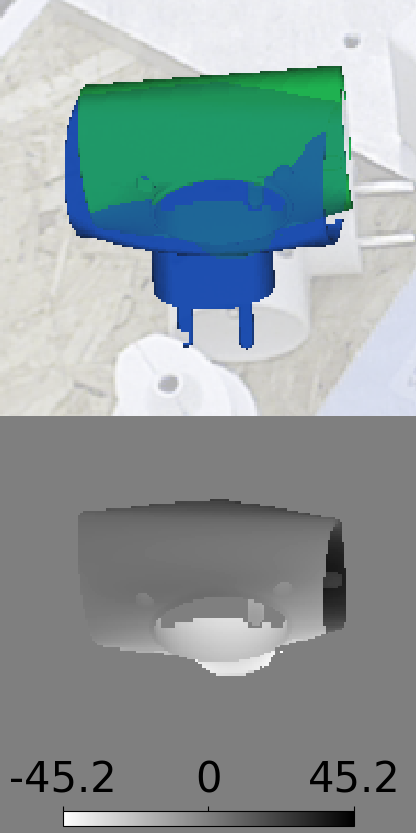} &
            \includegraphics[width=0.1185\columnwidth]{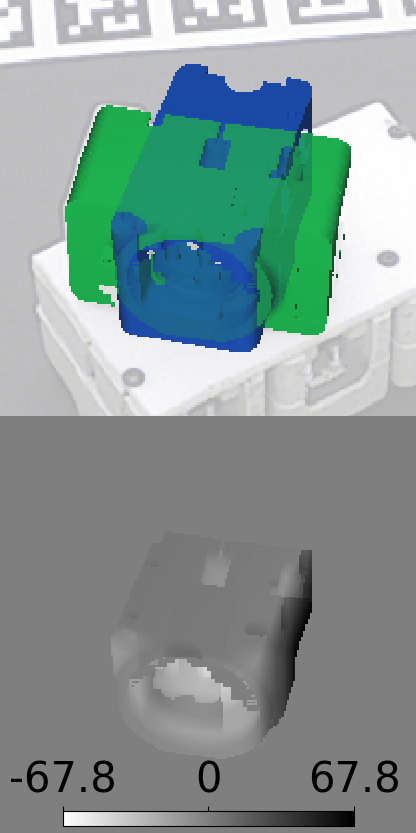} &
            \includegraphics[width=0.1185\columnwidth]{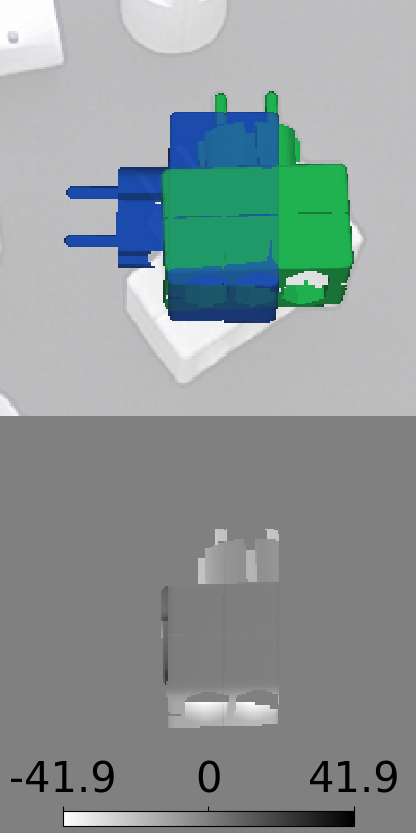} &
            \includegraphics[width=0.1185\columnwidth]{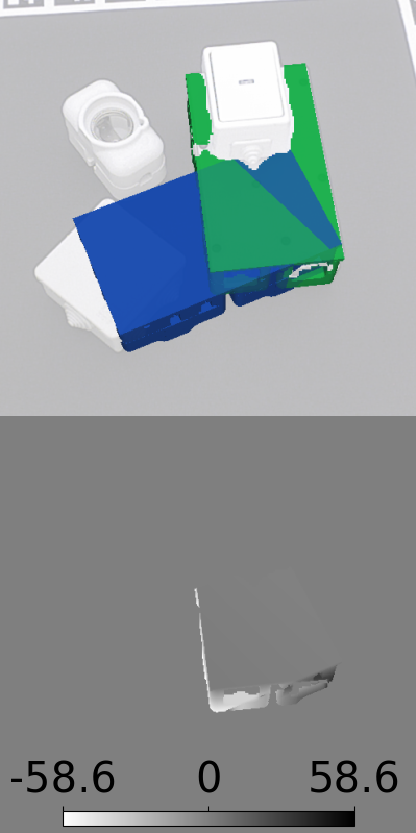} &
            \includegraphics[width=0.1185\columnwidth]{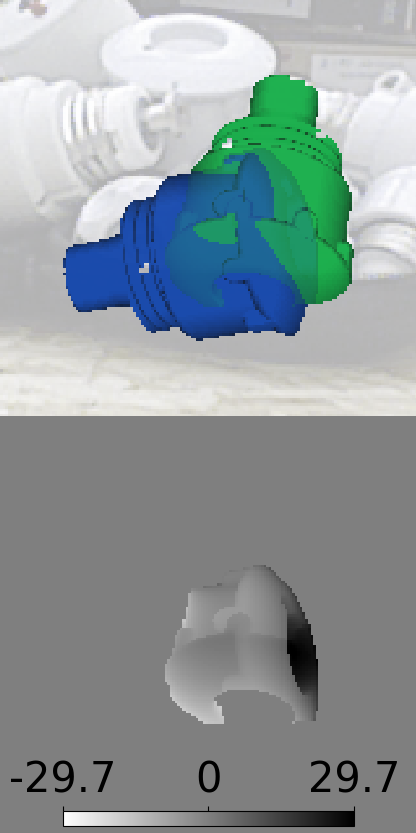} &
            \includegraphics[width=0.1185\columnwidth]{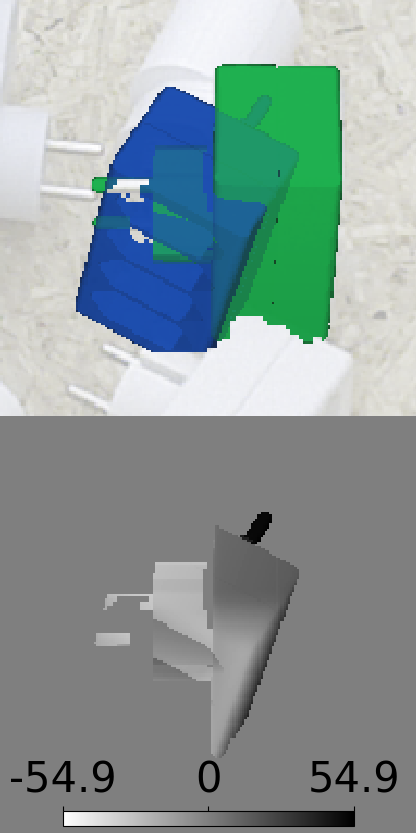} &
            \includegraphics[width=0.1185\columnwidth]{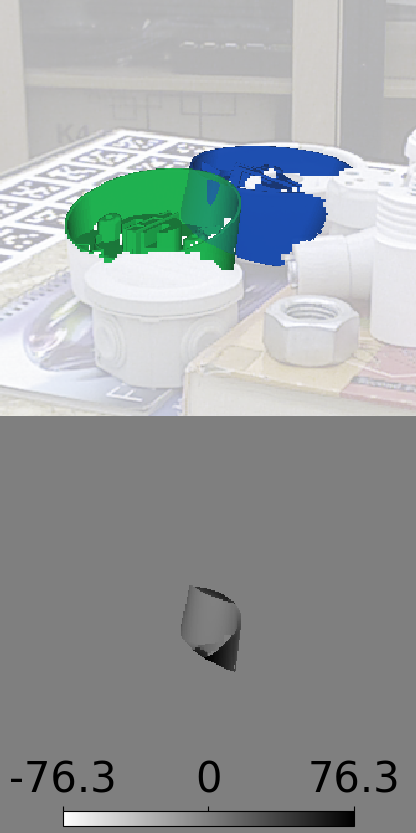} &
            \includegraphics[width=0.1185\columnwidth]{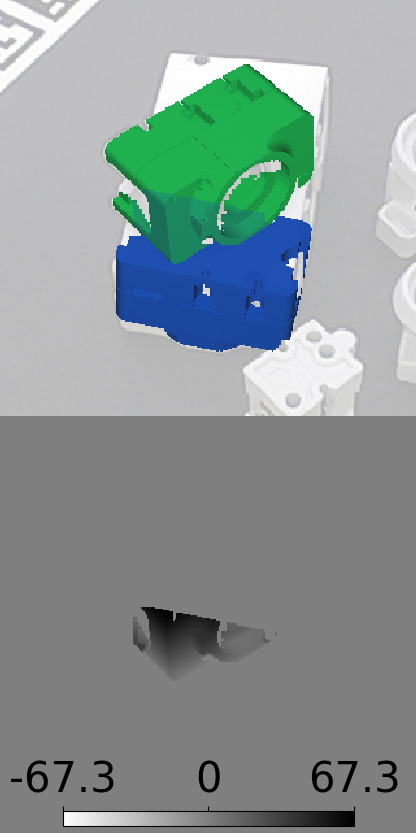} \\
        \end{tabular}
        \caption{
        	\textbf{Comparison of VSD and ADI.}
        	The VSD errors (bold, $\tau=20\,$mm, $\delta=15\,$mm) are compared with the ADI errors (in mm, the threshold of correctness $\theta_\mathrm{AD}$ is after the slash) on example pose estimates sorted by increasing VSD.
        	Top: Cropped and brightened test images overlaid with
            renderings of the model in the estimated pose $\hat{\mathbf{P}}$ in blue, and in the ground-truth pose $\bar{\mathbf{P}}$ in green. Only the part of the model surface that falls into the respective visibility mask is shown.
            Bottom: Distance differences $D_{\Delta}$. Case (b) is considered in Figure~\ref{fig:bop_vsd_components}.
        }
	    \vspace{-1.25ex}
        \label{fig:bop_vsd_examples}
        
        \endgroup
    \end{center}
\end{figure}

\customparagraph{Comparison of VSD and ADI.}
The most widely used pose-error functions have been ADD and ADI~\cite{hinterstoisser2012accv,hodan2016evaluation}
calculated as the average distance from vertices of the model $M$ in the ground-truth pose $\bar{\mathbf{P}}$ to vertices of $M$ in the estimated pose $\hat{\mathbf{P}}$.
The distance is measured to the corresponding vertices if all views of the object are distinguishable (ADD) and to the closest vertices otherwise (ADI).
The estimated pose~$\hat{\mathbf{P}}$ is considered correct if $e \leq \theta_\mathrm{AD} = 0.1 d$, where $e$ is
ADD or ADI,
and $d$ is the object diameter.

As discussed in Section~\ref{sec:rel_mesuring_error}, ADI may yield unintuitively low errors for poses that are clearly distinguishable due to a many-to-one vertex matching that may be established by the search for the closest vertex, and due to the possibility to match the outer and the inner model parts.
This is shown in Figure~\ref{fig:bop_vsd_examples}, which compares VSD and ADI on example pose estimates of objects that have indistinguishable views and their models include the inner parts. Overall, (f)-(n) yield low ADI errors and satisfy the correctness criterion of Hinterstoisser \etal.
These estimates, together with estimates (o) and (p), would be marked as incorrect by requiring the VSD error to be at most $0.3$, which is the criterion of correctness used in the initial version of the benchmark~\cite{hodan2018bop}. In the current version, the accuracy scores are calculated over multiple thresholds (see Section~\ref{sec:bop_accuracy_score}).

\subsection{MSSD: Maximum Symmetry-Aware Surface Distance} \label{sec:bop_mssd}

The MSSD error is defined as the maximum distance between a vertex $\textbf{x} \in V_M$ of the object model $M$ in the ground-truth pose~$\bar{\mathbf{P}} = [\bar{\mathbf{R}}\,|\,\bar{\mathbf{t}}]$ and the same vertex $\textbf{x}$ in the estimated pose~$\hat{\mathbf{P}} = [\hat{\mathbf{R}}\,|\,\hat{\mathbf{t}}]$, while considering global symmetry transformations $S_M$ of the model $M$
(the~symmetry transformations are identified as described in Section~\ref{sec:symmetries}):
\begin{equation}
	e_{\text{MSSD}}\big(\hat{\mathbf{R}}, \hat{\mathbf{t}}, \bar{\mathbf{R}}, \bar{\mathbf{t}}, S_M, V_M\big) = \min_{[\textbf{R} \, | \, \textbf{t}] \in S_M} \, \max_{\textbf{x}
		\in V_M}
	\big\Vert \big(\hat{\textbf{R}}\textbf{x} + \hat{\textbf{t}}\big) - \big(\bar{\textbf{R}}\big(\textbf{R}\textbf{x} + \textbf{t}\big) + \bar{\textbf{t}}\big)
	\big\Vert_2
\end{equation}

\noindent\textbf{Properties of MSSD.}
The maximum distance between model vertices is relevant for robotic manipulation, where the maximum surface deviation strongly indicates the chance of a successful grasp. Moreover, compared to the average distance used in the pose-error functions ADD and ADI~\cite{hodan2016evaluation,hinterstoisser2012accv}, which tends to be dominated by higher-frequency surface parts such as the thread of a fuse, the maximum distance is less dependent on the geometry of the object model and the sampling density of its surface.

As the MSSD error is calculated over the entire model surface, misalignment of the invisible surface part is penalized. This may not be desirable for applications such as robotic manipulation with suction cups where only the alignment of the visible part is relevant (note that VSD evaluates only the visible surface part). Pose ambiguities due to global object symmetries are handled properly in MSSD by explicitly considering all the symmetry transformations.

\subsection{MSPD: Maximum Symmetry-Aware Projection Distance} \label{sec:bop_mspd}

The MSPD error is defined as the maximum distance between the 2D projection of a vertex $\textbf{x} \in V_M$ of the object model $M$ in the ground-truth pose~$\bar{\mathbf{P}} = [\bar{\mathbf{R}}\,|\,\bar{\mathbf{t}}]$ and the 2D projection of the same vertex $\textbf{x}$ in the estimated pose~$\hat{\mathbf{P}} = [\hat{\mathbf{R}}\,|\,\hat{\mathbf{t}}]$, while considering the global symmetry transformations $S_M$ ($\text{proj}(\cdot)$ is the 2D projection):
\begin{equation}
	e_{\text{MSSD}}\big(\hat{\mathbf{R}}, \hat{\mathbf{t}}, \bar{\mathbf{R}}, \bar{\mathbf{t}}, S_M, V_M\big) = \min_{[\textbf{R} \, | \, \textbf{t}] \in S_M} \, \max_{\textbf{x}
		\in V_M}
	\big\Vert \text{proj}\big(\hat{\textbf{R}}\textbf{x} + \hat{\textbf{t}}\big) - \text{proj}\big(\bar{\textbf{R}}\big(\textbf{R}\textbf{x} + \textbf{t}\big) + \bar{\textbf{t}}\big)
	\big\Vert_2
\end{equation}

\noindent\textbf{Properties of MSPD.}
Since MSPD does not evaluate the alignment along the optical ($Z$) axis and measures only the perceivable discrepancy, it is relevant for augmented reality applications and suitable for evaluating RGB-only methods, for which estimating the alignment along the optical axis is more challenging.
Compared to the ``2D Projection'' pose-error function proposed in~\cite{brachmann2014learning}, MSPD explicitly considers the global object symmetries and replaces the average by the maximum distance to increase the robustness against geometry and sampling of the object model.
As MSSD, MSPD penalizes misalignment of the invisible surface part.

\subsection{Identifying Global Object Symmetries} \label{sec:symmetries}

The set of global symmetry transformations of an object model $M$, which is used in
MSSD and MSPD, is identified in two steps. Firstly, a set of candidate symmetry transformations is defined as $S'_M = \{\textbf{S} \in SE(3) \;|\; h(V_M, \textbf{S}V_M) < \varepsilon \}$, where $h(\cdot\,, \cdot)$ is the Hausdorff distance, $V_M$ are 3D vertices of the model $M$ in the canonical pose, and $\textbf{S}V_M$ are the 3D vertices transformed by a rigid transformation $\textbf{S} = [\mathbf{R}\,|\,\mathbf{t}]$, which consists of a $3\times3$ rotation matrix $\textbf{R}$ and a $3\times1$ translation vector $\textbf{t}$.
The allowed deviation is bounded by $\varepsilon = \text{max}(15\,\text{mm}, 0.1d)$, where $d$ is the diameter of $M$ and the truncation at $15\,$mm avoids breaking symmetries by too small details.
Note that even though all symmetry transformations of objects with a finite size are 3D rotations around an axis of symmetry, we have no guarantees that the axes of symmetry of the model $M$ intersect the origin of the model coordinate system. Therefore, we need to consider elements of the special Euclidean group $SE(3)$ as candidates for the symmetry transformations. In practice, the set $S'_M$ is found by dense sampling of the rotational and translational spaces and covers both discrete and continuous (discretized) global symmetries of the model $M$.

In the second step, the final set of symmetry transformations $S_M$ is defined as a subset of $S'_M$ that consists of those transformations that cannot be resolved by the color texture of $M$. This was decided subjectively by the BOP organizers -- as mentioned in Section~\ref{sec:bop_vsd}, the object models included in BOP have baked shadows and reflections in textures, which makes it difficult to evaluate the color alignment programmatically.

\subsection{Accuracy Score} \label{sec:bop_accuracy_score}

An estimated pose is considered correct with respect to a pose-error function $e$, if $e < \theta_e$, where $e \in \{e_{\text{VSD}}, e_{\text{MSSD}}, e_{\text{MSPD}}\}$ and $\theta_e$ is a threshold of correctness. All three pose-error functions are used as each evaluates different qualitites of the pose estimates and is relevant for a different target application.

The fraction of annotated object instances for which a correct pose is estimated is referred to as recall. The average recall with respect to a function~$e$, denoted as $\text{AR}_e$, is defined as the average of recall rates calculated for multiple settings of the threshold $\theta_e$, and also for multiple settings of the misalignment tolerance $\tau$ in the case of $e_{\text{VSD}}$.
In particular, $\text{AR}_\text{VSD}$ is the average of recall rates calculated for the misalignment tolerance $\tau$ ranging from $5\%$ to $50\%$ of the object diameter with a step of $5\%$, and for the threshold $\theta_{\text{VSD}}$ ranging from $0.05$ to $0.5$ with a step of $0.05$.
$\text{AR}_\text{MSSD}$ is the average of recall rates calculated for $\theta_{\text{MSSD}}$ ranging from $5\%$ to $50\%$ of the object diameter with a step of $5\%$.
Finally, $\text{AR}_\text{MSPD}$ is the average of recall rates calculated for $\theta_{\text{MSPD}}$ ranging from $5r$ to $50r$ with a step of $5r$, where $r = w/640$ and $w$ is the image width in pixels.

The accuracy of a method on a dataset $D$ is measured by $\text{AR}_D = (\text{AR}_{\text{VSD}} + \text{AR}_{\text{MSSD}} + \text{AR}_{\text{MSPD}}) / 3$. The overall accuracy on the core datasets is then measured by $\text{AR}_{\text{Core}}$ defined as the average of the per-dataset $\text{AR}_D$ scores. In this way, each dataset is treated as a separate sub-challenge which avoids $\text{AR}_{\text{Core}}$ being dominated by larger datasets.

\section{Datasets} \label{sec:bop_datasets}

The BOP benchmark currently includes eleven datasets in a unified format. Each dataset includes 3D object models and training and test RGB-D images annotated with ground-truth 6D object poses. The 3D object models were created manually or using KinectFusion-like systems for 3D surface reconstruction~\cite{newcombe2011kinectfusion}. The training images are real or synthetic. All test images are real. The 3D object models are shown in Figure~\ref{fig:bop_overview_objects}, sample test images in Figure~\ref{fig:bop_overview_images}, and parameters of the datasets are provided in Table~\ref{tab:bop_dataset_params}.

Seven of the datasets were selected as core datasets (marked with a star in Table~\ref{tab:bop_dataset_params}). In the recent public challenges from 2019 and 2020, a method had to be evaluated on all core datasets to be considered for the main challenge awards~\cite{hodan2020bop}.
The core datasets include photorealistic training images generated by the approach described in Section~\ref{sec:synth_bop20}. Datasets T-LESS, TUD-L, and YCB-V include real training images and most datasets include also training images obtained by OpenGL rendering of the 3D object models on a black background.
The test images were captured in scenes with varied complexity, often with clutter and occlusion.
Besides the training and test images, the HB and ITODD datasets include also validation images -- in this case, the ground-truth poses are publicly available for the validation images but not for the test images.
A description of the individual datasets is provided in the following paragraphs.

\customparagraph{LM/LM-O~\cite{hinterstoisser2012accv,brachmann2014learning}.}
Linemod (LM) has been the most widely used dataset for 6D object pose estimation. It contains 15 texture-less
objects with discriminative color, shape, and size. Each object is associated with a set of test images showing one annotated object instance under significant clutter but only mild occlusion. Linemod-Occluded (LM-O) provides ground-truth annotation for all other instances of the modeled objects in one of the test sets, which introduces challenging
test cases with various levels of occlusion.

\customparagraph{T-LESS~\cite{hodan2017tless}.}
This dataset features 30 industry-relevant objects with no significant texture or discriminative color, with symmetries and mutual similarities in shape and size, and with some objects being a composition of other objects. T-LESS includes images from three different sensors and two types of 3D object models.
Only RGB-D images from the Primesense sensor are included in BOP. See Chapter~\ref{ch:tless} for details.

\customparagraph{ITODD~\cite{drost2017introducing}.}
The MVTec Industrial 3D Object Detection Dataset (ITODD) is focused on objects and settings typical for industrial bin picking and object inspection.
The dataset contains 28 mostly texture-less objects with different characteristics in terms of material properties (glossy \vs Lambertian), symmetry (no \vs discrete or continuous symmetries), shape (flat, long, compact, \etc), and size (diameters from $24$ to $270\,$mm).
The objects are captured in various arrangements with two industrial 3D sensors of two qualities (providing aligned grayscale and depth images) and three high-resolution grayscale sensors. Only images from the higher-quality 3D sensor are included in BOP.

\customparagraph{HB~\cite{kaskman2019homebreweddb}.}
The HomebrewedDB dataset (HB) contains high-quality reconstructed 3D models of 33 objects (17 toys, 8 industry-relevant, and 8 household objects) captured in 13 test scenes with two
RGB-D sensors.
The complexity of the scenes ranges from simple scenes with several isolated objects on a plain background to challenging scenes with heavily occluded objects and extensive clutter.
Only RGB-D images from the Primesense sensor captured in three scenes including 16 objects are currently included in BOP.

\begin{figure}[t!]
\begin{center}

\begingroup
\footnotesize
\begin{tabularx}{\textwidth}{ l *{9}{Y} }
	\toprule
	&
	&
	&
	\multicolumn{2}{c}{Train. im.} &
	\multicolumn{1}{c}{Val im.} &
	\multicolumn{2}{c}{Test im.} &
	\multicolumn{2}{c}{Test inst.} \\
	\cmidrule(l{2pt}r{2pt}){4-5} \cmidrule(l{2pt}r{2pt}){6-6} \cmidrule(l{2pt}r{2pt}){7-8} \cmidrule(l{2pt}r{2pt}){9-10}
	\multicolumn{1}{l}{Dataset} &
	\multicolumn{1}{c}{Core} &
	\multicolumn{1}{c}{Objects} &
	\multicolumn{1}{c}{Real} &
	\multicolumn{1}{c}{PBR} &
	\multicolumn{1}{c}{Real} &
	\multicolumn{1}{c}{All} &
	\multicolumn{1}{c}{Used} &
	\multicolumn{1}{c}{All} &
	\multicolumn{1}{c}{Used} \\
	\midrule
	LM \cite{hinterstoisser2012accv} & & 15 & \hspace{1ex}$\phantom{0000}$-- & 50000 & $\phantom{000}$-- & 18273 & 3000 & 18273 & 3000 \\
	LM-O \cite{brachmann2014learning} & $\ast$ & $\phantom{0}$8 & \hspace{1ex}$\phantom{0000}$-- & 50000 & $\phantom{000}$-- & $\phantom{0}$1214 & $\phantom{0}$200 & $\phantom{0}$9038 & 1445 \\
	T-LESS \cite{hodan2017tless} & $\ast$ & 30 & $\phantom{0}$37584 & 50000 & $\phantom{000}$-- & 10080 & 1000 & 67308 & 6423 \\
	ITODD \cite{drost2017introducing} & $\ast$ & 28 & \hspace{1ex}$\phantom{0000}$-- & 50000 & $\phantom{00}$54 & $\phantom{00}$721 & $\phantom{0}$721 & $\phantom{0}$3041 & 3041 \\
	HB \cite{kaskman2019homebreweddb} & $\ast$ & 33 & \hspace{1ex}$\phantom{0000}$-- & 50000 & 4420 & 13000 & $\phantom{0}$300 & 67542 & 1630 \\
	YCB-V \cite{xiang2017posecnn} & $\ast$ & 21 & 113198 & 50000 & $\phantom{000}$-- & 20738 & $\phantom{0}$900 & 98547 & 4123 \\
	RU-APC \cite{rennie2016dataset} & & 14 & \hspace{1ex}$\phantom{0000}$-- & $\phantom{0000}$-- & $\phantom{000}$-- & $\phantom{0}$5964 & 1380 & $\phantom{0}$5964 & 1380\\
	IC-BIN \cite{doumanoglou2016recovering} & $\ast$ & $\phantom{0}$2 & \hspace{1ex}$\phantom{0000}$-- & 50000 & $\phantom{000}$-- & $\phantom{00}$177 & $\phantom{0}$150 & $\phantom{0}$2176 & 1786 \\
	IC-MI \cite{tejani2014latent} & & $\phantom{0}$6 & \hspace{1ex}$\phantom{0000}$-- & $\phantom{0000}$-- & $\phantom{000}$-- & $\phantom{0}$2067 & $\phantom{0}$300 & $\phantom{0}$5318 & $\phantom{0}$800 \\
	TUD-L \cite{hodan2018bop} & $\ast$ & $\phantom{0}$3 & $\phantom{0}$38288 & 50000 & $\phantom{000}$-- & 23914 & $\phantom{0}$600 & 23914 & $\phantom{0}$600 \\
	TYO-L \cite{hodan2018bop} & & 21 & \hspace{1ex}$\phantom{0000}$-- & $\phantom{0000}$-- & $\phantom{000}$-- & $\phantom{0}$1670 & 1670 & $\phantom{0}$1670 & 1670 \\
	\bottomrule
\end{tabularx}
\captionof{table}{\label{tab:bop_dataset_params} \textbf{Parameters of the BOP datasets.} Most datasets include also training images obtained by OpenGL rendering of the 3D object models on a black background (not shown in the table).
	Extra PBR training images can be rendered by BlenderProc4BOP~\cite{denninger2019blenderproc,denninger2020blenderproc}.
	If a dataset includes both validation and test images, the ground-truth annotations are public only for the validation images. All test images are real. Column ``Test inst./All'' shows the number of annotated object instances for which at least $10\%$ of the projected surface area is visible in the test images. Columns ``Used'' show the number of test images and object instances used in the BOP Challenge 2019 and 2020.
}
\endgroup

\end{center}
\end{figure}

\customparagraph{YCB-V~\cite{xiang2017posecnn}.}
The original YCB-Video dataset provides 92 RGB-D video sequences (80 for training, 12 for test) with over 133K frames showing 21 household, both textured and texture-less, objects from the YCB dataset~\cite{calli2015ycb}. The objects are arranged in scenes with mild clutter and various levels of occlusion. For the BOP benchmark, we have manually selected 75 images with higher-quality ground-truth poses from each test video sequence.

\customparagraph{RU-APC~\cite{rennie2016dataset}.}
The Rutgers APC dataset includes 14 textured products from the Amazon Picking Challenge 2015~\cite{eppner2016lessons}, each associated with test images showing the product in a cluttered warehouse shelf. The camera was equipped with LED strips to ensure constant lighting. Ten objects from the original dataset which are non-rigid or poorly captured by the depth sensor were omitted.

\customparagraph{IC-MI/IC-BIN~\cite{tejani2014latent,doumanoglou2016recovering}.}
The dataset of Tejani \etal (IC-MI) provides 3D models of two texture-less and four textured household objects. The test images show multiple object instances under clutter and mild occlusion. The dataset of Dou\-ma\-no\-glou \etal (IC-BIN, ``scenario 2'' from the original dataset) includes test images which simulate a bin-picking scenario and show two objects from IC-MI in multiple instances under heavy occlusion. We have removed test images with low-quality ground-truth poses from both datasets and refined the ground-truth poses for the remaining images in IC-BIN.

\customparagraph{TUD-L/TYO-L~\cite{hodan2018bop}.}
These two datasets were introduced together with the initial version of the BOP benchmark~\cite{hodan2018bop} and include household objects captured under different settings of ambient and directional light. The TU Dresden Light dataset (TUD-L) contains training and test image sequences that show three moving objects under eight lighting conditions. The object poses were annotated by manually aligning the 3D object model with the first frame of the sequence and propagating the pose frame-by-frame by ICP. The Toyota Light dataset (TYO-L) contains 21 objects, each captured in multiple poses on a table with four different table cloths and five different lighting conditions.
The ground-truth object poses were obtained by estimating rough poses from manually established 2D-3D correspondences and refining the poses by ICP.
The images in both datasets are labeled by categorized lighting conditions.

\section{BOP Challenge 2017} \label{sec:bop_challenge_2017}

The BOP Challenge 2017~\cite{hodan2017sixd} (originally named SIXD Challenge~2017) was the first step in shaping the BOP benchmark. It was organized together with the 3rd International Workshop on Recovering 6D Object Pose~\cite{hodan2017r6d} at the ICCV 2017 conference.
Participants of the challenge were submitting the results of their methods by e-mail from May 11, 2017 to March 14, 2018.
The experimental setup is described in Section~\ref{sec:bop17_setup}, and the results are analyzed in Section~\ref{sec:bop17_results} and published in~\cite{hodan2018bop}.

\subsection{Evaluated Methods} \label{sec:bop17_methods}

The methods evaluated in the BOP Challenge 2017 cover all major research directions of the 6D object pose estimation field which were active at that time.
A general description of the methods is in Chapter~\ref{ch:related_work}, and a description of 
the evaluated versions, together with the setting of their key parameters, in the following paragraphs.
If not stated otherwise, the image-based methods were trained on the images obtained by OpenGL rendering of the 3D object models on a black background.

\customparagraph{Template Matching Methods.}
Hoda{\v{n}}-15~\cite{hodan2015detection} is the HashMatch template-matching method described in Chapter~\ref{ch:method_template} which applies an efficient cascade-style evaluation to each sliding window location.
The templates were generated by applying the full circle of in-plane rotations with $10^\circ$ step to a portion of the synthetic training images, resulting in 11--23K templates per object. Other parameters were set as described in~\cite{hodan2015detection}. Results of the method without the last refinement step are reported under the name Hoda{\v{n}}-15-nr.

\customparagraph{Methods Based on Point-Pair Features.}
Drost-10~\cite{drost2010model}, \ie, the method which introduced the point pair features, was evaluated using function \texttt{find\_surface\_model} from HALCON 13.0.2~\cite{halcon}. The sampling distance for the model and the scene was set to 3\% of the object diameter, 10\% of points were used as the reference points, and the surface normals were computed using the \texttt{mls} method. Points farther than 2\,m were discarded.

Drost-10-edge extends Drost-10 by detecting depth edges and favoring poses in which the model contours are aligned with the edges. A multi-modal refinement minimizes the surface distances and the distances from the reprojected model contours to the detected edges. The evaluation was performed using the same software and parameters as Drost-10 but with the parameter \texttt{train\_3d\_edges} being activated. %

In the evaluated version, Vidal-18~\cite{vidal2018method} sorts the 500 most-voted pose candidates by a surface fitting score and refines the 200 best candidates by a projective ICP.
For the final 10 candidates, they evaluate the consistency of the object surface and the object silhouette with the scene.
The sampling distance for the model and the scene
was set to 5\% of the object diameter, and 20\%~of the scene points were used as the reference points.

\customparagraph{Learning-Based Methods.}
Brachmann-14~\cite{brachmann2014learning} and Brachmann-16~\cite{brachmann2016uncertainty} use a random forest to predict, at each pixel of the input image, the object identity and the 3D object coordinates.
Brachmann-16~\cite{brachmann2016uncertainty} presents three improvements of Brachmann-14 -- see Section~\ref{sec:related_object_coordinates}.
The first two improvements  were disabled for the evaluation since we deal with RGB-D input, and it is known which objects are visible in the image.
The parameters of Brachmann-14 were set as: maximum feature offset: $20\,\textrm{px}$, features per tree node: 1000, training patches per object: 1.5M, number of trees: 3, size of the hypothesis pool: 210, refined hypotheses: 25.
The parameters of Brachmann-16 were set as: maximum feature offset: $10\,\textrm{px}$, features per tree node: 100, number of trees: 3, number of sampled hypotheses: 256, pixels drawn in each RANSAC iteration: 10K, inlier threshold: $1\,$cm. Real training images from TUD-L and T-LESS were used for both methods.

In the Tejani-14 method~\cite{tejani2014latent}, each patch of the input image is processed by a random forest and casts 6D
votes which are then aggregated to produce the final pose estimates.
The evaluated version omits the iterative
refinement step and does not perform ICP. The features and forest parameters were set as in~\cite{tejani2014latent}: number of trees: 10, maximum depth of a tree: 25, number of features in both the color gradient and the surface normal channel: 20, patch size: 1/2 the image, rendered images used to train each forest: 360.

Kehl-16~\cite{kehl2016deep} is another voting-based method which calculates auto-encoder features for the image patches and retrieves up to $k$ nearest neighbors (whose distance is below a threshold $t$) from a codebook, where each entry is associated with 6D pose votes.
The 6D hypotheses space is filtered to remove spurious votes, modes are identified by mean-shift and refined by ICP. The final hypotheses are verified in color, depth and surface normals to suppress false positives. The main parameters of the method were set as follows: patch size: $32\times32\,\textrm{px}$, patch sampling step: $6\,\textrm{px}$, $k$-nearest neighbors:~3, threshold $t$: $2$, number of extracted modes from the pose space: 8. Real training images were used for T-LESS.

\customparagraph{Methods Based on 3D Local Features.}
Buch-16~\cite{buch2016local} is a RANSAC-based method which iteratively samples three feature correspondences between the object model and the scene. The correspondences are obtained by matching 3D local shape descriptors and used to generate 6D pose candidates, whose quality is measured by the consensus set size. The final pose is refined by ICP.
The method achieved the state-of-the-art results on earlier object recognition datasets captured by LIDAR, but suffers from a cubic complexity in the number of correspondences. The number of RANSAC iterations was set to 10000, allowing only for a limited search in cluttered scenes. The method was evaluated with several descriptors:
153d SI~\cite{johnson1999using},
352d SHOT~\cite{salti2014shot},
30d ECSAD~\cite{jorgensen2015geometric},
and 1536d PPFH~\cite{buch2018local}.
None of the descriptors utilize color.

The Buch-17 method~\cite{buch2017rotational} is based on the observation that a correspondence between two oriented points on the object surface is constrained to cast votes in a 1-DoF rotational subgroup of the
special Euclidean group $SE(3)$ of object poses
(this observation is used also in methods based on point pair features). The time complexity of the method is thus linear in the number of correspondences. Kernel density estimation is used to efficiently combine the votes and estimate the 6D pose.
As Buch-16, the method relies on 3D local shape descriptors and refines the final pose estimate by ICP.
The parameters were set as in~\cite{buch2017rotational}: 60-angle tessellation was used for casting rotational votes, and  the translation/rotation bandwidth was set to 10\,mm/22.5$^\circ$.

\subsection{Experimental Setup} \label{sec:bop17_setup}

\noindent\textbf{SiSo Variant of 6D Object Localization.}
The methods were evaluated on the task of 6D localization of a \textbf{s}ingle \textbf{i}nstance of a \textbf{s}ingle \textbf{o}bject.
A test target was defined by a pair $(I, o)$, where $I$ is a test image and $o$ is an identifier of an object model visible in the image.
If multiple instances of the object model $o$ are visible in the image, then the pose of an arbitrary instance may have been reported. If multiple object models are shown in the image, and annotated with their ground truth poses, then each object model defined a different test target.
For example, if a test image shows three object models, each in two instances, then three test targets were defined and the pose of one of the two object instances had to be estimated for each test target.

The SiSo task reflects the industry-relevant bin-picking scenario where a robot needs to grasp a single arbitrary instance of the required object, \eg, a~component such as a bolt or a nut, and perform some operation with it. The difficulty of ``multiple instances, find one that you pick'' is close to ``find the instance in most favorable pose'' (least occlusion, unambiguous view). Most methods were expected, but not required, to report the most favorable pose, treating the rest as clutter.
The simpler SiSo variant of the 6D object localization task was chosen because it allowed evaluating all relevant methods out of the box. Since then, the state of the art has advanced and we have moved to the more challenging ViVo variant in the 2019 and 2020 editions of the challenge (Section~\ref{sec:bop_challenge_2020}).

\customparagraph{Pose Error and Accuracy Score.}
Compared to the evaluation methodology presented in Section~\ref{sec:bop_methodology} and used in the 2019 and 2020 editions of the challenge, the error of a 6D object pose estimate was in the 2017 edition measured with only the VSD pose-error function, and the correctness of the estimate was decided with a single threshold $\theta_{\text{VSD}} = 0.3$.
The tolerance $\delta$ used in the estimation of the visibility masks in VSD was set to $15\,$mm.
The accuracy of a method on a dataset was measured by the recall rate calculated as the fraction of test targets for which a correct pose was estimated. The overall accuracy was measured by the average of the per-dataset recall rates.

\customparagraph{Datasets.}
The methods were evaluated on datasets LM, LM-O, IC-MI, IC-BIN, T-LESS, RU-APC, and TUD-L.
Only subsets of test images
were used to remove redundancies and speed up the evaluation. The test targets were defined by object instances for which at least $10\%$ of the projected surface area is visible in the test images.

\customparagraph{Training and Test Rules.}
A method had to use fixed
hyper-parameters across all objects and datasets. For training, a method may have used the provided 3D models and training images, and rendered extra training images using the models. However, not a single pixel of test images may have been used,
nor the individual ground-truth poses or
masks provided for the test images.
Ranges of the azimuth and elevation camera angles, and a range of the camera-object distances calculated from the ground-truth poses in test images is the only information about the test set that may have been used for training.

\subsection{Results} \label{sec:bop17_results}

\begin{figure}[t!]
	\begin{center}
		\vspace{-1.5ex}
		\scriptsize
		\begin{tabular}{ c c c }
			\includegraphics[width=0.31\columnwidth]{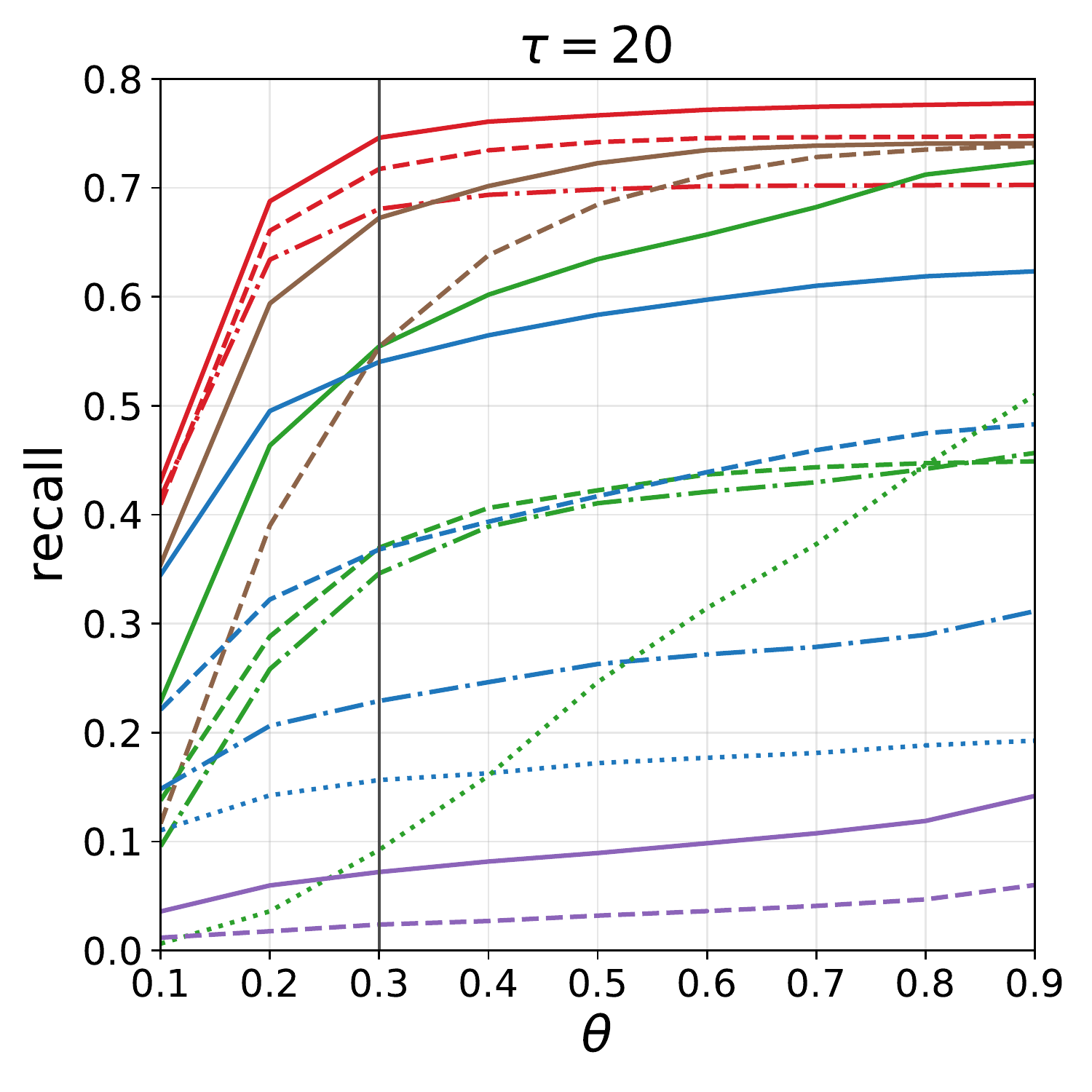} &
			\includegraphics[width=0.31\columnwidth]{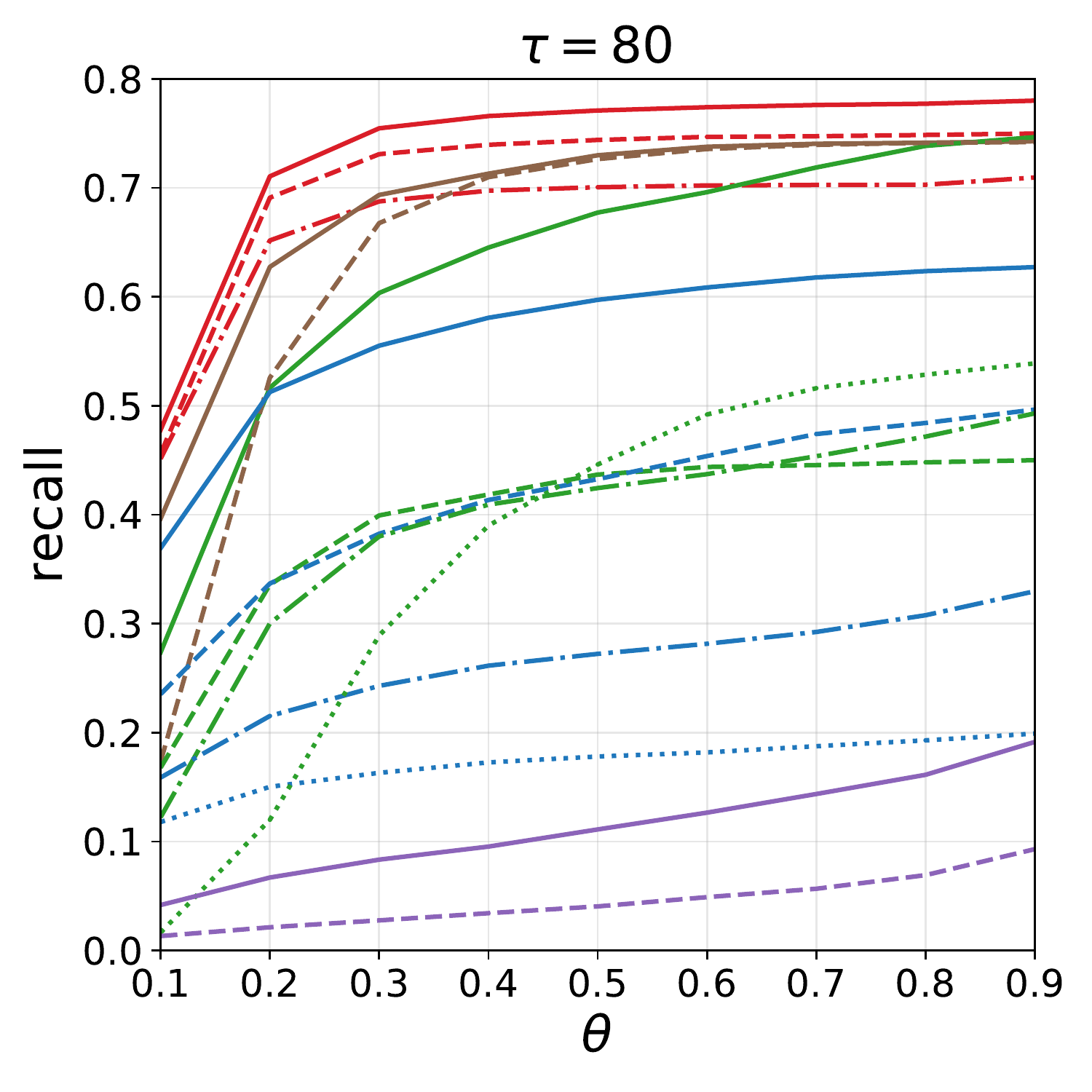} &
			\includegraphics[width=0.31\columnwidth]{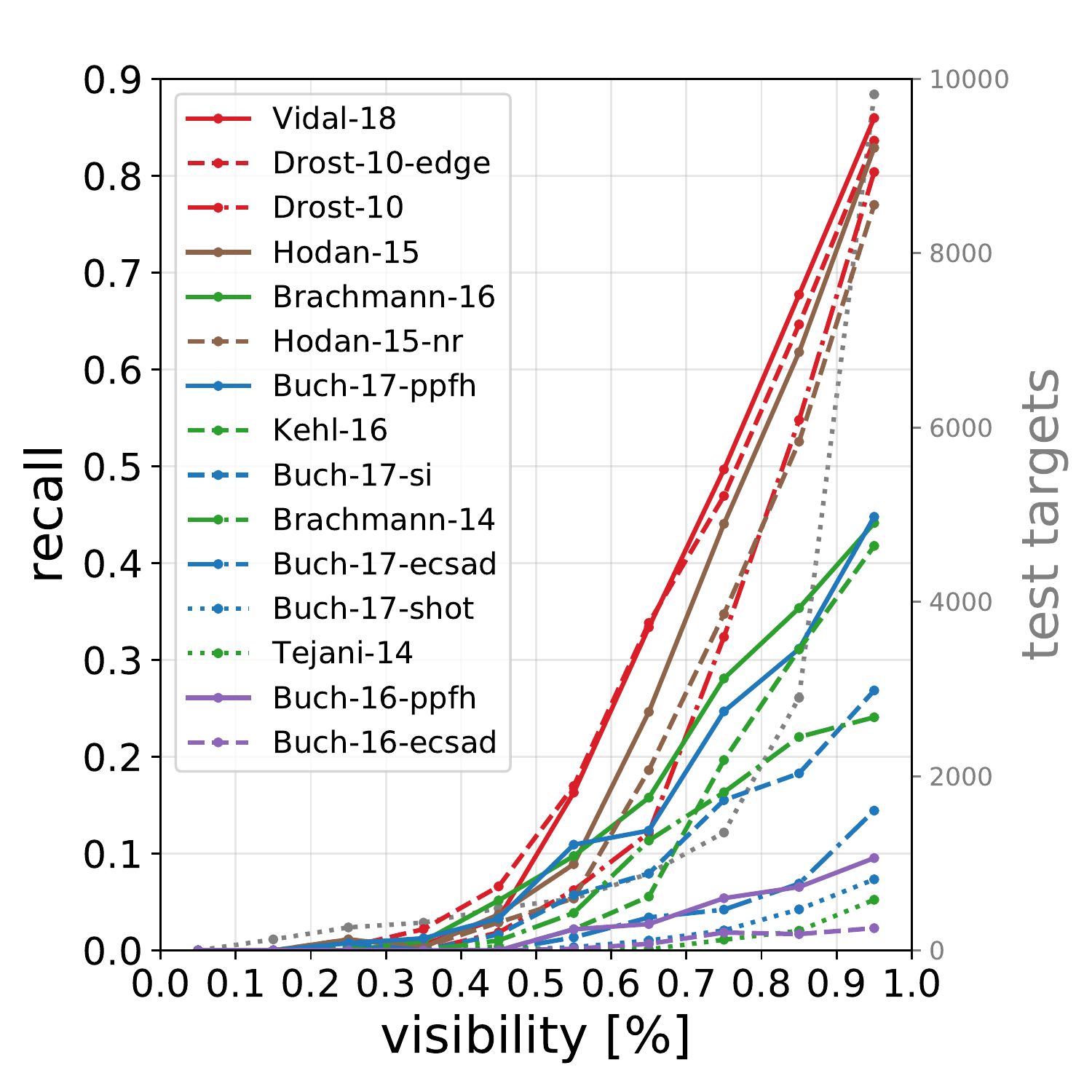} \\
		\end{tabular}
		\vspace{-1.5ex}
		\caption{
			\textbf{Recall with respect to} $\tau$\textbf{,} $\theta_{\text{VSD}}$\textbf{, and the visible surface fraction.}
			\label{tab:bop17_eval_curves} Left/middle: Average of the per-dataset recall rates for the misalignment tolerance~$\tau$ fixed to $20\,\text{mm}$ and $80\,\text{mm}$, and for different values of the correctness threshold~$\theta_{\text{VSD}}$. The curves do not change much for $\tau > 80\,\text{mm}$. Right: The recall rates \wrt the visible fraction of the target object. If more instances of the object are present in a test image, the largest visible fraction is considered.}
	\end{center}
\end{figure}

The accuracy scores of the evaluated methods for the misalignment tolerance $\tau=20\,$mm and the threshold of correctness $\theta_{\text{VSD}}=0.3$ are shown in Table~\ref{tab:bop_2017_results}.
Ranking of the methods according to the accuracy score is mostly stable across the datasets. Methods based on point-pair features perform best. Vidal-18 is the top-performing method with the average recall of 74.6\%, followed by Drost-10-edge, Drost-10, and the template-matching method Hoda{\v{n}}-15, all with the average recall above 67\%. Brachmann-16 is the best learning-based method, with 55.4\%, and Buch-17-ppfh is the best method based on 3D local features, with 54.0\%. Scores of Buch-16-si and Buch-16-shot are inferior to the other variants of this method and not presented.
Figure~\ref{tab:bop17_eval_curves} shows the average of the per-dataset scores for different values of $\tau$ and $\theta$. If the misalignment tolerance $\tau$ is increased from $20\,$mm to $80\,$mm, the scores increase only slightly for most methods. Similarly, the scores increase only slowly for $\theta > 0.3$. This suggests that poses estimated by most methods are either of a high quality or totally off, \ie, it is a hit or miss.

Table~\ref{tab:bop_2017_results} also shows accuracy scores of EPOS (Chapter~\ref{ch:method_epos}) on datasets LM-O, IC-BIN, T-LESS and TUD-L.\footnote{The object poses were originally estimated by EPOS for the BOP Challenge 2020 and here evaluated by the methodology of the BOP Challenge 2017. Results for some datasets are missing as not all datasets from 2017 were included among the core datasets in 2020 on which EPOS was trained.}
EPOS outperforms most participants of the BOP Challenge 2017 on these datasets.
Compared to the highest scores achieved on the datasets, the score of EPOS is 5\% lower on TUD-L and 8\% lower on IC-BIN, but 10\% higher on T-LESS and 19\% higher on LM-O. These are strong results considering that EPOS relies only on RGB image channels while the other methods also use the depth image channel.

\begingroup
\setlength{\tabcolsep}{3pt} %
\renewcommand{\arraystretch}{1.045} %

\begin{table}[t!]
	\begin{center}
		
		\vspace{1.0ex}
		
		\scriptsize
		\begin{tabularx}{0.8\textwidth}{ r l *{7}{Y} Y Y }
			\toprule
			\# &
			Method &
			\rotatebox[origin=c]{90}{Avg.} &
			\rotatebox[origin=c]{90}{LM} &
			\rotatebox[origin=c]{90}{LM-O} &
			\rotatebox[origin=c]{90}{IC-MI} &
			\rotatebox[origin=c]{90}{IC-BIN} &
			\rotatebox[origin=c]{90}{T-LESS} &
			\rotatebox[origin=c]{90}{RU-APC} &
			\rotatebox[origin=c]{90}{TUD-L} &
			\rotatebox[origin=c]{90}{Time} \\
			\midrule
			
			1 & Vidal-18~\cite{vidal2018method} & \cellcolor{avgcol!74.60}74.60 & \cellcolor{arcol!87.83}87.83 & \cellcolor{arcol!59.31}59.31 & \cellcolor{arcol!95.33}95.33 & \cellcolor{arcol!96.50}96.50 & \cellcolor{arcol!66.51}66.51 & \cellcolor{arcol!36.52}36.52 & \cellcolor{arcol!80.17}80.17 & $\phantom{0}$\cellcolor{timecol!4.7}4.7 \\
			2 & Drost-10-edge~\cite{drost2010model} & \cellcolor{avgcol!71.73}71.73 & \cellcolor{arcol!79.13}79.13 & \cellcolor{arcol!54.95}54.95 & \cellcolor{arcol!94.00}94.00 & \cellcolor{arcol!92.00}92.00 & \cellcolor{arcol!67.50}67.50 & \cellcolor{arcol!27.17}27.17 & \cellcolor{arcol!87.33}87.33 & \cellcolor{timecol!21.5}21.5 \\
			3 & Drost-10~\cite{drost2010model} & \cellcolor{avgcol!68.06}68.06 & \cellcolor{arcol!82.00}82.00 & \cellcolor{arcol!55.36}55.36 & \cellcolor{arcol!94.33}94.33 & \cellcolor{arcol!87.00}87.00 & \cellcolor{arcol!56.81}56.81 & \cellcolor{arcol!22.25}22.25 & \cellcolor{arcol!78.67}78.67 & $\phantom{0}$\cellcolor{timecol!2.3}2.3 \\
			4 & Hoda{\v{n}}-15~\cite{hodan2015detection} & \cellcolor{avgcol!67.23}67.23 & \cellcolor{arcol!87.10}87.10 & \cellcolor{arcol!51.42}51.42 & \cellcolor{arcol!95.33}95.33 & \cellcolor{arcol!90.50}90.50 & \cellcolor{arcol!63.18}63.18 & \cellcolor{arcol!37.61}37.61 & \cellcolor{arcol!45.50}45.50 & \cellcolor{timecol!13.5}13.5 \\
			5 & Brachmann-16~\cite{brachmann2016uncertainty} & \cellcolor{avgcol!55.44}55.44 & \cellcolor{arcol!75.33}75.33 & \cellcolor{arcol!52.04}52.04 & \cellcolor{arcol!73.33}73.33 & \cellcolor{arcol!56.50}56.50 & \cellcolor{arcol!17.84}17.84 & \cellcolor{arcol!24.35}24.35 & \cellcolor{arcol!88.67}88.67 & $\phantom{0}$\cellcolor{timecol!4.4}4.4 \\
			6 & Hoda{\v{n}}-15-nr~\cite{hodan2015detection} & \cellcolor{avgcol!55.40}55.40 & \cellcolor{arcol!69.83}69.83 & \cellcolor{arcol!34.39}34.39 & \cellcolor{arcol!84.67}84.67 & \cellcolor{arcol!76.00}76.00 & \cellcolor{arcol!62.70}62.70 & \cellcolor{arcol!32.39}32.39 & \cellcolor{arcol!27.83}27.83 & \cellcolor{timecol!12.3}12.3 \\
			7 & Buch-17-ppfh~\cite{buch2017rotational} & \cellcolor{avgcol!54.02}54.02 & \cellcolor{arcol!56.60}56.60 & \cellcolor{arcol!36.96}36.96 & \cellcolor{arcol!95.00}95.00 & \cellcolor{arcol!75.00}75.00 & \cellcolor{arcol!25.10}25.10 & \cellcolor{arcol!20.80}20.80 & \cellcolor{arcol!68.67}68.67 & \cellcolor{timecol!14.2}14.2 \\
			8 & Kehl-16~\cite{kehl2016deep} & \cellcolor{avgcol!36.97}36.97 & \cellcolor{arcol!58.20}58.20 & \cellcolor{arcol!33.91}33.91 & \cellcolor{arcol!65.00}65.00 & \cellcolor{arcol!44.00}44.00 & \cellcolor{arcol!24.60}24.60 & \cellcolor{arcol!25.58}25.58 & $\phantom{0}$\cellcolor{arcol!7.50}7.50 & $\phantom{0}$\cellcolor{timecol!1.8}1.8 \\
			9 & Buch-17-si~\cite{buch2017rotational} & \cellcolor{avgcol!36.81}36.81 & \cellcolor{arcol!33.33}33.33 & \cellcolor{arcol!20.35}20.35 & \cellcolor{arcol!67.33}67.33 & \cellcolor{arcol!59.00}59.00 & \cellcolor{arcol!13.34}13.34 & \cellcolor{arcol!23.12}23.12 & \cellcolor{arcol!41.17}41.17 & \cellcolor{timecol!15.9}15.9 \\
			10 & Brachmann-14~\cite{brachmann2014learning} & \cellcolor{avgcol!34.61}34.61 & \cellcolor{arcol!67.60}67.60 & \cellcolor{arcol!41.52}41.52 & \cellcolor{arcol!78.67}78.67 & \cellcolor{arcol!24.00}24.00 & $\phantom{0}$\cellcolor{arcol!0.25}0.25 & \cellcolor{arcol!30.22}30.22 & $\phantom{0}$\cellcolor{arcol!0}0.00 & $\phantom{0}$\cellcolor{timecol!1.4}1.4 \\
			11 & Buch-17-ecsad~\cite{buch2017rotational} & \cellcolor{avgcol!22.90}22.90 & \cellcolor{arcol!13.27}13.27 & $\phantom{0}$\cellcolor{arcol!9.62}9.62 & \cellcolor{arcol!40.67}40.67 & \cellcolor{arcol!59.00}59.00 & $\phantom{0}$\cellcolor{arcol!7.16}7.16 & $\phantom{0}$\cellcolor{arcol!6.59}6.59 & \cellcolor{arcol!24.00}24.00 & $\phantom{0}$\cellcolor{timecol!5.9}5.9 \\
			12 & Buch-17-shot~\cite{buch2017rotational} & \cellcolor{avgcol!15.64}15.64 & $\phantom{0}$\cellcolor{arcol!5.97}5.97 & $\phantom{0}$\cellcolor{arcol!1.45}1.45 & \cellcolor{arcol!43.00}43.00 & \cellcolor{arcol!38.50}38.50 & $\phantom{0}$\cellcolor{arcol!3.83}3.83 & $\phantom{0}$\cellcolor{arcol!0.07}0.07 & \cellcolor{arcol!16.67}16.67 & $\phantom{0}$\cellcolor{timecol!6.7}6.7 \\
			13 & Tejani-14~\cite{tejani2014latent} & $\phantom{0}$\cellcolor{avgcol!9.23}9.23 & \cellcolor{arcol!12.10}12.10 & $\phantom{0}$\cellcolor{arcol!4.50}4.50 & \cellcolor{arcol!36.33}36.33 & \cellcolor{arcol!10.00}10.00 & $\phantom{0}$\cellcolor{arcol!0.13}0.13 & $\phantom{0}$\cellcolor{arcol!01.52}1.52 & $\phantom{0}$\cellcolor{arcol!0}0.00 & $\phantom{0}$\cellcolor{timecol!1.4}1.4 \\
			14 & Buch-16-ppfh~\cite{buch2016local} & $\phantom{0}$\cellcolor{avgcol!7.20}7.20 & $\phantom{0}$\cellcolor{arcol!8.13}8.13 & $\phantom{0}$\cellcolor{arcol!2.28}2.28 & \cellcolor{arcol!20.00}20.00 & $\phantom{0}$\cellcolor{arcol!2.50}2.50 & $\phantom{0}$\cellcolor{arcol!7.81}7.81 & $\phantom{0}$\cellcolor{arcol!8.99}8.99 & $\phantom{0}$\cellcolor{arcol!0.67}0.67 & \cellcolor{timecol!47.1}47.1 \\
			15 & Buch-16-ecsad~\cite{buch2016local} & $\phantom{0}$\cellcolor{avgcol!2.38}2.38 & $\phantom{0}$\cellcolor{arcol!3.70}3.70 & $\phantom{0}$\cellcolor{arcol!0.97}0.97 & $\phantom{0}$\cellcolor{arcol!3.67}3.67 & $\phantom{0}$\cellcolor{arcol!4.00}4.00 & $\phantom{0}$\cellcolor{arcol!1.24}1.24 & $\phantom{0}$\cellcolor{arcol!2.90}2.90 & $\phantom{0}$\cellcolor{arcol!0.17}0.17 & \cellcolor{timecol!39.1}39.1 \\
			
			& & & & & & & & & & \\
			
			 & EPOS (Chapter~\ref{ch:method_epos}) & -- & -- & \cellcolor{arcol!78.28}78.28 & -- & \cellcolor{arcol!88.50}88.50 & \cellcolor{arcol!77.78}77.78 & -- & \cellcolor{arcol!83.67}83.67 & -- \\
			
			\bottomrule
		\end{tabularx}
		
		\captionof{table}{\label{tab:bop_2017_results}
			\textbf{Results of the BOP Challenge 2017.} Recall rates (in \%) for the misalignment tolerance $\tau=20\,\text{mm}$ and the threshold of correctness $\theta_{\text{VSD}}=0.3$.
			The recall rate is the percentage of test targets for which a correct object pose was estimated.
			The methods are sorted by the average of the per-dataset recall rates.
			The right-most column shows the average processing time per test target (in seconds). The last row shows results of the EPOS method (Chapter~\ref{ch:method_epos}).
		}
	\end{center}
\end{table}

\endgroup

The average running times per test target are reported in Table~\ref{tab:bop_2017_results}. However, the methods were evaluated on different computers
and the presented running times are therefore not directly comparable.
Moreover, the methods were optimized primarily for accuracy, not for speed.
For example, we evaluated Drost-10 with several parameter settings and observed that the running time can be lowered by a factor of $\mytilde5$ to $0.5\,$s with only a relatively small drop of the average recall from $68.1\%$ to $65.8\%$. However, in Table~\ref{tab:bop_2017_results} we present the result with the highest recall. Brachmann-14 could be sped up by sub-sampling the 3D object models and Hoda{\v{n}}-15 by using less object templates.

Occlusion is a big challenge for the methods, as shown by the accuracy scores dropping swiftly already at low levels of occlusion (Figure~\ref{tab:bop17_eval_curves}, right). The big gap between LM and LM-O scores provide further evidence. All methods perform on LM by at least 30\% better than on LM-O, which includes the same but partially occluded objects. Inspection of estimated poses on T-LESS test images confirms the weak performance for occluded objects.
Scores on TUD-L show that varying lighting conditions present a serious challenge for methods that rely on synthetic RGB training images, which were generated with fixed lighting. Methods relying only on depth information (\eg, Vidal-18, Drost-10) are noticeably more robust under such conditions.
Note that Brach\-mann-16 achieved a high score on TUD-L despite relying on RGB images because it used real training images, which were captured under the same range of lighting conditions as the test images.
Methods based on 3D local features and learning-based methods have very low scores
on T-LESS, which is likely caused by the object symmetries and similarities.
All methods perform poorly on RU-APC, which is likely because of a higher level of noise in the depth images.

\section{BOP Challenge 2019 and 2020} \label{sec:bop_challenge_2020}

The BOP Challenge 2019 was organized together with the 5th International Workshop on Recovering 6D Object Pose~\cite{hodan2019r6d} at ICCV 2019, and the BOP Challenge 2020 together with the 6th International Workshop on Recovering 6D Object Pose~\cite{hodan2020r6d} at ECCV 2020.
Participants were submitting their results to the online evaluation system at \texttt{\href{http://bop.felk.cvut.cz/}{bop.felk.cvut.cz}}, from July 30 to October 21 in 2019, and from June 5 to August 19 in 2020. The results are analyzed in Section~\ref{sec:bop20_results} and published in~\cite{hodan2020bop}.

In the BOP Challenge 2019, methods using the depth image channel, which were mostly based on the point pair features (PPF's)~\cite{drost2010model}, clearly outperformed methods relying only on the RGB channels, all of which were based on deep neural networks (DNN's).
The PPF-based methods match pairs of oriented 3D points between the point cloud
of the test scene and the 3D object model, and aggregate the matches via a voting scheme. As each pair is described by only the distance and relative orientation of the two points, PPF-based methods can be effectively trained directly on the 3D object models, without the need to synthesize any training images. In contrast, DNN-based methods require a large number of annotated training images, which have been typically obtained by OpenGL rendering of the 3D object models
with
random backgrounds~\cite{kehl2017ssd,rad2017bb8,hinterstoisser2017pre,dwibedi2017cut}. However, as
shown in Chapter~\ref{ch:synthesis}, the evident domain gap between these ``render\;\&\;paste'' training images and real test images presumably limits the potential of the DNN-based methods.

To reduce the gap between the synthetic and real domains and thus to bring fresh air to the DNN-based methods, we created BlenderProc4BOP~\cite{denninger2019blenderproc,denninger2020blenderproc,blenderproc4bop}, an open-source and light-weight physically-based renderer (PBR). %
Furthermore, to reduce the entry barrier of the challenge and to standardize the training set, the participants were provided with 350K pre-rendered PBR images described in Section~\ref{sec:synth_bop20}.

In 2020, the DNN-based methods finally caught up with the PPF-based methods -- five methods outperformed Vidal-Sensors18~\cite{vidal2018method}, the PPF-based winner from 2017 and 2019. Three of the top five methods, including the top-performing one, are single-view variants of CosyPose, a DNN-based method by Labb{\'e} \etal~\cite{labbe2020cosypose}. Strong data augmentation, similar to~\cite{sundermeyer2019augmented}, was identified as one of the key ingredients of this method. The second is a hybrid DNN+PPF method by K\"onig and Drost~\cite{koenig2020hybrid}, and the fourth is Pix2Pose, a DNN-based method by Park \etal~\cite{park2019pix2pose}. The first two methods used RGB-D image channels, while the third method achieved strong results with RGB channels only.

Methods achieved noticeably higher accuracy scores when trained on PBR images than when trained on ``render\;\&\;paste'' images. Although adding real training images yielded even higher scores, competitive results were achieved with PBR images only -- out of the 26 evaluated methods, the fifth was trained only on the PBR images. Interestingly, the increased photorealism from the PBR images led to clear improvements of also the CosyPose method, despite the strong data augmentation applied to the training images.

\subsection{Evaluated Methods} \label{sec:bop20_methods}

In total, 26 methods were evaluated on all seven core datasets.
Results of 11 methods were received in 2019 and results of 15 methods in 2020 (column Year in Table~\ref{tab:bop_2020_results}).
Compared to the ``classical'' methods evaluated in the 2017 edition of the challenge, which mostly do not involve any machine learning techniques and are prevailed by methods based on the point pair features, the majority of methods evaluated in the 2019 and 2020 editions reflect the current trend in computer vision and utilize deep neural networks. The following paragraphs provide a brief overview of the evaluated methods. Details are provided in Chapter~\ref{ch:related_work} and in the documentation of submissions on the project website.

\customparagraph{PPF-Based Methods.}
The method by Drost \etal~\cite{drost2010model}, which introduced the point~pair features, is evaluated in four versions, including the original version with two parameter settings and two extensions utilizing intensity and depth edges for voting and refinement.
Evaluated is also the method by Vidal \etal~\cite{vidal2018method}. %

\customparagraph{DNN+PPF-Based Methods.}
The hybrid method by F{\'e}lix and Neves~\cite{rodrigues2019deep,raposo2017using} segments object instances by Mask R-CNN~\cite{he2017mask}, selects the corresponding part of the 3D point cloud of the test scene for each instance mask, estimates the object pose using the point pair features~\cite{drost2010model}, and refines the pose with ICP.
The method of K\"onig and Drost~\cite{koenig2020hybrid} follows the same pipeline, with the instance masks predicted by RetinaMask~\cite{fu2019retinamask} or Mask R-CNN~\cite{he2017mask}, whichever performs better on the validation set.

\customparagraph{DNN-Based Methods.}
The EPOS method described in Chapter~\ref{ch:method_epos}, Pix2Pose by Park \etal~\cite{park2019pix2pose}, DPOD by Zakharov \etal~\cite{zakharov2019dpod}, and the method by Liu \etal~\cite{liu2020leaping} predict 2D-3D correspondences with deep neural networks and solve for the object pose using variants of the P\emph{n}P-RANSAC algorithm. The CDPN method by Li \etal~\cite{li2019cdpn} estimates the 3D rotation from 2D-3D correspondences and the 3D translation by directly regressing a scale-invariant 3D translation vector. The CosyPose method by Labb{\'e} \etal~\cite{labbe2020cosypose} first predicts 2D bounding boxes of the objects using Mask R-CNN~\cite{he2017mask}, and then applies to each box a DNN model for coarse pose estimation followed by a DNN model for iterative refinement.
The AAE method by Sundermeyer \etal~\cite{sundermeyer2019augmented} localizes the objects with 2D bounding boxes and calculates a descriptor of each detected region by the so-called Augmented Autoencoder, which consists of an encoder mapping the image region to a latent descriptor space and a decoder mapping the descriptor to a denoised reconstruction of the input image region.
The later method by Sundermeyer \etal~\cite{sundermeyer2020multi} shares a single encoder among multiple objects, which dramatically improves the scalability of the method.
The PointVoteNet2 method by Hagelskj{\ae}r and Buch~\cite{hagelskjaer2020pointvotenet} estimates the object poses from a point cloud using a PointNet backbone~\cite{qi2017pointnet}.

\subsection{Experimental Setup}

Both 2019 and 2020 editions of the challenge followed the evaluation methodology described in Section~\ref{sec:bop_methodology}.
A method had to be evaluated at least on the seven core datasets (LM-O, T-LESS, TUD-L, IC-BIN, ITODD, HB, YCB-V) to be considered for the main challenge awards~\cite{hodan2020bop}.
Only subsets of test images were used to remove redundancies and speed up the evaluation, and only object instances for which at least $10\%$ of the projected surface area is visible in the test images were to be localized.
The participants had to follow the same rules for training and test as in the 2017 edition.

\begingroup
\setlength{\tabcolsep}{3pt} %
\renewcommand{\arraystretch}{1.045} %

\begin{table}[t!]
\begin{center}
\scriptsize
\begin{tabularx}{0.85\linewidth}{r l Y Y Y Y Y Y Y Y Y}
\toprule
\# &
Method &
\rotatebox[origin=c]{90}{Avg.} &
\rotatebox[origin=c]{90}{LM-O} &
\rotatebox[origin=c]{90}{T-LESS} &
\rotatebox[origin=c]{90}{TUD-L} &
\rotatebox[origin=c]{90}{IC-BIN} &
\rotatebox[origin=c]{90}{ITODD} &
\rotatebox[origin=c]{90}{HB} &
\rotatebox[origin=c]{90}{YCB-V} &
\rotatebox[origin=c]{90}{Time} \\
\midrule
1 & CosyPose-ECCV20-Synt+Real-ICP~\cite{labbe2020cosypose} & \cellcolor{avgcol!69.8}69.8 & \cellcolor{arcol!71.4}71.4 & \cellcolor{arcol!70.1}70.1 & \cellcolor{arcol!93.9}93.9 & \cellcolor{arcol!64.7}64.7 & \cellcolor{arcol!31.3}31.3 & \cellcolor{arcol!71.2}71.2 & \cellcolor{arcol!86.1}86.1 & \cellcolor{timecol!13.74}13.74 \\
2 & K\"onig-Hybrid-DL-PointPairs~\cite{koenig2020hybrid} & \cellcolor{avgcol!63.9}63.9 & \cellcolor{arcol!63.1}63.1 & \cellcolor{arcol!65.5}65.5 & \cellcolor{arcol!92.0}92.0 & \cellcolor{arcol!43.0}43.0 & \cellcolor{arcol!48.3}48.3 & \cellcolor{arcol!65.1}65.1 & \cellcolor{arcol!70.1}70.1 & $\phantom{0}$\cellcolor{timecol!0.63}0.63 \\
3 & CosyPose-ECCV20-Synt+Real~\cite{labbe2020cosypose} & \cellcolor{avgcol!63.7}63.7 & \cellcolor{arcol!63.3}63.3 & \cellcolor{arcol!72.8}72.8 & \cellcolor{arcol!82.3}82.3 & \cellcolor{arcol!58.3}58.3 & \cellcolor{arcol!21.6}21.6 & \cellcolor{arcol!65.6}65.6 & \cellcolor{arcol!82.1}82.1 & $\phantom{0}$\cellcolor{timecol!0.45}0.45 \\
4 & Pix2Pose-BOP20\_w/ICP-ICCV19~\cite{park2019pix2pose} & \cellcolor{avgcol!59.1}59.1 & \cellcolor{arcol!58.8}58.8 & \cellcolor{arcol!51.2}51.2 & \cellcolor{arcol!82.0}82.0 & \cellcolor{arcol!39.0}39.0 & \cellcolor{arcol!35.1}35.1 & \cellcolor{arcol!69.5}69.5 & \cellcolor{arcol!78.0}78.0 & $\phantom{0}$\cellcolor{timecol!4.84}4.84 \\
5 & CosyPose-ECCV20-PBR~\cite{labbe2020cosypose} & \cellcolor{avgcol!57.0}57.0 & \cellcolor{arcol!63.3}63.3 & \cellcolor{arcol!64.0}64.0 & \cellcolor{arcol!68.5}68.5 & \cellcolor{arcol!58.3}58.3 & \cellcolor{arcol!21.6}21.6 & \cellcolor{arcol!65.6}65.6 & \cellcolor{arcol!57.4}57.4 & $\phantom{0}$\cellcolor{timecol!0.47}0.47 \\
6 & Vidal-Sensors18~\cite{vidal2018method} & \cellcolor{avgcol!56.9}56.9 & \cellcolor{arcol!58.2}58.2 & \cellcolor{arcol!53.8}53.8 & \cellcolor{arcol!87.6}87.6 & \cellcolor{arcol!39.3}39.3 & \cellcolor{arcol!43.5}43.5 & \cellcolor{arcol!70.6}70.6 & \cellcolor{arcol!45.0}45.0 & $\phantom{0}$\cellcolor{timecol!3.22}3.22 \\
7 & CDPNv2\_BOP20-RGB-ICP~\cite{li2019cdpn} & \cellcolor{avgcol!56.8}56.8 & \cellcolor{arcol!63.0}63.0 & \cellcolor{arcol!46.4}46.4 & \cellcolor{arcol!91.3}91.3 & \cellcolor{arcol!45.0}45.0 & \cellcolor{arcol!18.6}18.6 & \cellcolor{arcol!71.2}71.2 & \cellcolor{arcol!61.9}61.9 & $\phantom{0}$\cellcolor{timecol!1.46}1.46 \\
8 & Drost-CVPR10-Edges~\cite{drost2010model} & \cellcolor{avgcol!55.0}55.0 & \cellcolor{arcol!51.5}51.5 & \cellcolor{arcol!50.0}50.0 & \cellcolor{arcol!85.1}85.1 & \cellcolor{arcol!36.8}36.8 & \cellcolor{arcol!57.0}57.0 & \cellcolor{arcol!67.1}67.1 & \cellcolor{arcol!37.5}37.5 & \cellcolor{timecol!87.57}87.57 \\
9 & CDPNv2\_BOP20-PBR-ICP~\cite{li2019cdpn} & \cellcolor{avgcol!53.4}53.4 & \cellcolor{arcol!63.0}63.0 & \cellcolor{arcol!43.5}43.5 & \cellcolor{arcol!79.1}79.1 & \cellcolor{arcol!45.0}45.0 & \cellcolor{arcol!18.6}18.6 & \cellcolor{arcol!71.2}71.2 & \cellcolor{arcol!53.2}53.2 & $\phantom{0}$\cellcolor{timecol!1.49}1.49 \\
10 & CDPNv2\_BOP20-RGB~\cite{li2019cdpn} & \cellcolor{avgcol!52.9}52.9 & \cellcolor{arcol!62.4}62.4 & \cellcolor{arcol!47.8}47.8 & \cellcolor{arcol!77.2}77.2 & \cellcolor{arcol!47.3}47.3 & \cellcolor{arcol!10.2}10.2 & \cellcolor{arcol!72.2}72.2 & \cellcolor{arcol!53.2}53.2 & $\phantom{0}$\cellcolor{timecol!0.94}0.94 \\
11 & Drost-CVPR10-3D-Edges~\cite{drost2010model} & \cellcolor{avgcol!50.0}50.0 & \cellcolor{arcol!46.9}46.9 & \cellcolor{arcol!40.4}40.4 & \cellcolor{arcol!85.2}85.2 & \cellcolor{arcol!37.3}37.3 & \cellcolor{arcol!46.2}46.2 & \cellcolor{arcol!62.3}62.3 & \cellcolor{arcol!31.6}31.6 & \cellcolor{timecol!80.06}80.06 \\
12 & Drost-CVPR10-3D-Only~\cite{drost2010model} & \cellcolor{avgcol!48.7}48.7 & \cellcolor{arcol!52.7}52.7 & \cellcolor{arcol!44.4}44.4 & \cellcolor{arcol!77.5}77.5 & \cellcolor{arcol!38.8}38.8 & \cellcolor{arcol!31.6}31.6 & \cellcolor{arcol!61.5}61.5 & \cellcolor{arcol!34.4}34.4 & $\phantom{0}$\cellcolor{timecol!7.70}7.70 \\
13 & CDPN\_BOP19-RGB~\cite{li2019cdpn} & \cellcolor{avgcol!47.9}47.9 & \cellcolor{arcol!56.9}56.9 & \cellcolor{arcol!49.0}49.0 & \cellcolor{arcol!76.9}76.9 & \cellcolor{arcol!32.7}32.7 & $\phantom{0}$\cellcolor{arcol!6.7}6.7 & \cellcolor{arcol!67.2}67.2 & \cellcolor{arcol!45.7}45.7 & $\phantom{0}$\cellcolor{timecol!0.48}0.48 \\
14 & CDPNv2\_BOP20-PBR~\cite{li2019cdpn} & \cellcolor{avgcol!47.2}47.2 & \cellcolor{arcol!62.4}62.4 & \cellcolor{arcol!40.7}40.7 & \cellcolor{arcol!58.8}58.8 & \cellcolor{arcol!47.3}47.3 & \cellcolor{arcol!10.2}10.2 & \cellcolor{arcol!72.2}72.2 & \cellcolor{arcol!39.0}39.0 & $\phantom{0}$\cellcolor{timecol!0.98}0.98 \\
15 & leaping from 2D to 6D~\cite{liu2020leaping} & \cellcolor{avgcol!47.1}47.1 & \cellcolor{arcol!52.5}52.5 & \cellcolor{arcol!40.3}40.3 & \cellcolor{arcol!75.1}75.1 & \cellcolor{arcol!34.2}34.2 & $\phantom{0}$\cellcolor{arcol!7.7}7.7 & \cellcolor{arcol!65.8}65.8 & \cellcolor{arcol!54.3}54.3 & $\phantom{0}$\cellcolor{timecol!0.42}0.42 \\
16 & EPOS-BOP20-PBR~\cite{hodan2020epos} & \cellcolor{avgcol!45.7}45.7 & \cellcolor{arcol!54.7}54.7 & \cellcolor{arcol!46.7}46.7 & \cellcolor{arcol!55.8}55.8 & \cellcolor{arcol!36.3}36.3 & \cellcolor{arcol!18.6}18.6 & \cellcolor{arcol!58.0}58.0 & \cellcolor{arcol!49.9}49.9 & $\phantom{0}$\cellcolor{timecol!1.87}1.87 \\
17 & Drost-CVPR10-3D-Only-Faster~\cite{drost2010model} & \cellcolor{avgcol!45.4}45.4 & \cellcolor{arcol!49.2}49.2 & \cellcolor{arcol!40.5}40.5 & \cellcolor{arcol!69.6}69.6 & \cellcolor{arcol!37.7}37.7 & \cellcolor{arcol!27.4}27.4 & \cellcolor{arcol!60.3}60.3 & \cellcolor{arcol!33.0}33.0 & $\phantom{0}$\cellcolor{timecol!1.38}1.38 \\
18 & F{\'e}lix\&Neves-ICRA17-IET19~\cite{rodrigues2019deep,raposo2017using} & \cellcolor{avgcol!41.2}41.2 & \cellcolor{arcol!39.4}39.4 & \cellcolor{arcol!21.2}21.2 & \cellcolor{arcol!85.1}85.1 & \cellcolor{arcol!32.3}32.3 & $\phantom{0}$\cellcolor{arcol!6.9}6.9 & \cellcolor{arcol!52.9}52.9 & \cellcolor{arcol!51.0}51.0 & \cellcolor{timecol!55.78}55.78 \\
19 & Sundermeyer-IJCV19+ICP~\cite{sundermeyer2019augmented} & \cellcolor{avgcol!39.8}39.8 & \cellcolor{arcol!23.7}23.7 & \cellcolor{arcol!48.7}48.7 & \cellcolor{arcol!61.4}61.4 & \cellcolor{arcol!28.1}28.1 & \cellcolor{arcol!15.8}15.8 & \cellcolor{arcol!50.6}50.6 & \cellcolor{arcol!50.5}50.5 & $\phantom{0}$\cellcolor{timecol!0.86}0.86 \\
20 & Zhigang-CDPN-ICCV19~\cite{li2019cdpn} & \cellcolor{avgcol!35.3}35.3 & \cellcolor{arcol!37.4}37.4 & \cellcolor{arcol!12.4}12.4 & \cellcolor{arcol!75.7}75.7 & \cellcolor{arcol!25.7}25.7 & $\phantom{0}$\cellcolor{arcol!7.0}7.0 & \cellcolor{arcol!47.0}47.0 & \cellcolor{arcol!42.2}42.2 & $\phantom{0}$\cellcolor{timecol!0.51}0.51 \\
21 & PointVoteNet2~\cite{hagelskjaer2020pointvotenet} & \cellcolor{avgcol!35.1}35.1 & \cellcolor{arcol!65.3}65.3 & $\phantom{0}$\cellcolor{arcol!0.4}0.4 & \cellcolor{arcol!67.3}67.3 & \cellcolor{arcol!26.4}26.4 & $\phantom{0}$\cellcolor{arcol!0.1}0.1 & \cellcolor{arcol!55.6}55.6 & \cellcolor{arcol!30.8}30.8 & \cellcolor{timecol!0}- \\
22 & Pix2Pose-BOP20-ICCV19~\cite{park2019pix2pose} & \cellcolor{avgcol!34.2}34.2 & \cellcolor{arcol!36.3}36.3 & \cellcolor{arcol!34.4}34.4 & \cellcolor{arcol!42.0}42.0 & \cellcolor{arcol!22.6}22.6 & \cellcolor{arcol!13.4}13.4 & \cellcolor{arcol!44.6}44.6 & \cellcolor{arcol!45.7}45.7 & $\phantom{0}$\cellcolor{timecol!1.22}1.22 \\
23 & Sundermeyer-IJCV19~\cite{sundermeyer2019augmented} & \cellcolor{avgcol!27.0}27.0 & \cellcolor{arcol!14.6}14.6 & \cellcolor{arcol!30.4}30.4 & \cellcolor{arcol!40.1}40.1 & \cellcolor{arcol!21.7}21.7 & \cellcolor{arcol!10.1}10.1 & \cellcolor{arcol!34.6}34.6 & \cellcolor{arcol!37.7}37.7 & $\phantom{0}$\cellcolor{timecol!0.19}0.19 \\
24 & SingleMultiPathEncoder-CVPR20~\cite{sundermeyer2020multi} & \cellcolor{avgcol!24.1}24.1 & \cellcolor{arcol!21.7}21.7 & \cellcolor{arcol!31.0}31.0 & \cellcolor{arcol!33.4}33.4 & \cellcolor{arcol!17.5}17.5 & $\phantom{0}$\cellcolor{arcol!6.7}6.7 & \cellcolor{arcol!29.3}29.3 & \cellcolor{arcol!28.9}28.9 & $\phantom{0}$\cellcolor{timecol!0.19}0.19 \\
25 & Pix2Pose-BOP19-ICCV19~\cite{park2019pix2pose} & \cellcolor{avgcol!20.5}20.5 & $\phantom{0}$\cellcolor{arcol!7.7}7.7 & \cellcolor{arcol!27.5}27.5 & \cellcolor{arcol!34.9}34.9 & \cellcolor{arcol!21.5}21.5 & $\phantom{0}$\cellcolor{arcol!3.2}3.2 & \cellcolor{arcol!20.0}20.0 & \cellcolor{arcol!29.0}29.0 & $\phantom{0}$\cellcolor{timecol!0.79}0.79 \\
26 & DPOD (synthetic)~\cite{zakharov2019dpod} & \cellcolor{avgcol!16.1}16.1 & \cellcolor{arcol!16.9}16.9 & $\phantom{0}$\cellcolor{arcol!8.1}8.1 & \cellcolor{arcol!24.2}24.2 & \cellcolor{arcol!13.0}13.0 & $\phantom{0}$\cellcolor{arcol!0.0}0.0 & \cellcolor{arcol!28.6}28.6 & \cellcolor{arcol!22.2}22.2 & $\phantom{0}$\cellcolor{timecol!0.23}0.23 \\

\\
\end{tabularx}

\begin{tabularx}{0.85\linewidth}{r l l l l l l L l}
\toprule
\# &
Method &
Year &
PPF &
DNN &
Train &
...type &
Test &
Refine \\
\midrule    
1 & CosyPose-ECCV20-Synt+Real-ICP~\cite{labbe2020cosypose} & \cellcolor{ccol2!100}2020 & - & \cellcolor{ccol2!100}3/set & \cellcolor{ccol2!100}rgb & pbr{\tiny +}real & rgb-d & \cellcolor{ccol2!100}rgb{\tiny +}icp \\
2 & K\"onig-Hybrid-DL-PointPairs~\cite{koenig2020hybrid} & \cellcolor{ccol2!100}2020 & \cellcolor{ccol2!100}yes & \cellcolor{ccol2!100}1/set & \cellcolor{ccol2!100}rgb & syn{\tiny +}real & rgb-d & \cellcolor{ccol2!100}icp \\
3 & CosyPose-ECCV20-Synt+Real~\cite{labbe2020cosypose} & \cellcolor{ccol2!100}2020 & - & \cellcolor{ccol2!100}3/set & \cellcolor{ccol2!100}rgb & pbr{\tiny +}real & \cellcolor{ccol2!100}rgb & \cellcolor{ccol2!100}rgb \\
4 & Pix2Pose-BOP20\_w/ICP-ICCV19~\cite{park2019pix2pose} & \cellcolor{ccol2!100}2020 & - & 1/obj & \cellcolor{ccol2!100}rgb & pbr{\tiny +}real & rgb-d & \cellcolor{ccol2!100}icp \\
5 & CosyPose-ECCV20-PBR~\cite{labbe2020cosypose} & \cellcolor{ccol2!100}2020 & - & \cellcolor{ccol2!100}3/set & \cellcolor{ccol2!100}rgb & \cellcolor{ccol2!100}pbr & \cellcolor{ccol2!100}rgb & \cellcolor{ccol2!100}rgb \\
6 & Vidal-Sensors18~\cite{vidal2018method} & 2019 & \cellcolor{ccol2!100}yes & - & - & - & d & \cellcolor{ccol2!100}icp \\
7 & CDPNv2\_BOP20-RGB-ICP~\cite{li2019cdpn} & \cellcolor{ccol2!100}2020 & - & 1/obj & \cellcolor{ccol2!100}rgb & pbr{\tiny +}real & rgb-d & \cellcolor{ccol2!100}icp \\
8 & Drost-CVPR10-Edges~\cite{drost2010model} & 2019 & \cellcolor{ccol2!100}yes & - & - & - & rgb-d & \cellcolor{ccol2!100}icp \\
9 & CDPNv2\_BOP20-PBR-ICP~\cite{li2019cdpn} & \cellcolor{ccol2!100}2020 & - & 1/obj & \cellcolor{ccol2!100}rgb & \cellcolor{ccol2!100}pbr & rgb-d & \cellcolor{ccol2!100}icp \\
10 & CDPNv2\_BOP20-RGB~\cite{li2019cdpn} & \cellcolor{ccol2!100}2020 & - & 1/obj & \cellcolor{ccol2!100}rgb & pbr{\tiny +}real & \cellcolor{ccol2!100}rgb & - \\
11 & Drost-CVPR10-3D-Edges~\cite{drost2010model} & 2019 & \cellcolor{ccol2!100}yes & - & - & - & d & \cellcolor{ccol2!100}icp \\
12 & Drost-CVPR10-3D-Only~\cite{drost2010model} & 2019 & \cellcolor{ccol2!100}yes & - & - & - & d & \cellcolor{ccol2!100}icp \\
13 & CDPN\_BOP19-RGB~\cite{li2019cdpn} & \cellcolor{ccol2!100}2020 & - & 1/obj & \cellcolor{ccol2!100}rgb & pbr{\tiny +}real & \cellcolor{ccol2!100}rgb & - \\
14 & CDPNv2\_BOP20-PBR~\cite{li2019cdpn} & \cellcolor{ccol2!100}2020 & - & 1/obj & \cellcolor{ccol2!100}rgb & \cellcolor{ccol2!100}pbr & \cellcolor{ccol2!100}rgb & - \\
15 & leaping from 2D to 6D~\cite{liu2020leaping} & \cellcolor{ccol2!100}2020 & - & 1/obj & \cellcolor{ccol2!100}rgb & pbr{\tiny +}real & \cellcolor{ccol2!100}rgb & - \\
16 & EPOS-BOP20-PBR~\cite{hodan2020epos} & \cellcolor{ccol2!100}2020 & - & \cellcolor{ccol2!100}1/set & \cellcolor{ccol2!100}rgb & \cellcolor{ccol2!100}pbr & \cellcolor{ccol2!100}rgb & - \\
17 & Drost-CVPR10-3D-Only-Faster~\cite{drost2010model} & 2019 & \cellcolor{ccol2!100}yes & - & - & - & d & \cellcolor{ccol2!100}icp \\
18 & F{\'e}lix\&Neves-ICRA17-IET19~\cite{rodrigues2019deep,raposo2017using} & 2019 & \cellcolor{ccol2!100}yes & \cellcolor{ccol2!100}1/set & rgb-d & syn{\tiny +}real & rgb-d & \cellcolor{ccol2!100}icp \\
19 & Sundermeyer-IJCV19+ICP~\cite{sundermeyer2019augmented} & 2019 & - & 1/obj & \cellcolor{ccol2!100}rgb & syn{\tiny +}real & rgb-d & \cellcolor{ccol2!100}icp \\
20 & Zhigang-CDPN-ICCV19~\cite{li2019cdpn} & 2019 & - & 1/obj & \cellcolor{ccol2!100}rgb & syn{\tiny +}real & \cellcolor{ccol2!100}rgb & - \\
21 & PointVoteNet2~\cite{hagelskjaer2020pointvotenet} & \cellcolor{ccol2!100}2020 & - & 1/obj & rgb-d & \cellcolor{ccol2!100}pbr & rgb-d & \cellcolor{ccol2!100}icp \\
22 & Pix2Pose-BOP20-ICCV19~\cite{park2019pix2pose} & \cellcolor{ccol2!100}2020 & - & 1/obj & \cellcolor{ccol2!100}rgb & pbr{\tiny +}real & \cellcolor{ccol2!100}rgb & - \\
23 & Sundermeyer-IJCV19~\cite{sundermeyer2019augmented} & 2019 & - & 1/obj & \cellcolor{ccol2!100}rgb & syn{\tiny +}real & \cellcolor{ccol2!100}rgb & - \\
24 & SingleMultiPathEncoder-CVPR20~\cite{sundermeyer2020multi} & \cellcolor{ccol2!100}2020 & - & \cellcolor{ccol2!100}1/all & \cellcolor{ccol2!100}rgb & syn{\tiny +}real & \cellcolor{ccol2!100}rgb & - \\
25 & Pix2Pose-BOP19-ICCV19~\cite{park2019pix2pose} & 2019 & - & 1/obj & \cellcolor{ccol2!100}rgb & syn{\tiny +}real & \cellcolor{ccol2!100}rgb & - \\
26 & DPOD (synthetic)~\cite{zakharov2019dpod} & 2019 & - & \cellcolor{ccol2!100}1/scene & \cellcolor{ccol2!100}rgb & syn & \cellcolor{ccol2!100}rgb & - \\

\bottomrule
\end{tabularx}

\caption{\label{tab:bop_2020_results} \textbf{Results of the BOP Challenge 2019 and 2020.} The methods are ranked by the $\text{AR}_{\text{Core}}$ score (the third column of the upper table) which is the average of the per-dataset $\text{AR}_D$ scores (the following seven columns). The scores (in \%) are defined in Section~\ref{sec:bop_accuracy_score}. The last column of the upper table shows the average image processing time averaged over the datasets (in seconds). The lower table shows properties discussed in Sections~\ref{sec:bop20_results} and \ref{sec:bop20_rgb_types}.}
\vspace{-8ex}

\end{center}

\end{table}

\endgroup

\subsection{Results} \label{sec:bop20_results}

In 2020, methods based on deep neural networks finally caught up with methods based on point pair features -- four DNN-based and one DNN+PPF-based method from 2020 outperform the method of Vidal \etal~\cite{vidal2018method}, the PPF-based winner of the challenges from 2017 and 2019 (columns PPF and DNN in Table~\ref{tab:bop_2020_results}).
The neural networks are in all but one case applied only to the RGB channels, while the depth channel is often used for an ICP refinement
(columns Train, Test, Refine).
Only the PointVoteNet2 method~\cite{hagelskjaer2020pointvotenet} applies a neural network to the RGB and depth channels.
It is noteworthy that the overall third DNN-based method does not use the depth channel at all.

Three of the top five methods, including the top-performing one, are single-view variants of the CosyPose method.
The top variant,
with the $\text{AR}_{\text{Core}}$ score of $69.8\%$, additionally applies a depth-based ICP refinement which improves the score by $6.1\%$ (method \#1 \vs \#3 in Table~\ref{tab:bop_2020_results}).
One of the key ingredients of CosyPose is a strong data augmentation technique similar to~\cite{sundermeyer2018implicit}. As reported in~\cite{labbe2020cosypose}, using the augmentation for training the pose estimation models improved the
accuracy on T-LESS from $37.0\%$ to $63.8\%$. Access to a GPU cluster was also crucial as training of one network took $\mytilde$10 hours on 32 GPU's.

The second is a hybrid method by K\"onig and Drost with $\text{AR}_{\text{Core}}$ of $63.9\%$.
This method is noticeably faster than the top-performing CosyPose variant, mainly thanks to a highly optimized implementation of the ICP algorithm from HALCON~\cite{halcon}.

Another method that outperformed Vidal \etal is Pix2Pose with
$\text{AR}_{\text{Core}}$
of $59.1\%$.
The ICP refinement is crucial for this method as it improves the
score by an absolute $24.9\%$ and teleports the method from the 22nd to the 4th place. The importance of refinement was also demonstrated by other methods -- top nine methods applied ICP or an RGB-based refiner, similar to DeepIM~\cite{li2018deepim} (column Refine in Table~\ref{tab:bop_2020_results}).

Training a special DNN model per object has been a common practice in the field, also followed by most participants of the 2020 edition of the challenge. However, the CosyPose and K\"onig-Hybrid
methods showed that a single DNN model can be effectively shared among multiple objects (column DNN in Table~\ref{tab:bop_2020_results}). CosyPose trains three models per dataset -- one for detection, one for coarse pose estimation, and one for iterative pose refinement, whereas K\"onig-Hybrid %
trains only one model for instance segmentation.

Overall, the EPOS method trained on the PBR images placed 16th.
The better performing methods applied an RGB-based or depth-based post-refinement of the estimated poses, used real training images, trained one neural network per object instead of one network per dataset, or applied strong data augmentation.
As discussed in Chapter~\ref{ch:method_epos}, EPOS would likely benefit from these design choices as well.

\subsection{The Effectiveness of Photorealistic Training Images} \label{sec:bop20_rgb_types}

In 2020, most DNN-based methods were trained either only on the photorealistic (PBR) training images, or also on real training images which are available in datasets T-LESS, TUD-L, and YCB-V (column Train type in Table~\ref{tab:bop_2020_results})\footnote{Method \#2 used also synthetic training images obtained by cropping the objects from real validation images in the case of HB and ITODD and from OpenGL-rendered images in the case of other datasets, and pasting the cropped objects on images from Microsoft COCO~\cite{lin2014microsoft}. Method \#24 used PBR and real images for training Mask R-CNN~\cite{he2017mask} and OpenGL images for training a single Multi-path encoder. Two of the CosyPose variants (\#1 and \#3) also added the ``render\;\&\;paste'' synthetic images provided in the original YCB-V dataset, but these images were later found to have no effect on the accuracy score.}.
Although adding real training images yields higher scores (compare scores of methods \#3 and \#5 or \#10 and \#14 on T-LESS, TUD-L, and YCB-V in Table~\ref{tab:bop_2020_results}), competitive results can be achieved with PBR images only, as demonstrated by the overall fifth PBR-only variant of the CosyPose method. This is an important result considering that PBR-only training does not require any human effort for capturing and annotating real training images.

The PBR training images yield a noticeable improvement over the ``render\;\& paste''
synthetic images obtained by OpenGL rendering of the 3D object models on real photographs.
For example, the CDPN method with the same hyper-parameter settings improved by absolute $20.2\%$ on HB, by $19.5\%$ on LM-O, and by $7\%$ on IC-BIN when trained on 50K PBR images per dataset \vs 10K ``render\;\&\;paste'' images per object (compare methods \#13 and \#20 in Table~\ref{tab:bop_2020_results}).
As shown in Table~\ref{tab:cosy_data_res}, the CosyPose method improved by a significant $57.9\%$ (from $6.1\%$ to $64.0\%$) on T-LESS, by $19.0\%$ on TUD-L, and by $30.9\%$ on YCB-V when trained on 50K PBR images per dataset \vs 50K ``render\;\&\;paste v1'' images per dataset.
The ``render\;\&\;paste v1'' images used for training CosyPose were obtained by imitating the PBR images, \ie, the 3D object models were rendered in the same poses as in the PBR images and pasted on real backgrounds.

\begin{table}[t]
\centering

\begingroup

\setlength{\tabcolsep}{6pt} %
\renewcommand{\arraystretch}{1.0} %

\footnotesize
\begin{center}
\begin{tabularx}{0.85\textwidth}{ c c *{3}{Y} }
	\toprule
	Detection & Pose estim. & T-LESS & TUD-L & YCB-V \\
	\midrule
	PBR+Real & PBR+Real & 72.8 & 82.3 & 82.1 \\
	PBR & PBR & 64.0 & 68.5 & 57.4 \\
	PBR & Render\,\&\,paste v1 & 16.1 & 60.4 & 44.9 \\
	PBR & Render\,\&\,paste v2 & 60.0 & 58.9 & 58.5 \\
	Render\,\&\,paste v1 & Render\,\&\,paste v1 & 6.1 & 49.5 & 26.5 \\
	Render\,\&\,paste v2 & Render\,\&\,paste v2 & 45.3 & 42.4 & 25.7 \\
	\bottomrule
\end{tabularx}
\caption{\label{tab:cosy_data_res} \textbf{The effect of different training images}. Shown are the $\text{AR}_{\text{Core}}$ scores achieved by the CosyPose method~\cite{labbe2020cosypose} when different types of images were used for training its object detection (\ie Mask R-CNN~\cite{he2017mask}) and pose estimation stage.
The ``render\;\&\;paste v1'' images were obtained by OpenGL rendering of the 3D object models on random real photographs.
The ``render\;\&\;paste v2'' images were obtained similarly, but the CAD models of T-LESS objects were assigned a random surface texture instead of a random gray value, the background of most images was assigned a synthetic texture, and 1M instead of 50K images were generated.
Interestingly, the increased photorealism brought by the PBR images yields noticeable improvements despite the strong data augmentation applied by CosyPose to the training images.
}
\end{center}
\endgroup
\end{table}

As an additional experiment, we trained the CosyPose method on another variant of the ``render\;\&\;paste'' images, generated as in~\cite{labbe2020cosypose} and referred to as ``render\;\&\;paste v2''. The main differences compared to the ``render\;\&\;paste v1'' variant described in the previous paragraph are: (a) the CAD models of T-LESS objects were assigned a random surface texture instead of a random gray value, (b) the background was assigned a real photograph in 30\% images and a synthetic texture in 70\% images, and (c) 1M instead of 50K images were generated.
As shown in Table~\ref{tab:cosy_data_res}, ``render\;\&\;paste v2'' images yield a noticeable improvement of $39.2\%$ over ``render\;\&\;paste v1'' on T-LESS, but no improvement on TUD-L ($-7.1\%$) and YCB-V ($-0.8\%$).
This may suggest that randomizing the surface texture of the texture-less CAD models of T-LESS objects improves the generalization of the network by forcing the network to focus more on shape than on lower-level patterns, as found in~\cite{geirhos2018imagenet}.
When generating the PBR images, which yield the highest accuracy on T-LESS, the CAD models were assigned a random gray value, as in ``render\;\&\;paste v1'', but the effect of randomizing the surface texture may have been achieved by randomizing the PBR material (Section~\ref{sec:synth_bop20}) -- further investigation is needed to clearly answer these questions.
The importance of both the objects and the background being synthetic, as suggested in~\cite{hinterstoisser2019annotation}, was not confirmed in this experiment -- ``render\;\&\;paste v1'' images with only real backgrounds achieved higher scores than ``render\;\&\;paste v2'' images on TUD-L and YCB-V.
However, the first ten convolutional layers of Mask R-CNN (``conv1'' and ``conv2\_x'' of ResNet-50~\cite{he2016deep}) used for object detection in the CosyPose method were pre-trained on Microsoft COCO~\cite{lin2014microsoft} but not fine-tuned, whereas all layers were fine-tuned in~\cite{hinterstoisser2019annotation}.
The benefit of having 1M \vs 50K images is indecisive since 50K PBR images were sufficient to achieve high scores.

Both types of ``render\;\&\;paste'' images are far inferior compared to the PBR images, which yield an average improvement of $35.9\%$ over ``render\;\&\;paste v1'' and $25.5\%$ over ``render\;\&\;paste v2'' images (Table~\ref{tab:cosy_data_res}).
Interestingly, the increased photorealism brought by the PBR images is important despite the strong data augmentation which the CosyPose method applies to the training images. Since the poses of objects in PBR and ``render\;\&\;paste v1'' images are identical, the ray-tracing rendering technique, PBR materials and objects realistically embedded in synthetic environments seem to be the decisive factors for successful ``sim2real'' transfer~\cite{denninger2020blenderproc}.

We also observed that the PBR images are more important for training DNN models for object detection/segmentation (\eg, Mask R-CNN~\cite{he2017mask}) than for training DNN models for pose estimation from the detected regions (Table~\ref{tab:cosy_data_res}). In the case of CosyPose, if the detection model is trained on PBR images and the later two models for pose estimation are trained on the ``render\;\&\;paste v2'' instead of the PBR images, the accuracy drops moderately ($64.0\%$ to $60.0\%$ on T-LESS, $68.5\%$ to $58.9\%$ on TUD-L) or does not change much ($57.4\%$ \vs $58.5\%$ on YCB-V). However, if also the detection model is trained on the ``render\;\&\;paste v1'' or ``render\;\&\;paste v2'' images, the accuracy drops severely (the low accuracy achieved with ``render\;\&\;paste v1'' on T-LESS was discussed earlier).

	\chapter{Conclusion} \label{ch:conclusion}

We addressed the problem of estimating the 6D pose of specific rigid objects from a single RGB or RGB-D input image. We primarily focused on the problem of 6D object localization, where the identifiers of visible object instances are provided together with the input image, and assumed that the 3D object models are available.

Starting on the shoulders of giants who had worked on
this problem for decades, we made
contributions that address limitations of the existing methods, enable effective training of learning-based methods without
capturing
any real images, introduce an industry-relevant dataset with new challenges, and set a standard for benchmarking.

First, we proposed EPOS, a method based on a deep neural network that predicts 2D-3D correspondences between densely sampled pixels of the input RGB image and 3D locations on the object model, and solves for the object poses using an efficient variant of the P\emph{n}P-RANSAC algorithm.
The key idea is to represent an object by a controllable number of compact surface fragments, which allows handling object symmetries by predicting multiple potential 2D-3D correspondences at each pixel and ensures a consistent number and uniform coverage of candidate 3D locations on objects of any type. In the BOP Challenge 2019, EPOS outperformed all RGB and most RGB-D and D methods on the T-LESS and LM-O datasets. On the YCB-V dataset, it was superior to all competitors, with a large margin over the second-best RGB method.

Second, we presented HashMatch, an RGB-D template matching method that applies a cascade of evaluation stages to each sliding window location, which avoids exhaustive matching against all templates.
The key stage of the cascade is a voting procedure based on hashing, which substantially reduces the number of candidate templates (usually by three orders of magnitude) before a more expensive template verification stage, and makes the time complexity of the method largely unaffected by the number of templates. The method was fourth out of the 15 participants in the BOP Challenge 2017 and was successfully deployed in DARWIN, a robotic assembly project.

Third, we presented ObjectSynth, an approach to synthesize photorealistic training images of 3D object models.
The 3D models of objects are arranged in 3D models of complete indoor scenes with realistic materials and lighting, plausible geometric configurations of objects and cameras are generated by physics simulation, and a high degree of visual realism is achieved by physically-based rendering (PBR).
The generated images were shown to yield significant improvements in 2D object detection and 6D object pose estimation, compared to the commonly used training images synthesized by rendering object models on top of random photographs.
A refined version of the proposed synthesis approach was implemented in BlenderProc4BOP, a new open-source and light-weight physically-based renderer and procedural data generator. BlenderProc4BOP was used to synthesize photorealistic training images for the participants of the BOP Challenge 2020. Although adding real training images yielded even higher scores, competitive results were achieved with the photorealistic training images only -- out of the 26 evaluated methods, the fifth method was trained only on the photorealistic images.

Fourth, we created T-LESS, a publicly available dataset for training and testing methods for 6D object pose estimation. The dataset includes 3D models and training and test RGB-D images of thirty commodity electrical parts. Recognition and pose estimation of these objects is challenging because the objects have no significant texture or discriminative color, exhibit symmetries and similarities in shape and size, and some objects are a composition of others.
Objects exhibiting similar properties are common in industrial environments and T-LESS is the first dataset to include such objects.

Fifth, we lead the BOP benchmark with the goal to capture the status quo in the field and systematically measure the progress in the future. The benchmark currently comprises of eleven datasets in a unified format which cover various practical scenarios, an evaluation methodology with three new pose-error functions which address limitations of the previously used functions, an online evaluation system which is open for continuous submission of new results and reports the current state of the art, and public challenges held at the International Workshops on Recovering 6D Object Pose which we organize annually at the ICCV and ECCV conferences.
The recent BOP Challenge 2020 showed that methods based on neural networks finally caught up with methods based on point pair features, which were dominating previous editions of the challenge. Although the top-performing methods rely on RGB-D image channels, strong results were achieved with RGB channels only, also thanks to the provided photorealistic training images.

As research
never ends, other interesting problems in the field of object pose estimation remain open, with methods tackling some of them starting to emerge. Examples include pose estimation of piece-wise rigid objects~\cite{michel2015pose}, flexible objects~\cite{alp2018densepose},
or object categories~\cite{wang2019normalized,tian2020shape,chen2020learning,chen2020category}, and learning to recognize objects in a model-free~\cite{cai2020reconstruct,park2020latentfusion,park2020neural}, few-shot~\cite{park2020latentfusion,park2020neural}, weakly-supervised~\cite{shao2020pfrl}, or a self-supervised manner~\cite{wang2020self6d,sock2020introducing}.

	\backmatter
	
	\begingroup
	\pretolerance=-1
	\tolerance=400
	\spacing{1.0}
	
	\addcontentsline{toc}{chapter}{Bibliography}
	\printbibliography[notcategory=fullcited]

@article{hodan2015efficient,
  title={Efficient Texture-less Object Detection for Augmented Reality Guidance},
  author={Hoda{\v{n}}, Tom{\'a}{\v{s}} and Damen, Dima and Mayol-Cuevas, Walterio and Matas, Ji{\v{r}}{\'\i}},
  journal={ISMARW},
  year={2015}
}

@article{hodan2015detection,
  title={Detection and Fine {3D} Pose Estimation of Texture-less Objects in {RGB-D} Images},
  author={Hoda{\v{n}}, Tom{\'a}{\v{s}} and Zabulis, Xenophon and Lourakis, Manolis and Obdr{\v{z}}{\'a}lek, {\v{S}}t{\v{e}}p{\'a}n and Matas, Ji{\v{r}}{\'\i}},
  journal={IROS},
  year={2015}
}

@article{hodan2016evaluation,
  title={On Evaluation of {6D} Object Pose Estimation},
  author={Hoda{\v{n}}, Tom{\'a}{\v{s}} and Matas, Ji{\v{r}}{\'\i} and Obdr{\v{z}}{\'a}lek, {\v{S}}t{\v{e}}p{\'a}n},
  journal={ECCVW},
  year={2016}
}

@article{hodan2017tless,
  title={{T-LESS}: An {RGB-D} Dataset for {6D} Pose Estimation of Texture-less Objects},
  author={Hoda{\v{n}}, Tom{\'a}{\v{s}} and Haluza, Pavel and Obdr{\v{z}}{\'a}lek, {\v{S}}t{\v{e}}p{\'a}n and Matas, Ji{\v{r}}{\'\i} and Lourakis, Manolis and Zabulis, Xenophon},
  journal={WACV},
  year={2017}
}

@article{hodan2018bop,
  title={{BOP}: Benchmark for {6D} Object Pose Estimation},
  author={Hoda{\v{n}}, Tom{\'a}{\v{s}} and Michel, Frank and Brachmann, Eric and Kehl, Wadim and Glent Buch, Anders and Kraft, Dirk and Drost, Bertram and Vidal, Joel and Ihrke, Stephan and Zabulis, Xenophon and Sahin, Caner and Manhardt, Fabian and Tombari, Federico and Kim, Tae-Kyun and Matas, Ji{\v{r}}{\'i} and Rother, Carsten},
  journal={ECCV},
  year={2018}
}

@article{hodan2020bop,
  title={{BOP} Challenge 2020 on {6D} Object Localization},
  author={Hoda{\v{n}}, Tom{\'a}{\v{s}} and Sundermeyer, Martin and Drost, Bertram and Labb{\'e}, Yann and Brachmann, Eric and Michel, Frank and Rother, Carsten and Matas, Ji{\v{r}}{\'i}},
  journal={ECCVW},
  year={2020}
}

@article{hodan2018workshop,
  title={A Summary of the 4th International Workshop on Recovering {6D} Object Pose},
  author={Hoda{\v{n}}, Tom{\'a}{\v{s}} and Kouskouridas, Rigas and Kim, Tae-Kyun and Tombari, Federico and Bekris, Kostas and Drost, Bertram and Groueix, Thibault and Walas, Krzysztof and Lepetit, Vincent and Leonardis, Ales and Steger, Carsten and Michel, Frank and Sahin, Caner and Rother, Carsten and Matas, Ji{\v{r}}{\'i}},
  journal={ECCVW},
  year={2018}
}

@article{hodan2019photorealistic,
  title={Photorealistic Image Synthesis for Object Instance Detection},
  author={Hoda{\v{n}}, Tom{\'a}{\v{s}} and Vineet, Vibhav and Gal, Ran and Shalev, Emanuel and Hanzelka, Jon and Connell, Treb and Urbina, Pedro and Sinha, Sudipta and  Guenter, Brian},
  journal={ICIP},
  year={2019}
}

@article{hodan2020epos,
  title={{EPOS}: Estimating {6D} Pose of Objects with Symmetries},
  author={Hoda{\v{n}}, Tom{\'a}{\v{s}} and Bar{\'a}th, D{\'a}niel and Matas, Ji{\v{r}}{\'i}},
  journal={CVPR},
  year={2020}
}

@article{patel2020learning,
  title={Learning Surrogates via Deep Embedding},
  author={Patel, Yash and Hoda{\v{n}}, Tom{\'a}{\v{s}} and Matas, Ji{\v{r}}{\'i}},
  journal={ECCV},
  year={2020}
}

@article{denninger2020blenderproc,
  title={{BlenderProc:} Reducing the Reality Gap with Photorealistic Rendering},
  author={Denninger, Maximilian and Sundermeyer, Martin and Winkelbauer, Dominik and Olefir, Dmitry and Hoda{\v{n}}, Tom{\'a}{\v{s}} and Zidan, Youssef and Elbadrawy, Mohamad and Knauer, Markus and Katam, Harinandan and Lodhi, Ahsan},
  journal={RSS Workshops},
  year={2020}
}

@book{prince2012computer,
	title={Computer vision: models, learning, and inference},
	author={Prince, Simon JD},
	year={2012},
	publisher={Cambridge University Press}
}

@phdthesis{roberts1963machine,
  title={Machine perception of three-dimensional solids},
  author={Roberts, Lawrence G},
  year={1963},
  school={Massachusetts Institute of Technology}
}

@article{lowe1999object,
  title={Object recognition from local scale-invariant features.},
  author={Lowe, David G and others},
  journal={ICCV},
  year={1999}
}

@article{collet2011moped,
  title={The {MOPED} framework: {O}bject recognition and pose estimation for manipulation},
  author={Collet, Alvaro and Martinez, Manuel and Srinivasa, Siddhartha S},
  journal={IJRR},
  volume={},
  number={},
  pages={},
  year={2011},
  publisher={}
}

@article{brunelli2009template,
  title={Template matching techniques in computer vision: {T}heory and practice},
  author={Brunelli, Roberto},
  year={2009},
  journal={John Wiley \& Sons}
}

@article{hinterstoisser2011linemod,
  author  =  {Hinterstoisser, S. Holzer, S. and Cagniart, C. and Ilic, S. and Konolige, K. and Navab, N. and Lepetit, V.},
  title   =  {Multimodal Templates for Real-Time Detection of Texture-less Objects in Heavily Cluttered Scenes},
  journal =  {ICCV},
  year    =  {2011}
}

@article{hinterstoisser2012accv,
  author  =  {Hinterstoisser, S. and Lepetit, V. and Ilic, S. and Holzer, S. and Bradski, G. and Konolige, K. and Navab, N.},
  title   =  {Model Based Training, Detection and Pose Estimation of Texture-Less 3{D} Objects in Heavily Cluttered Scenes},
  journal =  {ACCV},
  year    =  {2012}
}

@article{hinterstoisser2012pami,
  author  =  {Hinterstoisser, S. and Cagniart, C. and Ilic, S. and Sturm, P. and Navab, N. and Fua, P. and Lepetit, V. },
  title   =  {Gradient Response Maps for Real-Time Detection of Texture-Less Objects},
  journal =  {TPAMI},
  year    =  {2012},
}

@article{drost2010model,
  title={Model globally, match locally: {E}fficient and robust {3D} object recognition},
  author={Drost, Bertram and Ulrich, Markus and Navab, Nassir and Ilic, Slobodan},
  journal={CVPR},
  pages={},
  year={2010},
  organization={}
}

@article{hinterstoisser2016going,
  title={Going further with point pair features},
  author={Hinterstoisser, Stefan and Lepetit, Vincent and Rajkumar, Naresh and Konolige, Kurt},
  journal={ECCV},
  pages={},
  year={2016},
  organization={}
}

@article{vidal2018method,
  title={A Method for {6D} Pose Estimation of Free-Form Rigid Objects Using Point Pair Features on Range Data},
  author={Vidal, Joel and Lin, Chyi-Yeu and Llad{\'o}, Xavier and Mart{\'\i}, Robert},
  journal={Sensors},
  volume={},
  number={},
  pages={},
  year={2018},
  publisher={}
}

@article{guo2016comprehensive,
  title={A comprehensive performance evaluation of {3D} local feature descriptors},
  author={Guo, Yulan and Bennamoun, Mohammed and Sohel, Ferdous and Lu, Min and Wan, Jianwei and Kwok, Ngai Ming},
  journal={IJCV},
  year={2016}
}

@article{brachmann2014learning,
  title={Learning {6D} object pose estimation using {3D} object coordinates},
  author={Brachmann, Eric and Krull, Alexander and Michel, Frank and Gumhold, Stefan and Shotton, Jamie and Rother, Carsten},
  journal={ECCV},
  pages={},
  year={2014},
  organization={}
}

@article{brachmann2016uncertainty,
  title={Uncertainty-Driven {6D} Pose Estimation of Objects and Scenes from a Single {RGB} Image},
  author={Brachmann, Eric and Michel, Frank and Krull, Alexander and Yang, Michael Ying and Gumhold, Stefan and Rother, Carsten},
  journal={CVPR},
  year={2016},
}

@article{krull2015learning,
  title={Learning analysis-by-synthesis for {6D} pose estimation in {RGB-D} images},
  author={Krull, Alexander and Brachmann, Eric and Michel, Frank and Ying Yang, Michael and Gumhold, Stefan and Rother, Carsten},
  journal={ICCV},
  pages={},
  year={2015}
}

@article{rad2017bb8,
  title={{BB8:} {A} scalable, accurate, robust to partial occlusion method for predicting the {3D} poses of challenging objects without using depth},
  author={Rad, Mahdi and Lepetit, Vincent},
  journal={ICCV},
  year={2017}
}

@article{pavlakos20176,
  title={{6-DoF} object pose from semantic keypoints},
  author={Pavlakos, Georgios and Zhou, Xiaowei and Chan, Aaron and Derpanis, Konstantinos G and Daniilidis, Kostas},
  journal={ICRA},
  pages={},
  year={2017},
  organization={}
}

@article{oberweger2018making,
  title={Making deep heatmaps robust to partial occlusions for {3D} object pose estimation},
  author={Oberweger, Markus and Rad, Mahdi and Lepetit, Vincent},
  journal={ECCV},
  pages={},
  year={2018}
}

@article{tekin2018real,
  title={Real-time seamless single shot {6D} object pose prediction},
  author={Tekin, Bugra and Sinha, Sudipta N and Fua, Pascal},
  journal={CVPR},
  pages={},
  year={2018}
}

@article{tremblay2018deep,
  title={Deep object pose estimation for semantic robotic grasping of household objects},
  author={Tremblay, Jonathan and To, Thang and Sundaralingam, Balakumar and Xiang, Yu and Fox, Dieter and Birchfield, Stan},
  journal={CoRL},
  year={2018}
}

@article{tremblay2018training,
  title={Training deep networks with synthetic data: Bridging the reality gap by domain randomization},
  author={Tremblay, Jonathan and Prakash, Aayush and Acuna, David and Brophy, Mark and Jampani, Varun and Anil, Cem and To, Thang and Cameracci, Eric and Boochoon, Shaad and Birchfield, Stan},
  journal={CVPRW},
  year={2018}
}

@article{fu2019deephmap++,
  title={{DeepHMap++:} {C}ombined Projection Grouping and Correspondence Learning for Full {DoF} Pose Estimation},
  author={Fu, Mingliang and Zhou, Weijia},
  journal={Sensors},
  year={2019}
}

@article{park2020latentfusion,
	title={{LatentFusion}: End-to-End Differentiable Reconstruction and Rendering for Unseen Object Pose Estimation},
	author={Park, Keunhong and Mousavian, Arsalan and Xiang, Yu and Fox, Dieter},
	journal={CVPR},
	year={2020}
}

@article{redmon2016you,
	title={You only look once: Unified, real-time object detection},
	author={Redmon, Joseph and Divvala, Santosh and Girshick, Ross and Farhadi, Ali},
	journal={CVPR},
	year={2016}
}

@article{liu2016ssd,
	title={{SSD}: Single shot multibox detector},
	author={Liu, Wei and Anguelov, Dragomir and Erhan, Dumitru and Szegedy, Christian and Reed, Scott and Fu, Cheng-Yang and Berg, Alexander C},
	journal={ECCV},
	year={2016}
}

@article{hu2019segmentation,
  title={Segmentation-driven {6D} object pose estimation},
  author={Hu, Yinlin and Hugonot, Joachim and Fua, Pascal and Salzmann, Mathieu},
  journal={CVPR},
  pages={},
  year={2019}
}

@article{peng2019pvnet,
  title={{PVNet:} {P}ixel-wise Voting Network for {6DoF} Pose Estimation},
  author={Peng, Sida and Liu, Yuan and Huang, Qixing and Zhou, Xiaowei and Bao, Hujun},
  journal={CVPR},
  pages={},
  year={2019}
}

@article{jafari2018ipose,
  title={{iPose:} {I}nstance-aware {6D} pose estimation of partly occluded objects},
  author={Jafari, Omid Hosseini and Mustikovela, Siva Karthik and Pertsch, Karl and Brachmann, Eric and Rother, Carsten},
  journal={ACCV},
  pages={},
  year={2018},
  organization={}
}

@article{nigam2018detect,
  title={Detect Globally, Label Locally: {L}earning Accurate {6-DOF} Object Pose Estimation by Joint Segmentation and Coordinate Regression},
  author={Nigam, Apurv and Penate-Sanchez, Adrian and Agapito, Lourdes},
  journal={RAL},
  volume={},
  number={},
  pages={},
  year={2018},
  publisher={}
}

@article{zakharov2019dpod,
  title={{DPOD:} {6D} Pose Object Detector and Refiner},
  author={Zakharov, Sergey and Shugurov, Ivan and Ilic, Slobodan},
  journal={ICCV},
  year={2019}
}

@article{park2019pix2pose,
  title={{Pix2Pose:} {P}ixel-Wise Coordinate Regression of Objects for {6D} Pose Estimation},
  author={Park, Kiru and Patten, Timothy and Vincze, Markus},
  journal={ICCV},
  pages={},
  year={2019}
}

@article{li2019cdpn,
  title={{CDPN:} {C}oordinates-Based Disentangled Pose Network for Real-Time {RGB}-Based {6-DoF} Object Pose Estimation},
  author={Li, Zhigang and Wang, Gu and Ji, Xiangyang},
  journal={ICCV},
  pages={},
  year={2019}
}

@article{xiang2017posecnn,
  title={{PoseCNN:} {A} convolutional neural network for {6D} object pose estimation in cluttered scenes},
  author={Xiang, Yu and Schmidt, Tanner and Narayanan, Venkatraman and Fox, Dieter},
  journal={RSS},
  year={2018}
}

@article{calli2015ycb,
	title={The {YCB} object and model set: Towards common benchmarks for manipulation research},
	author={Calli, Berk and Singh, Arjun and Walsman, Aaron and Srinivasa, Siddhartha and Abbeel, Pieter and Dollar, Aaron M},
	journal={ICAR},
	year={2015},
}

@article{li2018unified,
  title={A unified framework for multi-view multi-class object pose estimation},
  author={Li, Chi and Bai, Jin and Hager, Gregory D},
  journal={ECCV},
  pages={},
  year={2018}
}

@article{mahendran2018mixed,
	title={A mixed classification-regression framework for {3D} pose estimation from {2D} images},
	author={Mahendran, Siddharth and Ali, Haider and Vidal, Rene},
	journal={BMVC},
	year={2018}
}

@article{manhardt2019explaining,
  title={Explaining the Ambiguity of Object Detection and {6D} Pose from Visual Data},
  author={Manhardt, Fabian and Arroyo, Diego Martin and Rupprecht, Christian and Busam, Benjamin and Navab, Nassir and Tombari, Federico},
  journal={ICCV},
  year={2019}
}

@article{bui2017x,
	title={{X-Ray PoseNet}: 6 {DoF} pose estimation for mobile {X-Ray} devices},
	author={Bui, Mai and Albarqouni, Shadi and Schrapp, Michael and Navab, Nassir and Ilic, Slobodan},
	journal={WACV},
	year={2017}
}

@article{kendall2015posenet,
	title={{PoseNet:} A convolutional network for real-time {6-DOF} camera relocalization},
	author={Kendall, Alex and Grimes, Matthew and Cipolla, Roberto},
	journal={ICCV},
	year={2015}
}

@article{kehl2017ssd,
  title={{SSD-6D:} {M}aking {RGB}-based {3D} detection and {6D} pose estimation great again},
  author={Kehl, Wadim and Manhardt, Fabian and Tombari, Federico and Ilic, Slobodan and Navab, Nassir},
  journal={ICCV},
  pages={},
  year={2017}
}

@article{corona2018pose,
  title={Pose estimation for objects with rotational symmetry},
  author={Corona, Enric and Kundu, Kaustav and Fidler, Sanja},
  journal={IROS},
  pages={},
  year={2018},
  organization={}
}

@article{sundermeyer2019augmented,
  title={Augmented Autoencoders: {I}mplicit {3D} Orientation Learning for {6D} Object Detection},
  author={Sundermeyer, Martin and Marton, Zoltan-Csaba and Durner, Maximilian and Triebel, Rudolph},
  journal={IJCV},
  pages={},
  year={2019},
  publisher={}
}

@misc{bop19challenge,
  title={{BOP} {C}hallenge 2019},
  author={Hoda{\v{n}}, Tom{\'a}{\v{s}} and Brachmann, Eric and Drost, Bertram and Michel, Frank and Sundermeyer, Martin and Matas, Ji{\v{r}}{\'\i} and Rother, Carsten},
  year={},
  howpublished = {\url{https://bop.felk.cvut.cz/challenges/bop-challenge-2019/}}
}

@article{sock2018multi,
  title={Multi-task deep networks for depth-based {6D} object pose and joint registration in crowd scenarios},
  author={Sock, Juil and Kim, Kwang In and Sahin, Caner and Kim, Tae-Kyun},
  journal={BMVC},
  year={2018}
}

@article{wang2019densefusion,
  title={{DenseFusion:} {6D} object pose estimation by iterative dense fusion},
  author={Wang, Chen and Xu, Danfei and Zhu, Yuke and Mart{\'\i}n-Mart{\'\i}n, Roberto and Lu, Cewu and Fei-Fei, Li and Savarese, Silvio},
  journal={CVPR},
  pages={},
  year={2019}
}

@article{pitteri2019object,
  title={On Object Symmetries and {6D} Pose Estimation from Images},
  author={Pitteri, Giorgia and Ramamonjisoa, Micha{\"e}l and Ilic, Slobodan and Lepetit, Vincent},
  journal={3DV},
  pages={},
  year={2019},
  organization={}
}

@article{crivellaro2017robust,
  title={Robust {3D} object tracking from monocular images using stable parts},
  author={Crivellaro, Alberto and Rad, Mahdi and Verdie, Yannick and Yi, Kwang Moo and Fua, Pascal and Lepetit, Vincent},
  journal={TPAMI},
  volume={},
  number={},
  pages={},
  year={2017},
  publisher={}
}

@article{alp2018densepose,
  title={{DensePose:} {D}ense human pose estimation in the wild},
  author={Alp G{\"u}ler, R{\i}za and Neverova, Natalia and Kokkinos, Iasonas},
  journal={CVPR},
  pages={},
  year={2018}
}

@article{drost2017introducing,
  title={Introducing {MVTec} {ITODD} -- {A} dataset for {3D} object recognition in industry},
  author={Drost, Bertram and Ulrich, Markus and Bergmann, Paul and Hartinger, Philipp and Steger, Carsten},
  journal={ICCVW},
  pages={},
  year={2017}
}

@article{fischler1981random,
  title={Random sample consensus: {A} paradigm for model fitting with applications to image analysis and automated cartography},
  author={Fischler, M. A. and Bolles, R. C.},
  journal={Communications of the ACM},
  year={1981},
  publisher={}
}

@article{lepetit2009epnp,
  title={{EPnP:} {A}n Accurate {O(n)} Solution to the {PnP} Problem},
  author={Lepetit, Vincent and Moreno-Noguer, Francesc and Fua, Pascal},
  journal={IJCV},
  volume={},
  number={},
  pages={},
  year={2009},
  publisher={}
}

@article{mitra2006partial,
  title={Partial and approximate symmetry detection for {3D} geometry},
  author={Mitra, Niloy J and Guibas, Leonidas J and Pauly, Mark},
  journal={ACM Transactions on Graphics},
  volume={},
  number={},
  pages={},
  year={2006},
  organization={}
}

@article{wang2019normalized,
	title={Normalized object coordinate space for category-level {6D} object pose and size estimation},
	author={Wang, He and Sridhar, Srinath and Huang, Jingwei and Valentin, Julien and Song, Shuran and Guibas, Leonidas J},
	journal={CVPR},
	year={2019}
}

@article{tombari2013bold,
  title={{BOLD} features to detect texture-less objects},
  author={Tombari, Federico and Franchi, Alessandro and Di Stefano, Luigi},
  journal={ICCV},
  pages={},
  year={2013}
}

@article{barath2019progx,
  title={Progressive-{X}: {E}fficient, Anytime, Multi-Model Fitting Algorithm},
  author={Bar{\'a}th, D{\'a}niel and Matas, Ji{\v{r}}{\'\i}},
  journal={ICCV},
  volume={},
  number={},
  pages={},
  year={2019}
}

@article{barath2018gcransac,
  title={Graph-{C}ut {RANSAC}},
  author={Bar{\'a}th, D{\'a}niel and Matas, Ji{\v{r}}{\'\i}},
  journal={CVPR},
  volume={},
  number={},
  year={2018}
}

@article{chum2005matching,
  title={Matching with {PROSAC} -- {P}rogressive sample consensus},
  author={Chum, Ondrej and Matas, Ji{\v{r}}{\'\i}},
  journal={CVPR)},
  volume={},
  pages={},
  year={2005},
  organization={}
}

@article{manhardt2018deep,
  title={Deep model-based {6D} pose refinement in {RGB}},
  author={Manhardt, Fabian and Kehl, Wadim and Navab, Nassir and Tombari, Federico},
  journal={ECCV},
  pages={},
  year={2018}
}

@article{li2018deepim,
  title={{DeepIM:} {D}eep iterative matching for {6D} pose estimation},
  author={Li, Yi and Wang, Gu and Ji, Xiangyang and Xiang, Yu and Fox, Dieter},
  journal={ECCV},
  pages={},
  year={2018}
}

@article{chen2018encoder,
  title={Encoder-decoder with atrous separable convolution for semantic image segmentation},
  author={Chen, Liang-Chieh and Zhu, Yukun and Papandreou, George and Schroff, Florian and Adam, Hartwig},
  journal={ECCV},
  pages={},
  year={2018}
}

@incollection{huber1992robust,
  title={Robust estimation of a location parameter},
  author={Huber, Peter J},
  booktitle={Breakthroughs in statistics},
  pages={},
  year={1992},
  publisher={}
}

@book{goodfellow2016deep,
  title={Deep learning},
  author={Goodfellow, Ian and Bengio, Yoshua and Courville, Aaron},
  year={2016},
  publisher={MIT press}
}

@article{isack2012energy,
  title={Energy-based geometric multi-model fitting},
  author={Isack, Hossam and Boykov, Yuri},
  journal={IJCV},
  volume={},
  number={},
  pages={},
  year={2012},
  publisher={}
}

@incollection{more1978levenberg,
  title={The {L}evenberg-{M}arquardt algorithm: {I}mplementation and theory},
  author={Mor{\'e}, Jorge J},
  booktitle={Numerical analysis},
  pages={},
  year={1978},
  publisher={}
}

@article{haber2011three,
  title={Three-Dimensional Proper and Improper Rotation Matrices},
  author={Haber, H.},
  journal={University of California, Santa Cruz Physics 116A Lecture Notes},
  volume={},
  year={2011}
}

@article{silberman2012indoor,
  title={Indoor segmentation and support inference from {RGBD} images},
  author={Silberman, Nathan and Hoiem, Derek and Kohli, Pushmeet and Fergus, Rob},
  journal={ECCV},
  pages={},
  year={2012},
  organization={}
}

@article{hinterstoisser2017pre,
  title={On Pre-Trained Image Features and Synthetic Images for Deep Learning},
  author={Hinterstoisser, Stefan and Lepetit, Vincent and Wohlhart, Paul and Konolige, Kurt},
  journal={ECCVW},
  year={2018}
}

@article{chollet2017xception,
  title={Xception: Deep learning with depthwise separable convolutions},
  author={Chollet, Fran{\c{c}}ois},
  journal={CVPR},
  pages={},
  year={2017}
}

@article{lin2014microsoft,
  title={Microsoft {COCO}: {C}ommon objects in context},
  author={Lin, Tsung-Yi and Maire, Michael and Belongie, Serge and Hays, James and Perona, Pietro and Ramanan, Deva and Doll{\'a}r, Piotr and Zitnick, C Lawrence},
  journal={ECCV},
  year={2014}
}

@article{rodrigues2019deep,
  title={Deep segmentation leverages geometric pose estimation in computer-aided total knee arthroplasty},
  author={Rodrigues, Pedro and Antunes, Michel and Raposo, Carolina and Marques, Pedro and Fonseca, Fernando and Barreto, Joao},
  journal={Healthcare Technology Letters},
  year={2019},
  publisher={}
}

@article{raposo2017using,
  title={Using 2 point+normal sets for fast registration of point clouds with small overlap},
  author={Raposo, Carolina and Barreto, Joao P},
  journal={ICRA},
  pages={},
  year={2017},
  organization={}
}

@article{rusinkiewicz2001efficient,
  title={Efficient variants of the {ICP} algorithm},
  author={Rusinkiewicz, Szymon and Levoy, Marc},
  journal={Third International Conference on 3-D Digital Imaging and Modeling},
  pages={},
  year={2001},
  organization={}
}

@article{wahl2003surflet,
	title={Surflet-pair-relation histograms: a statistical {3D}-shape representation for rapid classification},
	author={Wahl, Eric and Hillenbrand, Ulrich and Hirzinger, Gerd},
	journal={3DIM},
	year={2003}
}

@article{kim20113d,
	title={{3D} object recognition in range images using visibility context},
	author={Kim, Eunyoung and Medioni, Gerard},
	journal={IROS},
	year={2011}
}

@article{drost20123d,
	title={{3D} object detection and localization using multimodal point pair features},
	author={Drost, Bertram and Ilic, Slobodan},
	journal={Second International Conference on 3D Imaging, Modeling, Processing, Visualization \& Transmission},
	year={2012}
}

@article{birdal2015point,
	title={Point pair features based object detection and pose estimation revisited},
	author={Birdal, Tolga and Ilic, Slobodan},
	journal={3DV},
	year={2015}
}

@article{torr2002bayesian,
  title={Bayesian model estimation and selection for epipolar geometry and generic manifold fitting},
  author={Torr, Philip},
  journal={IJCV},
  volume={},
  number={},
  pages={},
  year={2002},
  publisher={}
}

@article{hesch2011direct,
  title={A direct least-squares {(DLS)} method for {PnP}},
  author={Hesch, Joel A and Roumeliotis, Stergios I},
  journal={ICCV},
  pages={},
  year={2011},
  organization={}
}

@article{kneip2011novel,
  title={A novel parametrization of the perspective-three-point problem for a direct computation of absolute camera position and orientation},
  author={Kneip, Laurent and Scaramuzza, Davide and Siegwart, Roland},
  journal={CVPR},
  pages={},
  year={2011},
  organization={}
}

@article{torr2000mlesac,
  title={{MLESAC}: {A} new robust estimator with application to estimating image geometry},
  author={Torr, Philip and Zisserman, Andrew},
  journal={CVIU},
  volume={},
  number={},
  pages={},
  year={2000},
  publisher={}
}

@inproceedings{torr1998robust,
	title={Robust computation and parametrization of multiple view relations},
	author={Torr, Philip and Zisserman, Andrew},
	booktitle={ICCV},
	year={1998}
}

@inproceedings{lebeda2012fixing,
	title={Fixing the locally optimized ransac--full experimental evaluation},
	author={Lebeda, Karel and Matas, Ji{\v{r}}{\'\i} and Chum, Ond{\v{r}}ej},
	booktitle={BMVC},
	year={2012}
}

@article{besl1992method,
  author  = {Besl, Paul J and McKay, Neil D},
  title   = "{A Method for Registration of 3-D Shapes}",
  journal = {TPAMI},
  year    = {1992},
}

@article{poli2007swarm,
  author={Poli, Riccardo and Kennedy, James and Blackwell, Tim},
  title="{Particle Swarm Optimization}",
  journal={Swarm Intelligence},
  volume={},
  number={},
  publisher={},
  pages={},
  year={2007},
  issn={},
}

@article{oikonomidis2011efficient,
  author    = {Iason Oikonomidis and Nikolaos Kyriazis and Antonis A. Argyros},
  title     = {Efficient model-based 3{D} tracking of hand articulations using {K}inect},
  journal   = {BMVC},
  pages     = {},
  year      = {2011},
}

@article{ivekovic2008human,
 author = {Ivekovi\v{c}, S. and Trucco, E. and Petillot, Y.},
 title = "{Human Body Pose Estimation with Particle Swarm Optimisation}",
 journal = {Evolutionary Computation},
 volume = {},
 number = {},
 year = {2008},
 pages = {},
 publisher = {},
 address = {},
}

@article{choi2012pose,
  author    = {C. Choi and H.I. Christensen},
  title     = "{3D Pose Estimation of Daily Objects Using an RGB-D Camera}",
  journal = {IROS},
  year      = {2012}
}

@article{cai2013fast,
  title={Fast detection of multiple textureless 3-{D} objects},
  author={Cai, Hongping and Werner, Tom{\'a}{\v{s}} and Matas, Ji{\v{r}}{\'\i}},
  journal={ICVS},
  year={2013},
}

@article{shotton2005contour,
	title={Contour-based learning for object detection},
	author={Shotton, Jamie and Blake, Andrew and Cipolla, Roberto},
	journal={ICCV},
	year={2005}
}

@article{damen2012real,
  author={D. Damen and P. Bunnun and A. Calway and W. Mayol-Cuevas},
  title={{Real-time Learning and Detection of 3D Texture-less Objects: A Scalable Approach}},
  journal={BMVC},
  year={2012},
}

@article{zhang1994iterative,
	title={Iterative point matching for registration of free-form curves and surfaces},
	author={Zhang, Zhengyou},
	journal={IJCV},
	year={1994},
}

@article{alexe2012measuring, 
  author={Alexe, Bogdan and Deselaers, Thomas and Ferrari, Vittorio}, 
  journal={TPAMI}, 
  title={Measuring the Objectness of Image Windows}, 
  year={2012}, 
  month={}, 
  volume={}, 
  number={},
  pages={}, 
  ISSN={},
}

@article{cheng2014bing,
  title={{BING}: Binarized normed gradients for objectness estimation at 300fps},
  author={Cheng, Ming-Ming and Zhang, Ziming and Lin, Wen-Yan and Torr, Philip},
  journal={CVPR},
  pages={},
  year={2014},
}

@article{zabulis2015object,
  author={Zabulis, Xenophon and Lourakis, Manolis and Koutlemanis, Panagiotis},
  title={{3D} Object Pose Refinement in Range Images},
  year={2015},
  journal={ICVS}
}

@article{taylor2012vitruvian,
	title={The {V}itruvian {M}anifold: {I}nferring dense correspondences for one-shot human pose estimation},
	author={Taylor, Jonathan and Shotton, Jamie and Sharp, Toby and Fitzgibbon, Andrew},
	journal={CVPR},
	year={2012}
}

@article{dollar2013structured,
  title={Structured forests for fast edge detection},
  author={Doll{\'a}r, Piotr and Zitnick, C Lawrence},
  journal={ICCV},
  pages={},
  year={2013}
}

@article{carmichael2002object,
  title={Object recognition by a cascade of edge probes},
  author={Carmichael, Owen and Hebert, Martial},
  journal={BMVC},
  year={2002}
}

@article{chia2010object,
  title={Object recognition by discriminative combinations of line segments and ellipses},
  author={Chia, Alex Yong-Sang and Rahardja, Susanto and Rajan, Deepu and Leung, Maylor Karhang},
  journal={CVPR},
  year={2010}
}

@article{opelt2006boundary,
  title={A boundary-fragment-model for object detection},
  author={Opelt, Andreas and Pinz, Axel and Zisserman, Andrew},
  journal={ECCV},
  year={2006}
}

@article{danielsson2009automatic,
  title={Automatic learning and extraction of multi-local features},
  author={Danielsson, Oscar and Carlsson, Stefan and Sullivan, Josephine},
  journal={ICCV},
  year={2009},
}

@article{leordeanu2007beyond,
  title={Beyond local appearance: Category recognition from pairwise interactions of simple features},
  author={Leordeanu, Marius and Hebert, Martial and Sukthankar, Rahul},
  journal={CVPR},
  year={2007},
}

@article{beis1999indexing,
  title={Indexing without invariants in 3d object recognition},
  author={Beis, Jeffrey S. and Lowe, David G.},
  journal={TPAMI},
  year={1999}
}

@article{lamdan1988geometric,
  title={Geometric hashing: A general and efficient model-based recognition scheme},
  author={Lamdan, Yehezkel and Wolfson, Haim J},
  journal={ICCV},
  year={1988}
}

@article{lourakis13model,
  author={Lourakis, Manolis and Zabulis, Xenophon},
  title={Model-Based Pose Estimation for Rigid Objects},
  journal={Computer Vision Systems},
  year={2013}
}

@article{mian2010repeatability,
  title={On the repeatability and quality of keypoints for local feature-based {3D} object retrieval from cluttered scenes},
  author={Mian, Ajmal and Bennamoun, Mohammed and Owens, Robyn},
  journal={IJCV},
  year={2010}
}

@article{hosang2015makes,
	title={What makes for effective detection proposals?},
	author={Hosang, Jan and Benenson, Rodrigo and Doll{\'a}r, Piotr and Schiele, Bernt},
	journal={TPAMI},
	year={2015}
}

@article{wohlhart2015learning,
  author={Wohlhart, Paul and Lepetit, Vincent},
  title={Learning Descriptors for Object Recognition and {3D} Pose Estimation},
  journal={CVPR},
  year={2015}
}

@article{kehl2015hashmod,
	title={Hashmod: A Hashing Method for Scalable {3D} Object Detection},
	author={Kehl, Wadim and Tombari, Federico and Navab, Nassir and Ilic, Slobodan and Lepetit, Vincent},
	journal={BMVC},
	year={2015}
}

@article{kehl2016deep,
  title={Deep learning of local {RGB-D} patches for {3D} object detection and {6D} pose estimation},
  author={Kehl, Wadim and Milletari, Fausto and Tombari, Federico and Ilic, Slobodan and Navab, Nassir},
  journal={ECCV},
  year={2016}
}

@article{tombari2011online,
title={Online learning for automatic segmentation of {3D} data},
author={Tombari, Federico and Di Stefano, Luigi and Giardino, Simone},
journal={IROS},
year={2011}
}

@article{georgakis2016multiview,
  title={Multiview {RGB-D} Dataset for Object Instance Detection},
  author={Georgakis, Georgios and Reza, Md Alimoor and Mousavian, Arsalan and Le, Phi-Hung and Kosecka, Jana},
  journal={3DV},
  year={2016},
  note={\url{cs.gmu.edu/~robot/gmu-kitchens.html}}
}

@article{steinbrucker2014fastfusion,
  author={Steinbr{\"u}cker, Frank and Sturm, J{\"u}rgen and Cremers, Daniel},
  journal={ICRA},
  title={Volumetric {3D} mapping in real-time on a {CPU}},
  year={2014},
  note={\url{github.com/tum-vision/fastfusion}},
}

@article{firman2016rgbd,
  title={RGBD datasets: Past, present and future},
  author={Firman, Michael},
  journal={CVPRW},
  year={2016}
}

@article{rios2013discriminatively,
  title={Discriminatively trained templates for {3D} object detection: A real time scalable approach},
  author={Rios-Cabrera, Reyes and Tuytelaars, Tinne},
  journal={ICCV},
  year={2013}
}

@article{tejani2014latent,
  author={Tejani, Alykhan and Tang, Danhang and Kouskouridas, Rigas and Kim, Tae-Kyun},
  title={Latent-Class Hough Forests for {3D} Object Detection and Pose Estimation},
  journal={ECCV},
  year={2014},
  note={\url{www.iis.ee.ic.ac.uk/rkouskou/research/LCHF.html}},
}

@article{xie2013multimodal,
  title={Multimodal blending for high-accuracy instance recognition},
  author={Xie, Ziang and Singh, Arjun and Uang, Justin and Narayan, Karthik S and Abbeel, Pieter},
  journal={IROS},
  year={2013},
  note={\url{rll.berkeley.edu/2013_IROS_ODP}}
}

@article{aldoma2014automation,
  title={Automation of “ground truth” annotation for multi-view {RGB-D} object instance recognition datasets},
  author={Aldoma, Aitor and F{\"a}ulhammer, Thomas and Vincze, Markus},
  journal={IROS},
  year={2014},
  note={\url{repo.acin.tuwien.ac.at/tmp/permanent/dataset_index.php}}
}

@article{papazov2010efficient,
	title={An efficient {RANSAC} for {3D} object recognition in noisy and occluded scenes},
	author={Papazov, Chavdar and Burschka, Darius},
	journal={ACCV},
	year={2010}
}

@article{rennie2016dataset,
  author={Rennie, Colin and Shome, Rahul and Bekris, Kostas E and De Souza, Alberto F},
  journal={RA-L},
  title={A Dataset for Improved {RGBD}-Based Object Detection and Pose Estimation for Warehouse Pick-and-Place},
  year={2016},
  note={\url{www.pracsyslab.org/rutgers_apc_rgbd_dataset}},
}

@article{eppner2016lessons,
  title={Lessons from the {Amazon} picking challenge: Four aspects of building robotic systems.},
  author={Eppner, Clemens and H{\"o}fer, Sebastian and Jonschkowski, Rico and Mart{\'\i}n-Mart{\'\i}n, Roberto and Sieverling, Arne and Wall, Vincent and Brock, Oliver},
  journal={Robotics: science and systems},
  year={2016}
}

@article{aldoma2012global,
  title={A global hypotheses verification method for {3D} object recognition},
  author={Aldoma, Aitor and Tombari, Federico and Di Stefano, Luigi and Vincze, Markus},
  journal={ECCV},
  year={2012},
  note={\url{users.acin.tuwien.ac.at/aaldoma/datasets/ECCV.zip}}
}

@article{singh2014bigbird,
  author={A. Singh and J. Sha and K. S. Narayan and T. Achim and P. Abbeel},
  journal={ICRA},
  title={{BigBIRD}: A large-scale {3D} database of object instances},
  year={2014},
  note={\url{rll.berkeley.edu/bigbird}},
}

@article{lai2011large,
  title={A large-scale hierarchical multi-view {RGB-D} object dataset},
  author={Lai, Kevin and Bo, Liefeng and Ren, Xiaofeng and Fox, Dieter},
  journal={ICRA},
  year={2011},
  note={\url{rgbd-dataset.cs.washington.edu}}
}

@article{schlette2014new,
  title={A new benchmark for pose estimation with ground truth from virtual reality},
  author={Schlette, Christian and others},
  journal={Production Engineering},
  year={2014},
  note={\url{www.mmi.rwth-aachen.de/exchange/data/pesi2014/benchmark.htm}}
}

@article{collins1985development,
  title={The development of a {European} benchmark for the comparison of assembly robot programming systems},
  author={Collins, K and Palmer, AJ and Rathmill, K},
  journal={Robot technology and applications},
  year={1985}
}

@article{salti2014shot,
  author={Samuele Salti and Federico Tombari and Luigi Di Stefano},
  title={{SHOT}: Unique signatures of histograms for surface and texture description},
  journal={CVIU},
  note={\url{www.vision.deis.unibo.it/research/80-shot}},
}

@article{mian2006three,
  title={Three-dimensional model-based object recognition and segmentation in cluttered scenes},
  author={Mian, Ajmal S and Bennamoun, Mohammed and Owens, Robyn},
  journal={TPAMI},
  year={2006},
  note={\url{staffhome.ecm.uwa.edu.au/~00053650/recognition.html}},
}

@article{zhong2009intrinsic,
	title={Intrinsic shape signatures: A shape descriptor for {3D} object recognition},
	author={Zhong, Yu},
	journal={ICCVW},
	year={2009}
}

@article{unnikrishnan2008multi,
	title={Multi-scale interest regions from unorganized point clouds},
	author={Unnikrishnan, Ranjith and Hebert, Martial},
	journal={CVPRW},
	year={2008}
}

@article{taati2007variable,
  title={Variable dimensional local shape descriptors for object recognition in range data},
  author={Taati, Babak and Bondy, Michel and Jasiobedzki, Piotr and Greenspan, Michael},
  journal={ICCV},
  year={2007},
  note={\url{rcvlab.ece.queensu.ca/~qridb/lsdPage.html}}
}

@article{bonde2014robust,
  author={Bonde, Ujwal and Badrinarayanan, Vijay and Cipolla, Roberto},
  title={Robust Instance Recognition in Presence of Occlusion and Clutter},
  journal={ECCV},
  year={2014},
  note={\url{sites.google.com/site/ujwalbonde/publications/downloads}},
}

@article{lim2013parsing,
  title={Parsing {IKEA} Objects: Fine Pose Estimation},
  author={Lim, Joseph J and Pirsiavash, Hamed and Torralba, Antonio},
  journal={ICCV},
  year={2013},
  note={\url{ikea.csail.mit.edu}}
}

@article{crivellaro2015novel,
  author    = {Alberto Crivellaro and Mahdi Rad and Yannick Verdie and Kwang Moo Yi and Pascal Fua and Vincent Lepetit},
  title     = {A Novel Representation of Parts for Accurate {3D} Object Detection and Tracking in Monocular Images},
  journal   = {ICCV},
  year      = {2015},
  note      = {\url{cvlab.epfl.ch/data/3d_object_tracking}},
}

@article{munoz2016fast,
  title={Fast 6D pose estimation for texture-less objects from a single RGB image},
  author={Mu{\~n}oz, Enrique and Konishi, Yoshinori and Murino, Vittorio and Del Bue, Alessio},
  journal={ICRA},
  year={2016},
}

@article{hsiao2014occlusion,
  title={Occlusion reasoning for object detection under arbitrary viewpoint},
  author={Hsiao, Edward and Hebert, Martial},
  journal={TPAMI},
  year={2014},
  note={\url{www.cs.cmu.edu/\~./hebert/occarbview.html}},
}

@article{michel2015pose,
  title={Pose estimation of kinematic chain instances via object coordinate regression},
  author={Michel, Frank and Krull, Alexander and Brachmann, Eric and Yang, Michael Ying and Gumhold, Stefan and Rother, Carsten},
  journal={BMVC},
  year={2015},
  note={\url{cvlab-dresden.de/iccv2015-articulation-challenge}}
}

@article{tu2009auto,
	title={Auto-context and its application to high-level vision tasks and {3D} brain image segmentation},
	author={Tu, Zhuowen and Bai, Xiang},
	journal={TPAMI},
	year={2009},
}

@article{wohlkinger20123dnet,
  title={{3DNet}: Large-scale object class recognition from {CAD} models},
  author={Wohlkinger, Walter and Aldoma, Aitor and Rusu, Radu B and Vincze, Markus},
  journal={ICRA},
  year={2012},
  note={\url{repo.acin.tuwien.ac.at/tmp/permanent/3d-net.org}}
}

@misc{walas2016uob,
  author = {Krzysztof Walas and Ale{\v{s}} Leonardis},
  title = {{UoB} Highly Occluded Object Challenge},
  year = {2016},
  note = {\url{https://www.cs.bham.ac.uk/~walask/uob_hooc/}},
}

@misc{darwinproject,
	title = {{DARWIN:} Dextrous Assembler Robot Working with embodied Intelligence},
	year = {2015},
	note = {\url{https://cordis.europa.eu/project/id/270138}},
}

@article{meger2011mobile,
  title={Mobile {3D} object detection in clutter},
  author={Meger, David and Little, James J},
  journal={IROS},
  year={2011},
  note={\url{www.cs.ubc.ca/labs/lci/vrs}}
}

@article{guo2013support,
  title={Support surface prediction in indoor scenes},
  author={Guo, Ruiqi and Hoiem, Derek},
  journal={ICCV},
  year={2013},
  note={\url{ttic.uchicago.edu/~rurtasun/rmrc/indoor.php}}
}

@article{janoch2013category,
  title={A category-level {3D} object dataset: Putting the {Kinect} to work},
  author={Janoch, Allison and Karayev, Sergey and Jia, Yangqing and Barron, Jonathan T and Fritz, Mario and Saenko, Kate and Darrell, Trevor},
  journal={Consumer Depth Cameras for Computer Vision},
  year={2013},
  note={\url{kinectdata.com}}
}

@article{song2015sun,
  title={Sun {RGB-D}: A {RGB-D} scene understanding benchmark suite},
  author={Song, Shuran and Lichtenberg, Samuel P and Xiao, Jianxiong},
  journal={CVPR},
  year={2015},
  note={\url{rgbd.cs.princeton.edu}}
}

@article{xiang2014beyond,
  title={Beyond {PASCAL}: A benchmark for {3D} object detection in the wild},
  author={Xiang, Yu and Mottaghi, Roozbeh and Savarese, Silvio},
  journal={Winter Conference on Applications of Computer Vision},
  year={2014}
}

@article{xiang2016objectnet3d,
  title={{ObjectNet3D}: A Large Scale Database for {3D} Object Recognition},
  author={Xiang, Yu and others},
  journal={ECCV},
  year={2016},
  note={\url{cvgl.stanford.edu/projects/objectnet3d}}
}

@book{bradski2008learning,
  title={Learning OpenCV: Computer vision with the OpenCV library},
  author={Bradski, Gary and Kaehler, Adrian},
  year={2008},
  publisher={O'Reilly Media, Inc.}
}

@article{wagner2007artoolkitplus,
  title={{ARToolKitPlus} for pose tracking on mobile devices},
  author={Wagner, Daniel and Schmalstieg, Dieter},
  journal={CVWW},
  year={2007}
}

@article{foix2011lock,
  title={Lock-in time-of-flight ({ToF}) cameras: a survey},
  author={Foix, Sergi and Alenya, Guillem and Torras, Carme},
  journal={Sensors Journal},
  year={2011}
}

@article{sturm2012benchmark,
  title={A benchmark for the evaluation of {RGB-D SLAM} systems},
  author={Sturm, J{\"u}rgen and Engelhard, Nikolas and Endres, Felix and Burgard, Wolfram and Cremers, Daniel},
  journal={IROS},
  year={2012}
}

@article{cignoni2008meshlab,
  journal={Eurographics Italian Chapter Conf.},
  title={{MeshLab}: an Open-Source Mesh Processing Tool},
  author={Cignoni, Paolo and Callieri, Marco and Corsini, Massimiliano and Dellepiane, Matteo and Ganovelli, Fabio and Ranzuglia, Guido},
  year={2008},
}

@article{thurrner1998computing,
  title={Computing vertex normals from polygonal facets},
  author={Th{\"u}rrner, Grit and W{\"u}thrich, Charles A},
  journal={Journal of Graphics Tools},
  year={1998}
}

@article{cignoni1998metro,
  title={Metro: measuring error on simplified surfaces},
  author={Cignoni, Paolo and Rocchini, Claudio and Scopigno, Roberto},
  journal={Computer Graphics Forum},
  year={1998},
  organization={Wiley Online Library}
}

@article{khoshelham2012accuracy,
  author = {Khoshelham, Kourosh and Elberink, Sander Oude},
  title = {Accuracy and Resolution of {Kinect} Depth Data for Indoor Mapping Applications},
  journal = {Sensors},
  year = {2012}
}

@book{kramer2012hacking,
	title={Hacking the Kinect},
	author={Kramer, Jeff and Burrus, Nicolas and Echtler, Florian and Daniel, Herrera C and Parker, Matt},
	volume={268},
	year={2012},
	publisher={Springer}
}

@article{scharstein2002taxonomy,
  title = {A Taxonomy and Evaluation of Dense Two-Frame Stereo Correspondence Algorithms},
  author = {Scharstein, Daniel and Szeliski, Richard},
  journal = {IJCV},
  year = {2002}
}

@article{scharstein2007learning,
  title = {Learning Conditional Random Fields for Stereo.},
  author = {Scharstein, Daniel and Pal, Chris},
  journal = {CVPR},
  year = {2007}
}

@article{everingham2010pascal,
  title={The {PASCAL} Visual Object Classes {(VOC)} Challenge},
  author={Everingham, Mark and Van Gool, Luc and Williams, Christopher KI and Winn, John and Zisserman, Andrew},
  journal={IJCV},
  year={2010}
}

@article{russakovsky2015imagenet,
  title={{I}mage{N}et Large Scale Visual Recognition Challenge},
  author={Russakovsky, Olga and Deng, Jia and Su, Hao and Krause, Jonathan and Satheesh, Sanjeev and Ma, Sean and Huang, Zhiheng and Karpathy, Andrej and Khosla, Aditya and Bernstein, Michael and others},
  journal={IJCV},
  year={2015}
}

@article{krizhevsky2012imagenet,
title = {{I}mage{N}et Classification with Deep Convolutional Neural Networks},
author = {Alex Krizhevsky and Sutskever, Ilya and Hinton, Geoffrey E.},
journal = {NIPS},
year = {2012}
}

@article{geiger2012kitti,
  title = {Are we ready for Autonomous Driving? {T}he {KITTI} Vision Benchmark Suite},
  author = {Andreas Geiger and Philip Lenz and Raquel Urtasun},
  journal = {CVPR},
  year = {2012}
}

@article{michel2017global,
  author = {Michel, Frank and Alexander Kirillov, Brachmann, Eric and Krull, Alexander and Gumhold, Stefan and Savchynskyy, Bogdan and Rother, Carsten},
  title = {Global Hypothesis Generation for {6D} Object Pose Estimation},
  journal={CVPR},
  year = {2017}
}

@book{morawiec2004orientations,
	title={Orientations and Rotations: Computations in Crystallographic Textures},
	author={Morawiec, Adam},
	year={2004},
	publisher={Springer Science \& Business Media}
}

@article{buch2016local,
  title={Local shape feature fusion for improved matching, pose estimation and {3D} object recognition},
  author={Buch, Anders G and Petersen, Henrik G and Kr{\"u}ger, Norbert},
  journal={SpringerPlus},
  year={2016}
}

@article{buch2017rotational,
  title={Rotational Subgroup Voting and Pose Clustering for Robust {3D} Object Recognition},
  author={Buch, Anders Glent and Kiforenko, Lilita and Kraft, Dirk},
  journal={ICCV},
  year={2017}
}

@article{buch2018local,
  title={Local Point Pair Feature Histogram for Accurate {3D} Matching},
  author={Anders Glent Buch and Dirk Kraft},
  journal={BMVC},
  year={2018}
}

@article{kiforenko2018performance,
	title={A performance evaluation of point pair features},
	author={Kiforenko, Lilita and Drost, Bertram and Tombari, Federico and Kr{\"u}ger, Norbert and Buch, Anders Glent},
	journal={CVIU},
	year={2018}
}

@article{johnson1999using,
  title={Using spin images for efficient object recognition in cluttered {3D} scenes},
  author={Johnson, Andrew E. and Hebert, Martial},
  journal={TPAMI},
  year={1999}
}

@article{jorgensen2015geometric,
  title={Geometric edge description and classification in point cloud data with application to {3D} object recognition},
  author={J{\o}rgensen, Troels Bo and Buch, Anders Glent and Kraft, Dirk},
  journal={VISAPP},
  year={2015}
}

@article{su2015render,
  title={{Render for CNN}: Viewpoint estimation in images using cnns trained with rendered {3D} model views},
  author={Su, Hao and Qi, Charles R and Li, Yangyan and Guibas, Leonidas J},
  journal={ICCV},
  year={2015}
}

@article{dosovitskiy2015flownet,
  title={Flownet: Learning optical flow with convolutional networks},
  author={Dosovitskiy, Alexey and Fischer, Philipp and Ilg, Eddy and Hausser, Philip and Hazirbas, Caner and Golkov, Vladimir and Van Der Smagt, Patrick and Cremers, Daniel and Brox, Thomas},
  journal={ICCV},
  year={2015}
}

@article{dwibedi2017cut,
  title={Cut, paste and learn: Surprisingly easy synthesis for instance detection},
  author={Dwibedi, Debidatta and Misra, Ishan and Hebert, Martial},
  journal={ICCV},
  year={2017}
}

@article{dvornik2018modeling,
  title={Modeling visual context is key to augmenting object detection datasets},
  author={Dvornik, Nikita and Mairal, Julien and Schmid, Cordelia},
  journal={ECCV},
  year={2018}
}

@article{movshovitz2016useful,
  title={How useful is photo-realistic rendering for visual learning?},
  author={Movshovitz-Attias, Yair and Kanade, Takeo and Sheikh, Yaser},
  journal={ECCV},
  year={2016},
}

@article{richter2017playing,
  title={Playing for benchmarks},
  author={Richter, Stephan R and Hayder, Zeeshan and Koltun, Vladlen},
  journal={ICCV},
  year={2017}
}

@article{richter2016playing,
  title={Playing for data: Ground truth from computer games},
  author={Richter, Stephan R and Vineet, Vibhav and Roth, Stefan and Koltun, Vladlen},
  journal={ECCV},
  year={2016}
}

@article{ros2016synthia,
  title={The {SYNTHIA} dataset: A large collection of synthetic images for semantic segmentation of urban scenes},
  author={Ros, German and Sellart, Laura and Materzynska, Joanna and Vazquez, David and Lopez, Antonio M},
  journal={CVPR},
  year={2016}
}

@article{gaidon2016virtual,
  author = {Gaidon, A and Wang, Q and Cabon, Y and Vig, E},
  title = {Virtual Worlds as Proxy for Multi-Object Tracking Analysis},
  journal = {CVPR},
  year = {2016}
}

@article{hand2016understanding,
  title={Understanding real world indoor scenes with synthetic data},
  author={Handa, Ankur and Patraucean, Viorica and Badrinarayanan, Vijay and Stent, Simon and Cipolla, Roberto},
  journal={CVPR},
  year={2016}
}

@article{zhang2017physically,
  title={Physically-based rendering for indoor scene understanding using convolutional neural networks},
  author={Zhang, Yinda and Song, Shuran and Yumer, Ersin and Savva, Manolis and Lee, Joon-Young and Jin, Hailin and Funkhouser, Thomas},
  journal={CVPR},
  year={2017}
}

@article{mccormac2017scenenet,
  title={{SceneNet} {RGB-D}: {C}an {5M} synthetic images beat generic {ImageNet} pre-training on indoor segmentation?},
  author={McCormac, John and Handa, Ankur and Leutenegger, Stefan and Davison, Andrew J},
  journal={ICCV},
  year={2017}
}

@article{tobin2017domain,
  title={Domain randomization for transferring deep neural networks from simulation to the real world},
  author={Tobin, Josh and Fong, Rachel and Ray, Alex and Schneider, Jonas and Zaremba, Wojciech and Abbeel, Pieter},
  journal={IROS},
  year={2017}
}

@article{li2018cgintrinsics,
  title={{CGIntrinsics}: Better intrinsic image decomposition through physically-based rendering},
  author={Li, Zhengqi and Snavely, Noah},
  journal={ECCV},
  year={2018}
}

@article{wood2015rendering,
  title={Rendering of eyes for eye-shape registration and gaze estimation},
  author={Wood, Erroll and Baltrusaitis, Tadas and Zhang, Xucong and Sugano, Yusuke and Robinson, Peter and Bulling, Andreas},
  journal={ICCV},
  year={2015}
}

@article{wood2016learning,
  title={Learning an appearance-based gaze estimator from one million synthesised images},
  author={Wood, Erroll and Baltru{\v{s}}aitis, Tadas and Morency, Louis-Philippe and Robinson, Peter and Bulling, Andreas},
  journal={Proceedings of the Ninth Biennial ACM Symposium on Eye Tracking Research \& Applications},
  year={2016}
}

@article{shrivastava2017learning,
  title={Learning from Simulated and Unsupervised Images through Adversarial Training.},
  author={Shrivastava, Ashish and Pfister, Tomas and Tuzel, Oncel and Susskind, Joshua and Wang, Wenda and Webb, Russell},
  journal={CVPR},
  year={2017}
}

@article{ren2017faster,
  title={{Faster R-CNN:} towards real-time object detection with region proposal networks},
  author={Ren, Shaoqing and He, Kaiming and Girshick, Ross and Sun, Jian},
  journal={TPAMI},
  year={2017}
}

@article{rozantsev2018sharing,
  title={Beyond sharing weights for deep domain adaptation},
  author={Rozantsev, Artem and Salzmann, Mathieu and Fua, Pascal},
  journal={TPAMI},
  year={2018}
}

@article{csurka2017comprehensive,
  title={A comprehensive survey on domain adaptation for visual applications},
  author={Csurka, Gabriela},
  journal={Domain adaptation in computer vision applications},
  year={2017}
}

@article{georgiev2018arnold,
  title={Arnold: A brute-force production path tracer},
  author={Georgiev, Iliyan and Ize, Thiago and Farnsworth, Mike and Montoya-Vozmediano, Ramon and King, Alan and Lommel, Brecht Van and Jimenez, Angel and Anson, Oscar and Ogaki, Shinji and Johnston, Eric and others},
  journal={TOG},
  year={2018}
}

@book{pharr2016physically,
  title={Physically based rendering: From theory to implementation},
  author={Pharr, Matt and Jakob, Wenzel and Humphreys, Greg},
  year={2016},
  publisher={Morgan Kaufmann}
}

@book{shreiner2009opengl,
  title={{OpenGL} programming guide: the official guide to learning {OpenGL}, versions 3.0 and 3.1},
  author={Shreiner, Dave},
  year={2009},
  publisher={Pearson Education}
}

@book{marschner2015fundamentals,
  title={Fundamentals of computer graphics},
  author={Marschner, Steve and Shirley, Peter},
  year={2015},
  publisher={CRC Press}
}

@article{yu2016summary,
  title={A Summary of Team {MIT}'s Approach to the {A}mazon {P}icking {C}hallenge 2015},
  author={Yu, Kuan-Ting and Fazeli, Nima and Chavan-Dafle, Nikhil and Taylor, Orion and Donlon, Elliott and Lankenau, Guillermo Diaz and Rodriguez, Alberto},
  journal={arXiv preprint arXiv:1604.03639},
  year={2016}
}

@article{gal2014flare,
  title={{FLARE}: Fast layout for augmented reality applications},
  author={Gal, Ran and Shapira, Lior and Ofek, Eyal and Kohli, Pushmeet},
  journal={ISMAR},
  year={2014}
}

@inproceedings{toldo2008robust,
	title={Robust multiple structures estimation with j-linkage},
	author={Toldo, Roberto and Fusiello, Andrea},
	booktitle={ECCV},
	year={2008},
}

@article{szegedy2017inception,
  title={Inception-v4, inception-resnet and the impact of residual connections on learning.},
  author={Szegedy, Christian and Ioffe, Sergey and Vanhoucke, Vincent and Alemi, Alexander A},
  journal={AAAI},
  year={2017}
}

@article{he2016deep,
  title={Deep residual learning for image recognition},
  author={He, Kaiming and Zhang, Xiangyu and Ren, Shaoqing and Sun, Jian},
  journal={CVPR},
  year={2016}
}

@article{huang2017speed,
  title={Speed/accuracy trade-offs for modern convolutional object detectors},
  author={Huang, Jonathan and Rathod, Vivek and Sun, Chen and Zhu, Menglong and Korattikara, Anoop and Fathi, Alireza and Fischer, Ian and Wojna, Zbigniew and Song, Yang and Guadarrama, Sergio and others},
  journal={CVPR},
  year={2017}
}

@article{divvala2009empirical,
  title={An empirical study of context in object detection},
  author={Divvala, Santosh K and Hoiem, Derek and Hays, James H and Efros, Alexei A and Hebert, Martial},
  journal={CVPR},
  year={2009}
}

@article{kaskman2019homebreweddb,
  title={{HomebrewedDB}: {RGB-D} Dataset for {6D} Pose Estimation of {3D} Objects},
  author={Kaskman, Roman and Zakharov, Sergey and Shugurov, Ivan and Ilic, Slobodan},
  journal={ICCVW},
  year={2019}
}

@article{qian20163d,
  title={3D reconstruction of transparent objects with position-normal consistency},
  author={Qian, Yiming and Gong, Minglun and Hong Yang, Yee},
  journal={CVPR},
  year={2016}
}

@article{wu2018full,
  title={Full 3D reconstruction of transparent objects},
  author={Wu, Bojian and Zhou, Yang and Qian, Yiming and Cong, Minglun and Huang, Hui},
  journal={ACM TOG},
  year={2018}
}

@article{godard2015multi,
  title={Multi-view reconstruction of highly specular surfaces in uncontrolled environments},
  author={Godard, Clement and Hedman, Peter and Li, Wenbin and Brostow, Gabriel J},
  journal={3DV},
  year={2015}
}

@article{labbe2020cosypose,
	title={{CosyPose:} Consistent multi-view multi-object {6D} pose estimation},
	author={Labb{\'e}, Yann and Carpentier, Justin and Aubry, Mathieu and Sivic, Josef},
	journal={ECCV},
	year={2020}
}

@article{koenig2020hybrid,
	title={A Hybrid Approach for 6DoF Pose Estimation},
	author={Koenig, Rebecca and Drost, Bertram},
	journal={ECCVW},
	year={2020}
}

@article{choi2012voting,
	title={Voting-based pose estimation for robotic assembly using a {3D} sensor},
	author={Choi, Changhyun and Taguchi, Yuichi and Tuzel, Oncel and Liu, Ming-Yu and Ramalingam, Srikumar},
	author={2012 IEEE International Conference on Robotics and Automation},
	year={2012}
}

@article{liu2020leaping,
	title={Leaping from {2D} Detection to Efficient {6DoF} Object Pose Estimation},
	author={Liu, Jinhui and Zou, Zhikang and Ye, Xiaoqing and Tan, Xiao and Ding, Errui and Xu, Feng and Yu, Xin},
	journal={ECCVW},
	year={2020}
}

@article{hagelskjaer2020pointvotenet,
	title={{PointVoteNet:} Accurate Object Detection And {6 DOF} Pose Estimation In Point Clouds},
	author={Hagelskj{\ae}r, Frederik and Buch, Anders Glent},
	journal={ICIP},
	year={2020}
}

@article{sundermeyer2020multi,
	title={Multi-path learning for object pose estimation across domains},
	author={Sundermeyer, Martin and Durner, Maximilian and Puang, En Yen and Marton, Zoltan-Csaba and Vaskevicius, Narunas and Arras, Kai O and Triebel, Rudolph},
	journal={CVPR},
	year={2020}
}

@misc{hodan2017sixd,
	title={{SIXD} {C}hallenge 2017},
	author={Hoda{\v{n}}, Tom{\'a}{\v{s}} and Michel, Frank and Sahin, Caner and Kim, Tae-Kyun and Matas, Ji{\v{r}}{\'i} and Rother, Carsten},
	howpublished={\url{http://cmp.felk.cvut.cz/sixd/challenge_2017/}},
	year={2017}
}

@article{denninger2019blenderproc,
	title={{BlenderProc}},
	author={Denninger, Maximilian and Sundermeyer, Martin and Winkelbauer, Dominik and Zidan, Youssef and Olefir, Dmitry and Elbadrawy, Mohamad and Lodhi, Ahsan and Katam, Harinandan},
	journal={arXiv preprint arXiv:1911.01911},
	year={2019}
}

@article{newcombe2011kinectfusion,
	title={{K}inect{F}usion: Real-time dense surface mapping and tracking},
	author={Newcombe, Richard and Izadi, Shahram and Hilliges, Otmar and Molyneaux, David and Kim, David and Davison, Andrew and Kohi, Pushmeet and Shotton, Jamie and Hodges, Steve and Fitzgibbon, Andrew},
	journal={ISMAR},
	year={2011}
}

@misc{hodan2017r6d,
	title={3rd International Workshop on Recovering {6D} Object Pose},
	author={Walas, Krzysztof and Hoda{\v{n}}, Tom{\'a}{\v{s}} and Kim, Tae-Kyun and Matas, Ji{\v{r}}{\'i} and Rother, Carsten and Michel, Frank and Lepetit, Vincent and Leonardis, Ales and Steger, Carsten and Sahin, Caner and Kouskouridas, Rigas},
	howpublished={\url{http://cmp.felk.cvut.cz/sixd/workshop_2017/}},
	year={2017}
}

@misc{hodan2019r6d,
	title={5th International Workshop on Recovering {6D} Object Pose},
	author={Hoda{\v{n}}, Tom{\'a}{\v{s}} and Kouskouridas, Rigas and Kim, Tae-Kyun and Matas, Ji{\v{r}}{\'i} and Rother, Carsten and Lepetit, Vincent and Leonardis, Ales and Walas, Krzysztof and Steger, Carsten and Brachmann, Eric and Drost, Bertram and Sock, Juil},
	howpublished={\url{http://cmp.felk.cvut.cz/sixd/workshop_2019/}},
	year={2019}
}

@misc{hodan2020r6d,
	title={6th International Workshop on Recovering {6D} Object Pose},
	author={Hoda{\v{n}}, Tom{\'a}{\v{s}} and Sundermeyer, Martin and Kouskouridas, Rigas and Kim, Tae-Kyun and Matas, Ji{\v{r}}{\'i} and Rother, Carsten and Lepetit, Vincent and Leonardis, Ales and Walas, Krzysztof and Steger, Carsten and Brachmann, Eric and Drost, Bertram and Sock, Juil},
	howpublished={\url{http://cmp.felk.cvut.cz/sixd/workshop_2020/}},
	year={2020}
}

@article{doumanoglou2016recovering,
	title={Recovering {6D} Object Pose and Predicting Next-Best-View in the Crowd},
	author={Doumanoglou, Andreas and Kouskouridas, Rigas and Malassiotis, Sotiris and Kim, Tae-Kyun},
	journal={CVPR},
	year={2016}
}

@article{oberweger2015training,
	title={Training a feedback loop for hand pose estimation},
	author={Oberweger, Markus and Wohlhart, Paul and Lepetit, Vincent},
	journal={ICCV},
	year={2015}
}

@misc{demes2020textures,
	title={{CC0 Textures}},
	author={Demes, Lennart},
	howpublished={\url{https://cc0textures.com/}},
	year={2020}
}

@article{geirhos2018imagenet,
	title={ImageNet-trained CNNs are biased towards texture; increasing shape bias improves accuracy and robustness},
	author={Geirhos, Robert and Rubisch, Patricia and Michaelis, Claudio and Bethge, Matthias and Wichmann, Felix A and Brendel, Wieland},
	journal={arXiv preprint arXiv:1811.12231},
	year={2018}
}

@article{bay2006surf,
	title={{SURF:} Speeded up robust features},
	author={Bay, Herbert and Tuytelaars, Tinne and Van Gool, Luc},
	journal={ECCV},
	year={2006}
}

@article{matas2004robust,
	title={Robust wide-baseline stereo from maximally stable extremal regions},
	author={Matas, Ji{\v{r}}{\'\i} and Chum, Ond{\v{r}}ej and Urban, Martin and Pajdla, Tom{\'a}{\v{s}}},
	journal={Image and vision computing},
	year={2004},
}

@article{muja2009fast,
	title={Fast approximate nearest neighbors with automatic algorithm configuration.},
	author={Muja, Marius and Lowe, David G},
	journal={VISAPP},
	year={2009}
}

@article{murase1995visual,
	title={Visual learning and recognition of {3-D} objects from appearance},
	author={Murase, Hiroshi and Nayar, Shree K},
	journal={IJCV},
	year={1995}
}

@article{leonardis1996dealing,
	title={Dealing with occlusions in the eigenspace approach},
	author={Leonardis, Ales and Bischof, Horst},
	journal={CVPR}
	year={1996}
}

@article{ponce2004toward,
	title={Toward true 3D object recognition},
	author={Ponce, Jean and Lazebnik, Svetlana and Rothganger, Fredrick and Schmid, Cordelia},
	journal={Reconnaissance de Formes et Intelligence Artificielle},
	year={2004}
}

@article{obdrzalek2006object,
	title={Object recognition using local affine frames},
	author={Obdr{\v{z}}{\'a}lek, {\v{S}}t{\v{e}}p{\'a}n and Matas, Ji{\v{r}}{\'\i}},
	journal={Proc. Research Reports of CMP, Czech Technical University},
	year={2006}
}

@article{jund2016freiburg,
	title={The {Freiburg} groceries dataset},
	author={Jund, Philipp and Abdo, Nichola and Eitel, Andreas and Burgard, Wolfram},
	journal={arXiv preprint arXiv:1611.05799},
	year={2016}
}

@article{guo20143d,
	title={{3D} object recognition in cluttered scenes with local surface features: a survey},
	author={Guo, Yulan and Bennamoun, Mohammed and Sohel, Ferdous and Lu, Min and Wan, Jianwei},
	journal={TPAMI},
	year={2014}
}

@article{matas2004object,
	title={Object recognition methods based on transformation covariant features},
	author={Matas, Ji{\v{r}}{\'\i} and Obdr{\v{z}}{\'a}lek, {\v{S}}t{\v{e}}p{\'a}n},
	journal={12th European Signal Processing Conference},
	year={2004}
}

@article{tombari2013performance,
	title={Performance evaluation of 3D keypoint detectors},
	author={Tombari, Federico and Salti, Samuele and Di Stefano, Luigi},
	journal={IJCV},
	year={2013}
}

@article{alexe2010object,
	title={What is an object?},
	author={Alexe, Bogdan and Deselaers, Thomas and Ferrari, Vittorio},
	journal={CVPR},
	year={2010}
}

@article{arteta2016counting,
	title={Counting in the wild},
	author={Arteta, Carlos and Lempitsky, Victor and Zisserman, Andrew},
	journal={ECCV},
	year={2016}
}

@misc{blender,
	title = {Blender -- a {3D} modelling and rendering package},
	author = {Blender Online Community},
	organization = {Blender Foundation},
	year = {2018},
	url = {http://www.blender.org},
}

@misc{inteldenoise,
	title = {Intel {O}pen {I}mage {D}enoise},
	year = {2020},
	url = {https://www.openimagedenoise.org/},
}

@misc{boptoolkit,
  title = {{BOP Toolkit}},
  author = {Hoda{\v{n}}, Tom{\'a}{\v{s}} and Sundermeyer, Martin},
  year = {2020},
  url = {https://github.com/thodan/bop_toolkit},
}

@misc{blenderproc4bop,
  title = {{BlenderProc4BOP}},
  year = {2020},
  url = {https://github.com/DLR-RM/BlenderProc/blob/master/README_BlenderProc4BOP.md},
}

@misc{halcon,
	title={{MVT}ec {HALCON}},
	author={},
	year={2020},
	url = {https://www.mvtec.com/halcon/},
}

@article{he2017mask,
	title={Mask {R-CNN}},
	author={He, Kaiming and Gkioxari, Georgia and Doll{\'a}r, Piotr and Girshick, Ross},
	journal={ICCV},
	year={2017}
}

@article{sundermeyer2018implicit,
	title={Implicit {3D} orientation learning for {6D} object detection from {RGB} images},
	author={Sundermeyer, Martin and Marton, Zoltan-Csaba and Durner, Maximilian and Brucker, Manuel and Triebel, Rudolph},
	journal={ECCV},
	year={2018}
}

@article{fu2019retinamask,
	title={RetinaMask: Learning to predict masks improves state-of-the-art single-shot detection for free},
	author={Fu, Cheng-Yang and Shvets, Mykhailo and Berg, Alexander C},
	journal={arXiv preprint arXiv:1901.03353},
	year={2019}
}

@article{he2020pvn3d,
	title={{PVN3D}: A deep point-wise {3D} keypoints voting network for {6DoF} pose estimation},
	author={He, Yisheng and Sun, Wei and Huang, Haibin and Liu, Jianran and Fan, Haoqiang and Sun, Jian},
	journal={CVPR},
	year={2020}
}

@article{do2019real,
	title={Real-time monocular object instance {6D} pose estimation},
	author={Do, Thanh-Toan and Pham, Trung and Cai, Ming and Reid, Ian},
	year={2019},
	journal={BMVC}
}

@article{hinterstoisser2019annotation,
	title={An annotation saved is an annotation earned: Using fully synthetic training for object detection},
	author={Hinterstoisser, Stefan and Pauly, Olivier and Heibel, Hauke and Martina, Marek and Bokeloh, Martin},
	journal={ICCVW},
	pages={},
	year={2019}
}

@article{qi2017pointnet,
	title={{PointNet}: Deep learning on point sets for {3D} classification and segmentation},
	author={Qi, Charles R and Su, Hao and Mo, Kaichun and Guibas, Leonidas J},
	journal={CVPR},
	year={2017}
}

@article{qi2017pointnet++,
	title={Pointnet++: Deep hierarchical feature learning on point sets in a metric space},
	author={Qi, Charles Ruizhongtai and Yi, Li and Su, Hao and Guibas, Leonidas J},
	journal={Advances in neural information processing systems},
	year={2017}
}

@article{qi2020imvotenet,
	title={{ImVoteNet}: Boosting {3D} object detection in point clouds with image votes},
	author={Qi, Charles R and Chen, Xinlei and Litany, Or and Guibas, Leonidas J},
	journal={CVPR},
	year={2020}
}

@article{zhou2019continuity,
	title={On the continuity of rotation representations in neural networks},
	author={Zhou, Yi and Barnes, Connelly and Lu, Jingwan and Yang, Jimei and Li, Hao},
	journal={CVPR},
	year={2019}
}

@article{canny1986computational,
	title={A computational approach to edge detection},
	author={Canny, John},
	journal={TPAMI},
	year={1986}
}

@article{lowe1987three,
	title={Three-dimensional object recognition from single two-dimensional images},
	author={Lowe, David G},
	journal={Artificial intelligence},
	year={1987}
}

@article{planche2017depthsynth,
	title={{DepthSynth}: Real-time realistic synthetic data generation from {CAD} models for {2.5D} recognition},
	author={Planche, Benjamin and Wu, Ziyan and Ma, Kai and Sun, Shanhui and Kluckner, Stefan and Lehmann, Oliver and Chen, Terrence and Hutter, Andreas and Zakharov, Sergey and Kosch, Harald and others},
	journal={3DV},
	year={2017}
}

@article{pitteri20203d,
	title={{3D} Object Detection and Pose Estimation of Unseen Objects in Color Images with Local Surface Embeddings},
	author={Pitteri, Giorgia and Bugeau, Aur{\'e}lie and Ilic, Slobodan and Lepetit, Vincent},
	journal={ACCV},
	year={2020}
}

@article{chen2020category,
	title={Category Level Object Pose Estimation via Neural Analysis-by-Synthesis},
	author={Chen, Xu and Dong, Zijian and Song, Jie and Geiger, Andreas and Hilliges, Otmar},
	journal={ECCV},
	year={2020}
}

@article{cai2020reconstruct,
	title={Reconstruct Locally, Localize Globally: A Model Free Method for Object Pose Estimation},
	author={Cai, Ming and Reid, Ian},
	journal={CVPR},
	year={2020}
}

@article{park2020neural,
	title={Neural Object Learning for {6D} Pose Estimation Using a Few Cluttered Images},
	author={Park, Kiru and Patten, Timothy and Vincze, Markus},
	journal={ECCV},
	year={2020}
}

@article{shao2020pfrl,
	title={{PFRL}: Pose-Free Reinforcement Learning for {6D} Pose Estimation},
	author={Shao, Jianzhun and Jiang, Yuhang and Wang, Gu and Li, Zhigang and Ji, Xiangyang},
	journal={CVPR},
	year={2020}
}

@article{tian2020shape,
	title={Shape Prior Deformation for Categorical {6D} Object Pose and Size Estimation},
	author={Tian, Meng and Ang, Marcelo H and Lee, Gim Hee},
	journal={ECCV},
	year={2020}
}

@article{chen2020learning,
	title={Learning canonical shape space for category-level {6D} object pose and size estimation},
	author={Chen, Dengsheng and Li, Jun and Wang, Zheng and Xu, Kai},
	journal={CVPR},
	year={2020}
}

@article{wang2020self6d,
	title={{Self6D}: Self-Supervised Monocular {6D} Object Pose Estimation},
	author={Wang, Gu and Manhardt, Fabian and Shao, Jianzhun and Ji, Xiangyang and Navab, Nassir and Tombari, Federico},
	journal={ECCV},
	year={2020}
}

@article{chen2020g2l,
	title={{G2L-Net}: Global to Local Network for Real-time {6D} Pose Estimation with Embedding Vector Features},
	author={Chen, Wei and Jia, Xi and Chang, Hyung Jin and Duan, Jinming and Leonardis, Ales},
	journal={CVPR},
	year={2020}
}

@article{sock2020introducing,
	title={Introducing Pose Consistency and Warp-Alignment for Self-Supervised {6D} Object Pose Estimation in Color Images},
	author={Sock, Juil and Garcia-Hernando, Guillermo and Armagan, Anil and Kim, Tae Kyun},
	journal={3DV},
	year={2020}
}

@article{ammirato2020symgan,
	title={{SymGAN}: Orientation estimation without annotation for symmetric objects},
	author={Ammirato, Phil and Tremblay, Jonathan and Liu, Ming-Yu and Berg, Alexander and Fox, Dieter},
	journal={WACV},
	year={2020}
}

@article{deng2019poserbpf,
	title={{PoseRBPF}: A {R}ao-{B}lackwellized particle filter for {6D} object pose tracking},
	author={Deng, Xinke and Mousavian, Arsalan and Xiang, Yu and Xia, Fei and Bretl, Timothy and Fox, Dieter},
	journal={RSS},
	year={2019}
}

@article{zakharov2019deceptionnet,
	title={{DeceptionNet}: Network-driven domain randomization},
	author={Zakharov, Sergey and Kehl, Wadim and Ilic, Slobodan},
	journal={ICCV},
	year={2019}
}

@article{bauer2020verefine,
	title={{VeREFINE}: Integrating Object Pose Verification With Physics-Guided Iterative Refinement},
	author={Bauer, Dominik and Patten, Timothy and Vincze, Markus},
	journal={RAL},
	year={2020}
}

@inproceedings{pitteri2019cornet,
	title={{CorNet}: Generic {3D} Corners for {6D} Pose Estimation of New Objects without Retraining},
	author={Pitteri, Giorgia and Ilic, Slobodan and Lepetit, Vincent},
	booktitle={ICCVW},
	year={2019}
}

@misc{unrealenginepbr,
	author = {{Epic Games}},
	title = {Physically Based Materials in Unreal Engine},
	url = {https://docs.unrealengine.com/en-US/Engine/Rendering/Materials/PhysicallyBased},
	date = {2020},
}

@misc{evermotion,
  title={Evermotion},
  author={},
  year={},
  howpublished = {\url{www.evermotion.org}}
}

@misc{nixsensor,
  title={{NIX} {C}olor {S}ensor},
  author={},
  year={},
  howpublished = {\url{www.nixsensor.com}}
}

@misc{muravision,
  title={{M}ura},
  author={},
  year={},
  howpublished = {\url{www.muravision.com}}
}
	\endgroup
	
\end{document}